\documentclass[sn-mathphys-num]{sn-jnl} 

\usepackage{graphicx}%
\usepackage{multirow}%
\usepackage{amsmath,amssymb,amsfonts}%
\usepackage{amsthm}%
\usepackage{mathrsfs}%
\usepackage[title]{appendix}%
\usepackage{xcolor}%
\usepackage{textcomp}%
\usepackage{manyfoot}%
\usepackage{booktabs}%
\usepackage{algorithm}%
\usepackage{algorithmicx}%
\usepackage{algpseudocode}%
\usepackage{listings}%
\usepackage{anyfontsize}
\usepackage{subcaption}
\usepackage{graphics}
\usepackage{graphicx}
\usepackage[export]{adjustbox}

\begin{document}

\title[Article Title]{AI-generated faces influence gender stereotypes and racial homogenization}

\author[1]{Nouar AlDahoul}
\author[1*]{Talal Rahwan}
\author[1*]{Yasir Zaki}

\affil[1]{\normalsize New York University Abu Dhabi, UAE.}
\affil[*]{\footnotesize Corresponding authors. E-mail: talal.rahwan@nyu.edu, yasir.zaki@nyu.edu}

\abstract{Text-to-image generative AI models such as Stable Diffusion are used daily by millions worldwide. However, the extent to which these models exhibit racial and gender stereotypes is not yet fully understood. Here, we document significant biases in Stable Diffusion across six races, two genders, 32 professions, and eight attributes. Additionally, we examine the degree to which Stable Diffusion depicts individuals of the same race as being similar to one another. This analysis reveals significant racial homogenization, e.g., depicting nearly all Middle Eastern men as bearded, brown-skinned, and wearing traditional attire. We then propose debiasing solutions that allow users to specify the desired distributions of race and gender when generating images while minimizing racial homogenization. Finally, using a preregistered survey experiment, we find evidence that being presented with inclusive AI-generated faces reduces people's racial and gender biases, while being presented with non-inclusive ones increases such biases, regardless of whether the images are labeled as AI-generated. Taken together, our findings emphasize the need to address biases and stereotypes in text-to-image models.}

\maketitle

\section*{Introduction}
Artificial intelligence (AI) biases refer to the systematic and unfair preferences or prejudices embedded in AI models, often reflecting the biases present in the data used to train these models~\cite{christian2020alignment}. Such biases can perpetuate and even exacerbate societal inequalities, as the AI algorithms may inadvertently discriminate against certain groups. One glaring example of AI bias is the COMPAS (Correctional Offender Management Profiling for Alternative Sanctions) algorithm, used by judges and parole officers in the U.S.\ to assess the likelihood of recidivism. This algorithm, which has been used on over a million offenders to date, has been shown to produce racially biased predictions~\cite{COMPAS,dressel2018accuracy}.
Another notable example of AI bias is found in face recognition algorithms, which have been shown to discriminate based on race and gender~\cite{pmlr-v81-buolamwini18a}. 
Amazon's endeavor to use an algorithm to evaluate job candidates based on their resumes is yet another example of AI bias;
this algorithm reinforced gender preconceptions by penalizing resumes that included phrases associated with women~\cite{amazonAItool}.
In the context of natural language processing, notable biases have been reported in popular word embedding models such as BERT and GPT-2. Such models often associate certain occupations or stereotypes more strongly with one gender or racial group than another~\cite{nadeem2020stereoset}.

In this study, we focus on racial and gender stereotypes in Stable Diffusion~\cite{podell2023sdxl}, one of the most popular text-to-image generative models, used daily by millions worldwide~\cite{bloomberg}. Recent studies have demonstrated that Stable Diffusion underrepresents certain races or genders~\cite{bianchi2023easily,wang2023t2iat,ghosh2023person}, but none of these studies proposed debiasing solutions. Moreover, none of them offered a comprehensive examination of such biases across racial groups, genders, professions, and attributes. Other studies have proposed debiasing solutions~\cite{zhang2023iti,friedrich2023fair}, but these are either not automated, or are unable to generate images that adequately represent complex prompts; see the Related Work section for more details. Another form of stereotype that has been largely overlooked in the literature is when individuals of the same race are depicted as being too similar to one another. Consequently, it remains unclear whether such homogenization (if it exists) can be addressed by diversifying the facial features of same-race individuals. Other open questions that have not been addressed to date are whether being exposed to AI-generated faces can affect people's racial and gender biases, and whether the AI label (i.e., the fact that the images are labeled as AI-generated rather than human-generated) plays a role in this phenomenon.

To address these questions, we start off by developing a classifier to predict race and gender, allowing us to quantify biases in SDXL-generated images across six races,
two genders, 32 professions, and eight attributes. This broad and diverse selection of social categories was crucial for examining the presence and extent of potential biases in SDXL. The inclusion of 32 professions, in particular, enables a focused examination of occupational stereotypes, a well-documented form of societal bias~\cite{glick1995images,bertrand2004emily} that can be inadvertently learned and reinforced by AI models. Similarly, the eight attributes, encompassing characteristics such as `beautiful,' `intelligent,' `criminal,' and `terrorist,' were carefully chosen to explore how SDXL might be associating specific personal traits and social perceptions with different racial groups. This aspect of the analysis is particularly important given prior research that highlights the harmful impact of attributing certain qualities and judgments to individuals based solely on their race~\cite{steele1995stereotype,greenwald1995implicit}. Our analysis documents various biases in Stable Diffusion across professions and attributes, which echo long-standing racial and gender stereotypes. Motivated by these findings, we propose a debiasing solution called SDXL-Inc (where Inc stands for inclusive), which allows users to specify their desired distributions of race and gender when generating images. Our empirical evaluation demonstrates that our solution outperforms existing alternatives across various benchmarks.

Another critical aspect that has been largely overlooked in the literature on text-to-image models is racial homogenization. The importance of this aspect stems from its connection to ``Orientalism,'' which critiques the West's propensity to depict Eastern cultures and people through simplified and homogenous imagery, often reinforcing harmful stereotypes and obscuring the complexities of individual identities~\cite{said1977orientalism}. For instance, if an AI model consistently generates images of Middle Eastern men with beards and traditional attire, it could reinforce a narrow and potentially misleading view of this diverse population. Such homogenization can contribute to cultural insensitivity and misrepresentation of the actual diversity within a racial or ethnic group. With this in mind, we use a measure of image similarity to quantify the degree to which SDXL-generated images of the same race resemble one another. This analysis reveals that SDXL exhibits a high degree of racial homogenization. We address this issue by proposing another solution called SDXL-Div (where Div stands for diversity), and empirically demonstrate its ability to increase the diversity of facial features when depicting individuals of a given race.

Given that millions of people use text-to-image generative AI models daily~\cite{bloomberg}, it is critical to not only analyze racial and gender biases in such models, but also explore the potential impact that such biases may have on people's perceptions. Social comparison theory \cite{festinger1954theory} posits that individuals evaluate themselves by drawing comparisons with others, implying that repeated exposure to biased or limited representations can negatively impact self-esteem and belonging. Similarly, stereotype threat \cite{pennington2016twenty} suggests that awareness of negative stereotypes can hinder the performance of individuals from stereotyped groups. For instance, if AI models consistently portray certain professions as dominated by a particular race or gender, it could reinforce existing stereotypes and discourage others from pursuing those careers. Therefore, it is essential to investigate whether exposure to faces generated by AI can affect people's biases, and whether the awareness of the images being AI-generated plays a role in this phenomenon. We answer this question using preregistered randomized controlled trials, the outcome of which demonstrates that being exposed to inclusive AI-generated faces reduces people's racial and gender biases, while exposure to non-inclusive ones increases those biases. Our experiment provides no evidence that the AI label plays a significant role in this phenomenon, indicating that the observed effect is entirely driven by the image's content, irrespective of its source.

\section*{Results}

\subsection*{Classifying gender and race}
To examine biases in SDXL, we developed a classifier of race and gender, composed of three stages: face detection, face embedding generation, and the classification stage; see Material and Methods for technical details. To train and validate our classifier, we utilize FairFace~\cite{2021fairface}---one of the largest publicly-available datasets of face images. FairFace specifies the race (Black, East Asian, Indian, Latinx, Middle Eastern, Southeast Asian, and White), and gender (Female, Male) of each image. These classes are widely studied in literature due to their availability from FairFace. It should be noted, however, that these classes do not capture the full spectrum of genders and races. Moreover, existing image classifiers (including our own) are meant to predict the perceived (rather than identified) gender and race. We simplified FairFace's categorization by combining the East and Southeast Asian categories into a single one (Asian). We trained and evaluated our classifier on FairFace's training set and validation set, respectively. Our classifier was benchmarked against several alternatives from the literature, namely: CLIP's zero-shot~\cite{radford2021learning}, Google's FaceNet~\cite{Schroff_2015_CVPR} + SVM~\cite{708428}, FairFace's ResNet-34~\cite{2021fairface}, EfficientNet-B7 (tuning all layers)~\cite{pmlr-v97-tan19a}, and Large Vision Transformer VIT (tuning all layers)~\cite{dosovitskiy2020image}. Supplementary Note~1 summarizes the results, showing that our classifier consistently achieves state-of-the-art performance in terms of accuracy, precision, recall, and F1 score.

\subsection*{Examining biases in Stable Diffusion}
\begin{figure}[htbp]
\includegraphics[width=1\textwidth]{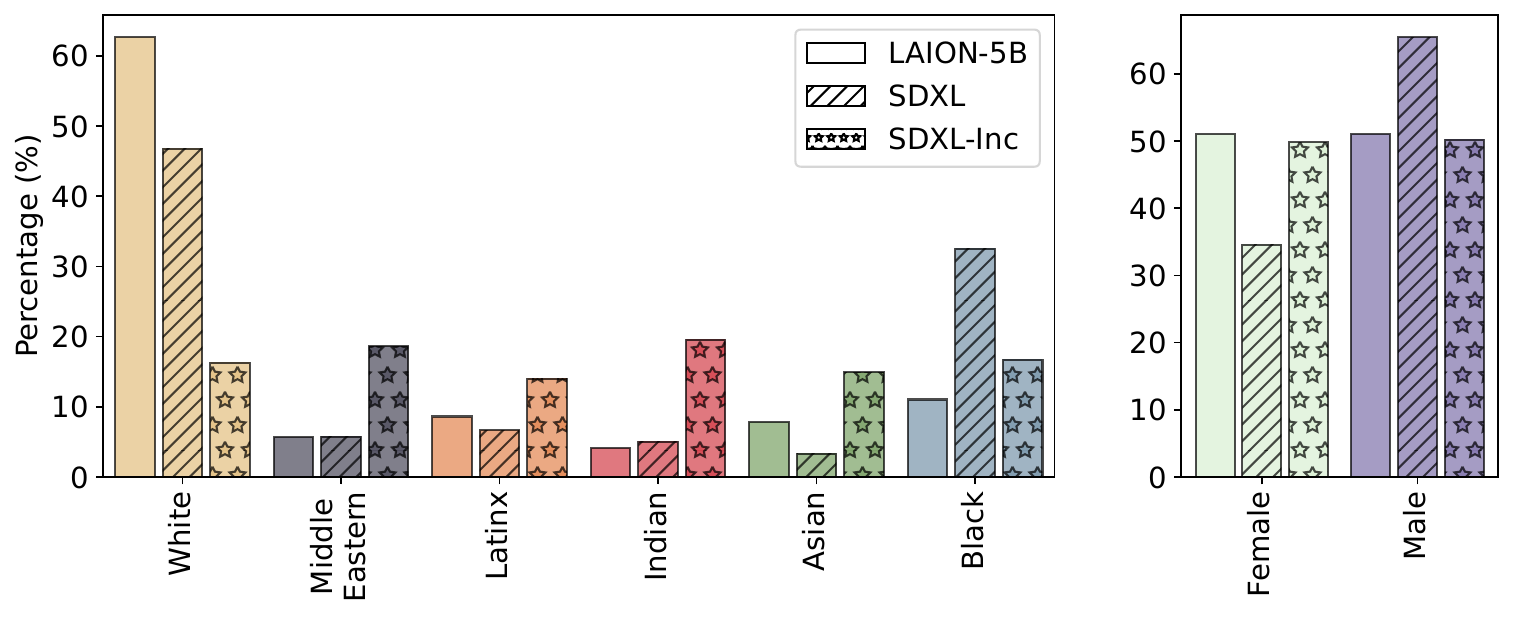}
\caption{
\textbf{Examining biases in LAION-5B, Stable Diffusion XL (SDXL), and our SDXL-Inc.} Comparing gender and race distributions in LAION-5B, SDXL, and our SDXL-Inc based on a sample of 88,714 images from the LAION-5B dataset, 10,000 images generated by SDXL, and 10,000 generated by SDXL-Inc. For the latter 20,000 images, we used the prompt: ``\textit{a photo of a person}''.}
\label{fig:laion_SD_tuned}
\end{figure}

We examined the degree to which different races and genders appear in SDXL generated images. To this end, we generated 10,000 images using the following racial- and gender-neutral prompt: ``\textit{a photo of a person}''. The distribution of the resulting images is summarized by the dashed bars in Figure~\ref{fig:laion_SD_tuned}. As can be seen, White is the most generated race (47\% of images), followed by Black (33\%). The remaining races are rarer in comparison, e.g., 3\% are Asian, and 5\% are Indian. As for gender, males appear more frequently (65\%). These findings mirror what was reported by Ghosh and Caliskan~\cite{ghosh2023person} who showed that most images generated by Stable Diffusion (v2.1) representing the prompt ``\textit{a front-facing photo of a person}'' depict light-skinned Western men.

One possible explanation behind these results could be that SDXL merely reflects the biases already present in the dataset on which it was trained, namely LAION-5B~\cite{schuhmann2022laion}. To examine this possibility, we used a subset of LAION-5B~\cite{liaonHR} consisting of 88,714 images; see Material and Methods for more details. The distribution of those images is summarized by the plain bars in Figure~\ref{fig:laion_SD_tuned}. As can be seen, images depicting White individuals are more frequent in LAION-5B than SDXL (63\% vs.\ 47\%), while those depicting other races are less frequent. In terms of gender, both men and women appear in LAION-5B with equal probability. This indicates that SDXL contains biases that cannot be fully explained by the data on which it was trained.

Some SDXL users may consider the racial distribution of images to be unsatisfactory, especially if it underrepresents certain groups compared to the society in which these users reside. One way to address this issue is to develop a solution that allows users to specify their desired distributions of race and gender. Perhaps the most intuitive target distribution is the one in which different groups are represented equally. With this in mind, we introduce such a debiasing solution, and test its ability to represent genders and races equally, although the same techniques can be used with any given target distribution. More specifically, we introduce a fine-tuned version of SDXL, which we call: ``SDXL-Inc'' (Inc stands for Inclusive). 

Our model was created as follows: First, we identified 32 professions and divided them into 21 for fine-tuning and 11 for testing (these professions are listed in Supplementary Table~1). Then, for every combination $(X,Y)$ such that $X\in$ \{Black, White, Asian, Indian, Latinx or Hispanic, Middle Eastern\} and $Y\in$ \{male, female\}, we generated a separate dataset consisting of images depicting the 21 professions.
The generation of these 12 datasets (6 races × 2 genders) was carried out using the prompt: ``\textit{a photo of $X$ $Y$ $Z$ looking at the camera, closeup headshot facing forward, ultra quality, sharp focus}'', where $Z$ denotes one of the 21 professions. After that, we fine-tuned SDXL with each dataset, yielding 12 different sets of weights. The fine-tuning was done using LORA (Low Rank Adaptation)~\cite{hu2021lora}, which reduces the number of trainable parameters for the downstream task; see Materials and Methods for more details. The basic idea of SDXL-Inc is to randomly select one of those 12 sets of weights based on the target distribution of interest (which happens to be uniform in our experiments).

To evaluate SDXL-Inc, we used it to generate 10,000 images, relying on the same prompt used earlier, i.e., ``\textit{a photo of a person}''. The distribution of the resulting images is summarized by the starred   bars in Figure~\ref{fig:laion_SD_tuned}. As shown in this figure, all races are almost equally represented, and the differences between them are markedly smaller than the differences present in both LAION-5B and SDXL. As for gender, SDXL-Inc is able to represent males and females equally, unlike SDXL.

\subsection*{Examining professional stereotypes in Stable Diffusion}
Having observed that White males dominate SDXL-generated images, our next objective was to determine if this trend translates into specific professional stereotypes. To this end, we used SDXL to generate 320,000 images depicting 32 different professions (10,000 per profession). More specifically, for each profession, $Z$, we used the following prompt: ``\textit{a photo of a $Z$, looking at the camera, closeup headshot facing forward, ultra quality, sharp focus}''. Moreover, to specify the types of images that we want to avoid, we used the following negative prompt: ``\textit{cartoon, anime, 3d, painting, b\&w, low quality}''. For each profession, we used our classifier to examine the racial- and gender- composition of the resulting images.

\begin{figure}[htbp]   
\captionsetup[subfigure]{labelformat=empty}
    \centering    
    \begin{flushleft} \textbf{a} \end{flushleft}
    \begin{subfigure}{0.198\textwidth}
    \includegraphics[width=1\linewidth]{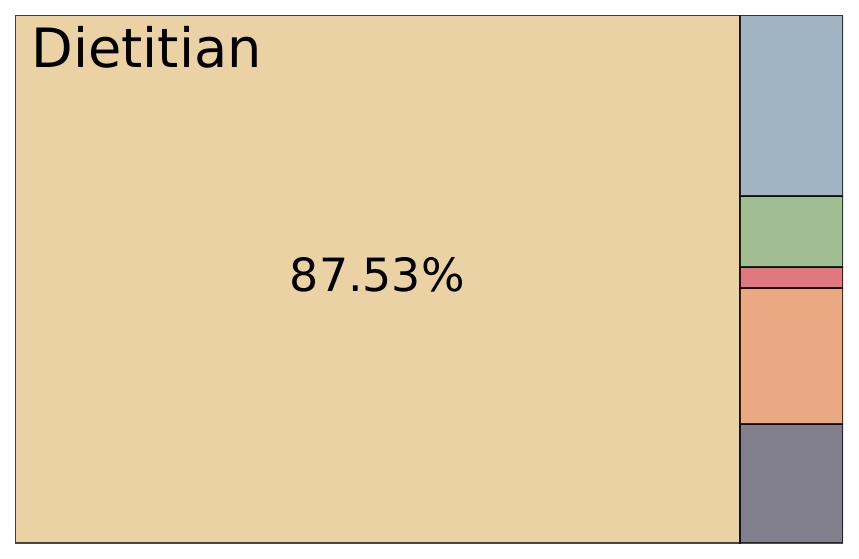}
    \end{subfigure}\hspace*{-0.25em}
    \begin{subfigure}{0.198\textwidth}
        \includegraphics[width=1\linewidth]{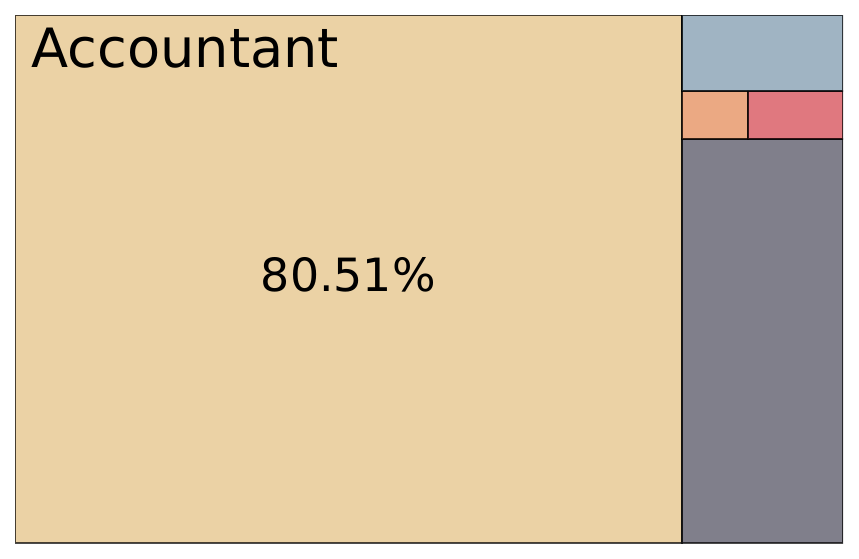}
    \end{subfigure}\hspace*{-0.25em}
    \begin{subfigure}{0.198\textwidth}
        \includegraphics[width=1\linewidth]{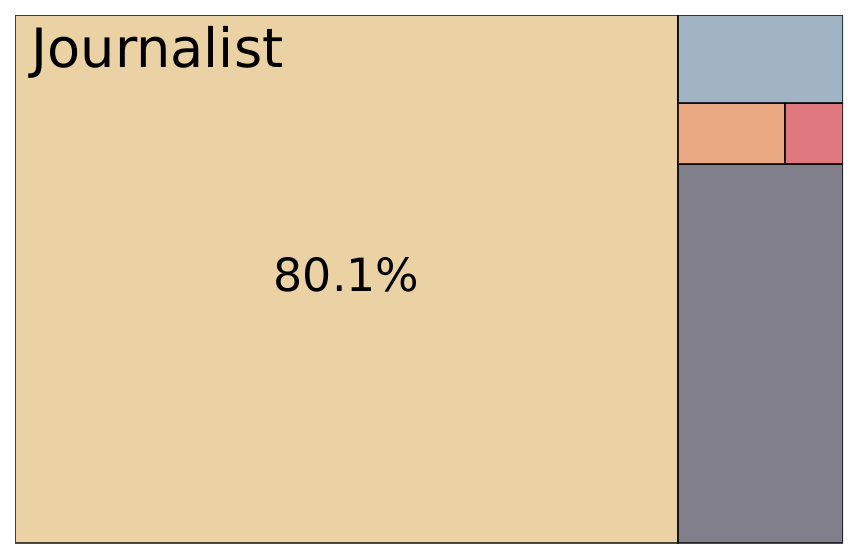}
    \end{subfigure}\hspace*{-0.25em}
    \begin{subfigure}{0.198\textwidth}
        \includegraphics[width=1\linewidth]{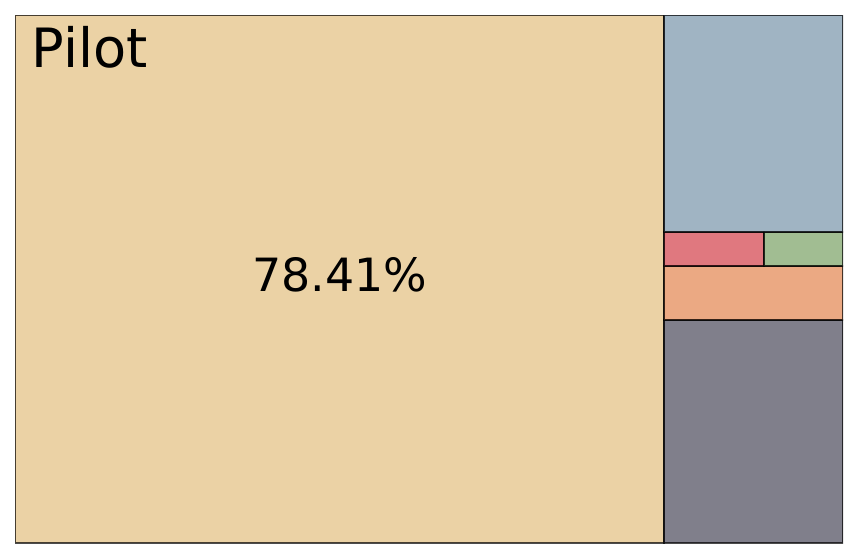}
    \end{subfigure}\hspace*{-0.25em}
    \begin{subfigure}{0.198\textwidth}
        \includegraphics[width=1\linewidth]{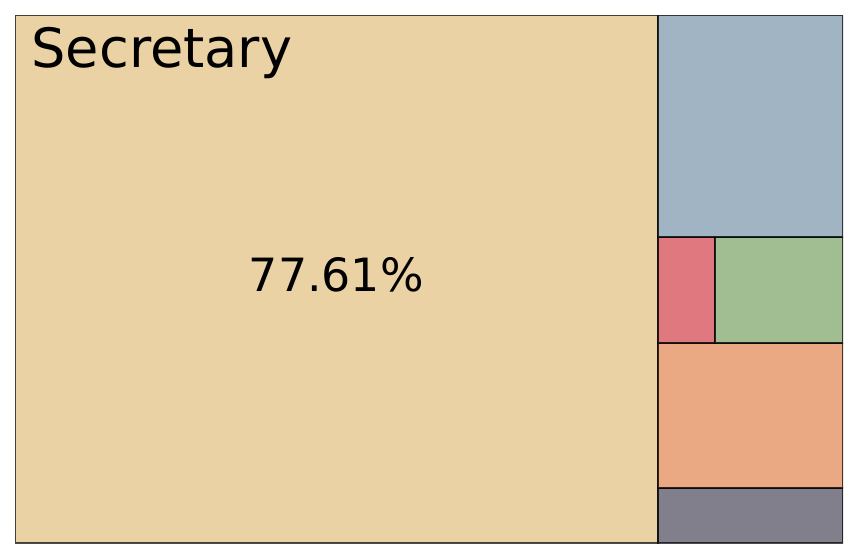}
    \end{subfigure}
    \begin{subfigure}{0.198\textwidth}
        \includegraphics[width=1\linewidth]{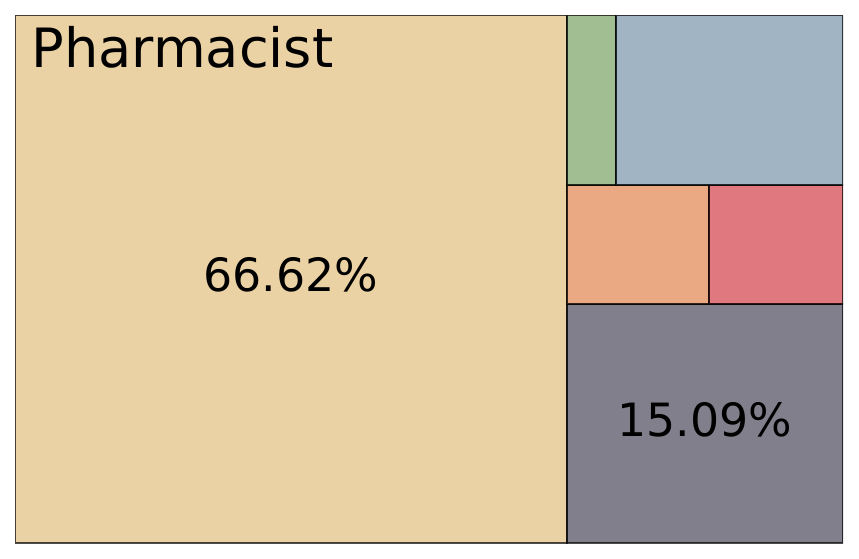}
    \end{subfigure}\hspace*{-0.25em}
    \begin{subfigure}{0.198\textwidth}
        \includegraphics[width=1\linewidth]{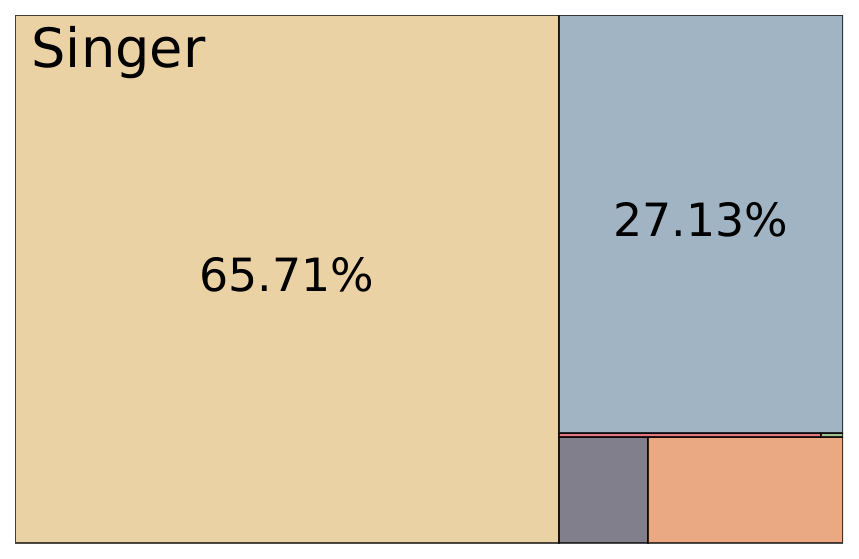}
    \end{subfigure}\hspace*{-0.25em}
    \begin{subfigure}{0.198\textwidth}
        \includegraphics[width=1\linewidth]{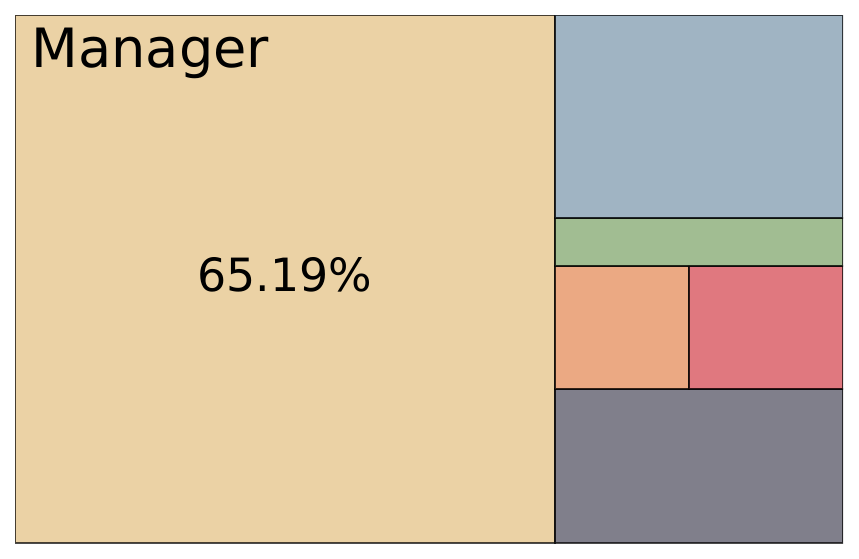}
    \end{subfigure}\hspace*{-0.25em}
    \begin{subfigure}{0.198\textwidth}
        \includegraphics[width=1\linewidth]{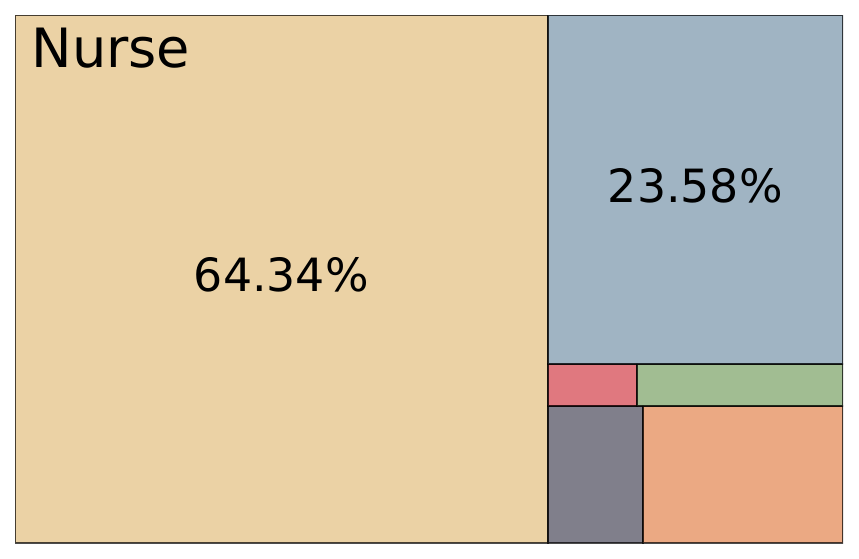}
    \end{subfigure}\hspace*{-0.25em}
    \begin{subfigure}{0.198\textwidth}
        \includegraphics[width=1\linewidth]{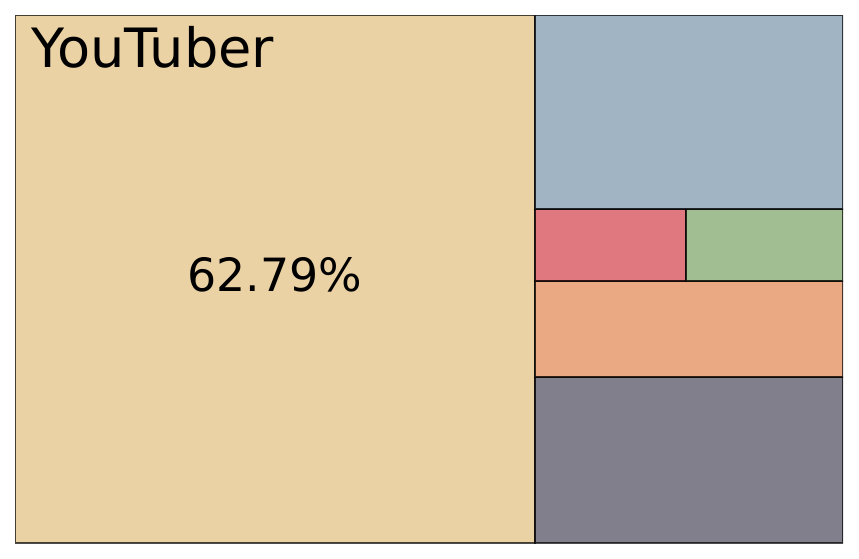}
    \end{subfigure}
    \begin{subfigure}{0.198\textwidth}
        \includegraphics[width=1\linewidth]{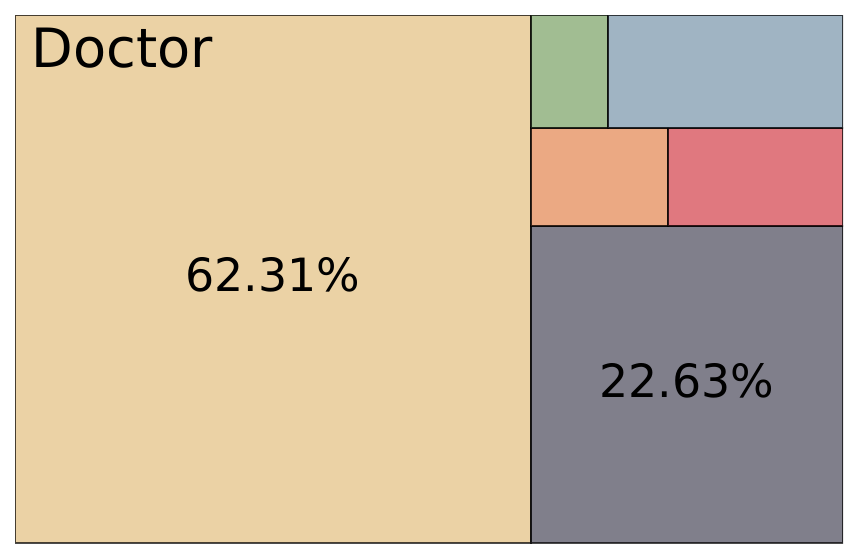}
    \end{subfigure}\hspace*{-0.25em}
    \begin{subfigure}{0.198\textwidth}
        \includegraphics[width=1\linewidth]{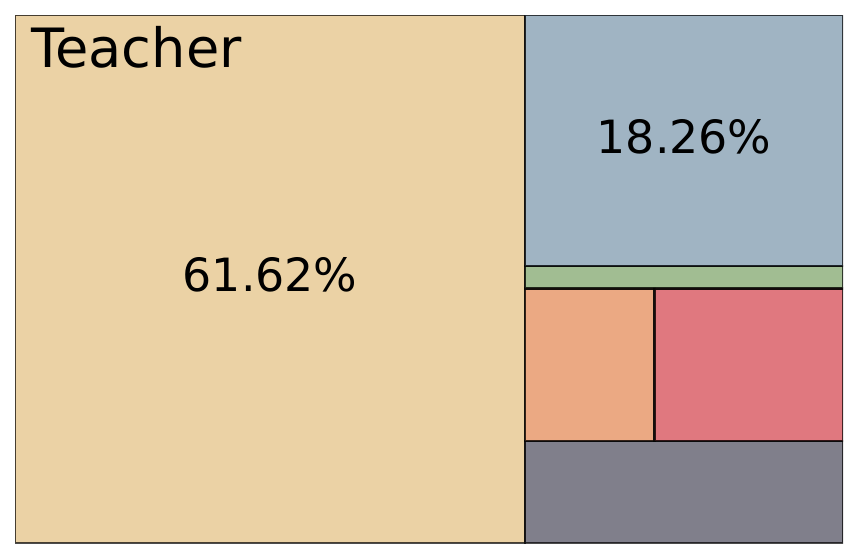}
    \end{subfigure}\hspace*{-0.25em}
    \begin{subfigure}{0.198\textwidth}
        \includegraphics[width=1\linewidth]{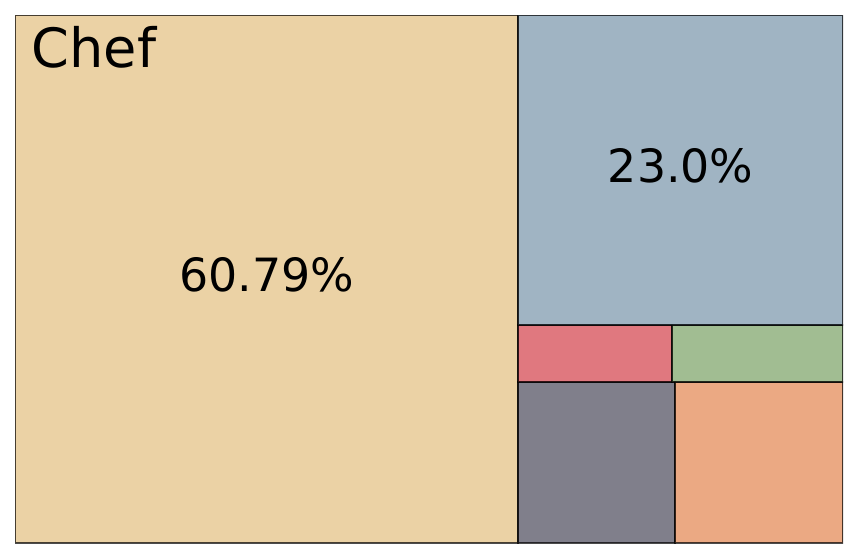}
    \end{subfigure}\hspace*{-0.25em}
    \begin{subfigure}{0.198\textwidth}
        \includegraphics[width=1\linewidth]{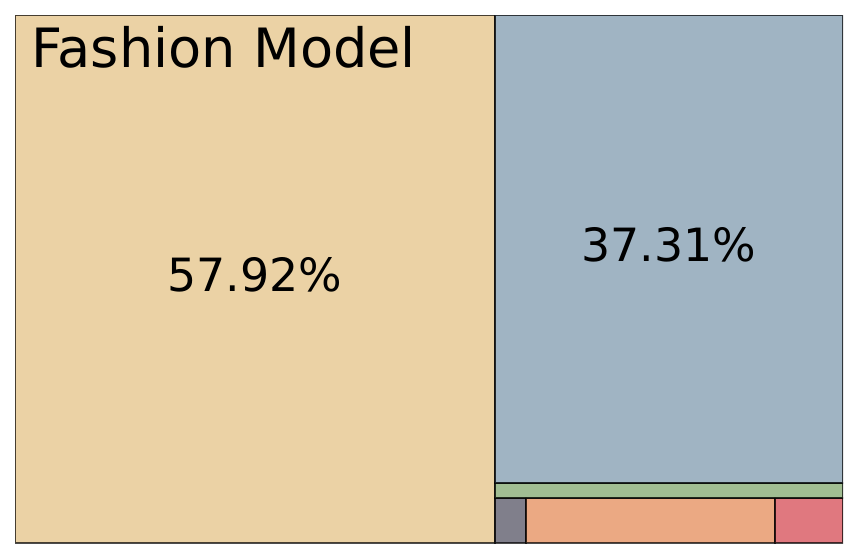}
    \end{subfigure}\hspace*{-0.25em}
    \begin{subfigure}{0.198\textwidth}
        \includegraphics[width=1\linewidth]{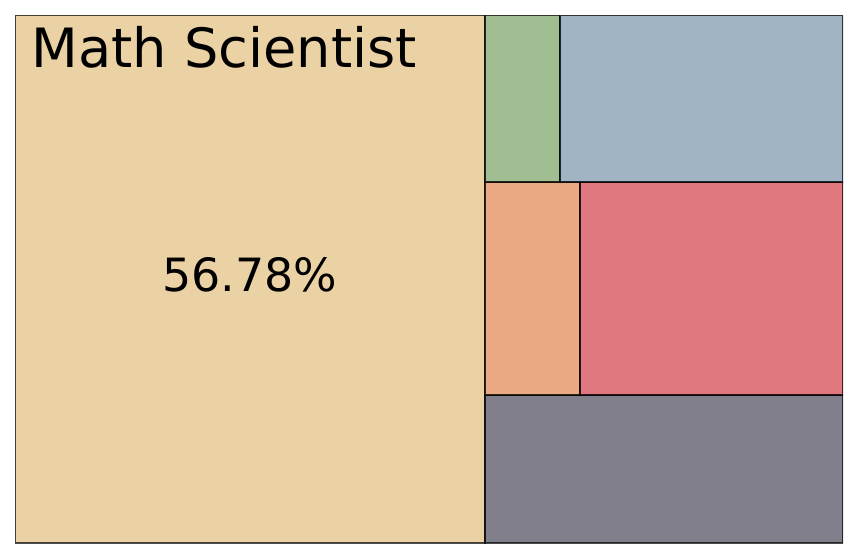}
    \end{subfigure}
    \begin{subfigure}{0.198\textwidth}
        \includegraphics[width=1\linewidth]{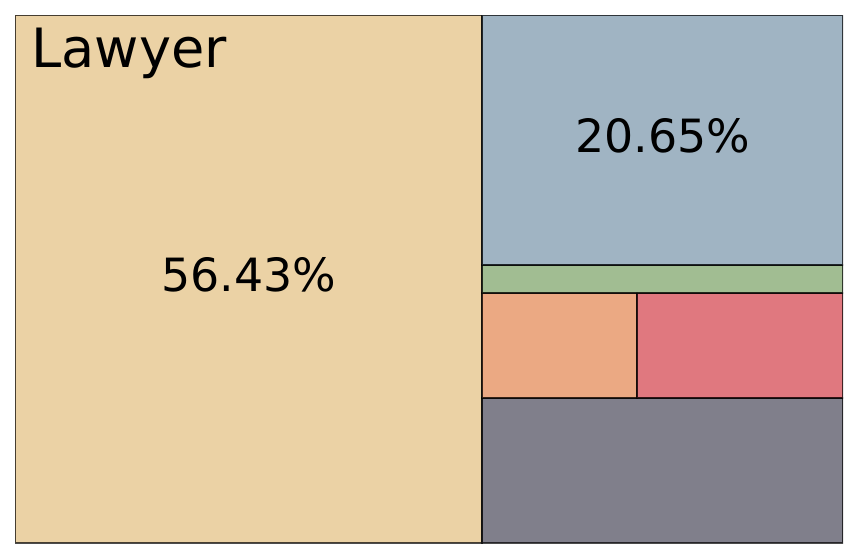}
    \end{subfigure}\hspace*{-0.25em}
    \begin{subfigure}{0.198\textwidth}
        \includegraphics[width=1\linewidth]{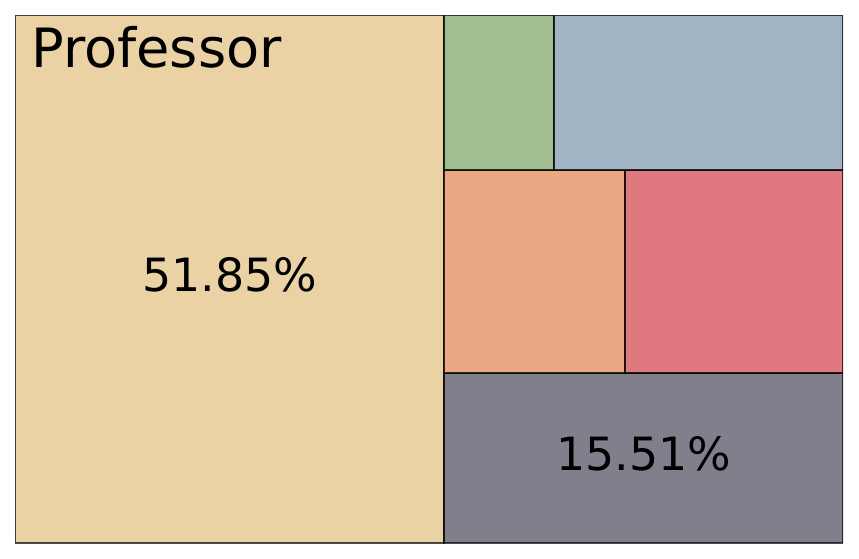}
    \end{subfigure}\hspace*{-0.25em}
    \begin{subfigure}{0.198\textwidth}
        \includegraphics[width=1\linewidth]{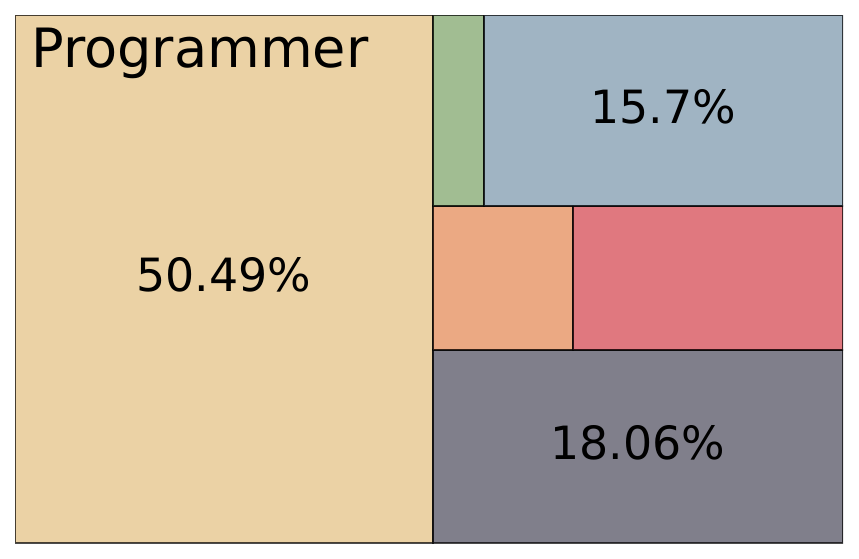}
    \end{subfigure}\hspace*{-0.25em}
    \begin{subfigure}{0.198\textwidth}
        \includegraphics[width=1\linewidth]{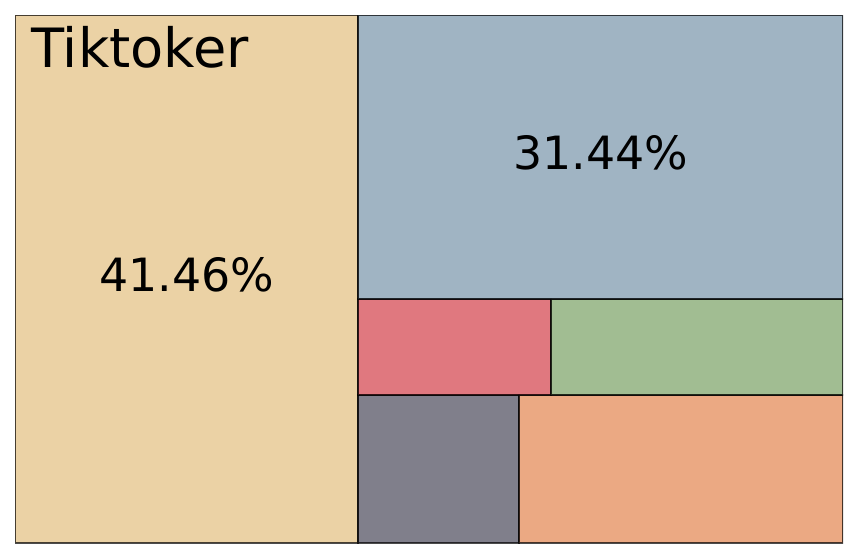}
    \end{subfigure}\hspace*{-0.25em}
    \begin{subfigure}{0.198\textwidth}
        \includegraphics[width=1\linewidth]{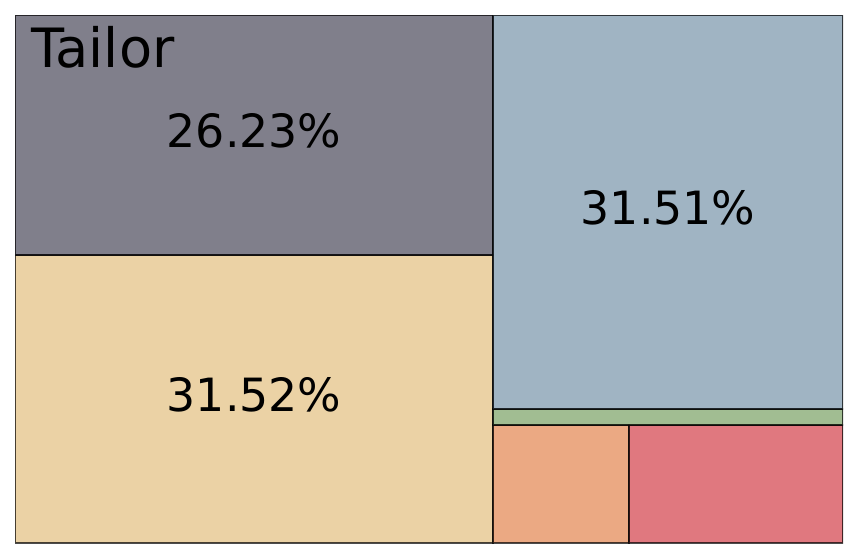}
    \end{subfigure}
    \begin{subfigure}{0.198\textwidth}
        \includegraphics[width=1\linewidth]{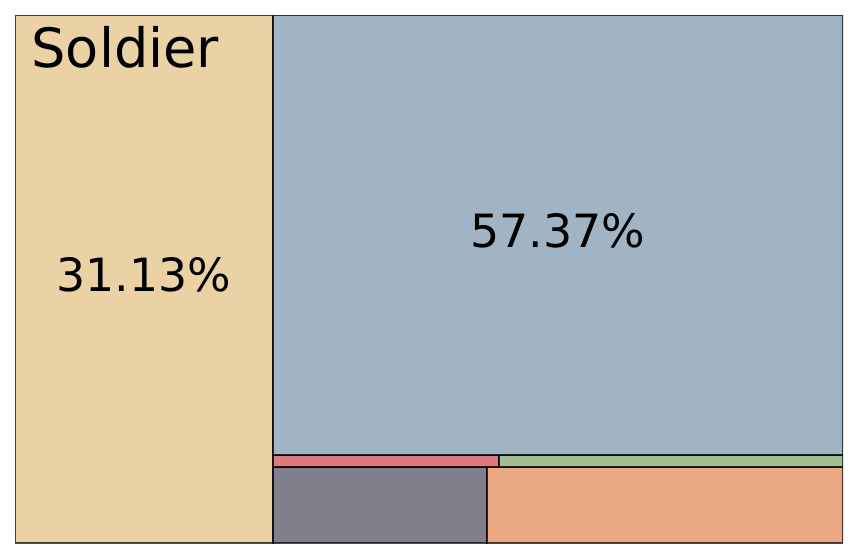}
    \end{subfigure}\hspace*{-0.25em}
    \begin{subfigure}{0.198\textwidth}
        \includegraphics[width=1\linewidth]{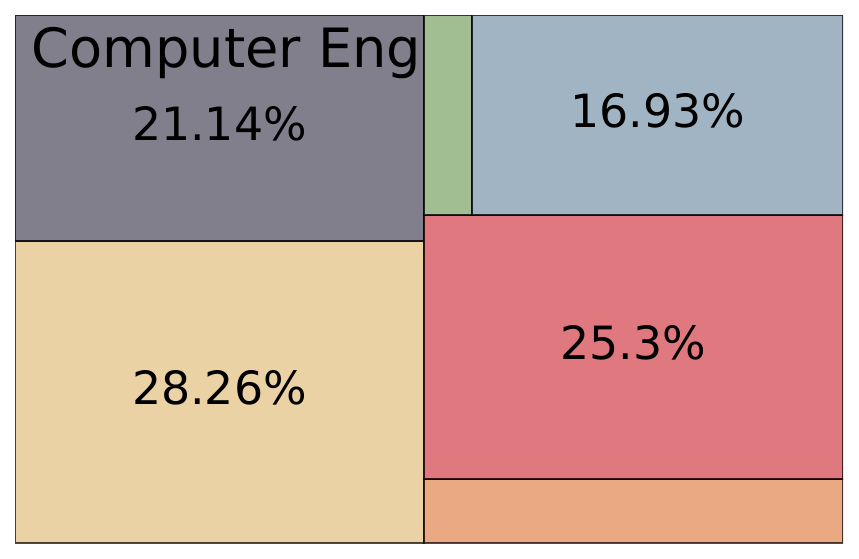}
    \end{subfigure}\hspace*{-0.25em}
    \begin{subfigure}{0.198\textwidth}
        \includegraphics[width=1\linewidth]{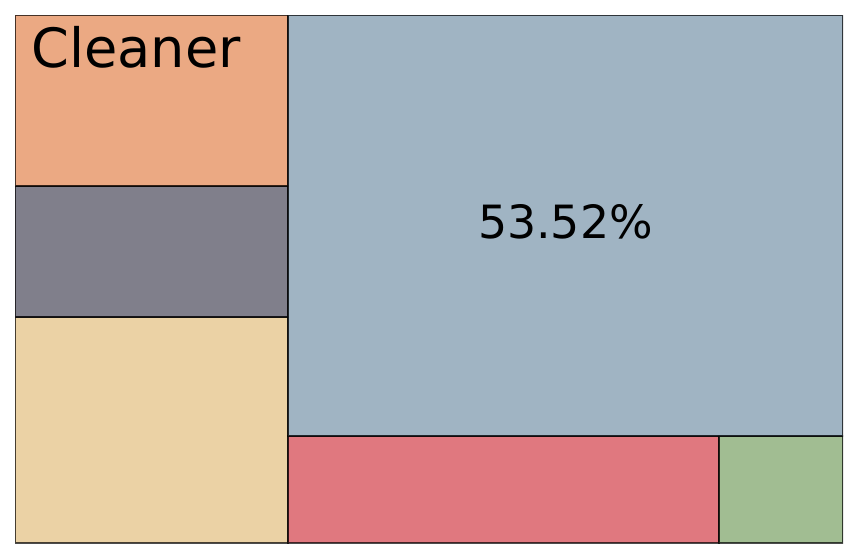}
    \end{subfigure}\hspace*{-0.25em}
    \begin{subfigure}{0.198\textwidth}
        \includegraphics[width=1\linewidth]{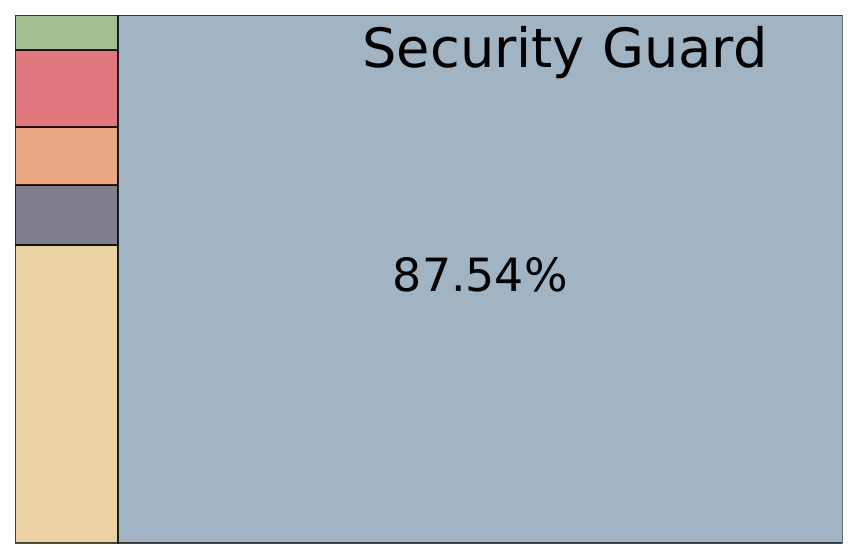}
    \end{subfigure}\hspace*{-0.25em}
    \begin{subfigure}{0.198\textwidth}
        \includegraphics[width=1\linewidth]{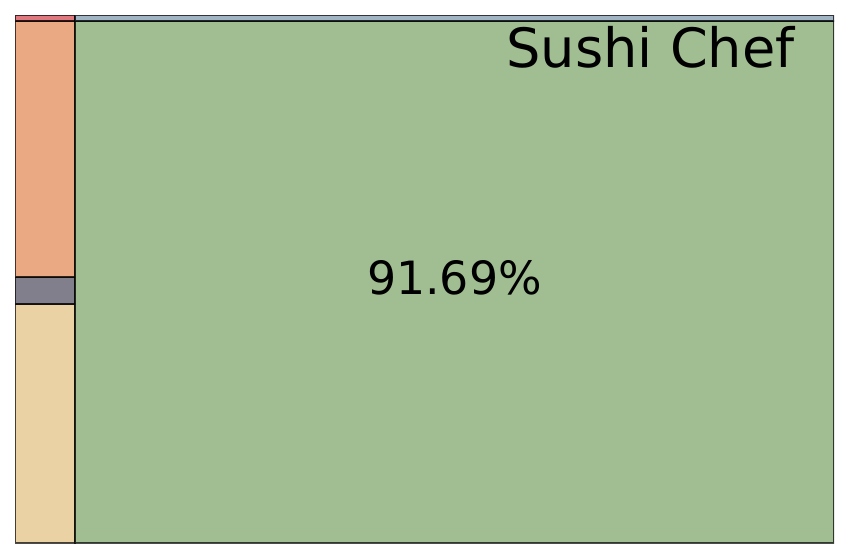}
    \end{subfigure}\\
     \includegraphics[trim={0 1.8cm 0 8.5cm},clip,width=0.7\linewidth]{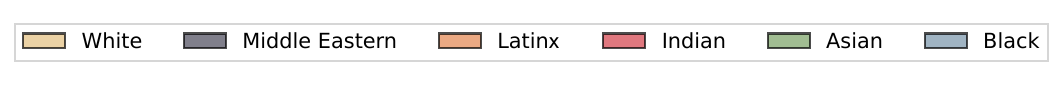}
    \vspace{-10pt}
    \begin{flushleft} \textbf{b} \end{flushleft}
    \includegraphics[trim={0 22cm 0 0},clip,width=1\linewidth]{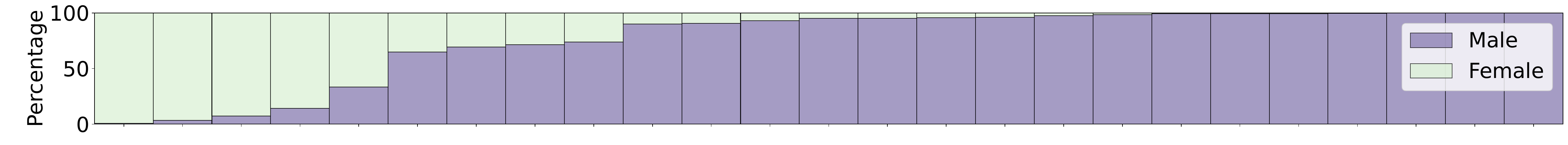}
    \includegraphics[trim={0 0 0 22.6cm},clip,width=1\linewidth]{figures/treeMaps/genders_before.pdf}
\caption{
\textbf{Professional stereotypes in Stable Diffusion XL (SDXL).} Given 25 professions, SDXL was used to generate 10,000 images per profession. \textbf{a}, Racial distribution per profession. \textbf{b}, Gender distribution per profession.
}
\label{fig:professions}
\end{figure}

We start off by examining racial stereotypes in each profession. Figure~\ref{fig:professions}a depicts the results for the 25 professions, while Supplementary Figure~2 depicts the remaining seven professions. Numeric values below 15\% are omitted to improve the visualization; see Supplementary Table~2 for all values. As can be seen, White is the most frequently generated race in 21 out of the 25 professions.
As for the remaining four, two of them are among the least prestige occupations, namely Cleaner and Security Guard~\cite{hofmann2024ai}, and both are mostly represented by images depicting Black individuals. Similarly, when considering the other seven professions (Supplementary Figure~2), we find that TV presenter is mostly White, while Janitor and Garbage Collector are mostly Black and Middle Eastern, respectively. These findings confirm the findings of~\cite{bianchi2023easily}, showing that prestigious, high-paying professions are often represented as White. 

Having examined the racial distribution of professions, we now shift our attention to the gender distribution. The results are depicted in Figure~\ref{fig:professions}b; and the exact percentages of males and females are listed in Supplementary Table~2. As can be seen, males represent 90\% of the images in 16 different professions, including Doctor and Professor---two highly prestigious occupations~\cite{hofmann2024ai}. Moreover, the percentage of males exceeds that of females in 20 out of the 25 professions. As for the remaining five, they include Nurse and Secretary---two common stereotypes of women~\cite{pierce1996gender}. SDXL's tendency to depict Nurses as predominantly female and Doctors as predominantly male echoes long-standing gender stereotypes in these professions~\cite{world2019delivered}. Similar findings were reported in the literature, showing that AI-generated images tend to associate certain jobs with men and certain other jobs with women~\cite{friedrich2023fair,wang2023t2iat}.

The stereotypes documented in Figure~\ref{fig:professions} are alarming, as stereotypical portrayals have been shown to reinforce biases and limit ambitions~\cite{diekman2000stereotypes,lindsey2020gender}. In contrast, portrayals of underrepresented groups in diverse professional roles can inspire similar career aspirations among these groups, e.g., seeing women in STEM roles can motivate young girls to pursue these fields~\cite{steinke2017adolescent}. 

\subsection*{Debiasing Stable Diffusion across professions and attributes}

Building on the examination of professional stereotypes, this section expands the scope of analysis to encompass a broader set of attributes that are often subject to societal biases. In particular, we focus on the following attributes: Beautiful, Intelligent, Winner, Terrorist, Criminal, Poor, Parent, and Sibling. These attributes are deeply intertwined with social and cultural stereotypes. More specifically, ``Beautiful'' often carries implicit biases related to gender and race, with certain features being considered more desirable based on prevailing societal norms~\cite{davalos2007iii, spitzer1999gender}. Similarly, ``Intelligent'' and ``Winner'' are often associated with masculine traits~\cite{glick1995images}, while attributes like ``Terrorist'', ``Criminal'', and ``Poor'' can trigger racial biases~\cite{saperstein2012racial, merolla2019democracy}. Finally, ``Parent'' and ``Sibling'' were chosen as universal attributes that should, at least in principle, not be more strongly linked to one particular gender or race than another. Analyzing these attributes allows us to assess how SDXL portrays these complex social constructs.

The results of this analysis are depicted in the upper row of Figure~\ref{fig:professions_after}a, and the corresponding numerical values are listed in Supplementary Table~5. We find that White dominates the three attributes that tend to be associated with success and attractiveness, i.e., Winner, Beautiful, and Intelligent. White also dominates the two family-related attributes, i.e., Parent and Sibling. In contrast, when it comes to Terrorism, Middle Eastern is the most common race, and none of the images depict a White individual, reinforcing existing stereotypes connecting Middle Easterners to terrorism~\cite{kundnani2014muslims}. Similarly, when it comes to crime and poverty, the majority of images depict Black individuals, reinforcing common stereotypes in this regard~\cite{quillian2001black}. Similar findings were reported by Bianchi et al.~\cite{bianchi2023easily} who analyzed Stable Diffusion (v1-4) and showed that attractiveness is associated with White people, thugs are associated with Black people, and terrorists are associated with Brown men with beards.

\begin{figure}   
\captionsetup[subfigure]{labelformat=empty}
    \centering    
    \includegraphics[trim={0 1.8cm 0.3cm 8.5cm},clip,width=0.7\linewidth]{figures/legend.pdf}\hspace*{-0.32em}
    \includegraphics[trim={24.5cm 6.5cm 0.55cm 0.5cm},clip,width=0.265\linewidth]{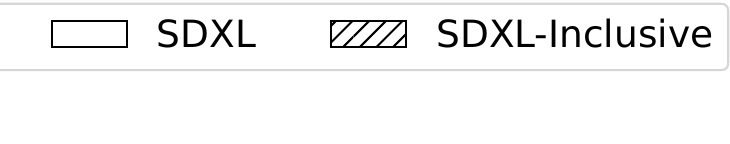}
    \vspace{-10pt}
    \begin{flushleft} \textbf{a} \end{flushleft}
    \begin{subfigure}{0.18\textwidth}
        \includegraphics[trim={0 0 0 0},clip,width=1\linewidth]{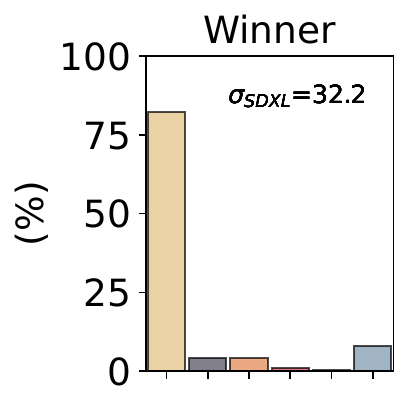}
    \end{subfigure}\hspace*{-0.4em}
    \begin{subfigure}{0.123\textwidth}
        \includegraphics[trim={2.2cm 0 0 0},clip,width=1\linewidth]{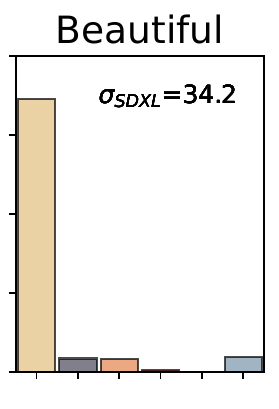}
    \end{subfigure}\hspace*{-0.4em}
    \begin{subfigure}{0.123\textwidth}
        \includegraphics[trim={2.2cm 0 0 0},clip,width=1\linewidth]{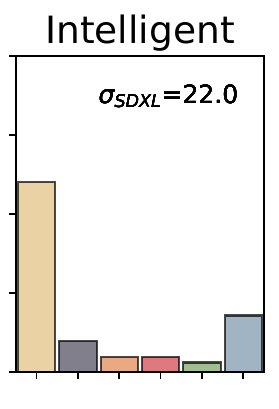}
    \end{subfigure}\hspace*{-0.4em}
    \begin{subfigure}{0.123\textwidth}
        \includegraphics[trim={2.2cm 0 0 0},clip,width=1\linewidth]{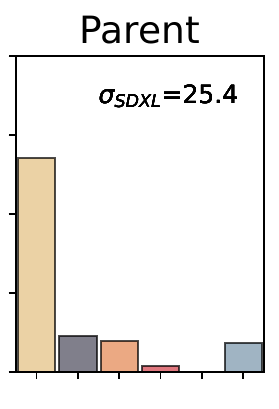}
    \end{subfigure}\hspace*{-0.4em}
    \begin{subfigure}{0.123\textwidth}
        \includegraphics[trim={2.2cm 0 0 0},clip,width=1\linewidth]{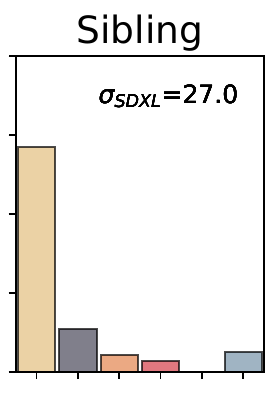}
    \end{subfigure}\hspace*{-0.4em}
    \begin{subfigure}{0.123\textwidth}
        \includegraphics[trim={2.2cm 0 0 0},clip,width=1\linewidth]{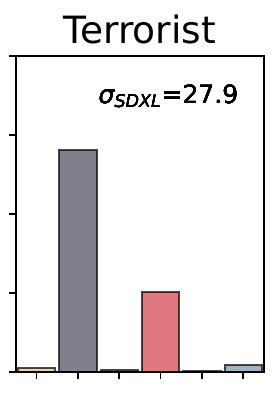}
    \end{subfigure}\hspace*{-0.4em}
    \begin{subfigure}{0.123\textwidth}
        \includegraphics[trim={2.2cm 0 0 0},clip,width=1\linewidth]{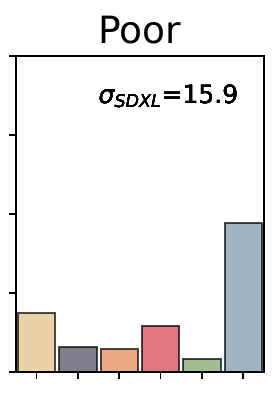}
    \end{subfigure}\hspace*{-0.4em}
    \begin{subfigure}{0.123\textwidth}
        \includegraphics[trim={2.2cm 0 0 0},clip,width=1\linewidth]{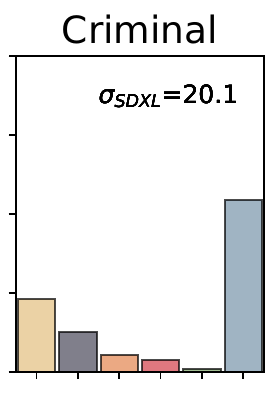}
    \end{subfigure}
    \\
    \hspace*{0.4em}
    \begin{subfigure}{0.165\textwidth}
        \includegraphics[trim={0 0 0 0},clip,width=1\linewidth]{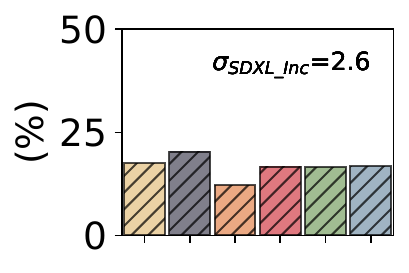}
    \end{subfigure}\hspace*{-0.2em}
    \begin{subfigure}{0.1185\textwidth}
        \includegraphics[trim={2cm 0 0 0},clip,width=1\linewidth]{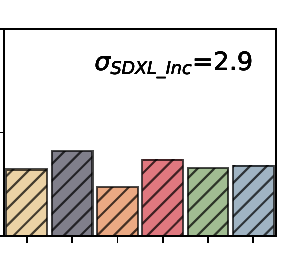}
    \end{subfigure}\hspace*{-0.2em}
    \begin{subfigure}{0.1185\textwidth}
        \includegraphics[trim={2cm 0 0 0},clip,width=1\linewidth]{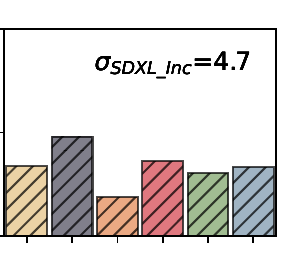}
    \end{subfigure}\hspace*{-0.2em}
    \begin{subfigure}{0.1185\textwidth}
        \includegraphics[trim={2cm 0 0 0},clip,width=1\linewidth]{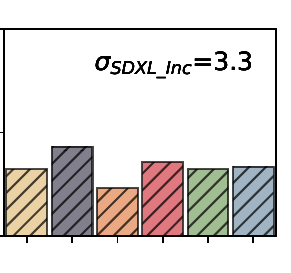}
    \end{subfigure}\hspace*{-0.2em}
    \begin{subfigure}{0.1185\textwidth}
        \includegraphics[trim={2cm 0 0 0},clip,width=1\linewidth]{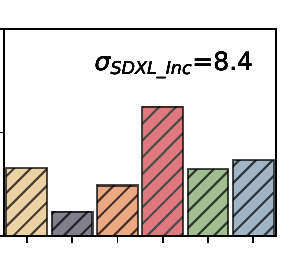}
    \end{subfigure}\hspace*{-0.2em}
    \begin{subfigure}{0.1185\textwidth}
        \includegraphics[trim={2cm 0 0 0},clip,width=1\linewidth]{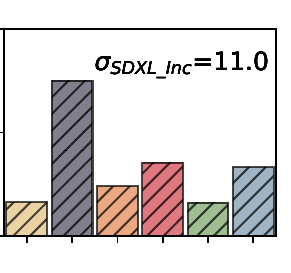}
    \end{subfigure}\hspace*{-0.2em}
    \begin{subfigure}{0.1185\textwidth}
        \includegraphics[trim={2cm 0 0 0},clip,width=1\linewidth]{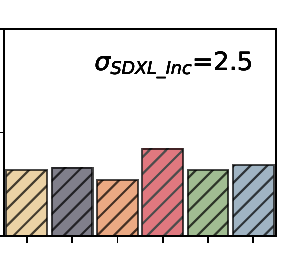}
    \end{subfigure}\hspace*{-0.2em}
    \begin{subfigure}{0.1185\textwidth}
        \includegraphics[trim={2cm 0 0 0},clip,width=1\linewidth]{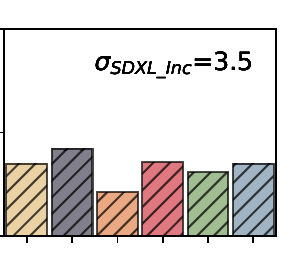}
    \end{subfigure}
    \begin{flushleft} \textbf{b} \end{flushleft}
    \begin{subfigure}{0.18\textwidth}
        \includegraphics[trim={0 0 0 0},clip,width=1\linewidth]{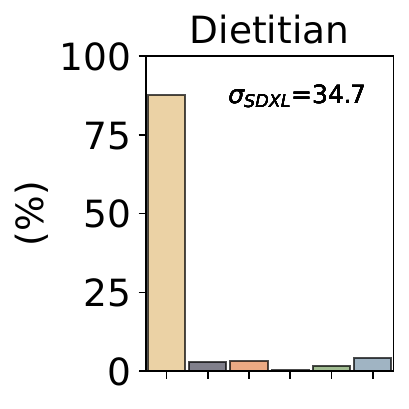}
    \end{subfigure}\hspace*{-0.4em}
    \begin{subfigure}{0.123\textwidth}
        \includegraphics[trim={2.2cm 0 0 0},clip,width=1\linewidth]{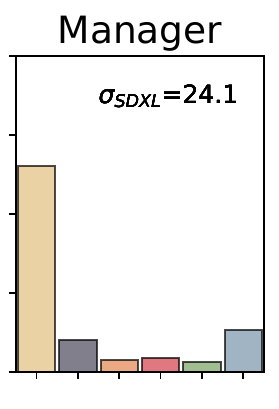}
    \end{subfigure}\hspace*{-0.4em}
    \begin{subfigure}{0.123\textwidth}
        \includegraphics[trim={2.2cm 0 0 0},clip,width=1\linewidth]{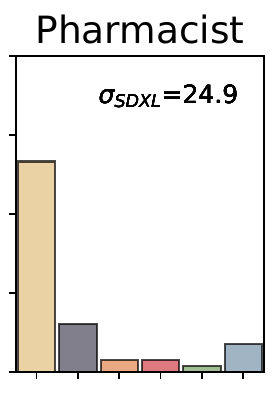}
    \end{subfigure}\hspace*{-0.4em}
    \begin{subfigure}{0.123\textwidth}
        \includegraphics[trim={2.2cm 0 0 0},clip,width=1\linewidth]{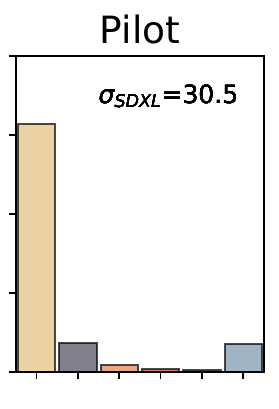}
    \end{subfigure}\hspace*{-0.4em}
    \begin{subfigure}{0.123\textwidth}
        \includegraphics[trim={2.2cm 0 0 0},clip,width=1\linewidth]{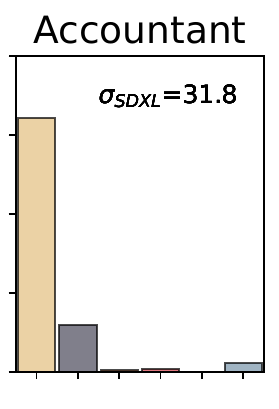}
    \end{subfigure}\hspace*{-0.4em}
    \begin{subfigure}{0.123\textwidth}
        \includegraphics[trim={2.2cm 0 0 0},clip,width=1\linewidth]{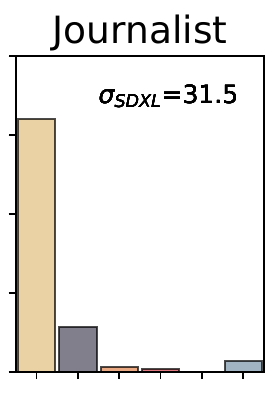}
    \end{subfigure}\hspace*{-0.4em}
    \begin{subfigure}{0.123\textwidth}
        \includegraphics[trim={2.2cm 0 0 0},clip,width=1\linewidth]{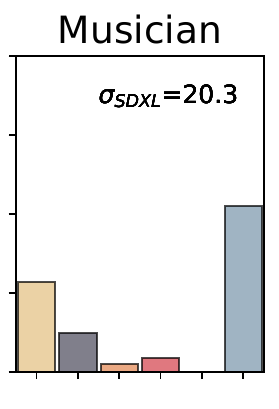}
    \end{subfigure}\hspace*{-0.4em}
    \begin{subfigure}{0.123\textwidth}
        \includegraphics[trim={2.2cm 0 0 0},clip,width=1\linewidth]{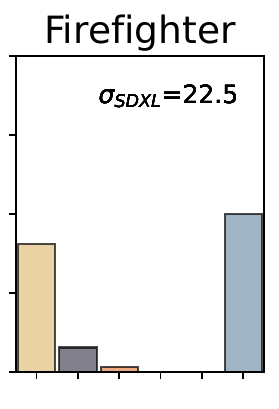}
    \end{subfigure}
    \\
    \hspace*{0.4em}
    \begin{subfigure}{0.165\textwidth}
        \includegraphics[trim={0 0 0 0},clip,width=1\linewidth]{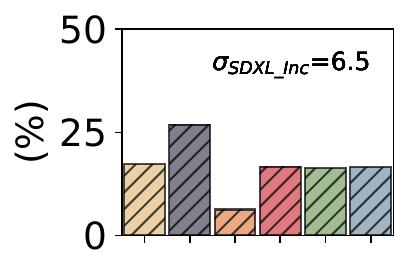}
    \end{subfigure}\hspace*{-0.2em}
    \begin{subfigure}{0.1185\textwidth}
        \includegraphics[trim={2cm 0 0 0},clip,width=1\linewidth]{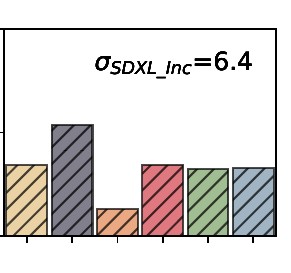}
    \end{subfigure}\hspace*{-0.2em}
    \begin{subfigure}{0.1185\textwidth}
        \includegraphics[trim={2cm 0 0 0},clip,width=1\linewidth]{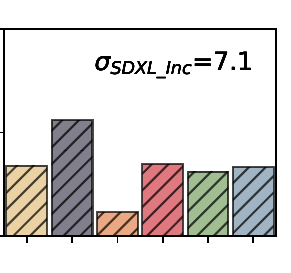}
    \end{subfigure}\hspace*{-0.2em}
    \begin{subfigure}{0.1185\textwidth}
        \includegraphics[trim={2cm 0 0 0},clip,width=1\linewidth]{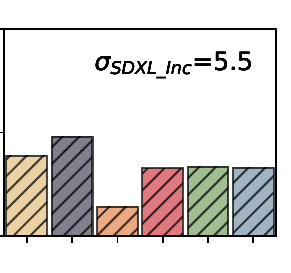}
    \end{subfigure}\hspace*{-0.2em}
    \begin{subfigure}{0.1185\textwidth}
        \includegraphics[trim={2cm 0 0 0},clip,width=1\linewidth]{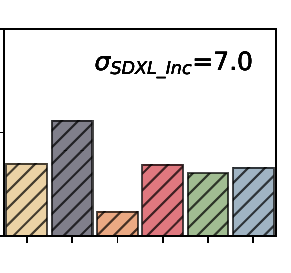}
    \end{subfigure}\hspace*{-0.2em}
    \begin{subfigure}{0.1185\textwidth}
        \includegraphics[trim={2cm 0 0 0},clip,width=1\linewidth]{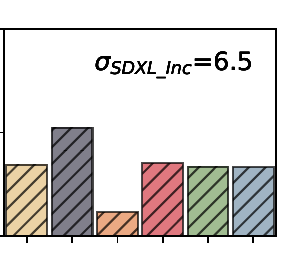}
    \end{subfigure}\hspace*{-0.2em}
    \begin{subfigure}{0.1185\textwidth}
        \includegraphics[trim={2cm 0 0 0},clip,width=1\linewidth]{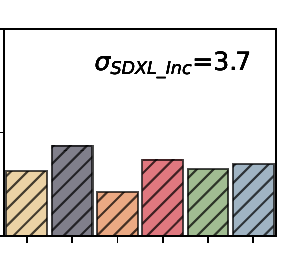}
    \end{subfigure}\hspace*{-0.2em}
    \begin{subfigure}{0.1185\textwidth}
        \includegraphics[trim={2cm 0 0 0},clip,width=1\linewidth]{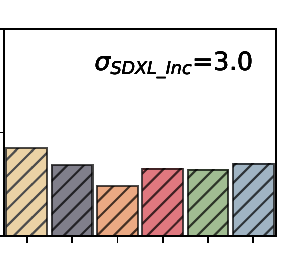}
    \end{subfigure}
    \\
    \begin{flushleft} \textbf{c} \end{flushleft}
    \vspace{-10pt}
    \includegraphics[trim={15cm 7.5cm 0.5cm 0.5cm},clip,width=0.5\linewidth]{figures/gender_comparison_lengend.pdf}
    \includegraphics[width=1\linewidth]{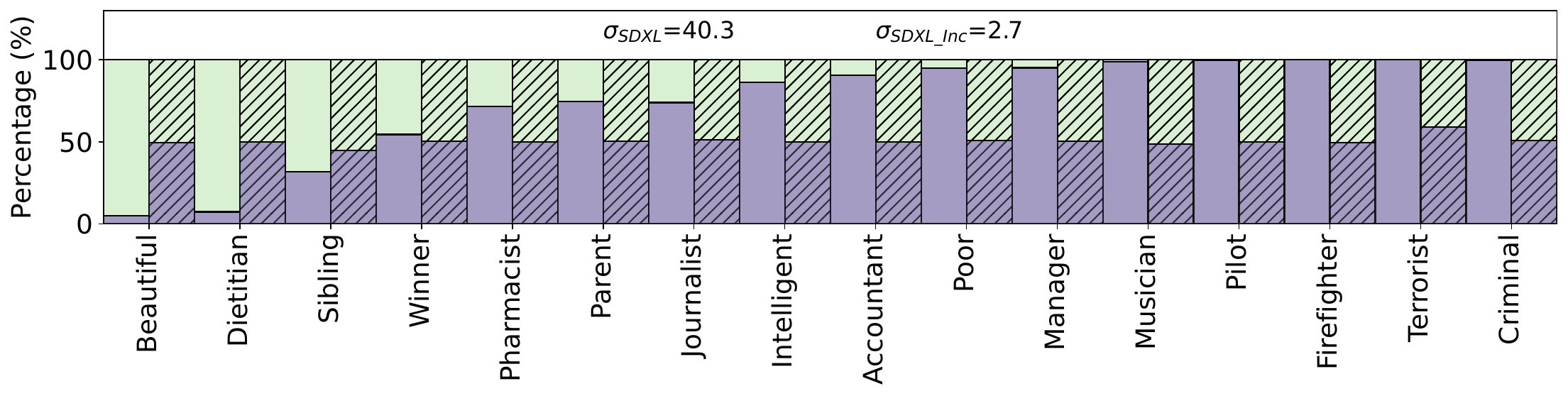}
\caption{
\textbf{Results of SDXL-Inc.} Given eight attributes and eight professions, both SDXL and SDXL-Inc were used to generate 10,000 images per profession and per attribute. \textbf{a}, Race distribution per attribute, with the upper row corresponding to SDXL, and the lower row corresponding to SDXL-Inc. \textbf{b}, The same as (\textbf{a}) but for professions instead of attributes. \textbf{c}, Gender distribution per profession and per attribute for SDXL and SDXL-Inc. The standard deviation(s) corresponding to each subplot is denoted by $\sigma$ followed by a subscript indicating the model.
}
\label{fig:professions_after}
\end{figure}

Having established that SDXL exhibits biases in terms of attributes, we now evaluate SDXL-Inc's ability to address these biases. To this end, we repeated the same procedure used earlier with SDXL, but using our SDXL-Inc model instead. The results for professions are depicted in the bottom row of Figure~\ref{fig:professions_after}a. As shown, races are represented more uniformly compared to SDXL, as evidenced by the substantial reduction in standard deviation ($\sigma$). Importantly, none of the eight attributes considered in this analysis was used during the fine-tuning phase, and yet SDXL-Inc was able to significantly reduce the racial biases related to them. This indicates that our solution can indeed be generalized beyond the features it was fine-tuned on.

Let us now reexamine professional stereotypes to further evaluate SDXL-Inc's debiasing capability. To this end, we selected four of the professions that were used during the fine-tuning phase (Dietitian, Manager, Pharmacist, and Pilot), as well as four professions that were not used during that phase (Accountant, Journalist, Musician, and Firefighter) to further assess the generalizability of SDXL-Inc. For each profession, Figure~\ref{fig:professions_after}b shows the racial distribution of images using SDXL (upper row) and SDXL-Inc (lower row). As can be seen, regardless of whether the profession is Black-dominated (Musician and Firefighter) or White-dominated, our solution is able to significantly reduce difference between races. Again, the reduction in standard deviation is substantial, reaching multiple folds across all professions. Similar trends are observed when examining the remaining 24 professions considered in our analysis; see Supplementary Figure~3a.

Finally, we evaluate SDXL-Inc's ability to address gender biases across the aforementioned attributes and professions. As can be seen in Figure~\ref{fig:professions_after}c, SDXL (solid bars) exhibits substantial biases, e.g., depicting the vast majority of beautiful individuals as female while depicting the vast majority of intelligent individuals as male. In contrast, our solution consistently produces an equal, or near-equal, split between female and male (dashed bars), regardless of the attribute or profession. These performance improvements are reflected by the vast reduction in standard deviation, from 40.3 to just 2.7. In fact, for the remaining 24 professions, the standard deviation corresponding to our solution drops even further, reaching 1.0; see Supplementary Figure 3b.

In addition to SDXL-Inc, we experimented with an alternative debiasing solution that follows a fundamentally different approach, which is to use a Large Language Model, namely GPT-4~\cite{chatgpt}, ``in-the-loop''; see Materials and Methods for technical details. The results of this solution are depicted in Supplementary Figure~4, and the corresponding numerical values are listed in Supplementary Table~6. As can be seen, this solution is also capable of drastically reducing the race and gender biases exhibited by SDXL.

\begin{figure}[htbp]   
\captionsetup[subfigure]{labelformat=empty}
    \centering    
    \begin{flushleft} \textbf{a} \end{flushleft}
    \begin{subfigure}{0.095\textwidth}
        \includegraphics[width=1\linewidth]{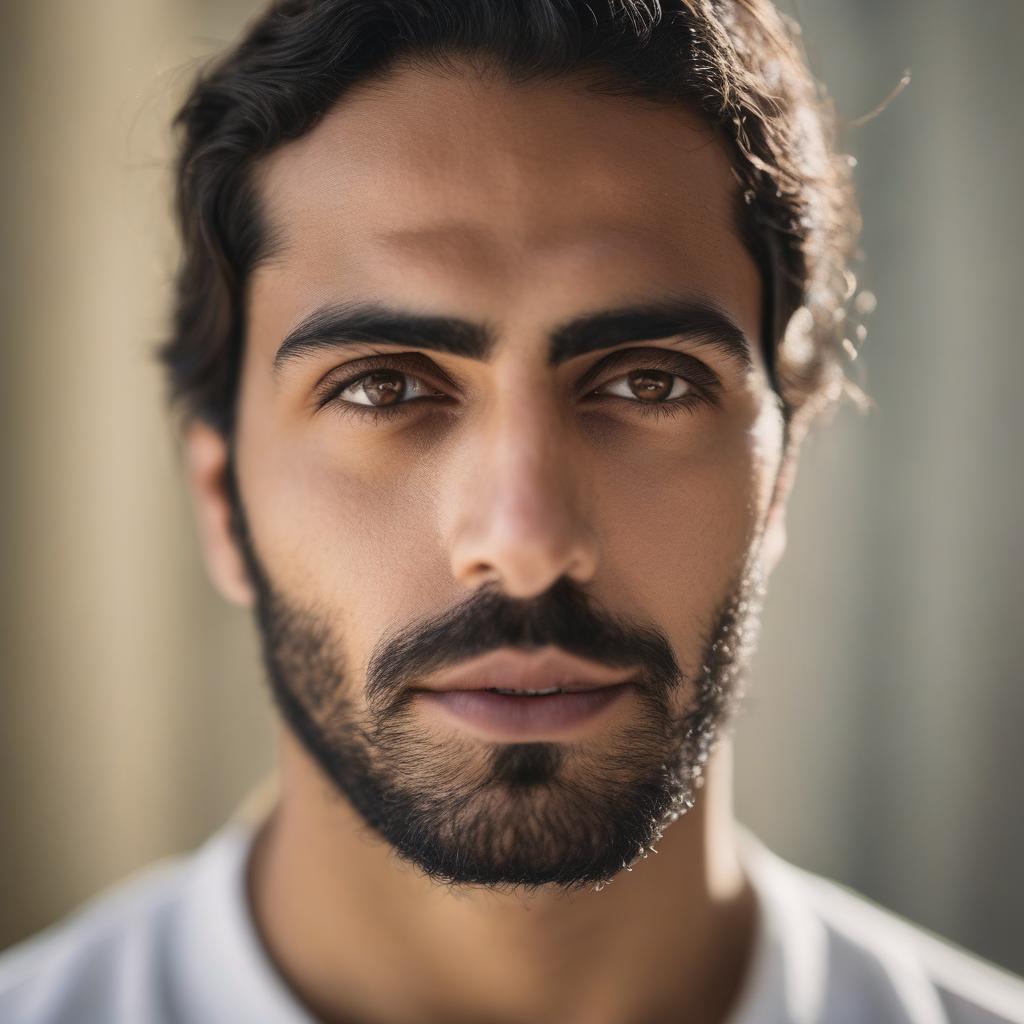}
    \end{subfigure}\hspace*{-0.25em}
    \begin{subfigure}{0.095\textwidth}
        \includegraphics[width=1\linewidth]{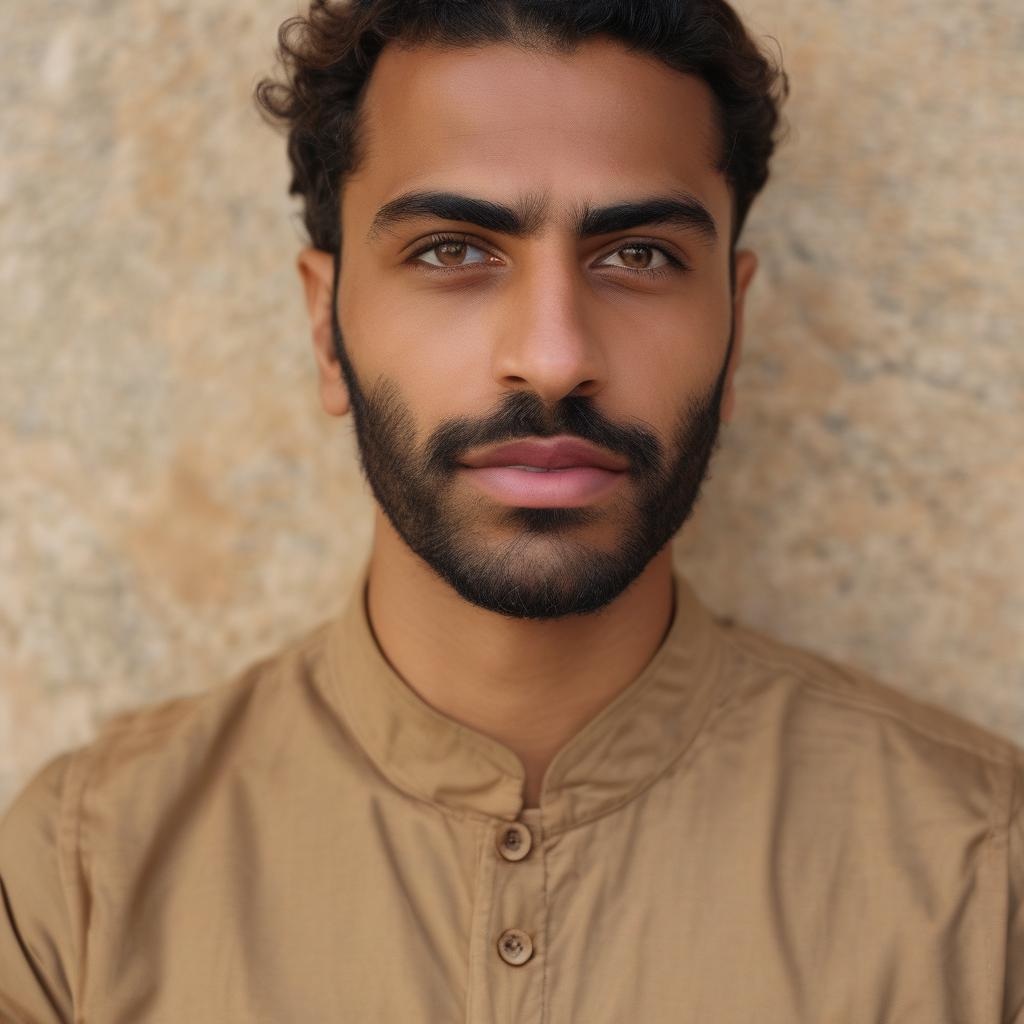}
    \end{subfigure}\hspace*{-0.25em}
    \begin{subfigure}{0.095\textwidth}
        \includegraphics[width=1\linewidth]{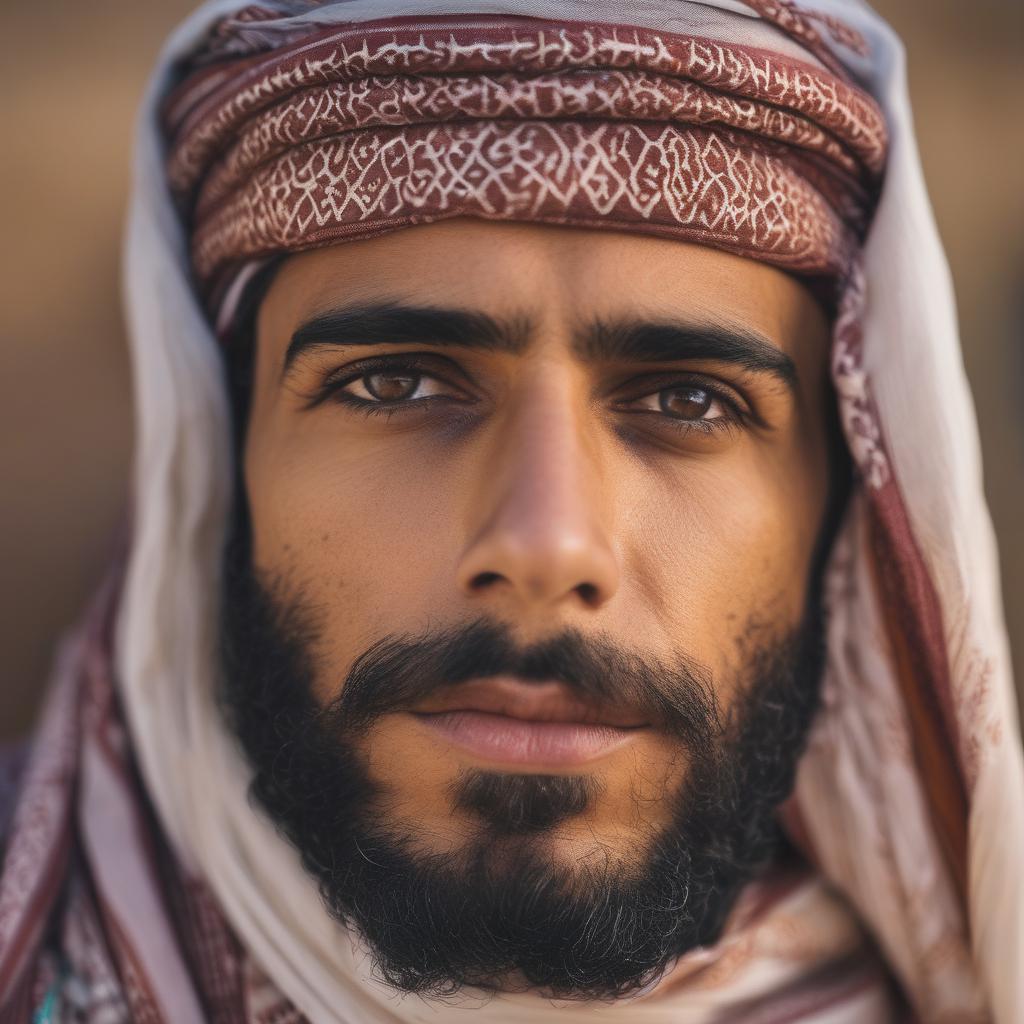}
    \end{subfigure}\hspace*{-0.25em}
    \begin{subfigure}{0.095\textwidth}
        \includegraphics[width=1\linewidth]{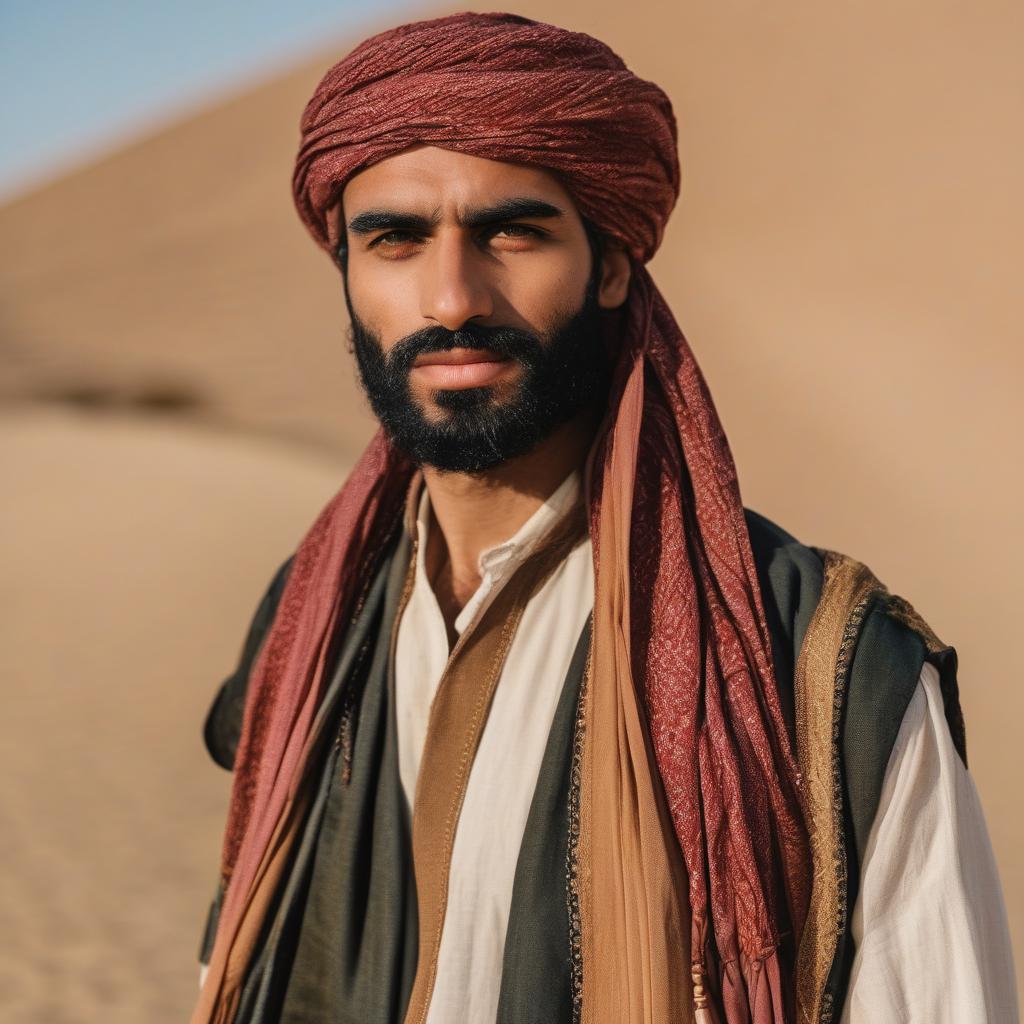}
    \end{subfigure}\hspace*{-0.25em}
    \begin{subfigure}{0.095\textwidth}
        \includegraphics[width=1\linewidth]{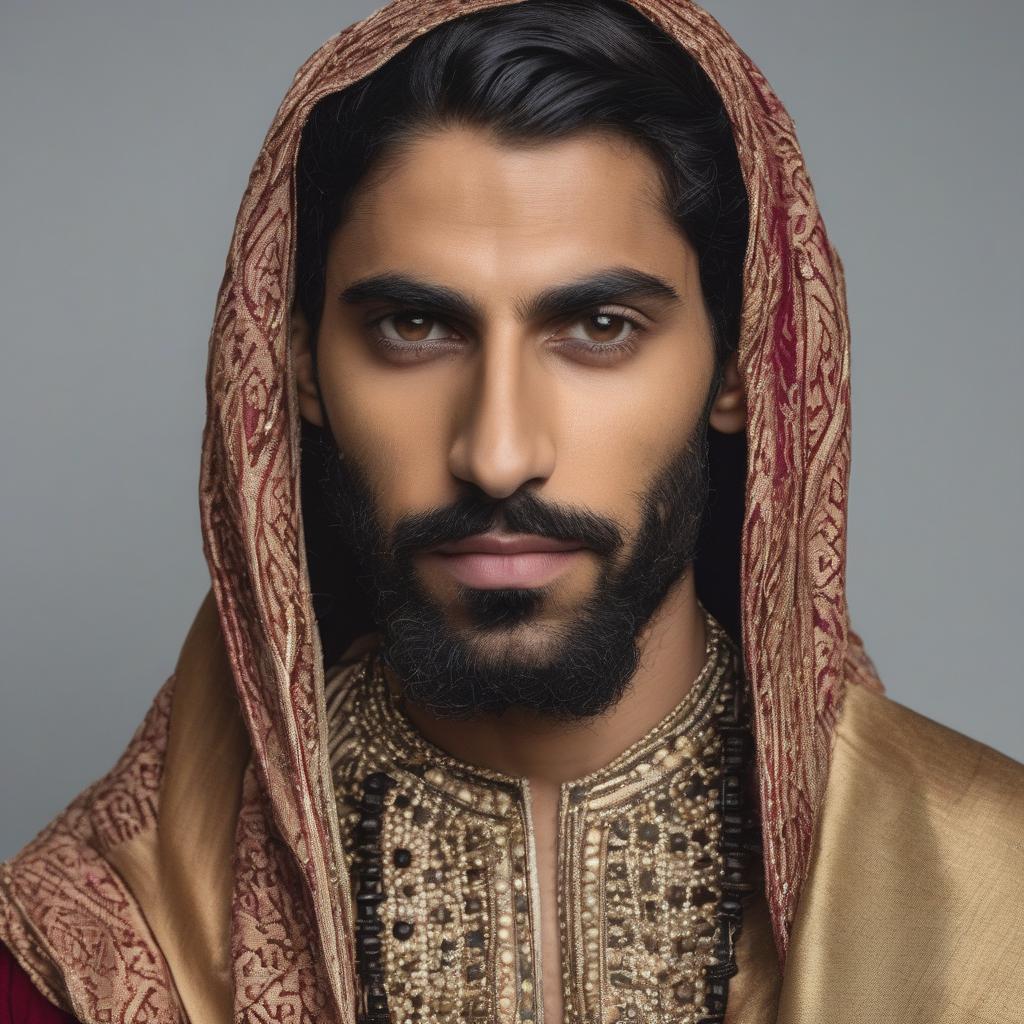}
    \end{subfigure}\hspace*{-0.25em}
    \begin{subfigure}{0.095\textwidth}
        \includegraphics[width=1\linewidth]{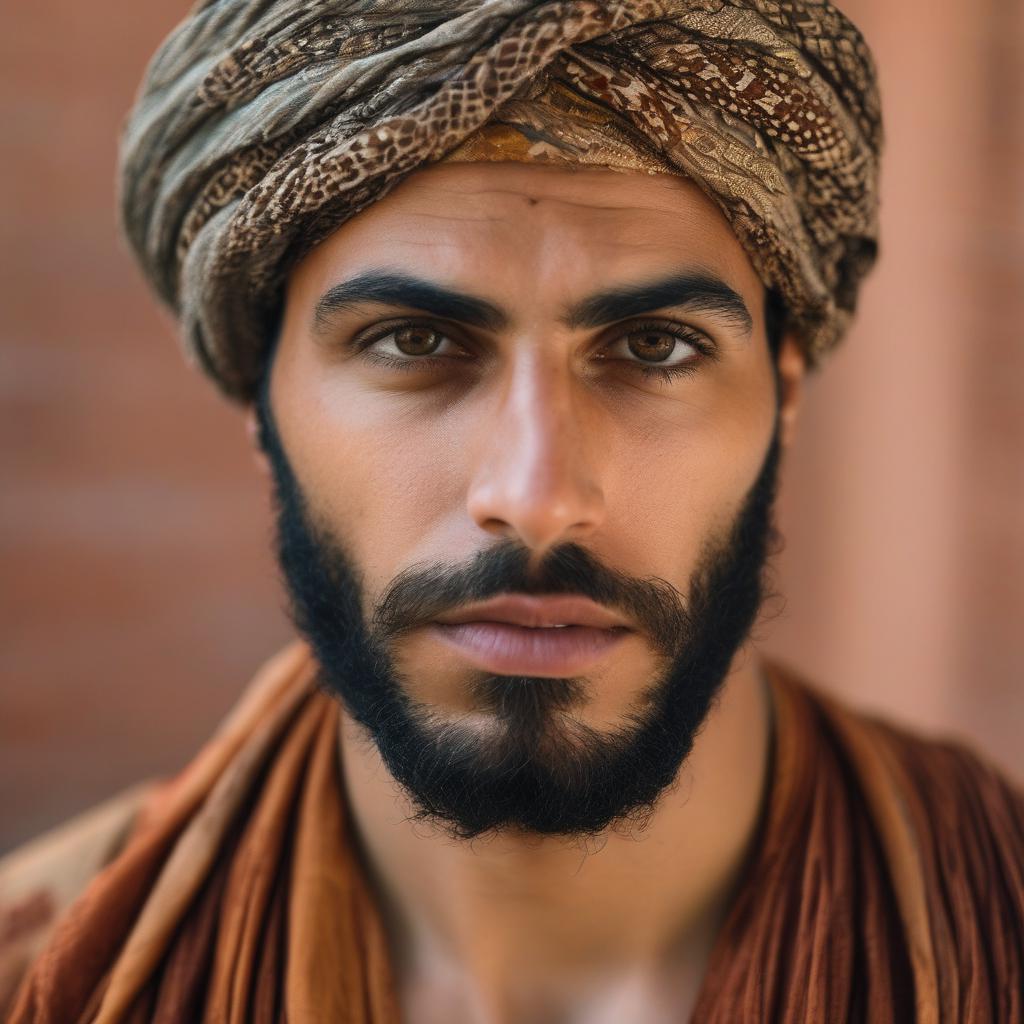}
    \end{subfigure}\hspace*{-0.25em}
    \begin{subfigure}{0.095\textwidth}
        \includegraphics[width=1\linewidth]{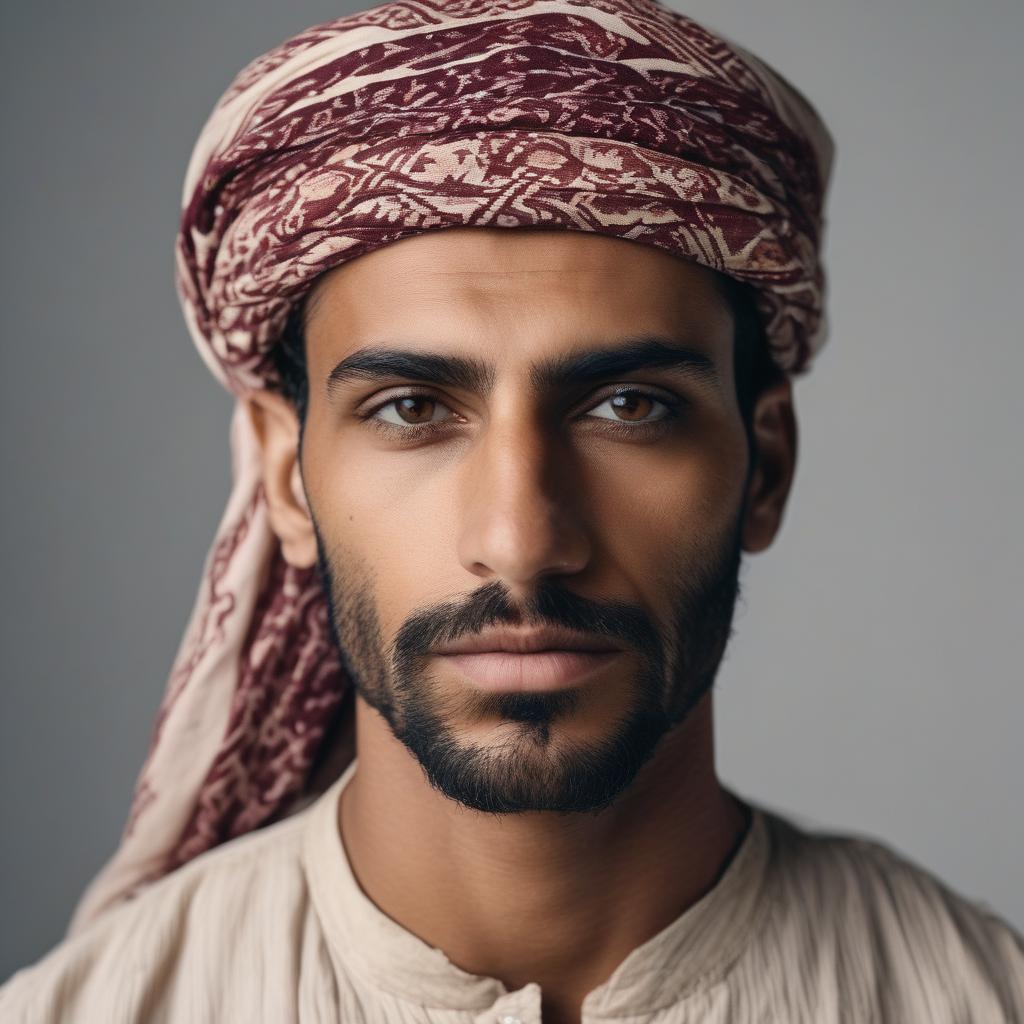}
    \end{subfigure}\hspace*{-0.25em}
    \begin{subfigure}{0.095\textwidth}
        \includegraphics[width=1\linewidth]{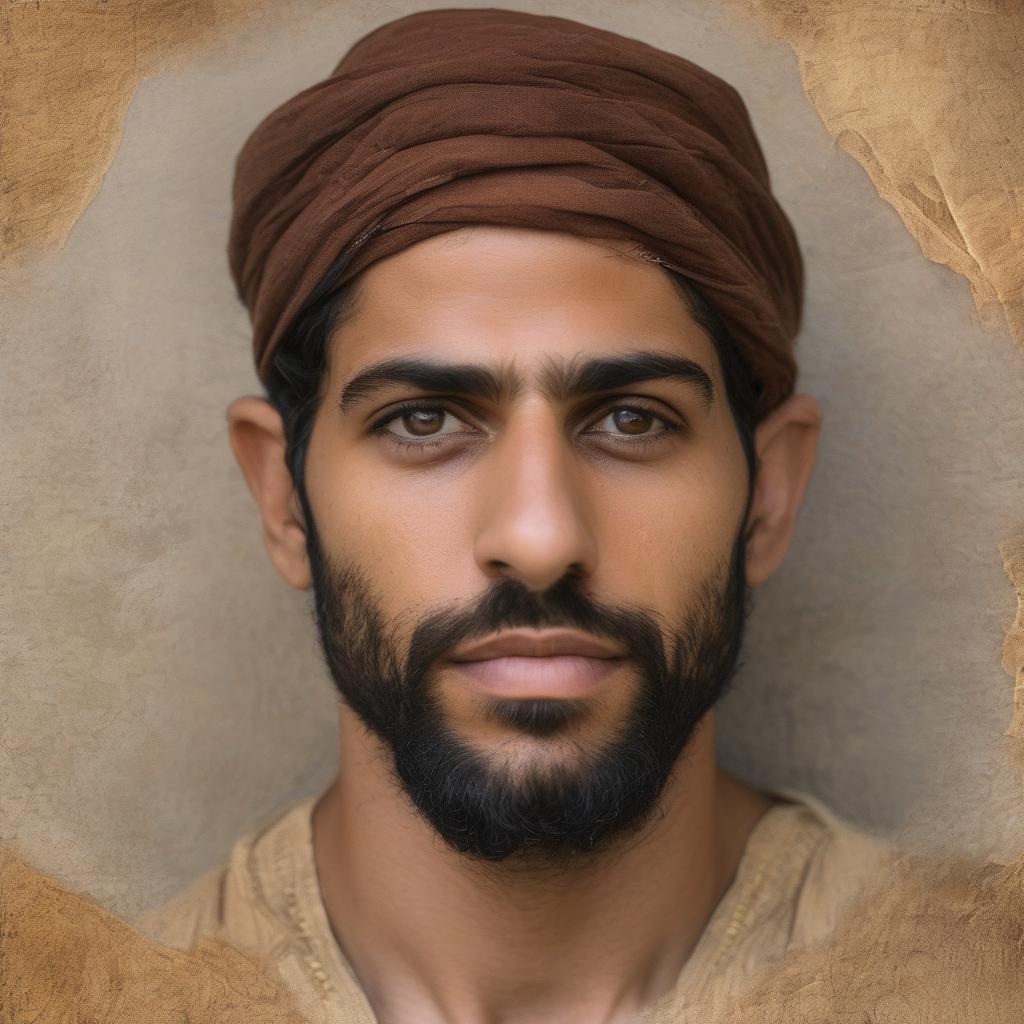}
    \end{subfigure}\hspace*{-0.25em}
    \begin{subfigure}{0.095\textwidth}
        \includegraphics[width=1\linewidth]{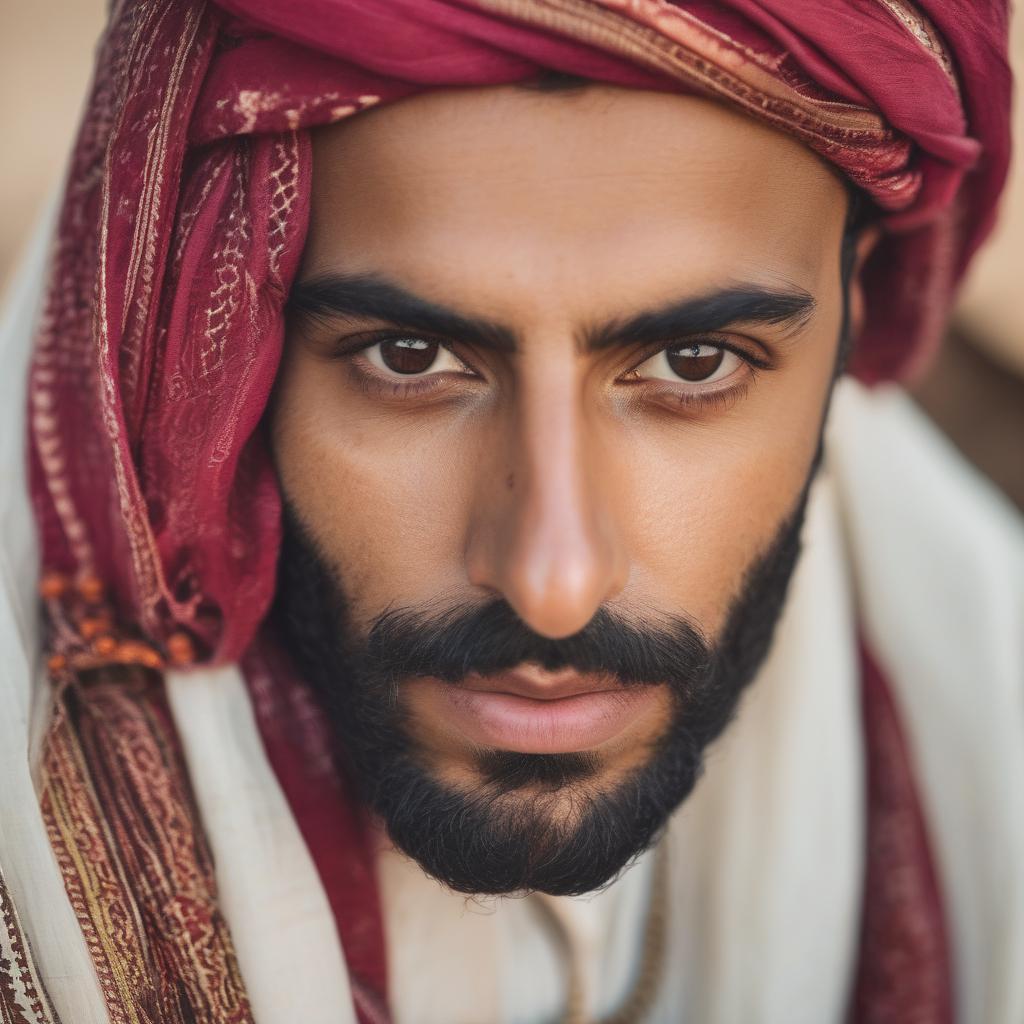}
    \end{subfigure}\hspace*{-0.25em}
    \begin{subfigure}{0.095\textwidth}
        \includegraphics[width=1\linewidth]{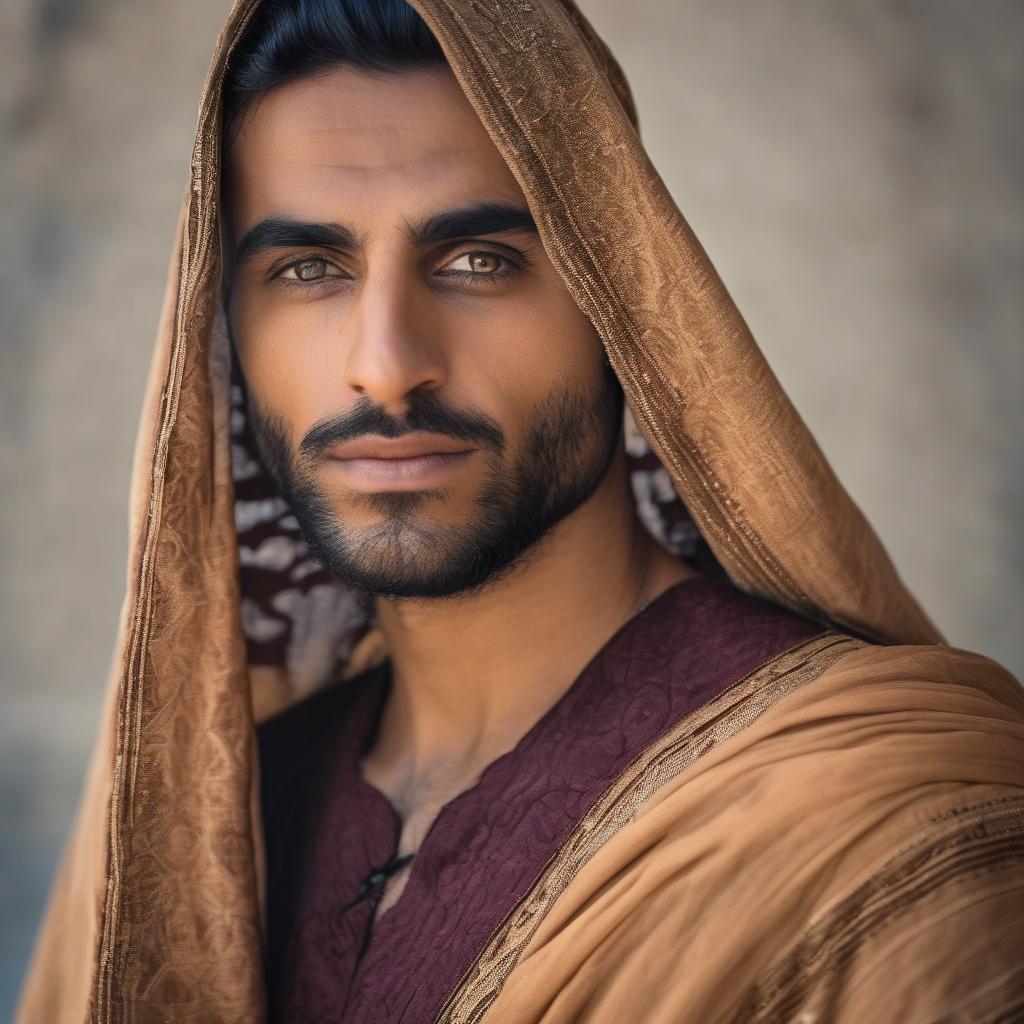}
    \end{subfigure}\\
    \begin{subfigure}{0.095\textwidth}
        \includegraphics[width=1\linewidth]{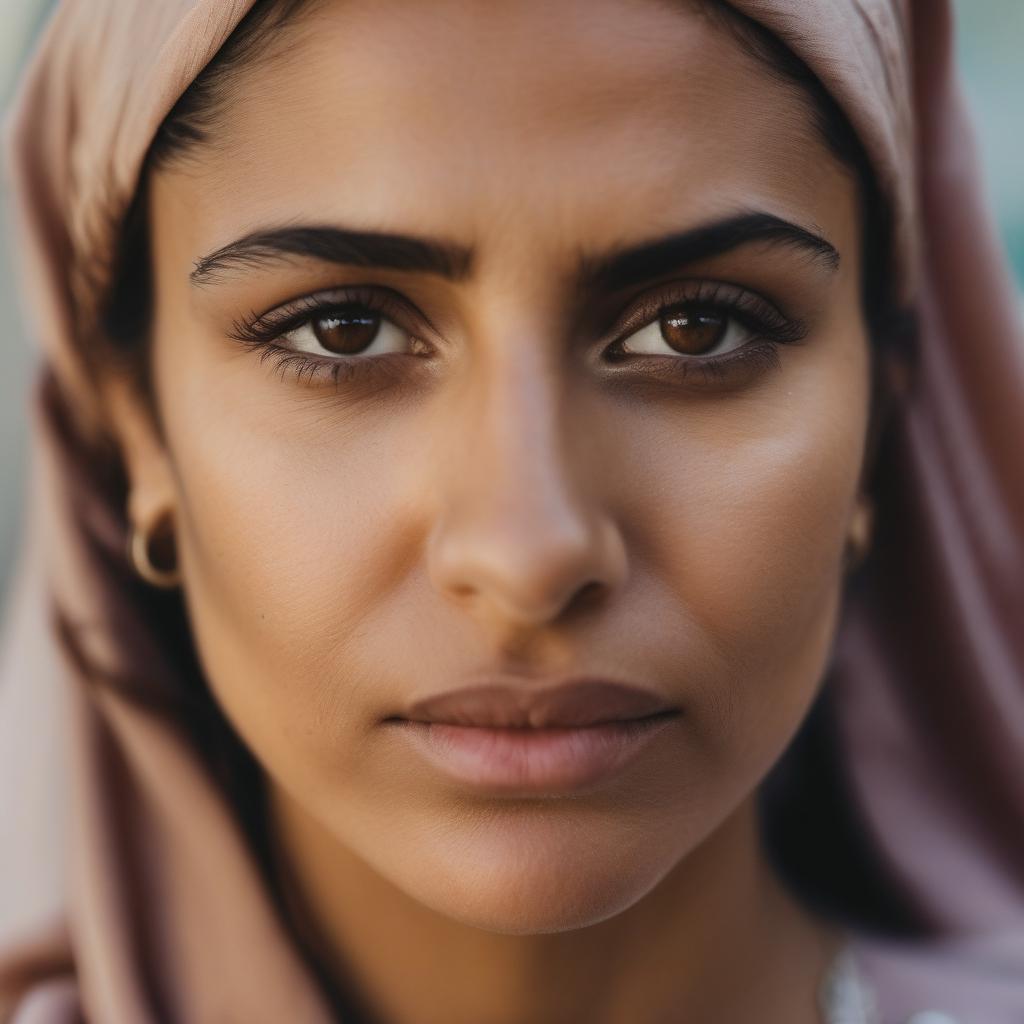}
    \end{subfigure}\hspace*{-0.25em}
    \begin{subfigure}{0.095\textwidth}
        \includegraphics[width=1\linewidth]{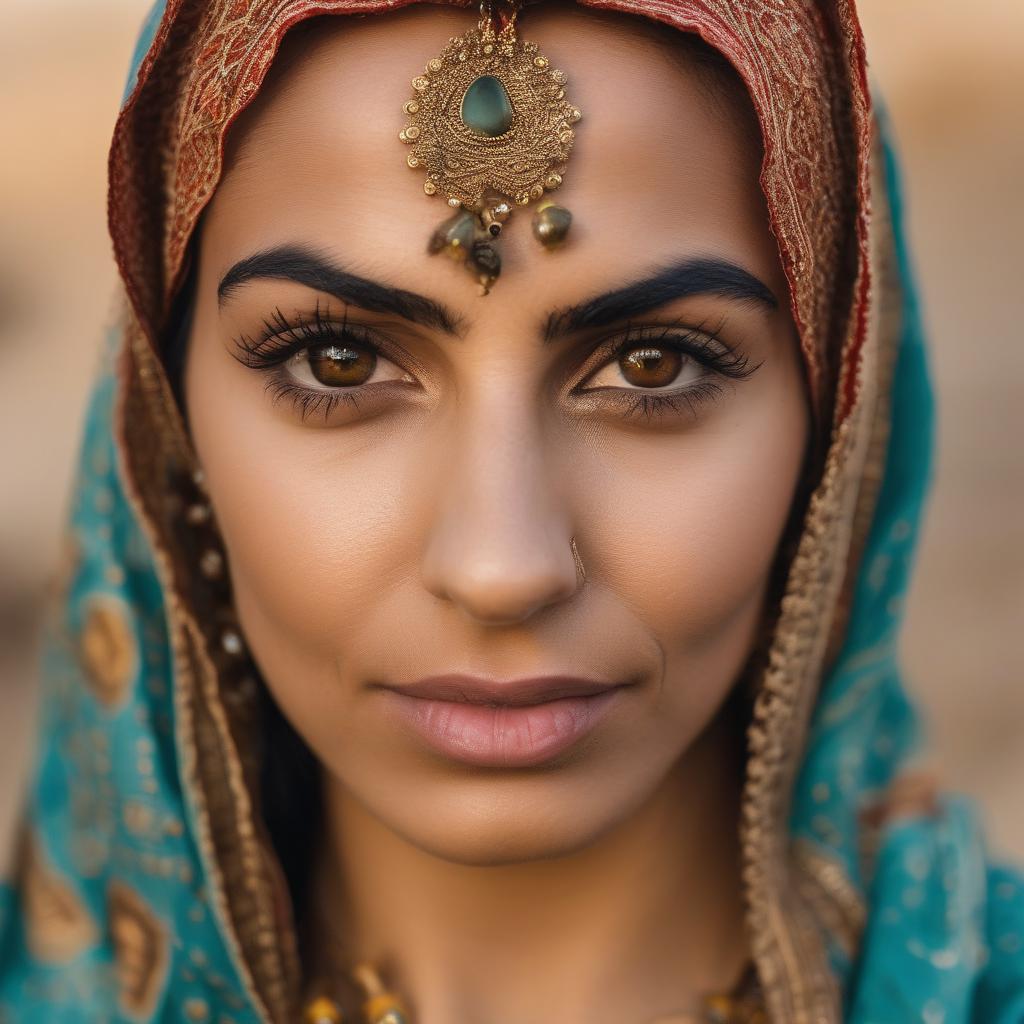}
    \end{subfigure}\hspace*{-0.25em}
    \begin{subfigure}{0.095\textwidth}
        \includegraphics[width=1\linewidth]{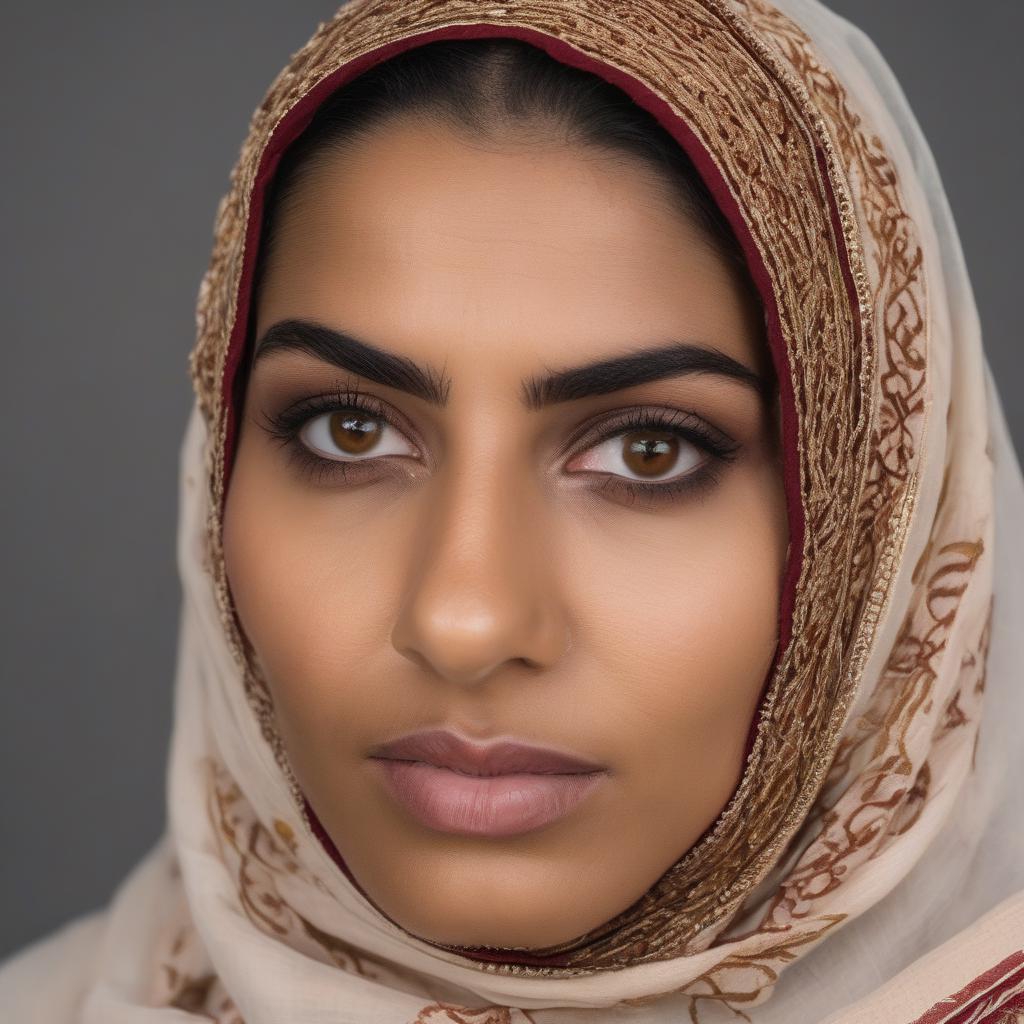}
    \end{subfigure}\hspace*{-0.25em}
    \begin{subfigure}{0.095\textwidth}
        \includegraphics[width=1\linewidth]{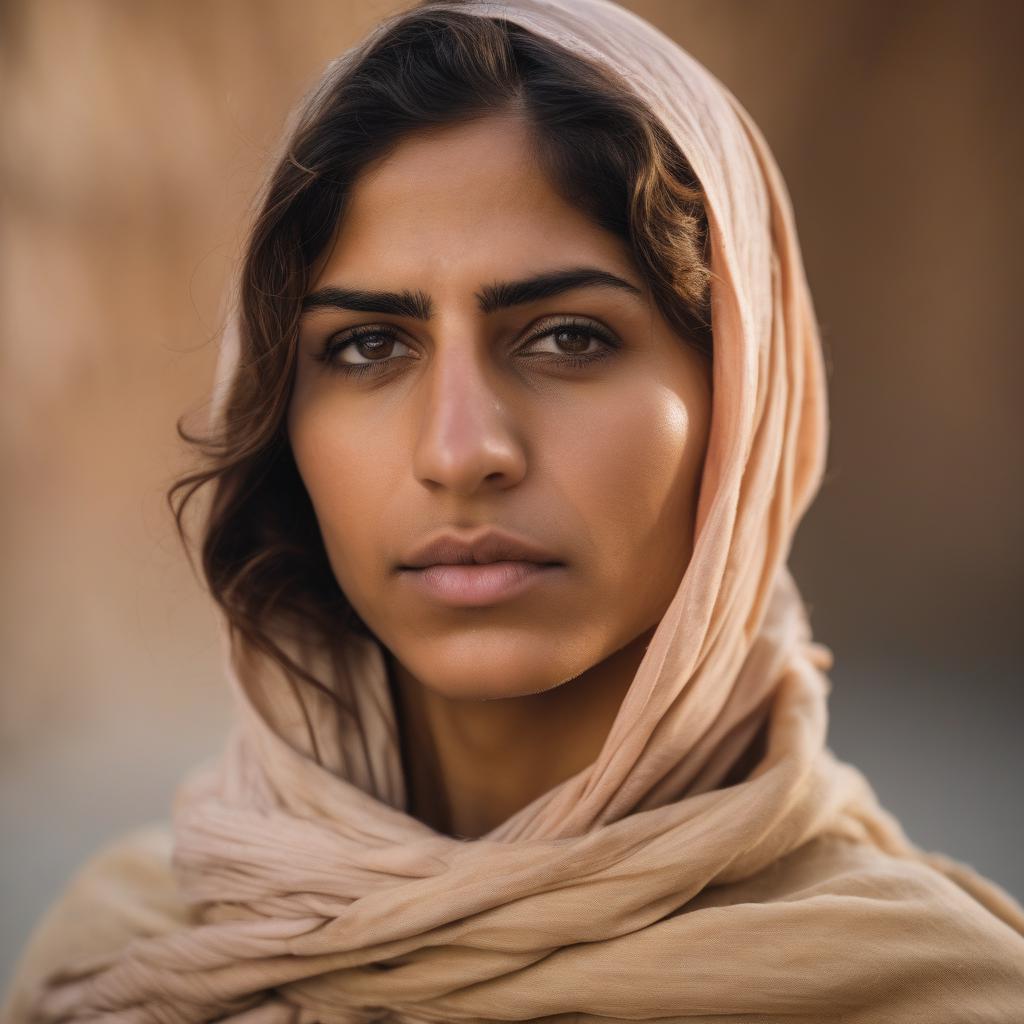}
    \end{subfigure}\hspace*{-0.25em}
    \begin{subfigure}{0.095\textwidth}
        \includegraphics[width=1\linewidth]{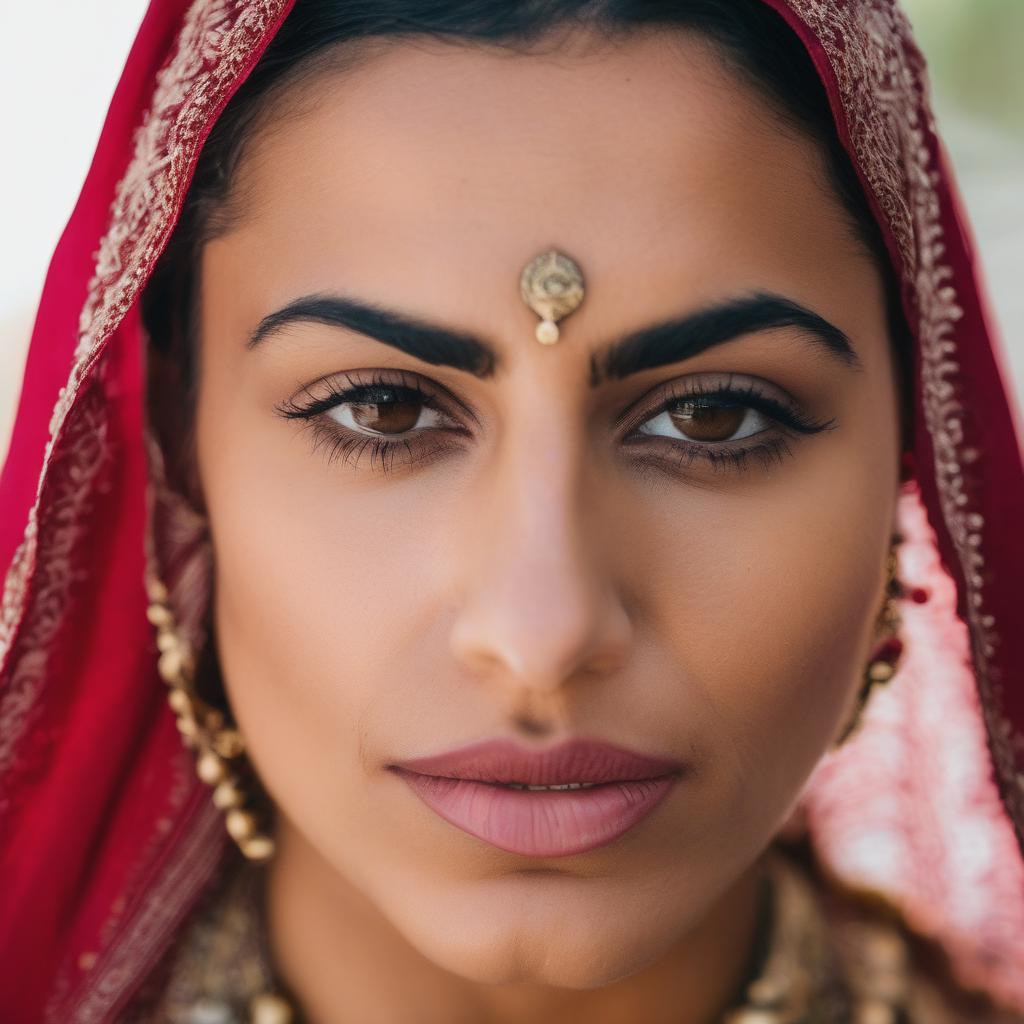}
    \end{subfigure}\hspace*{-0.25em}
    \begin{subfigure}{0.095\textwidth}
        \includegraphics[width=1\linewidth]{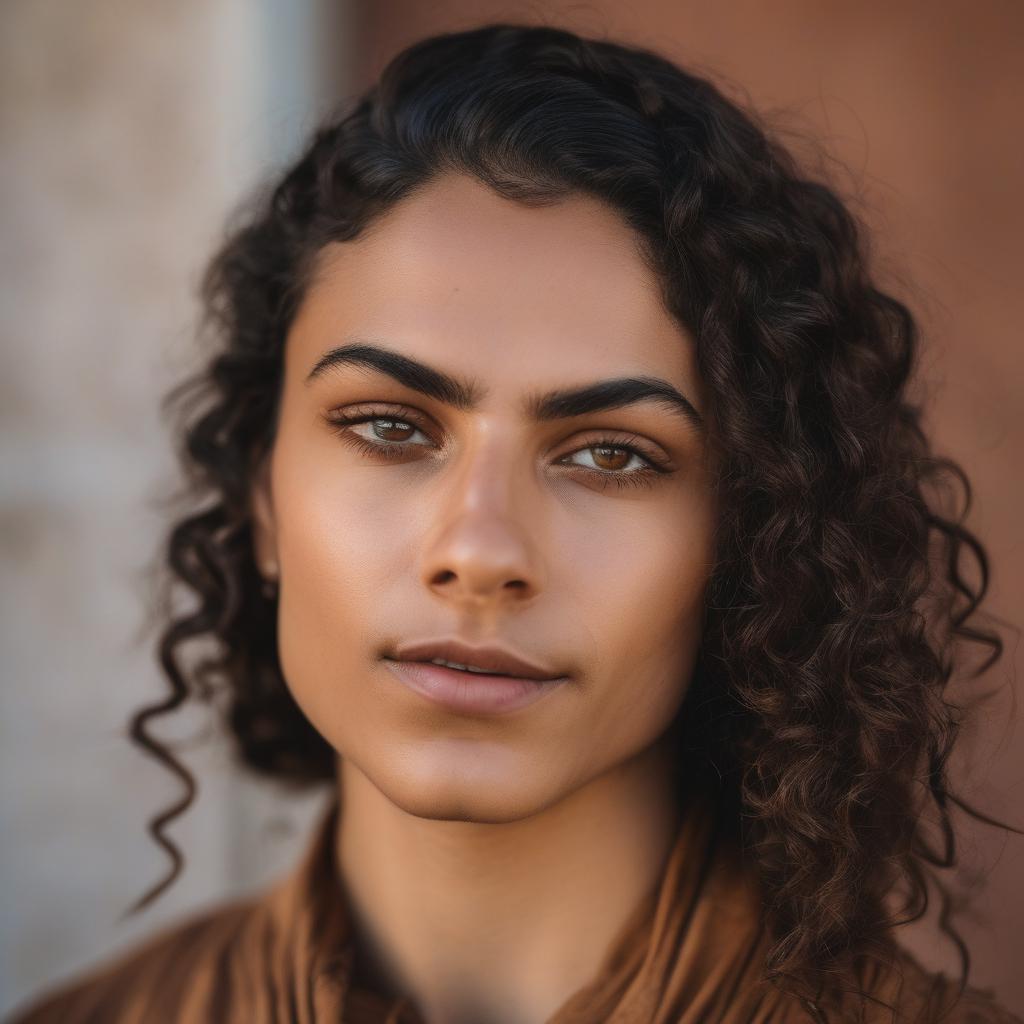}
    \end{subfigure}\hspace*{-0.25em}
    \begin{subfigure}{0.095\textwidth}
        \includegraphics[width=1\linewidth]{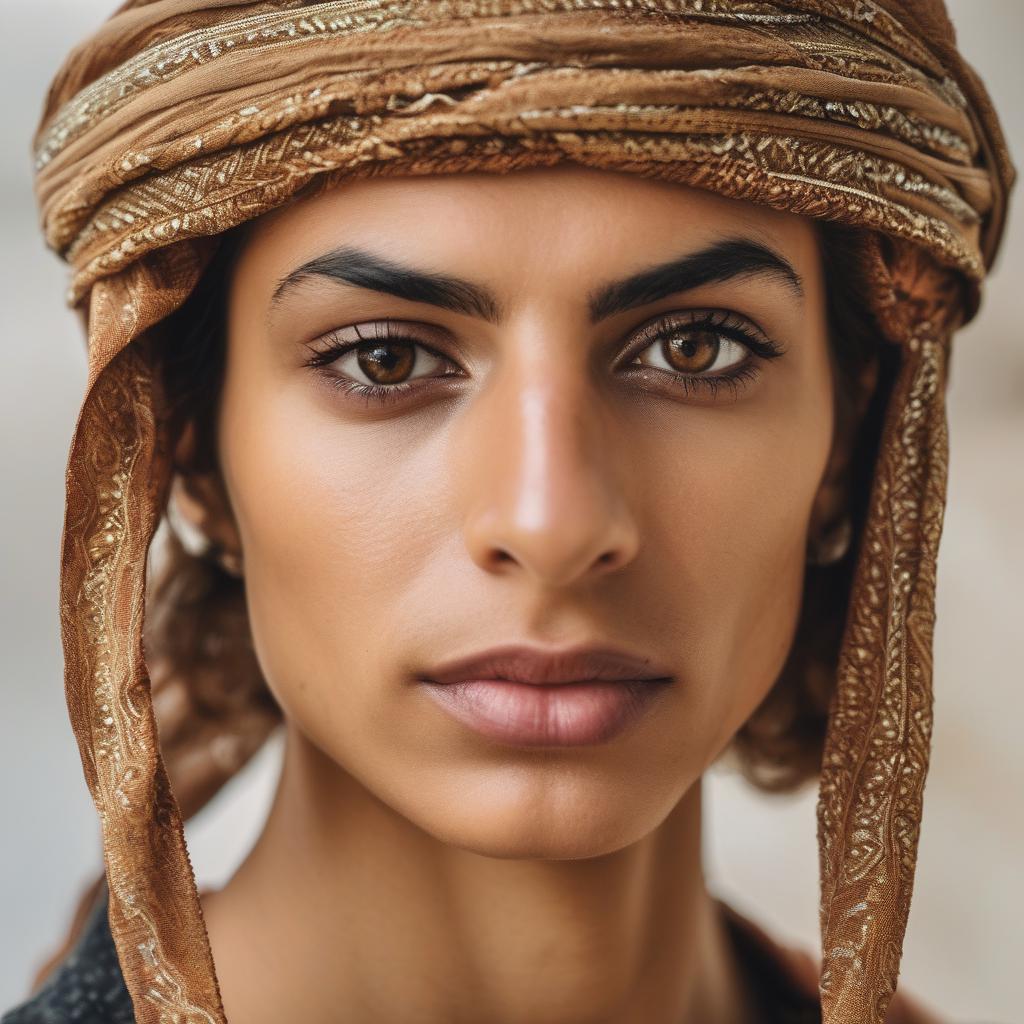}
    \end{subfigure}\hspace*{-0.25em}
    \begin{subfigure}{0.095\textwidth}
        \includegraphics[width=1\linewidth]{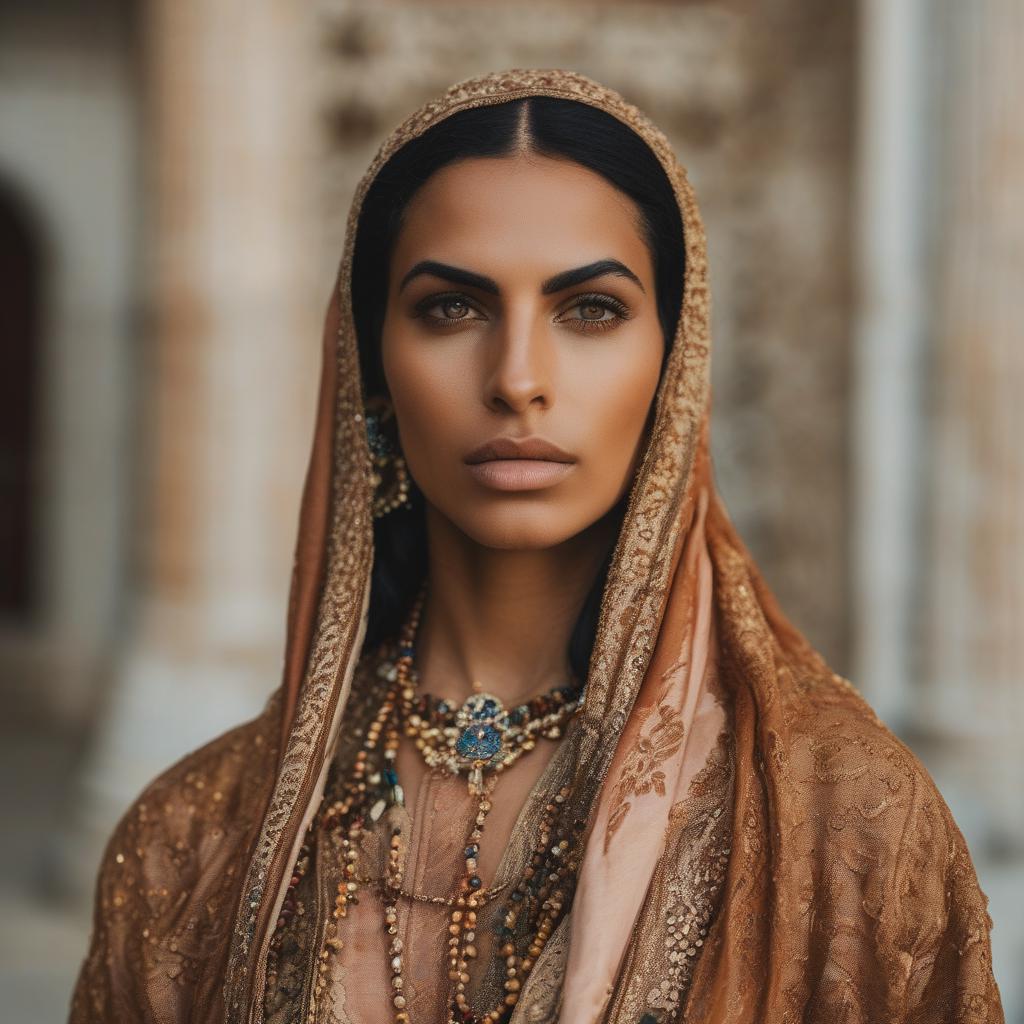}
    \end{subfigure}\hspace*{-0.25em}
    \begin{subfigure}{0.095\textwidth}
        \includegraphics[width=1\linewidth]{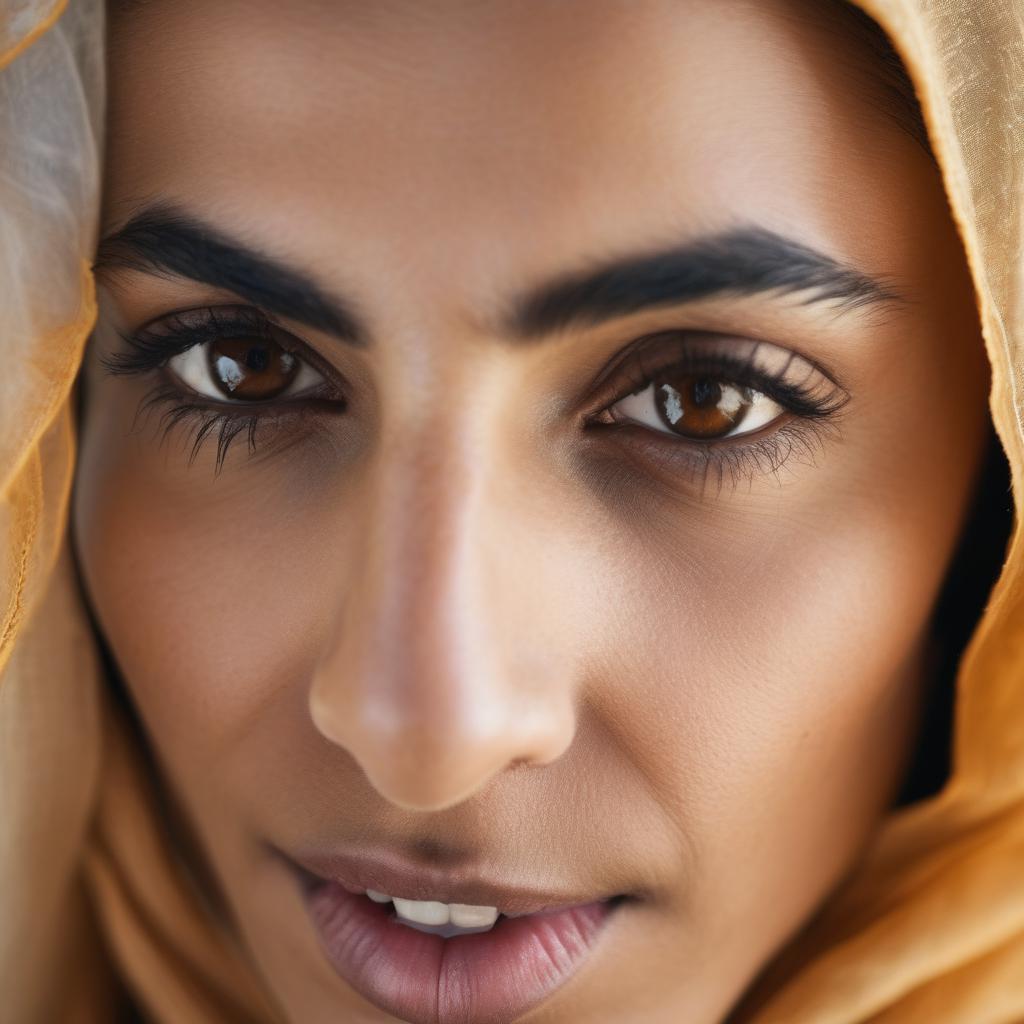}
    \end{subfigure}\hspace*{-0.25em}
    \begin{subfigure}{0.095\textwidth}
        \includegraphics[width=1\linewidth]{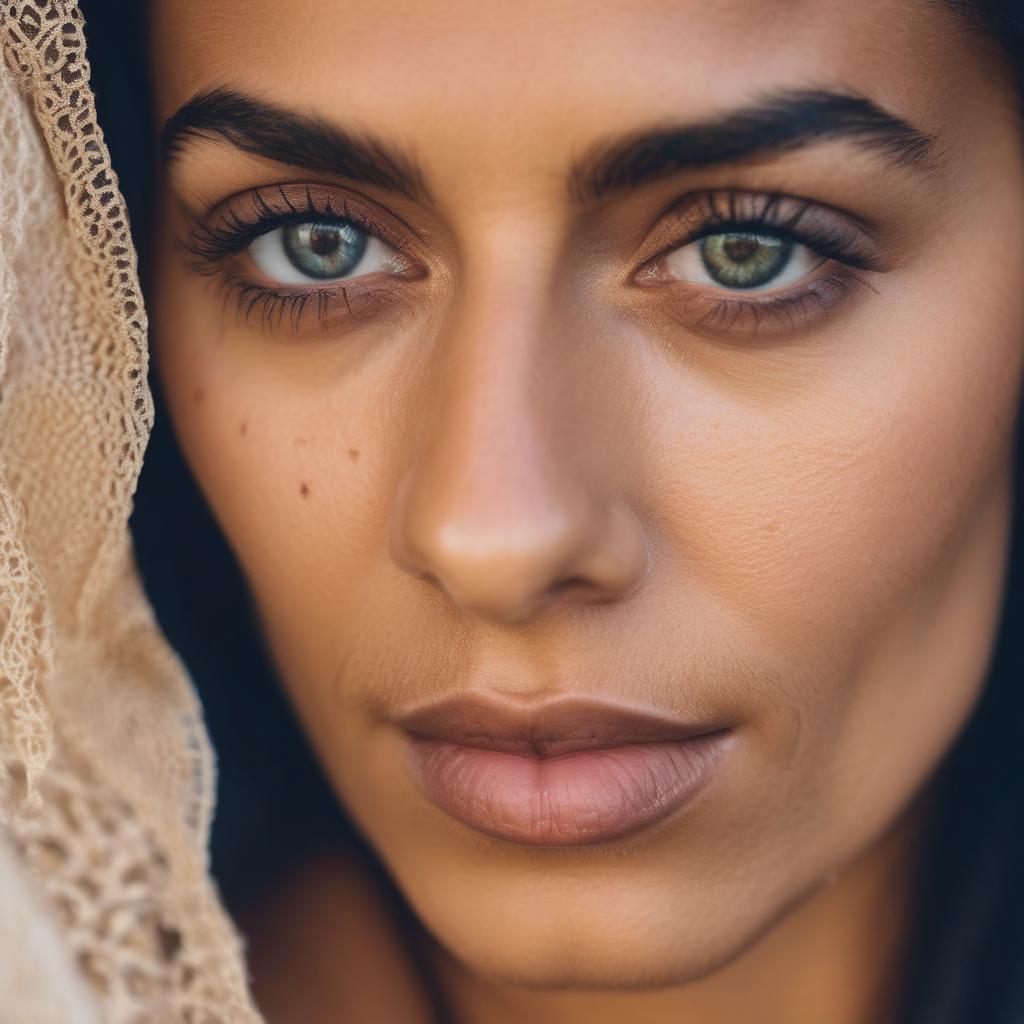}
    \end{subfigure}
    \begin{minipage}[b]{0.24\textwidth} 
        \includegraphics[width=\textwidth]{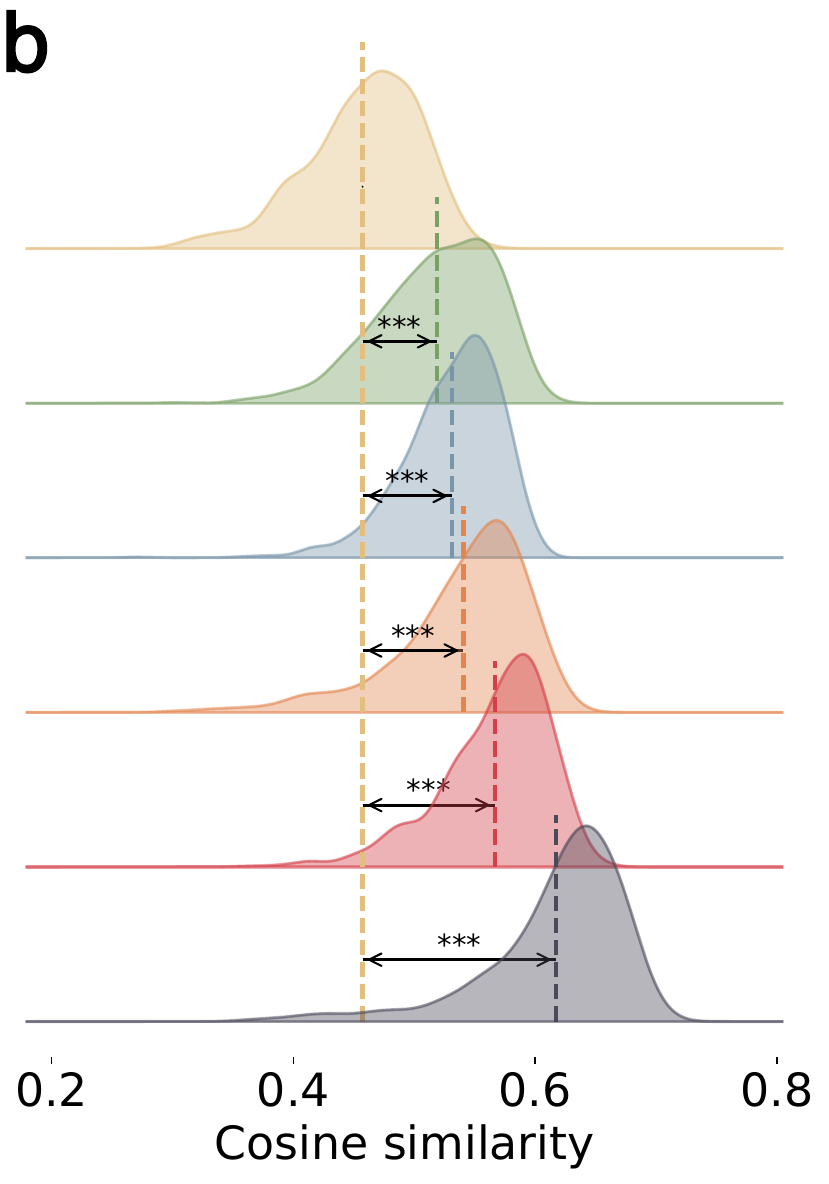} 
    \end{minipage}%
    \begin{minipage}[b]{0.75\textwidth} 
        \centering
        \begin{minipage}[b]{0.33\textwidth}
            \includegraphics[width=\textwidth]{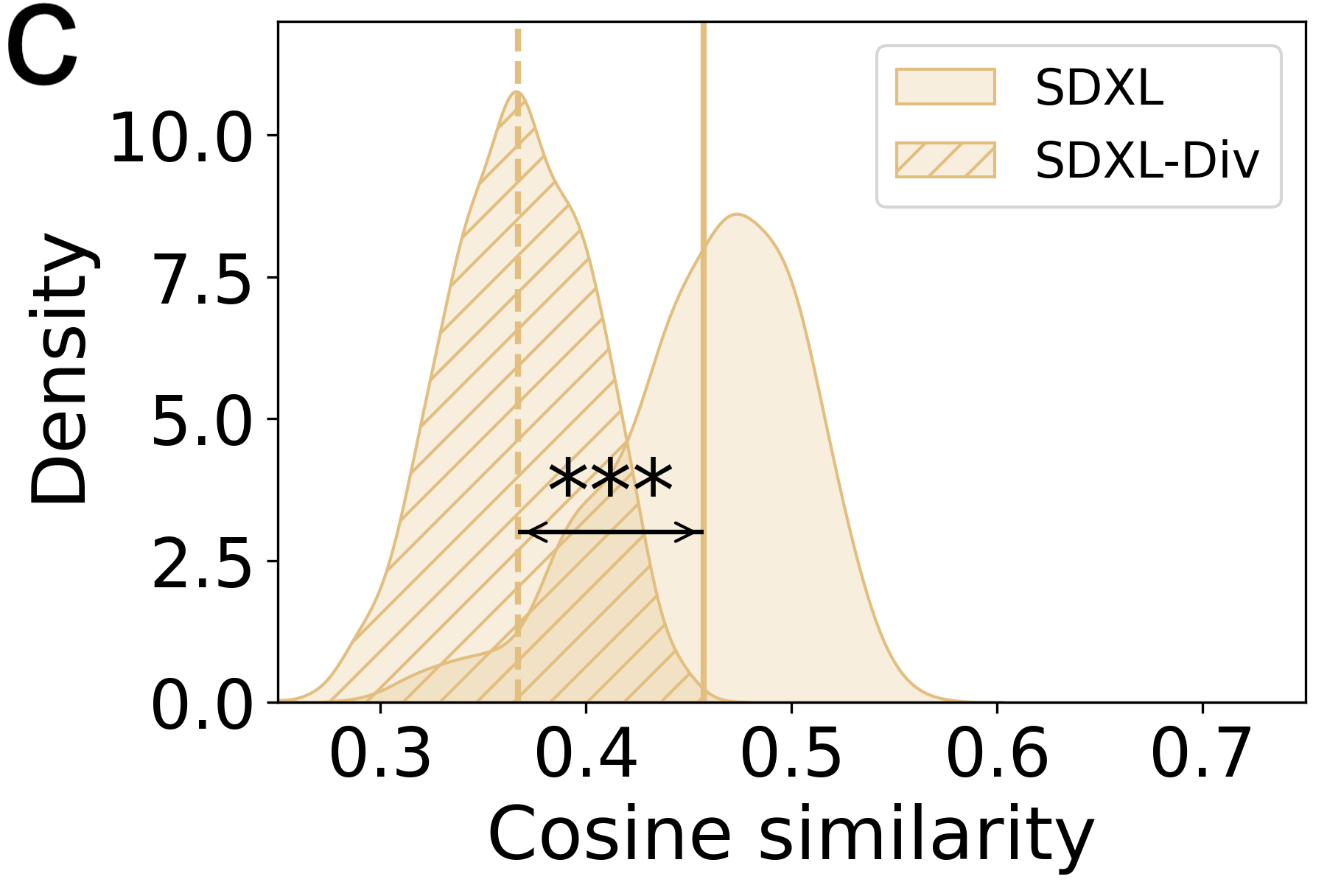} 
        \end{minipage}%
        \hfill
        \begin{minipage}[b]{0.33\textwidth}
            \includegraphics[width=\textwidth]{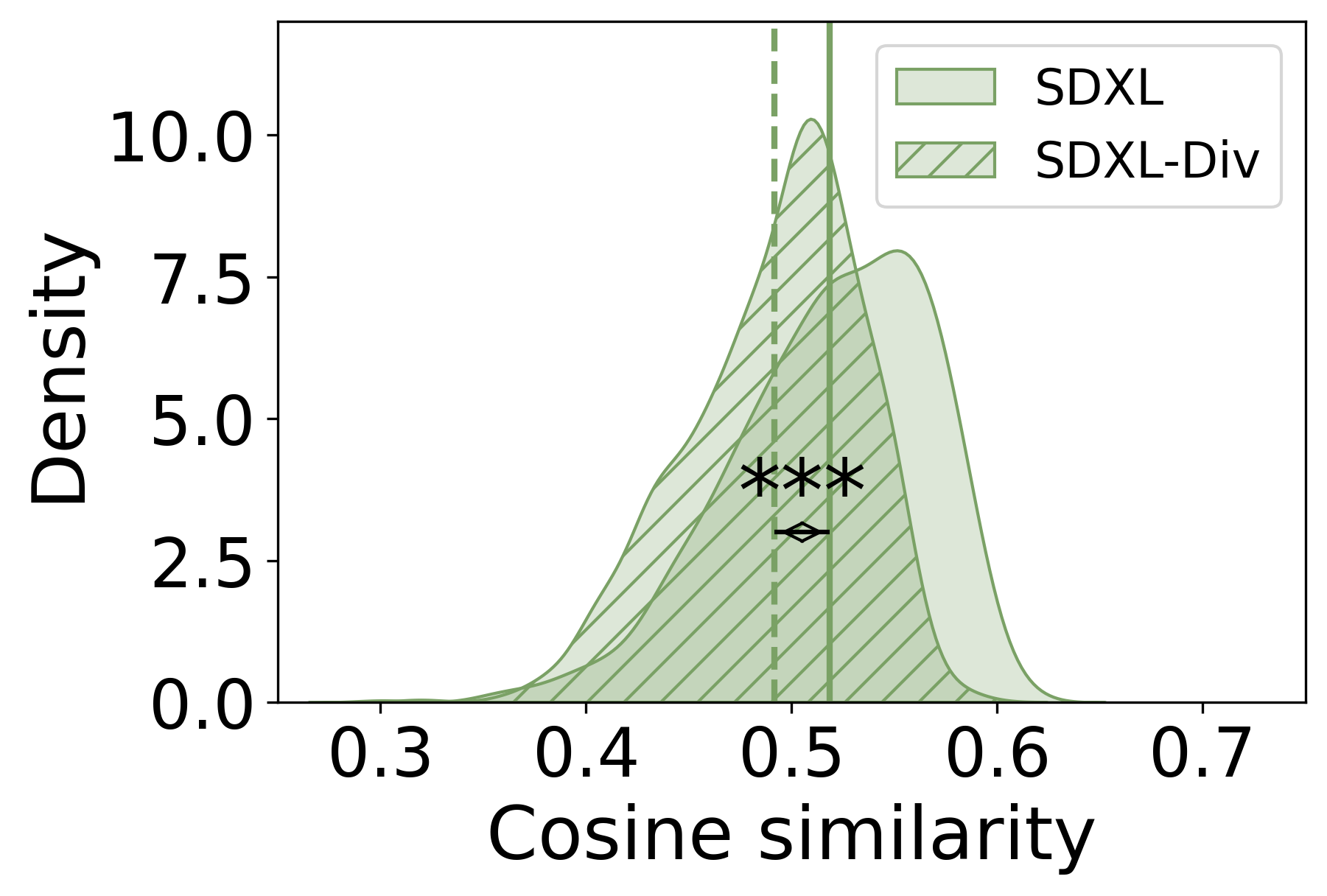} 
        \end{minipage}%
        \hfill
        \begin{minipage}[b]{0.33\textwidth}
            \includegraphics[width=\textwidth]{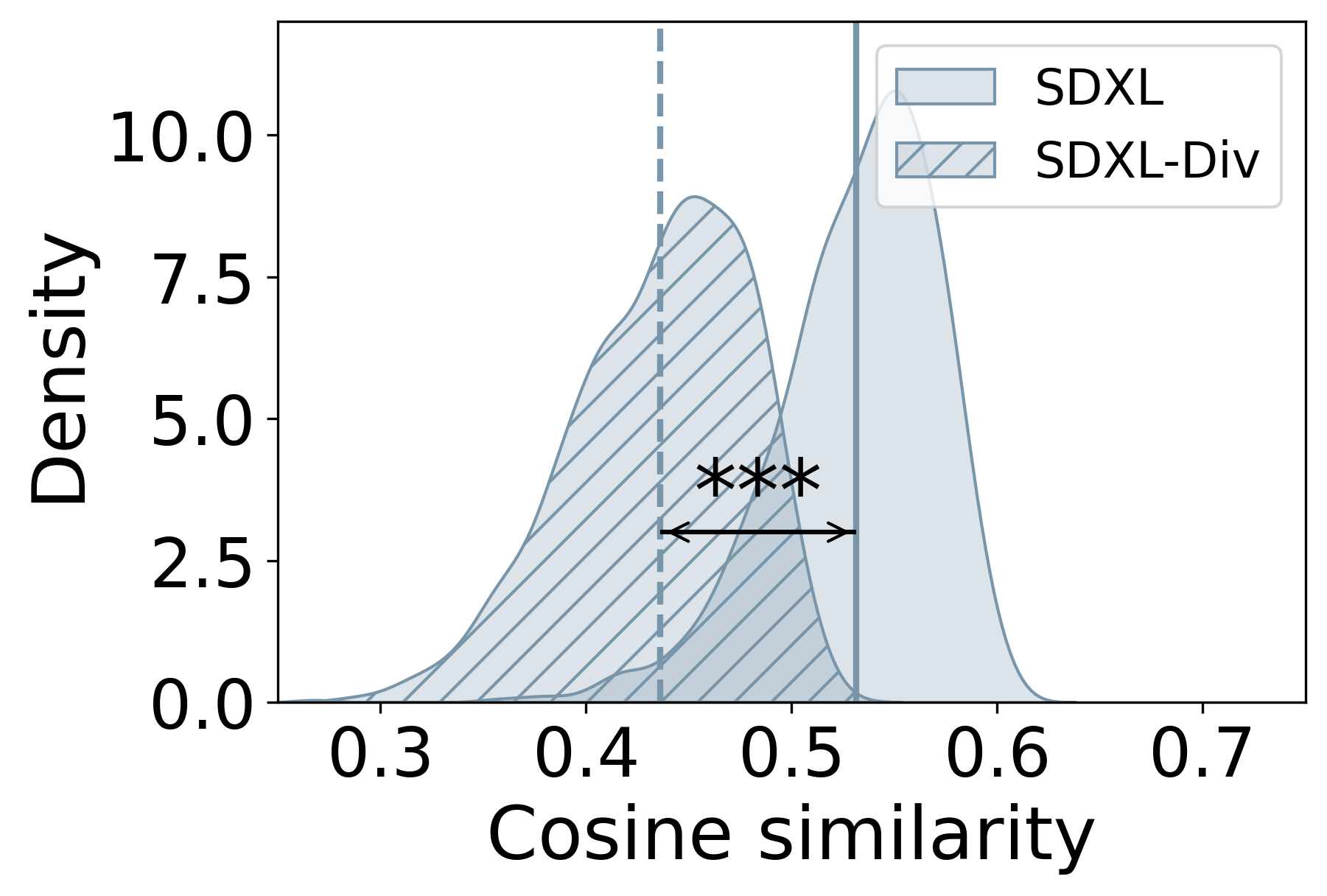} 
        \end{minipage}        
        \begin{minipage}[b]{0.33\textwidth}
            \includegraphics[width=\textwidth]{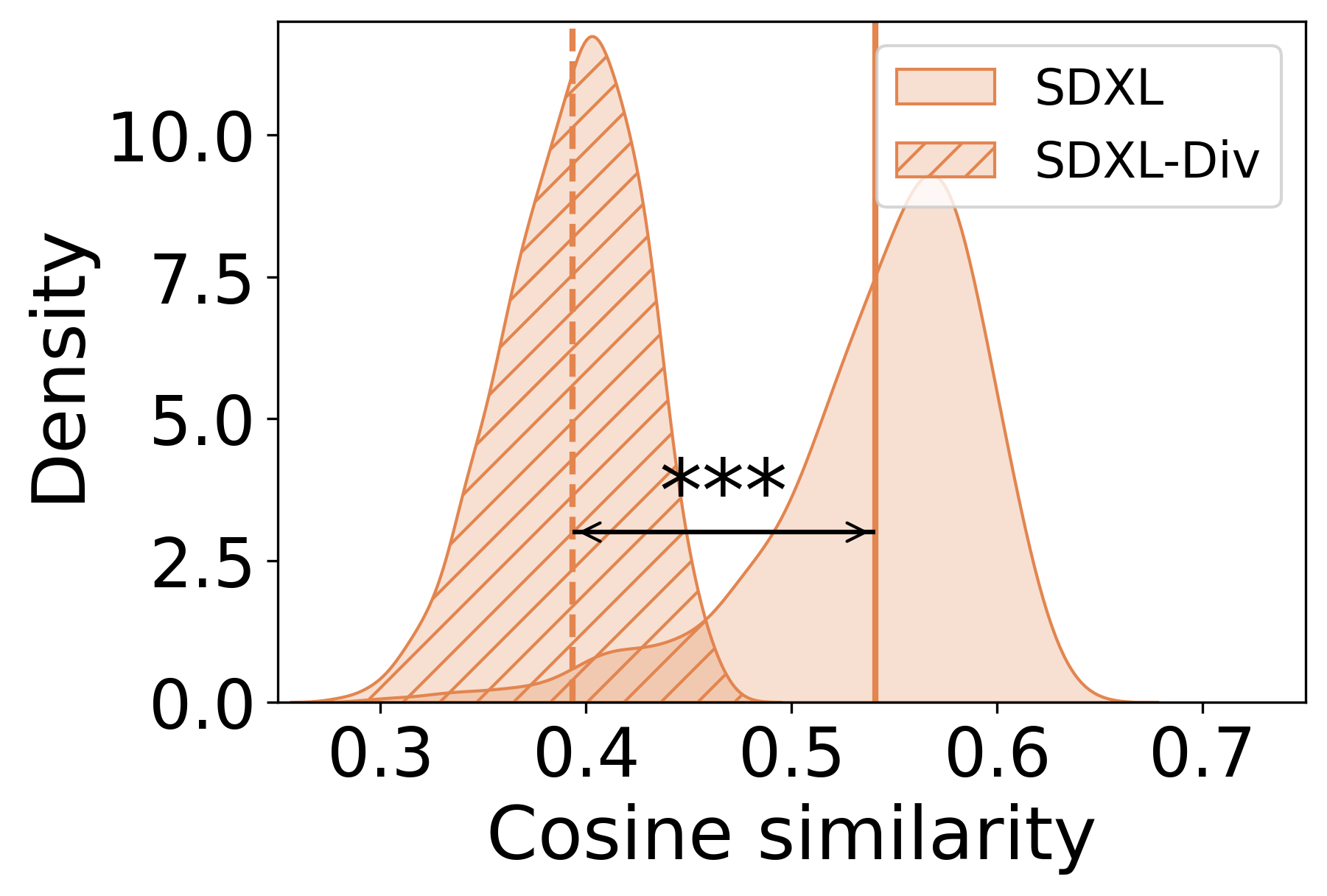} 
        \end{minipage}%
        \hfill
        \begin{minipage}[b]{0.33\textwidth}
            \includegraphics[width=\textwidth]{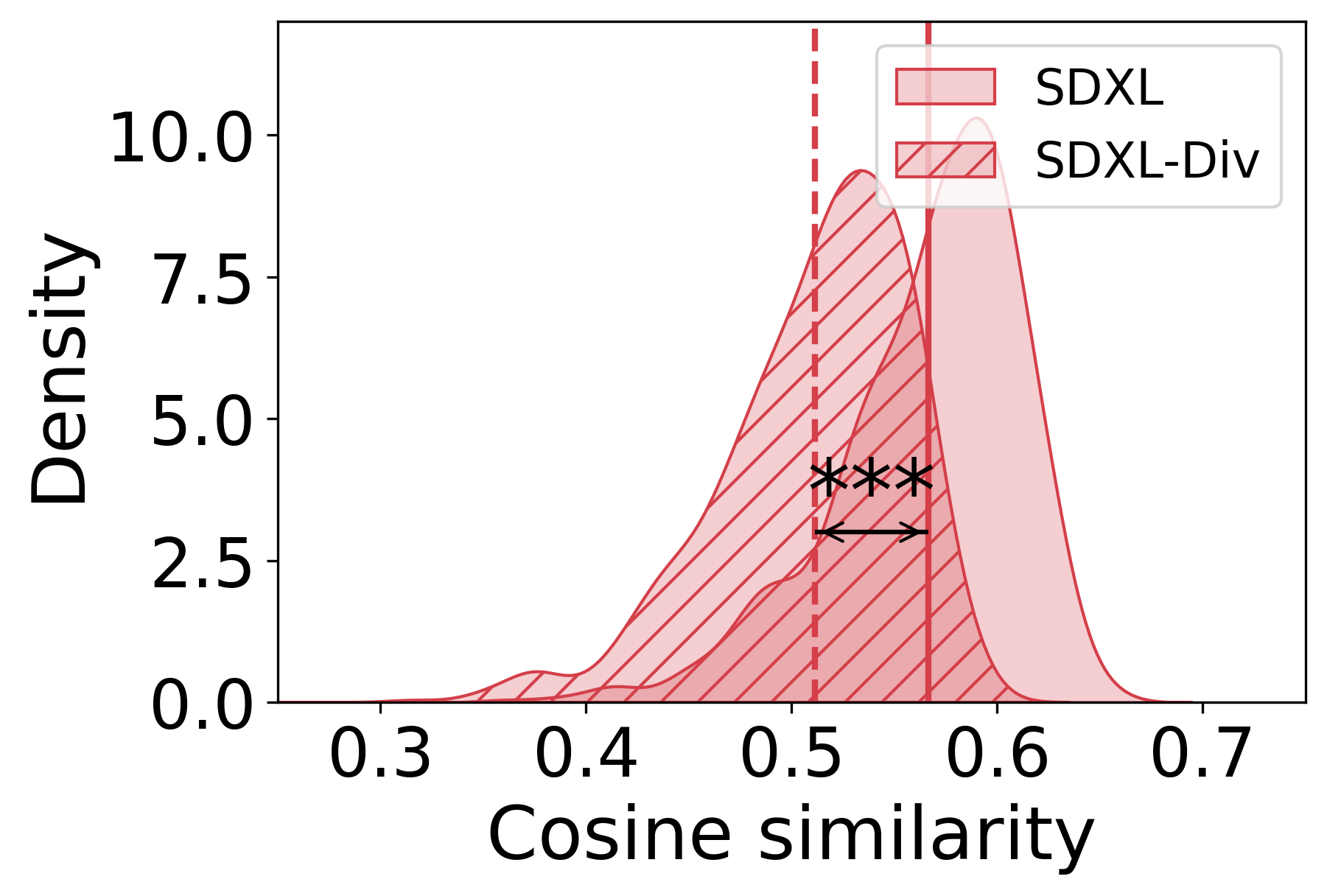} 
        \end{minipage}%
        \hfill
        \begin{minipage}[b]{0.33\textwidth}
            \includegraphics[width=\textwidth]{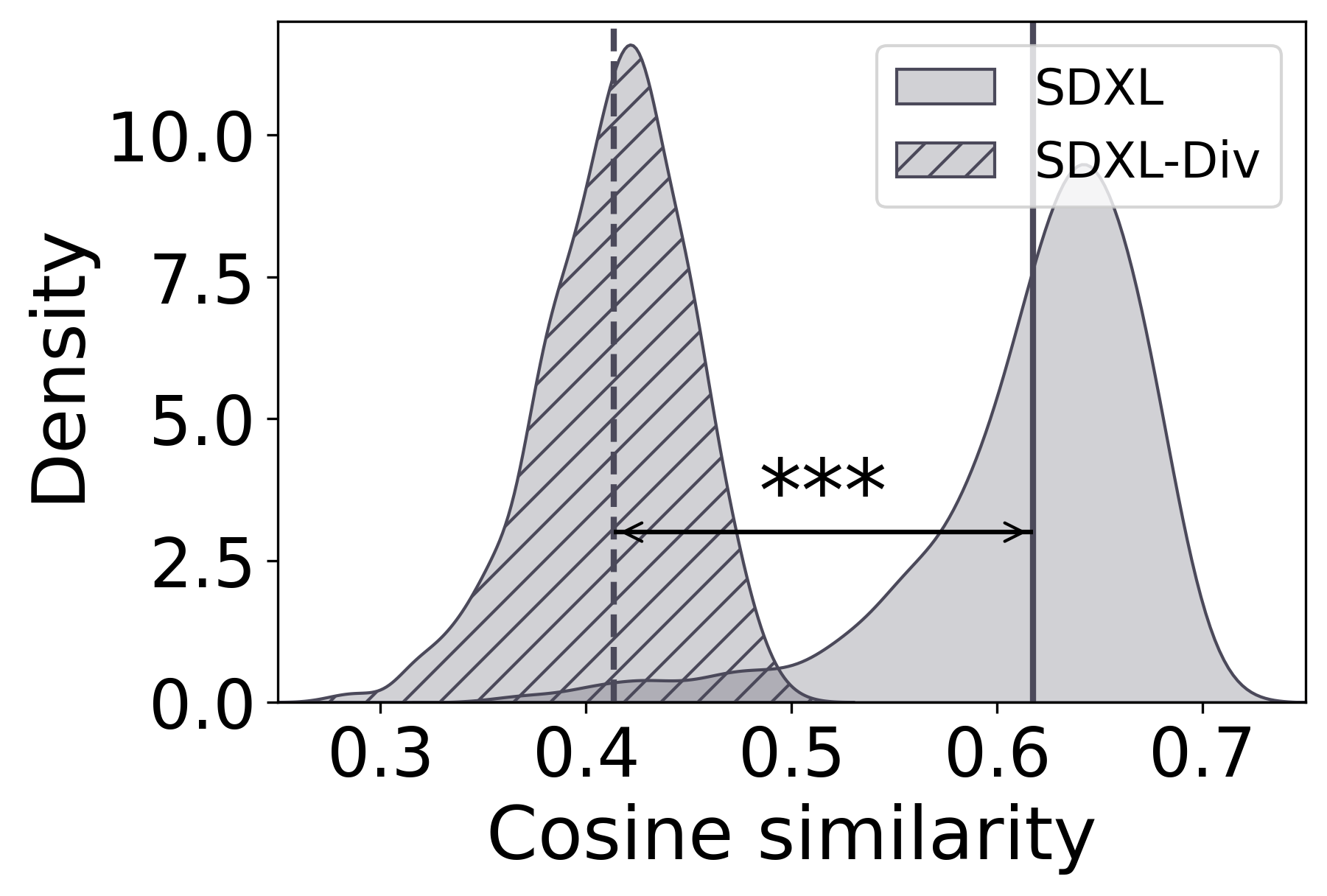} 
        \end{minipage}
    \end{minipage}
    \\
    \includegraphics[trim={0 1.8cm 0 8.5cm},clip,width=0.7\linewidth]{figures/legend.pdf}
    \vspace{-15pt}
    \begin{flushleft} \textbf{d} \end{flushleft}
    \begin{subfigure}{0.095\textwidth}
        \includegraphics[width=1\linewidth]{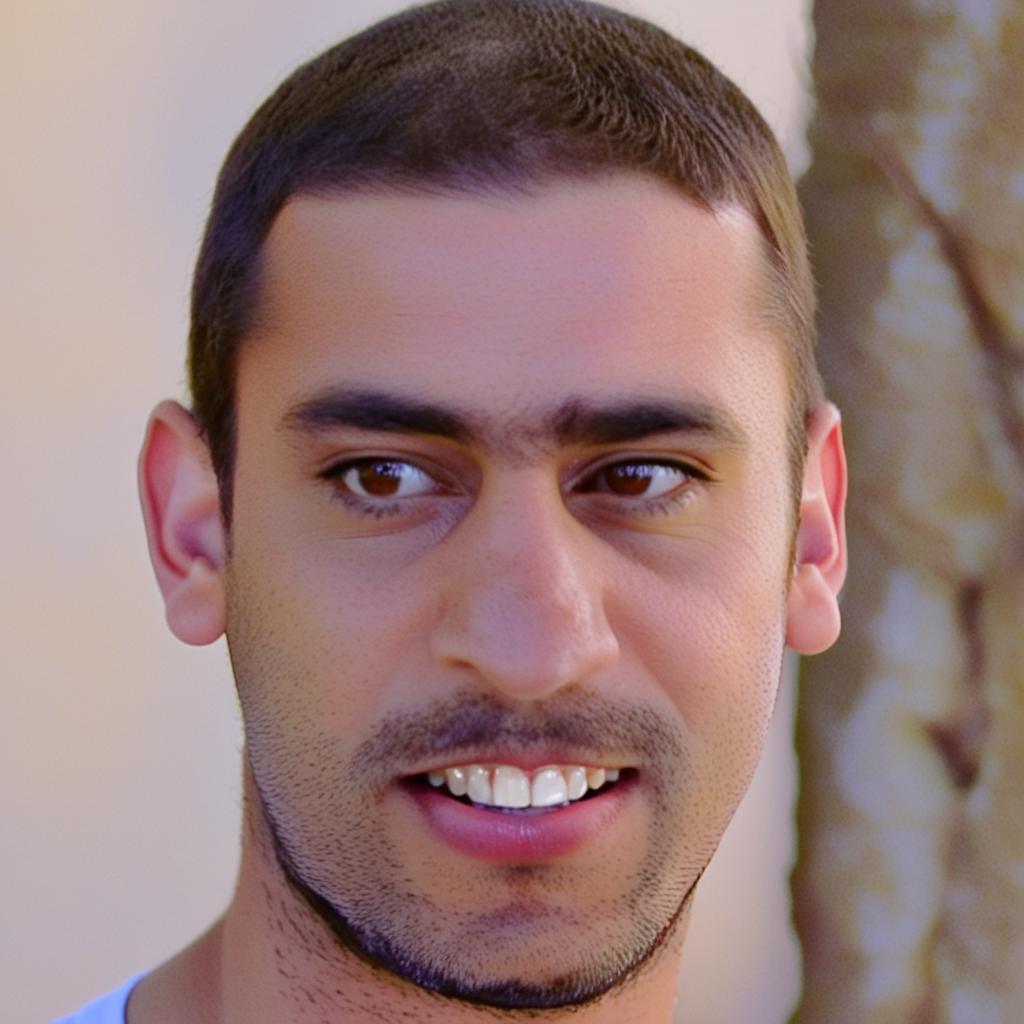}
    \end{subfigure}\hspace*{-0.25em}
    \begin{subfigure}{0.095\textwidth}
        \includegraphics[width=1\linewidth]{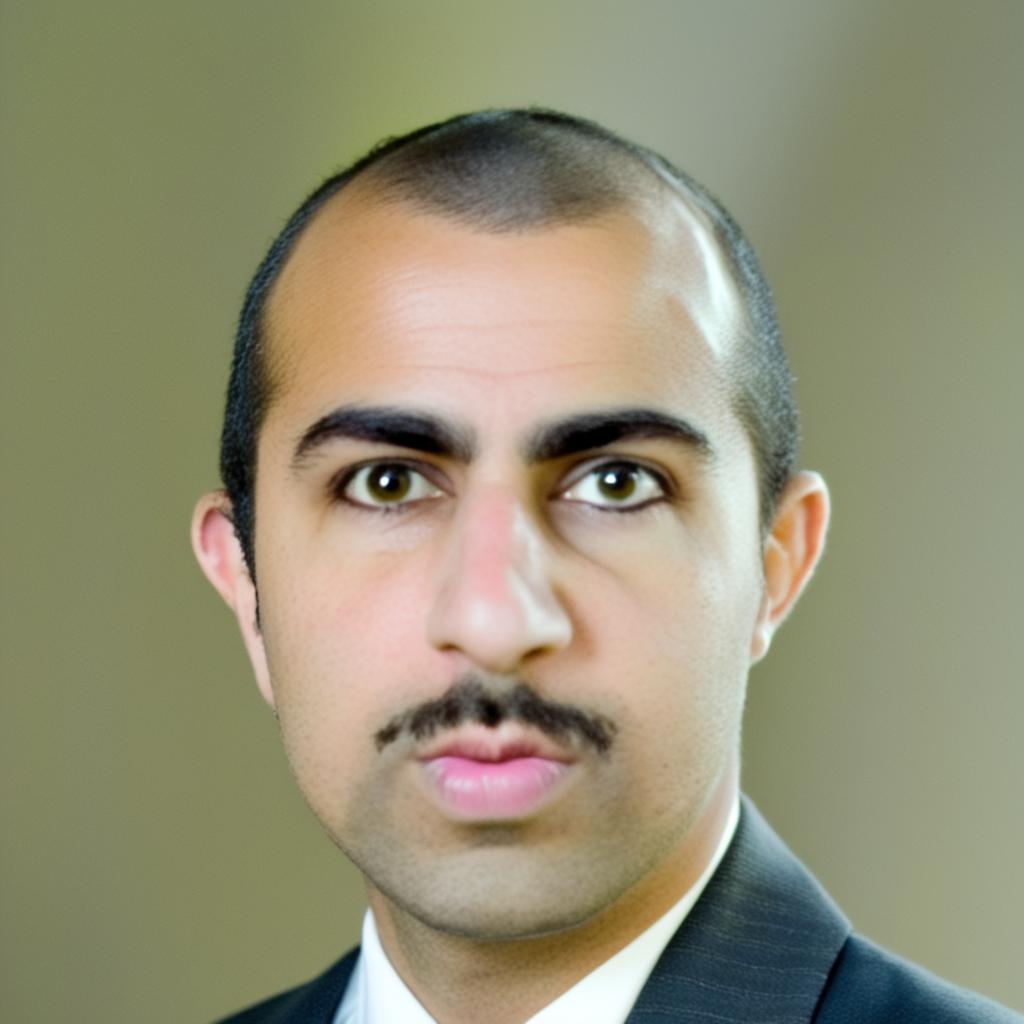}
    \end{subfigure}\hspace*{-0.25em}
    \begin{subfigure}{0.095\textwidth}
        \includegraphics[width=1\linewidth]{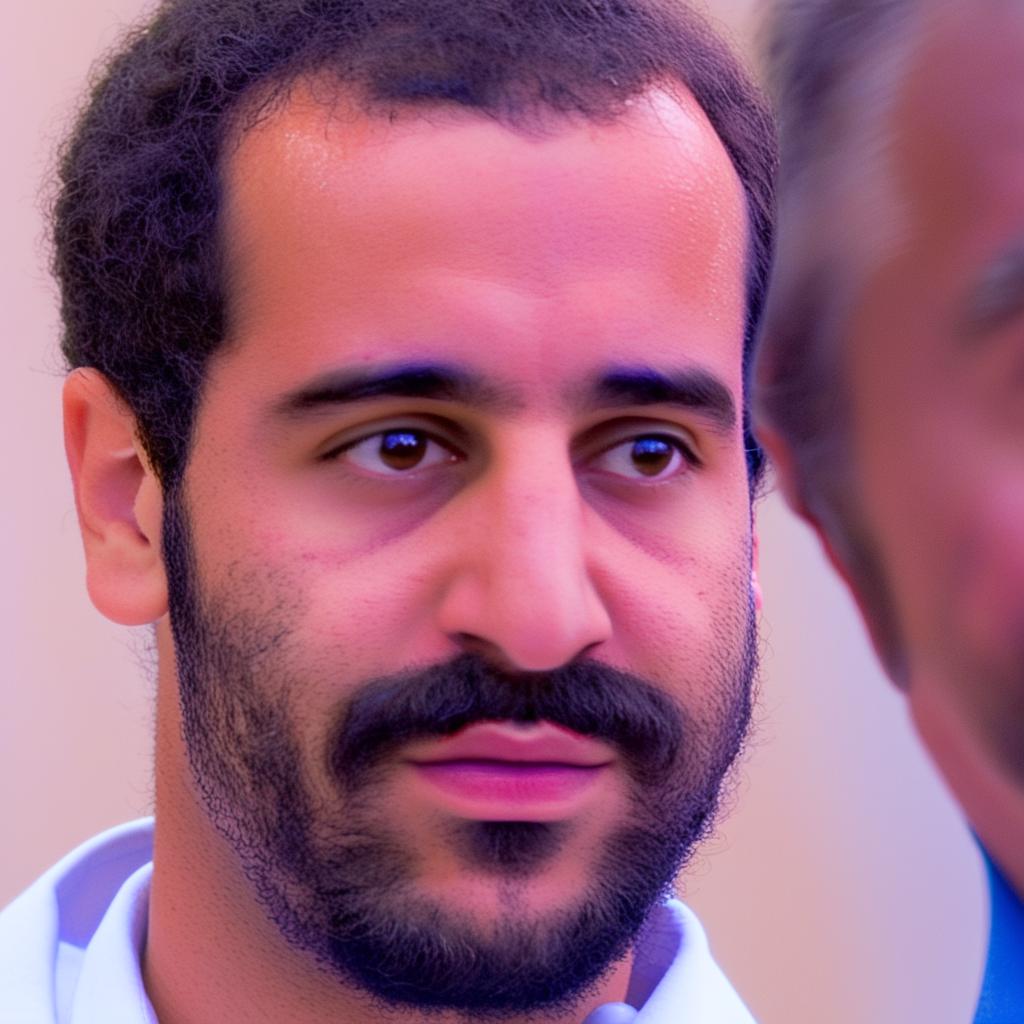}
    \end{subfigure}\hspace*{-0.25em}
    \begin{subfigure}{0.095\textwidth}
        \includegraphics[width=1\linewidth]{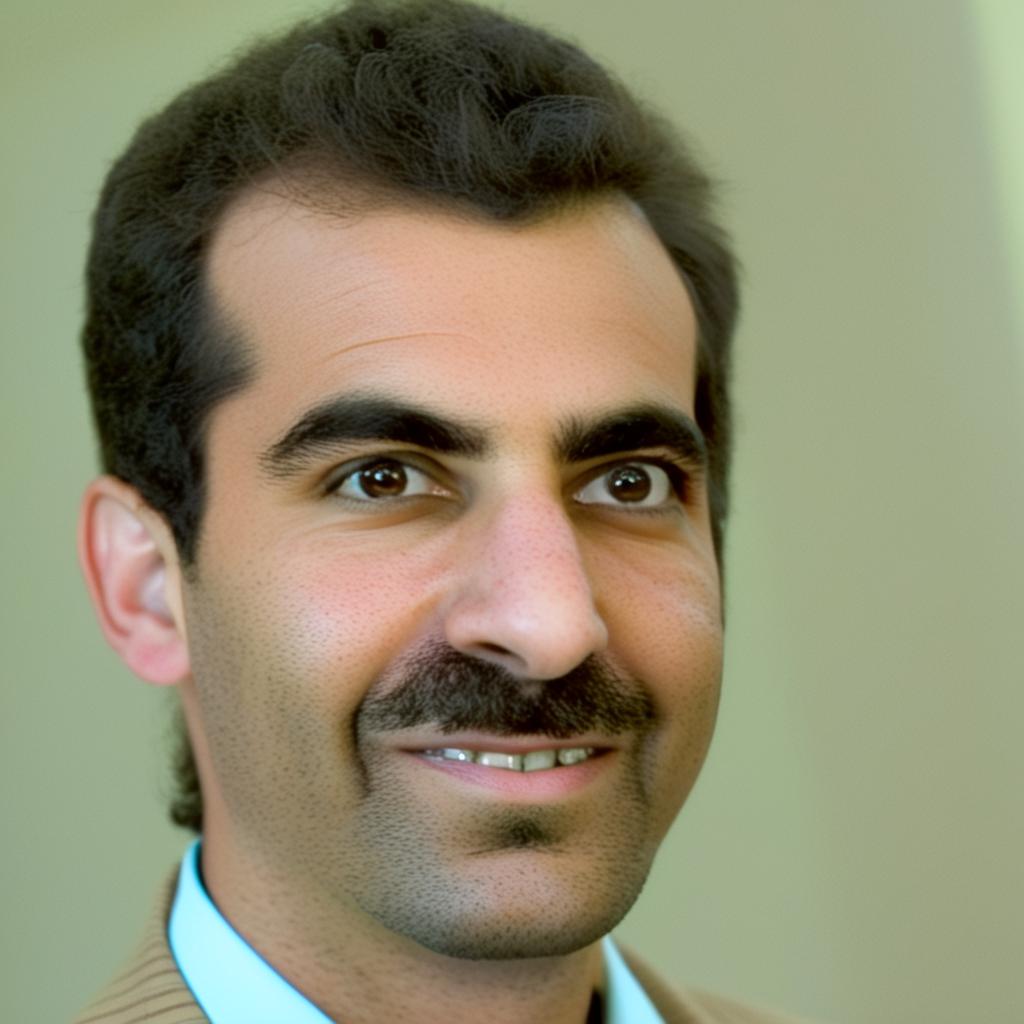}
    \end{subfigure}\hspace*{-0.25em}
    \begin{subfigure}{0.095\textwidth}
        \includegraphics[width=1\linewidth]{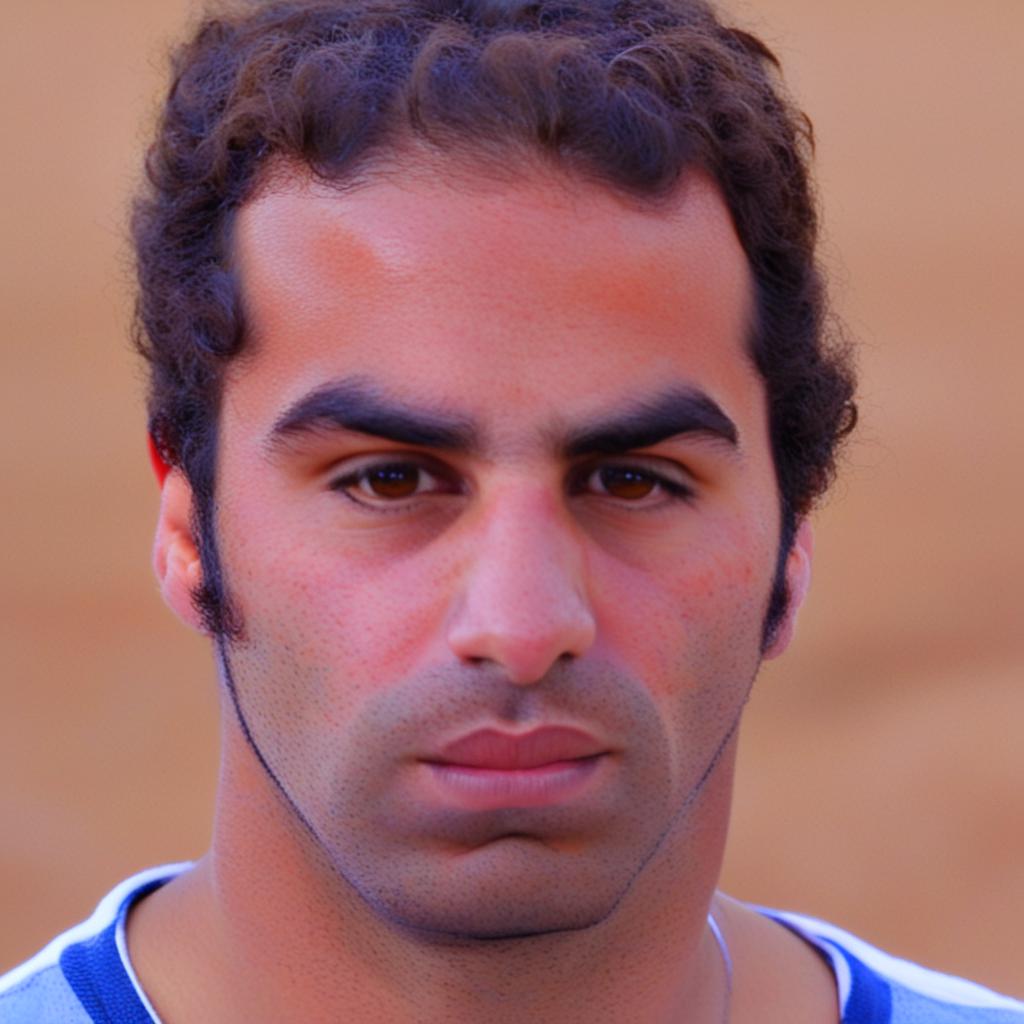}
    \end{subfigure}\hspace*{-0.25em}
    \begin{subfigure}{0.095\textwidth}
        \includegraphics[width=1\linewidth]{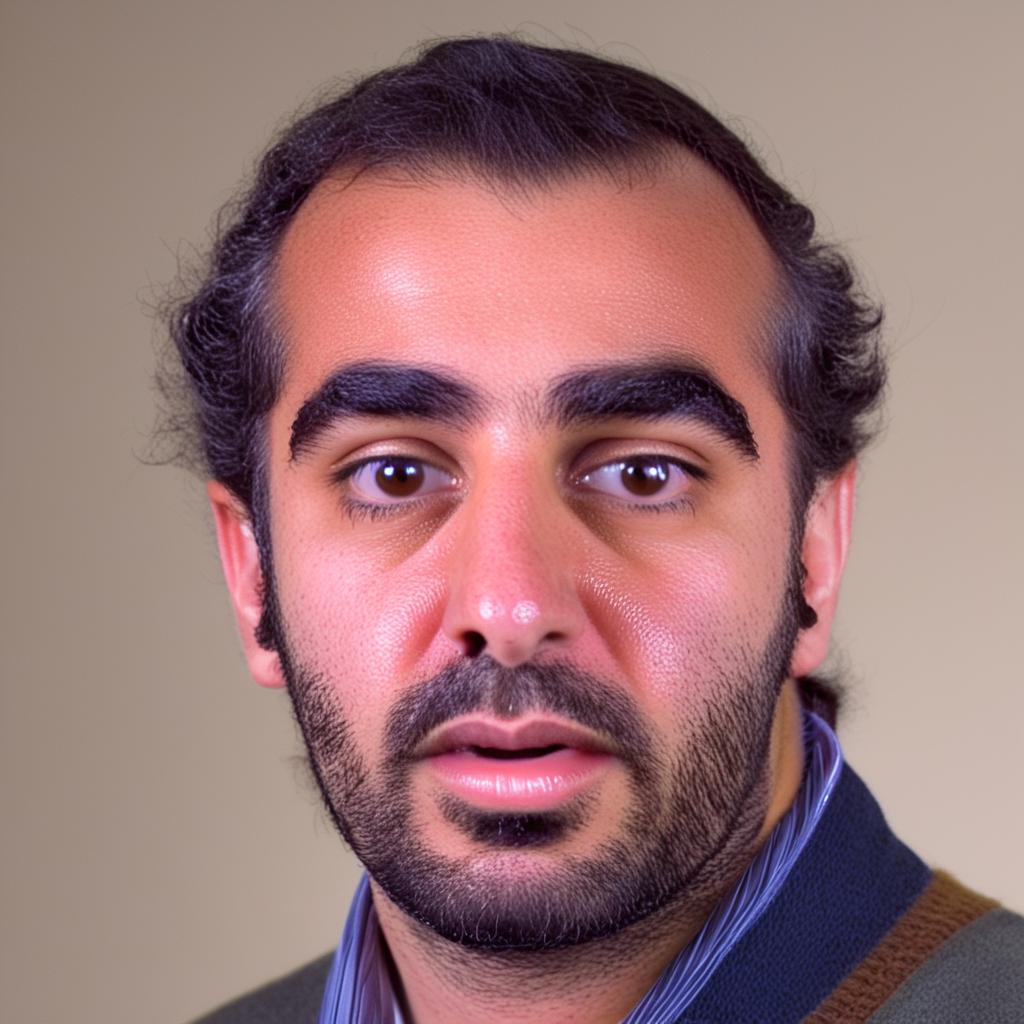}
    \end{subfigure}\hspace*{-0.25em}
    \begin{subfigure}{0.095\textwidth}
        \includegraphics[width=1\linewidth]{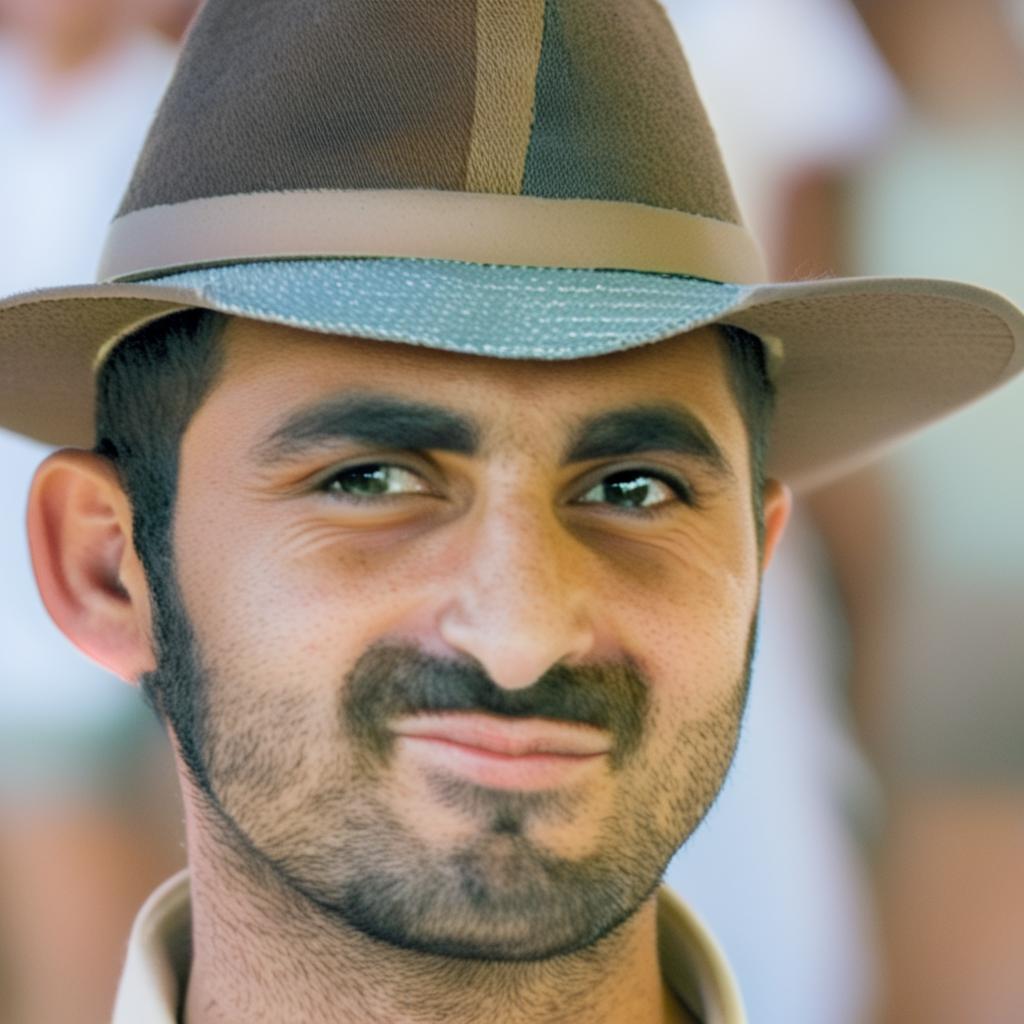}
    \end{subfigure}\hspace*{-0.25em}
    \begin{subfigure}{0.095\textwidth}
        \includegraphics[width=1\linewidth]{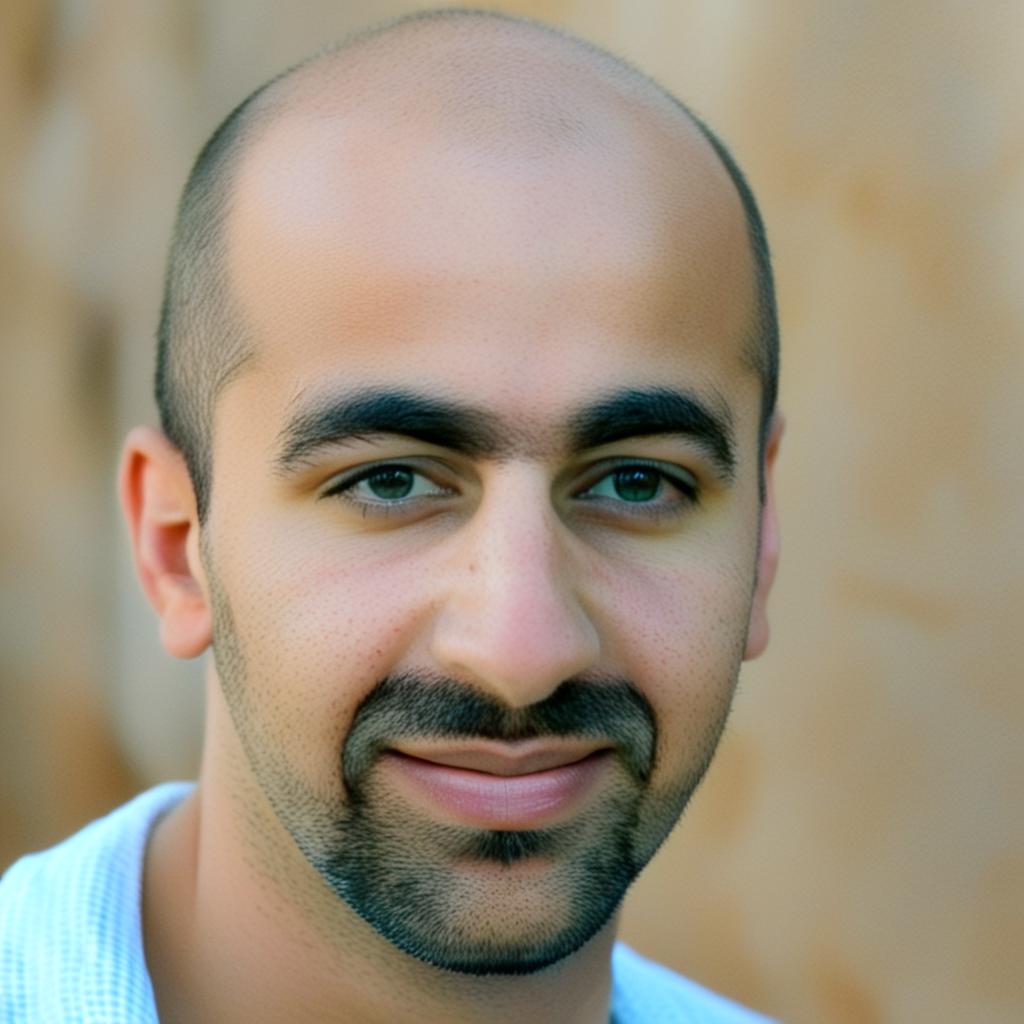}
    \end{subfigure}\hspace*{-0.25em}
    \begin{subfigure}{0.095\textwidth}
        \includegraphics[width=1\linewidth]{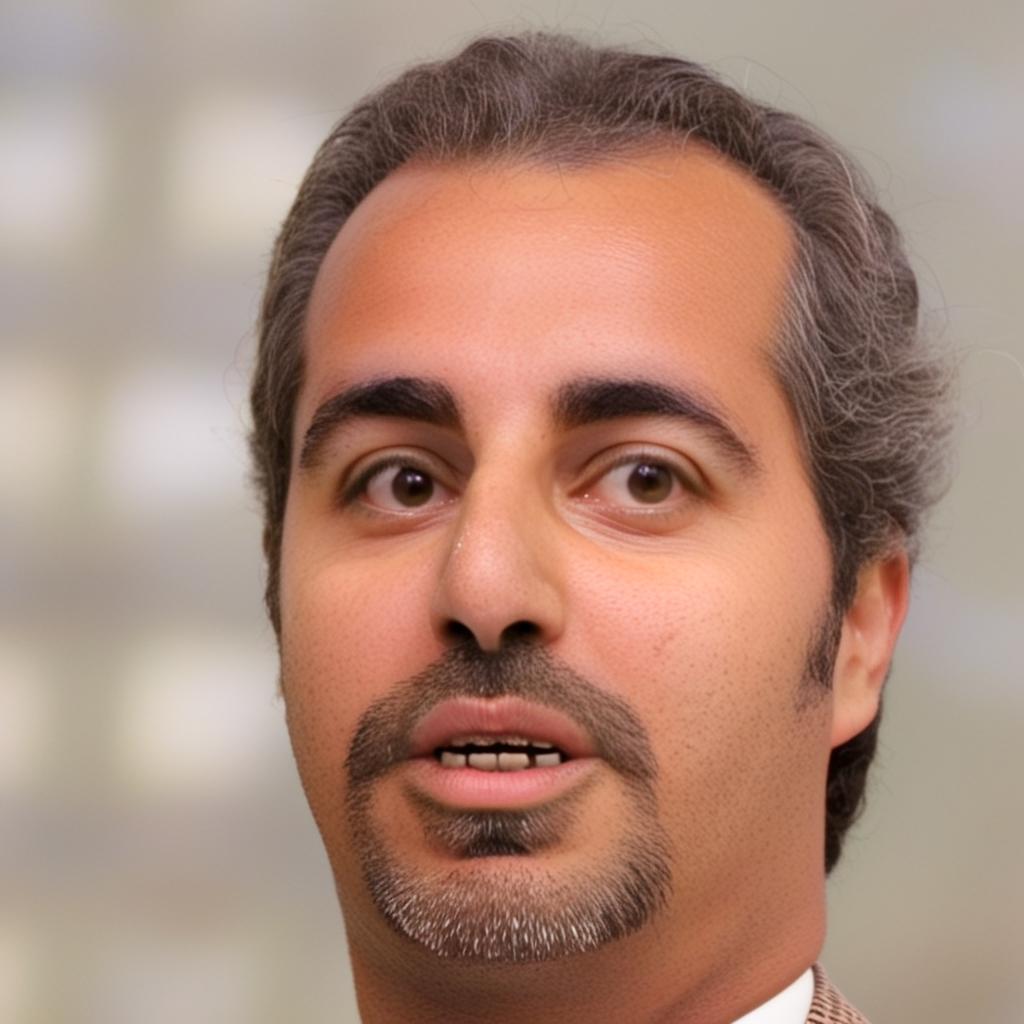}
    \end{subfigure}\hspace*{-0.25em}
    \begin{subfigure}{0.095\textwidth}
        \includegraphics[width=1\linewidth]{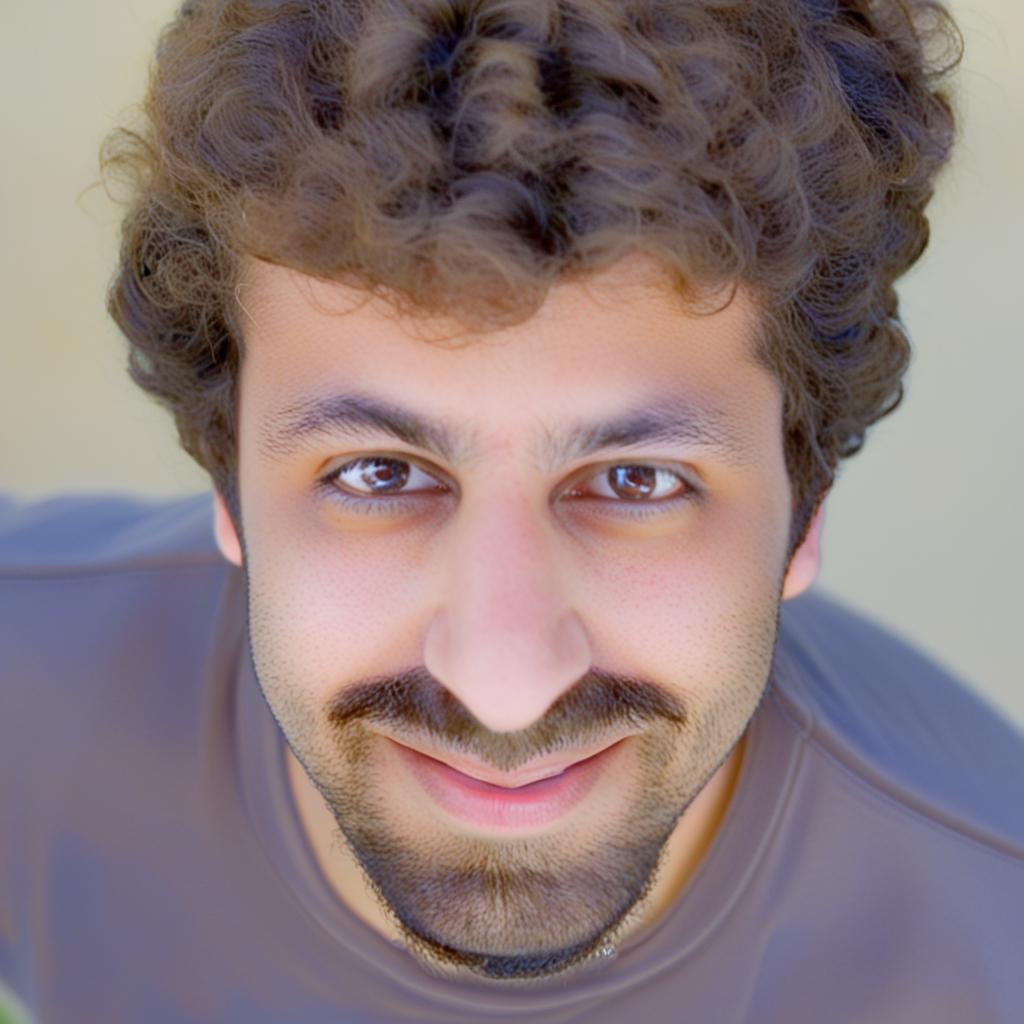}
    \end{subfigure}\\
    \begin{subfigure}{0.095\textwidth}
        \includegraphics[width=1\linewidth]{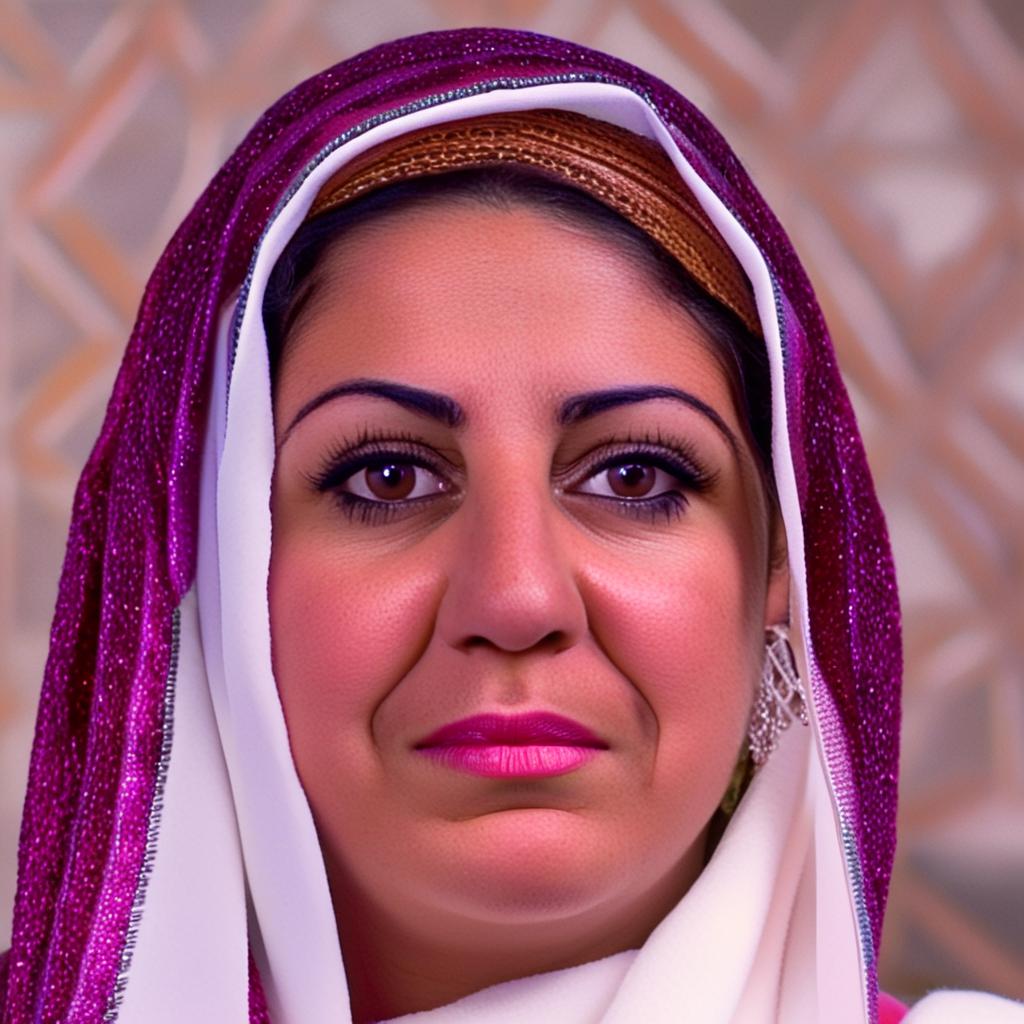}
    \end{subfigure}\hspace*{-0.25em}
    \begin{subfigure}{0.095\textwidth}
        \includegraphics[width=1\linewidth]{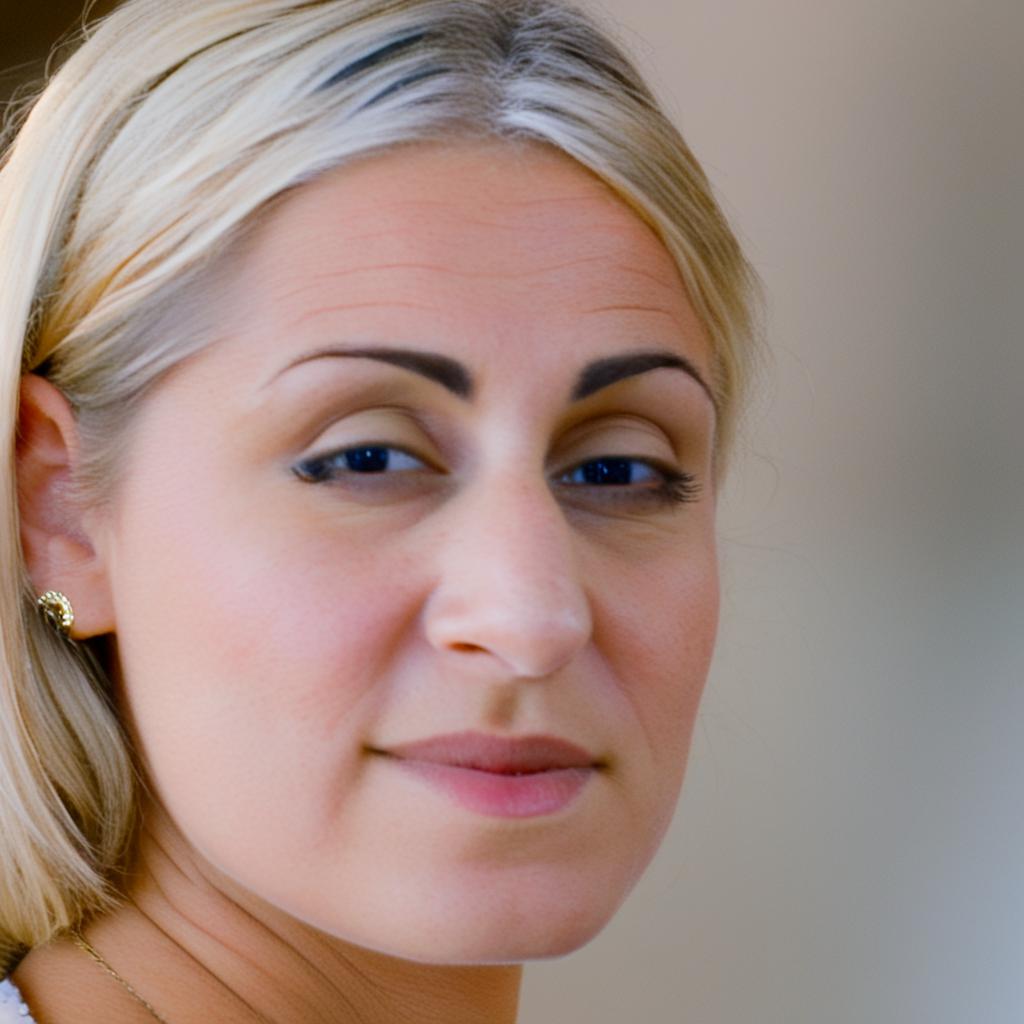}
    \end{subfigure}\hspace*{-0.25em}
    \begin{subfigure}{0.095\textwidth}
        \includegraphics[width=1\linewidth]{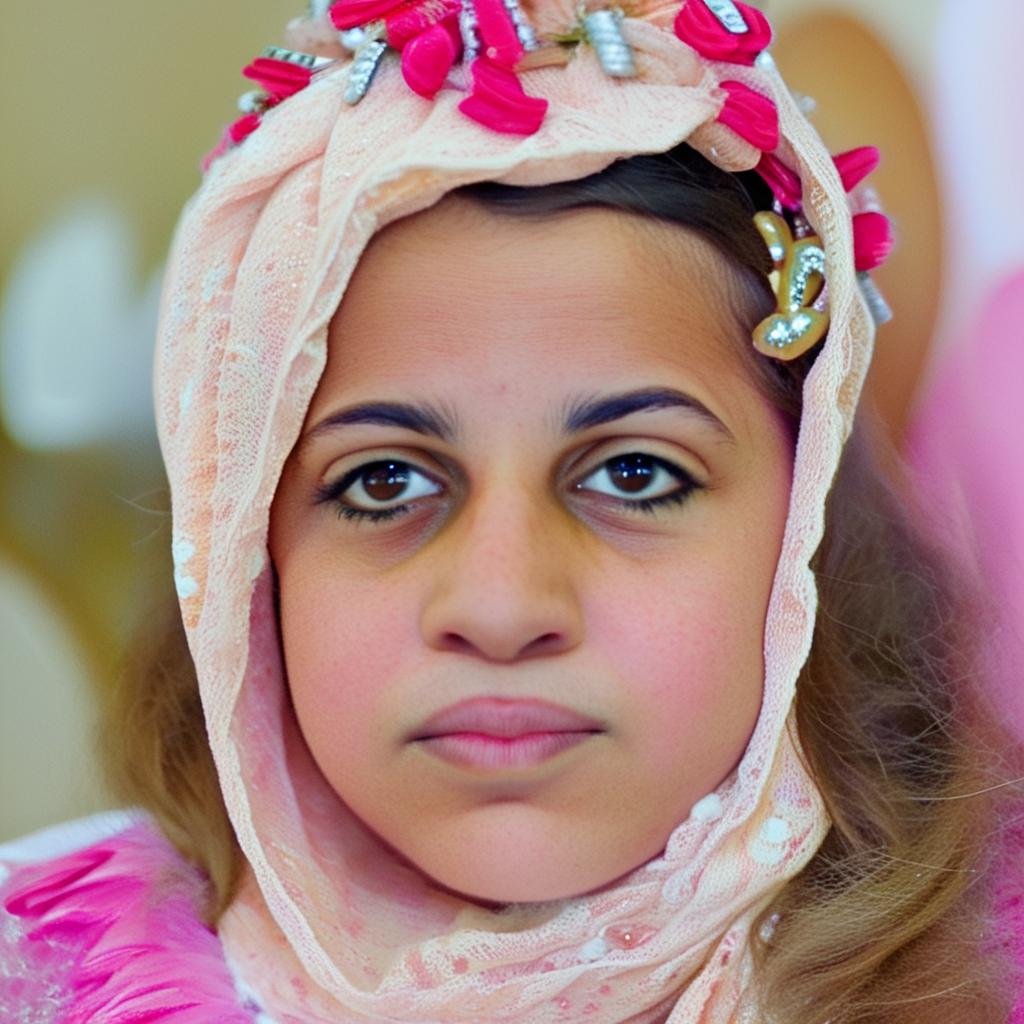}
    \end{subfigure}\hspace*{-0.25em}
    \begin{subfigure}{0.095\textwidth}
        \includegraphics[width=1\linewidth]{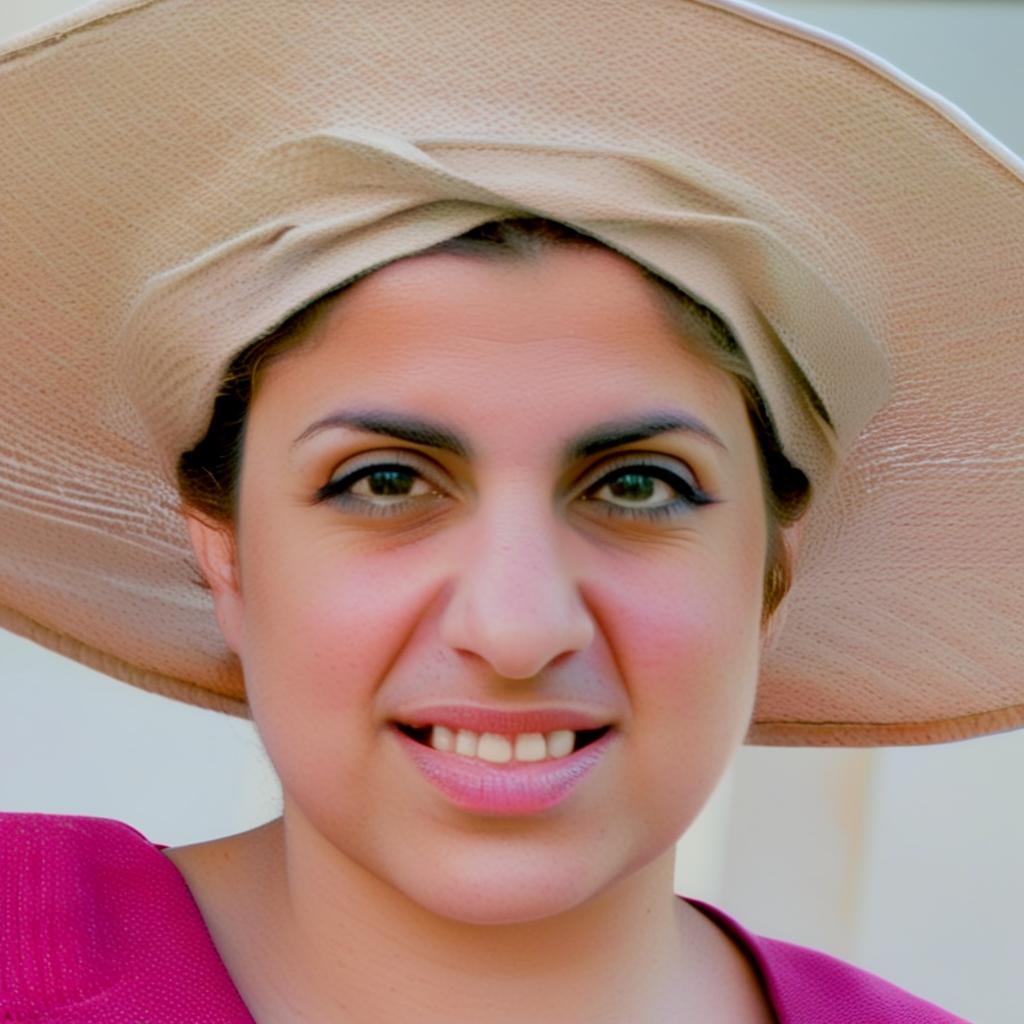}
    \end{subfigure}\hspace*{-0.25em}
    \begin{subfigure}{0.095\textwidth}
        \includegraphics[width=1\linewidth]{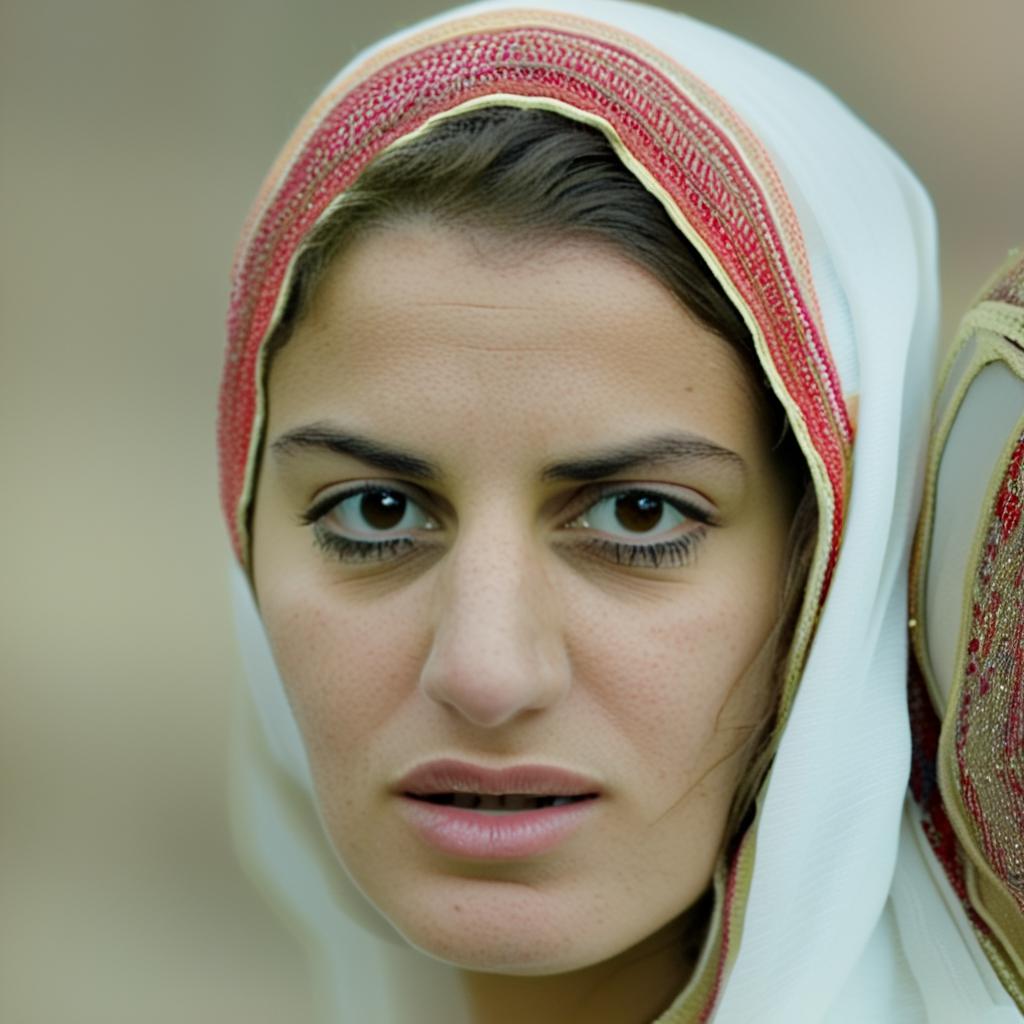}
    \end{subfigure}\hspace*{-0.25em}
    \begin{subfigure}{0.095\textwidth}
        \includegraphics[width=1\linewidth]{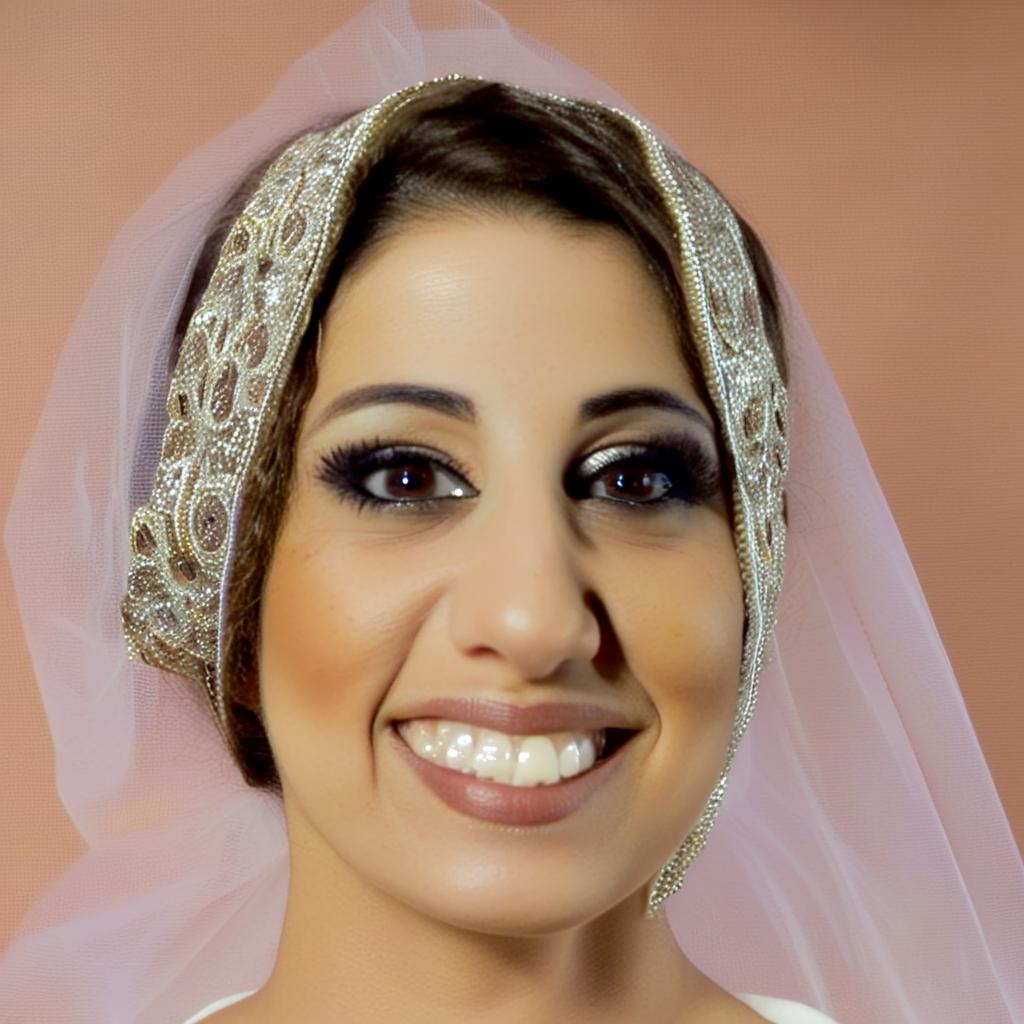}
    \end{subfigure}\hspace*{-0.25em}
    \begin{subfigure}{0.095\textwidth}
        \includegraphics[width=1\linewidth]{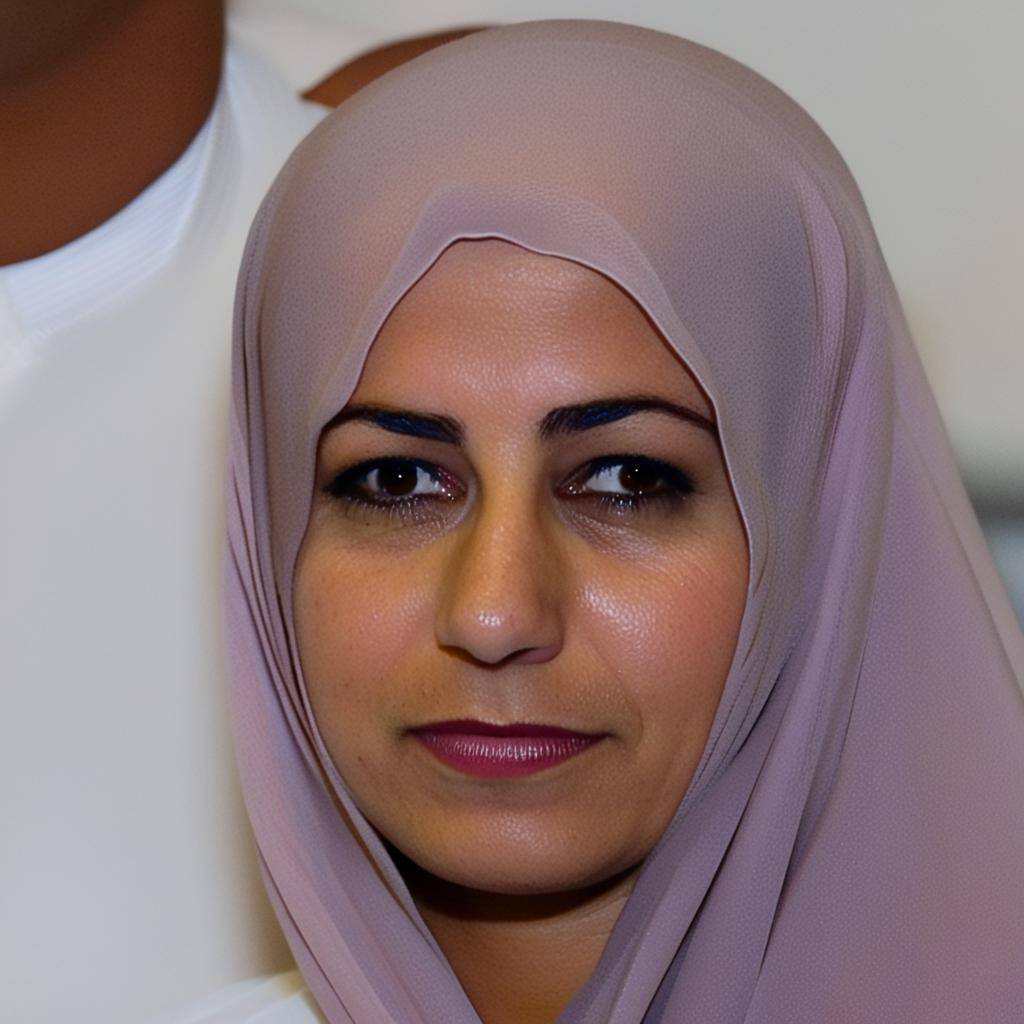}
    \end{subfigure}\hspace*{-0.25em}
    \begin{subfigure}{0.095\textwidth}
        \includegraphics[width=1\linewidth]{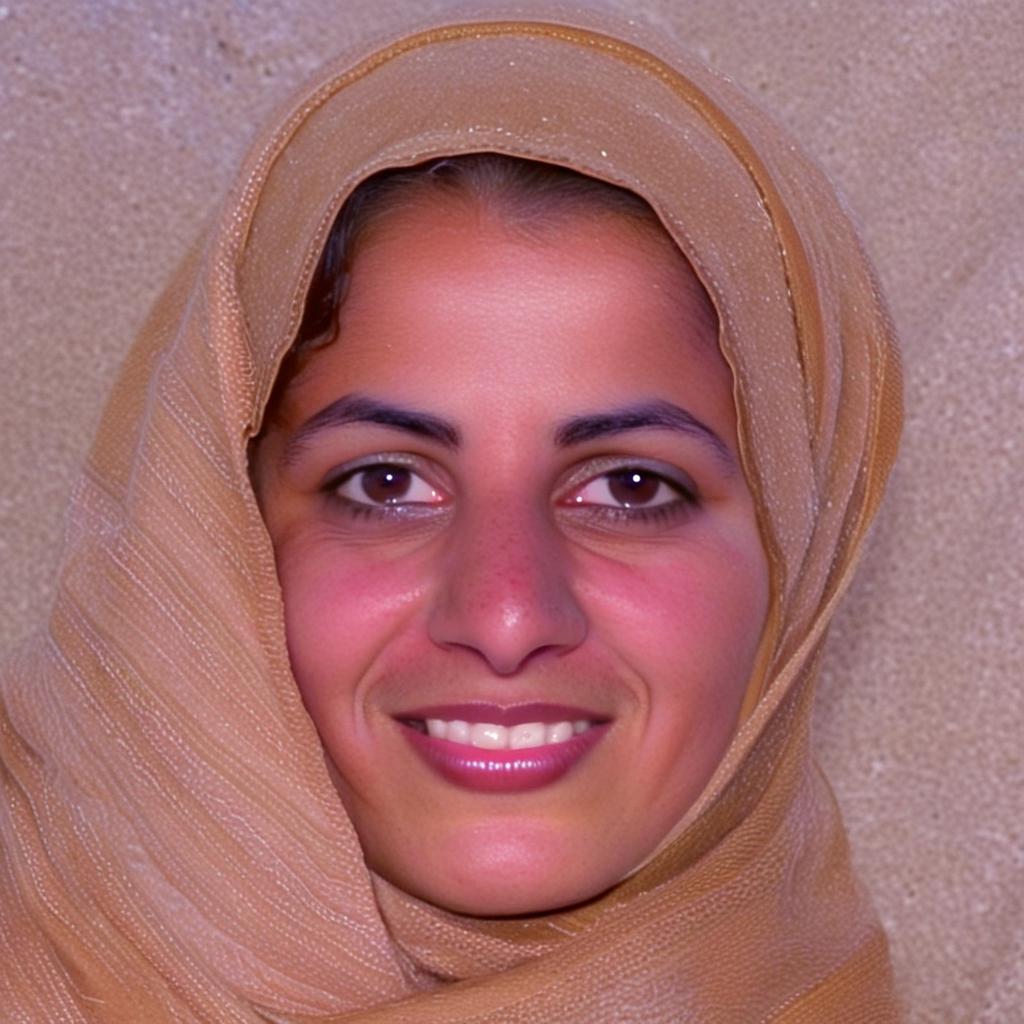}
    \end{subfigure}\hspace*{-0.25em}
    \begin{subfigure}{0.095\textwidth}
        \includegraphics[width=1\linewidth]{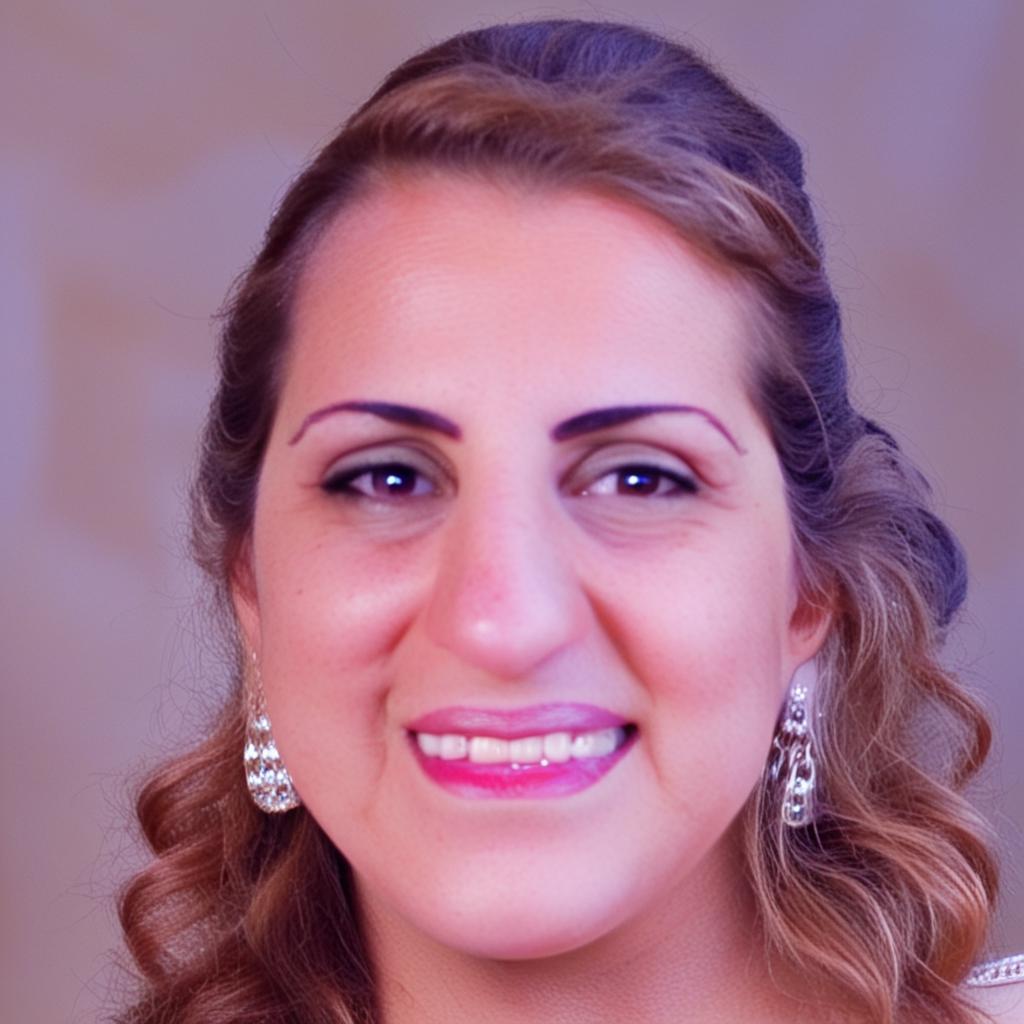}
    \end{subfigure}\hspace*{-0.25em}
    \begin{subfigure}{0.095\textwidth}
        \includegraphics[width=1\linewidth]{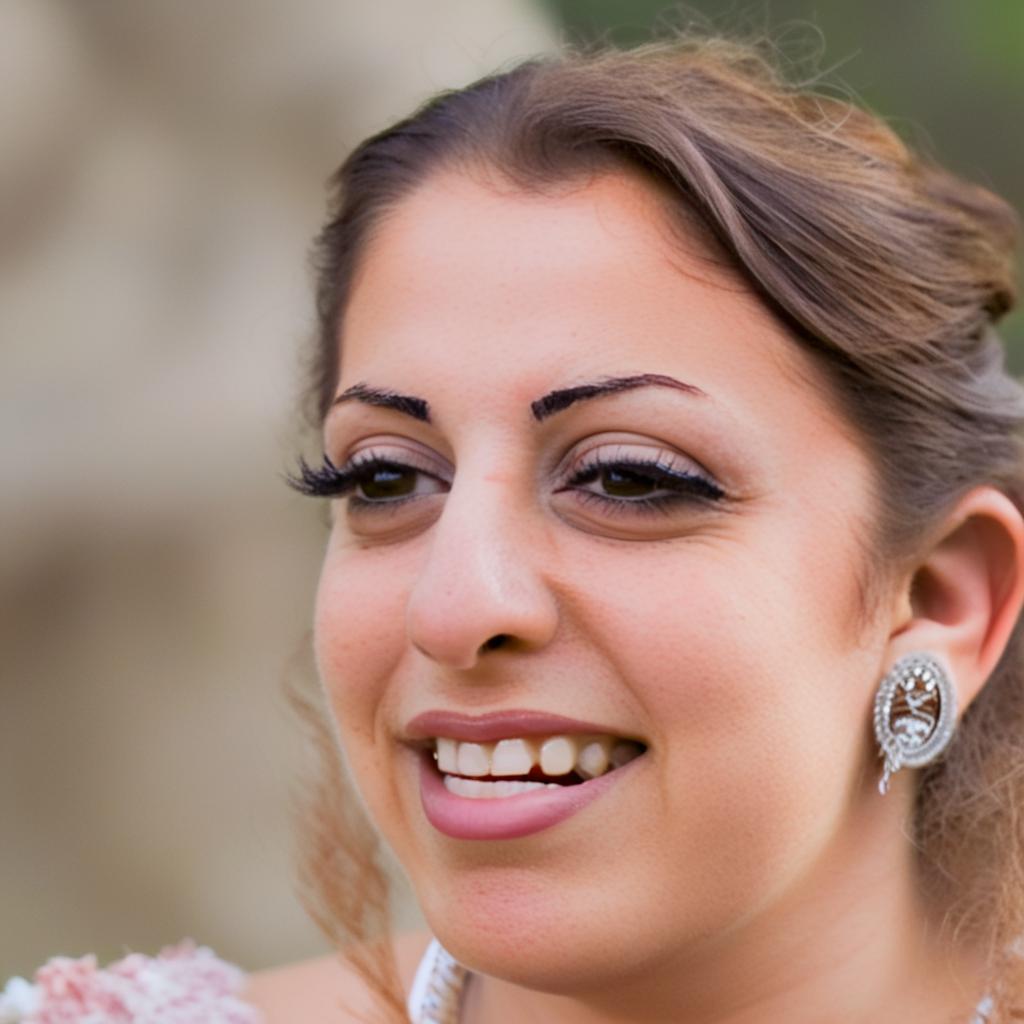}
    \end{subfigure}
    
\caption{
\textbf{Quantifying and addressing racial homogenization.} \textbf{a}, Sample images of Middle Eastern individuals generated using SDXL. \textbf{b}, For each race, the distribution of the average cosine similarity between a given image and all other images of that race (dashed lines represent the means). \textbf{c}, Comparing the distributions produced by SDXL (solid) to those produced by SDXL-Div (dashed). \textbf{d}, Sample images of Middle Eastern individuals generated using SDXL-Div. P values are from t-tests; *** $p<.001$.
}
\label{fig:diversity}
\end{figure}

\subsection*{Quantifying and addressing racial homogenization}
Motivated by Edward Said's concept of Orientalism, which critiques the West's tendency to represent Eastern cultures and people through reductive and homogenous imagery~\cite{said1977orientalism}, we examined the potential for SDXL to exhibit similar patterns of racial homogenization. Specifically, we investigated whether SDXL generated images of certain racial groups with overly similar facial features, echoing Orientalist representations that flatten the complexities of individual identities. To this end, we started off by manually examining a random sample of SDXL-generated images depicting Middle Eastern individuals. As shown in Figure~\ref{fig:diversity}a, such images depict men as being extremely similar to one another, e.g., they are all bearded and brown-skinned. Similarly, images representing women appear similar to each other, e.g., they have almost identical skin tones, and they all appear to be wearing traditional attire. While this random sample is indeed very small, it motivates a thorough examination of the phenomenon across racial and gender groups. Moreover, this sample highlights the need to develop a solution that increases the diversity of facial features when depicting individuals of a given race or gender.

To quantify the degree of racial homogenization in SDXL, we used it to generate 1,000 images per race, and obtained an embedding for each image using our racial classifier. Finally, for every image of a particular race, we measured the average cosine similarity between that image and all other images of that race. The result of this analysis is depicted in Figure~\ref{fig:diversity}b. As shown in this figure, the degree to which White individuals resemble one another is smaller than that of any given race. To put it differently, the facial diversity of White people is greater than that of other races. Interestingly, images depicting Middle Eastern individuals are the least diverse, which reinforces Orientalism. 

The above results motivate the development of a version of SDXL that generates images with greater facial diversity per race (compared to the original version). To this end, we downloaded from Flickr-Faces-HQ high-resolution (1024×1024) images of human faces with ``considerable variation in terms of age, race, and image background''~\cite{ffhq}. Since the dataset is unlabeled, we used our classifier to infer the race of each image; see Supplementary Table~4 for the number of images per race. The resulting labeled dataset was then used to fine-tune SDXL. Again, the fine-tuning was done using LORA~\cite{hu2021lora} to reduce the number of trainable parameters for the downstream task. We call the resulting model ``SDXL-Div'' (where Div is a shorthand for Diversity); see Materials and Methods for technical details. To evaluate this model, we used it to generate 1,000 images and for each one of them, we measured the average cosine similarity to all other images of that race.

As can be seen in Figure~\ref{fig:diversity}c, SDXL-Div is indeed able to increase the facial diversity of SDXL, regardless of the race. The greatest difference is observed in the case of Middle Eastern individuals (dropping the mean cosine similarity from 0.61 to 0.41) and Latinx individuals (from 0.54 to 0.39). To help the reader appreciate the difference in diversity between the two models, Figure~\ref{fig:diversity}d provides sample images of Middle Eastern men and women generated SDXL-Div. As can be seen, images generated using this model have markedly greater facial diversity compared to those generated using the former model.

\subsection*{Survey experiments}

Given the increasing pervasiveness of AI-generated images, it is crucial to understand how repeated exposure to such images may shape public perceptions, especially concerning sensitive topics like race and gender. To this end, we ran four  survey experiments. We estimated that to obtain a power of 0.8 to detect a medium effect size (Cohen's $d$) of 0.5 in a paired-sample comparison, a sample of 135 participants would be needed. We recruited participants on Prolific, with prescreening settings of US residents and English as native language. Additionally, we prohibited participation more than once in our experiment. These survey experiments were preregistered at AsPredicted (\url{https://aspredicted.org/3v3wt.pdf}). All experiments were approved by the Institutional Review Board at New York University Abu Dhabi under the category of Exempt or Expedited Research (HRPP-2024-53).

In each of the four studies, participants are presented with six AI-generated images, answer a few questions about each image, and then answer an overall question. Each study has four different conditions, depending on whether the images are inclusive or not, and whether participants are informed that the images are generated by AI or are produced by an artist. Next, we provide an overview of the four studies; see Supplementary Note~3 for the exact wording and images used therein.

Study~1 examines racial bias. In particular, given a profession $P \in \{\textnormal{chef}, \textnormal{dietitian}, \textnormal{journalist}\}$, participants are presented with six images depicting $P$. For each image, they are asked to determine the age, perceived gender, and perceived race of the person depicted in the image. Finally, after seeing all six images, participants answer the question $Q_1$: What percentage of $P$ in the US are White?
Here, the non-inclusive images are generated using SDXL and they all depict white individuals, whereas the inclusive images are generated using SDXL-Inc and each of the six races are depicted in a separate image. In both conditions, three images depict men, while the other three depict women. All images can be seen in Supplementary Figure~7.

Study~2 examines gender bias, and is similar to Study~1 apart from two modifications: (i) The professions are $\{\textnormal{accountant}, \textnormal{math scientist}, \textnormal{and} \textnormal{tailor}\}$; (ii) after seeing all six images, participants answer the question $Q_2$: What percentage of $P$ in the US are men?
Here, the non-inclusive images are SDXL-generated, and all depict men, while the inclusive images are generated by SDXL-Inc and half depict men while the other half depicts women. In both conditions, each of the six races is depicted in a separate image; see Supplementary Figure~9.

Study~3 examines racial homogenization. Specifically, participants are presented with six images depicting Middle Eastern men. For each image, they are asked to describe the age, skin tone, and facial hair of the man featured in the image. After seeing all six images, participants answer the question $Q_3$: What is your estimation of the percentage of Middle Eastern men who have beards?
Here, the non-inclusive images are SDXL-generated, and all depict bearded Middle Eastern men. In contrast, the inclusive ones are generated by SDXL-Div, and depict men with varying levels of facial hair; see Supplementary Figure~11

Study~4 also examines racial homogenization, but focuses on women instead of men. In particular, participants are presented with six images depicting Middle Eastern women, and are asked to describe the age, skin tone, and head cover of the woman featured in the image. Finally, participants answer the question $Q_4$: What is your estimation of the percentage of Middle Eastern women who wear headcovers? Here, the non-inclusive images are generated by SDXL, and depict Middle Eastern women that are all wearing a head cover. On the other hand, the inclusive ones are generated by SDXL-Div and depict Middle Eastern women, half of whom are wearing a head cover while the other half are showing their hair.

\begin{figure}[!htbp]
\centering
\includegraphics[width=0.48\textwidth]{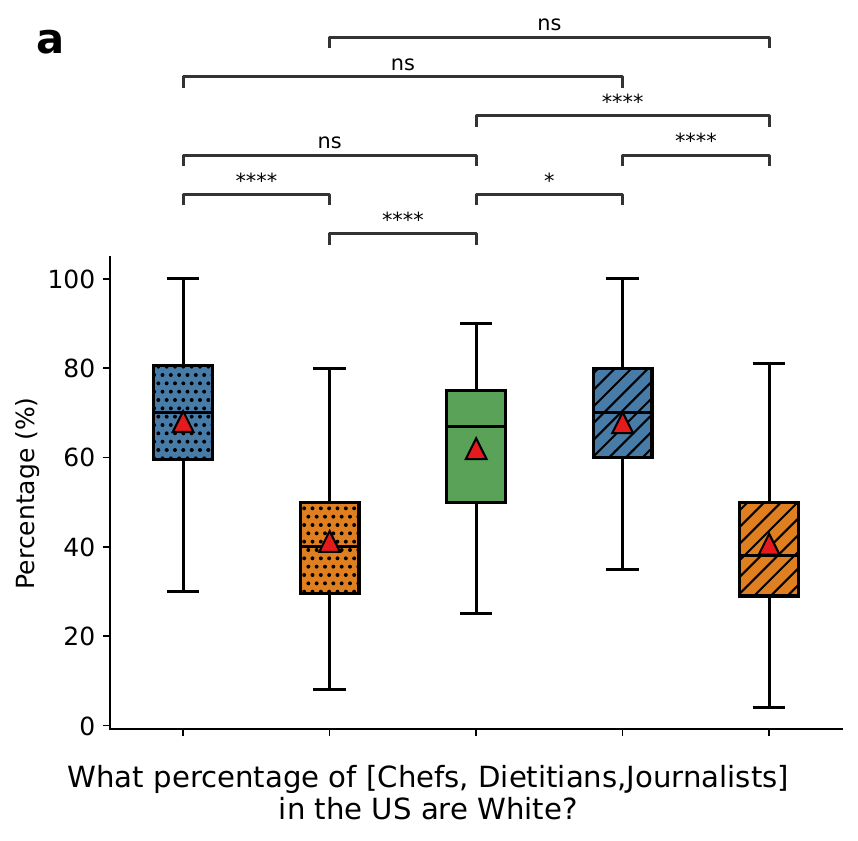}\quad
\includegraphics[width=0.48\textwidth]{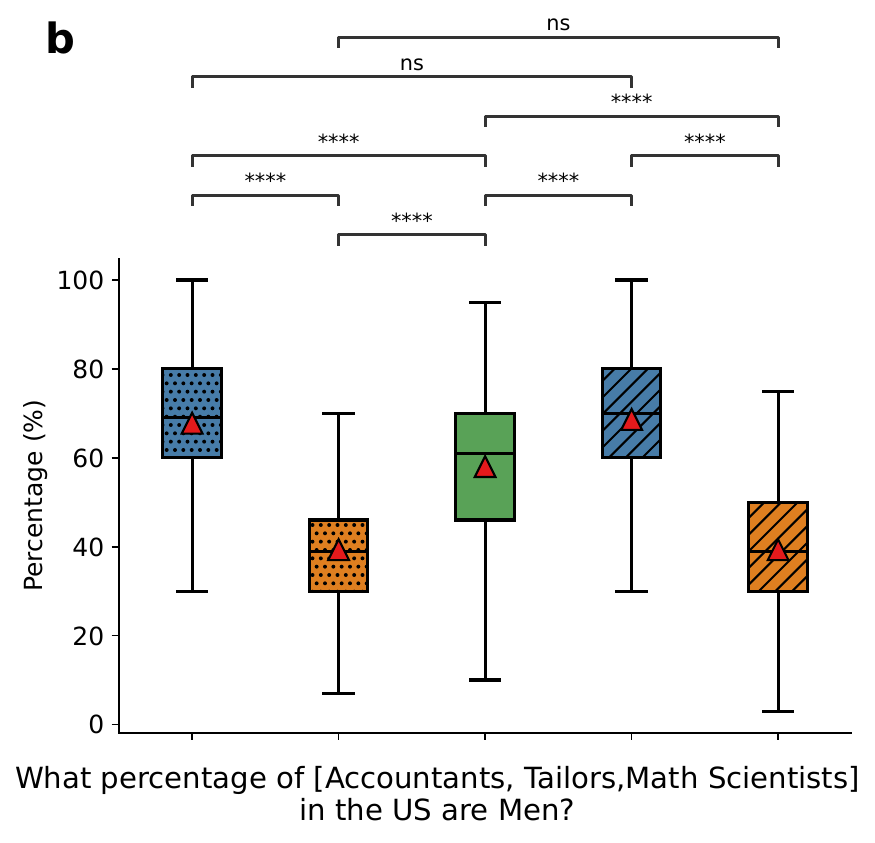}
\includegraphics[width=0.48\textwidth]{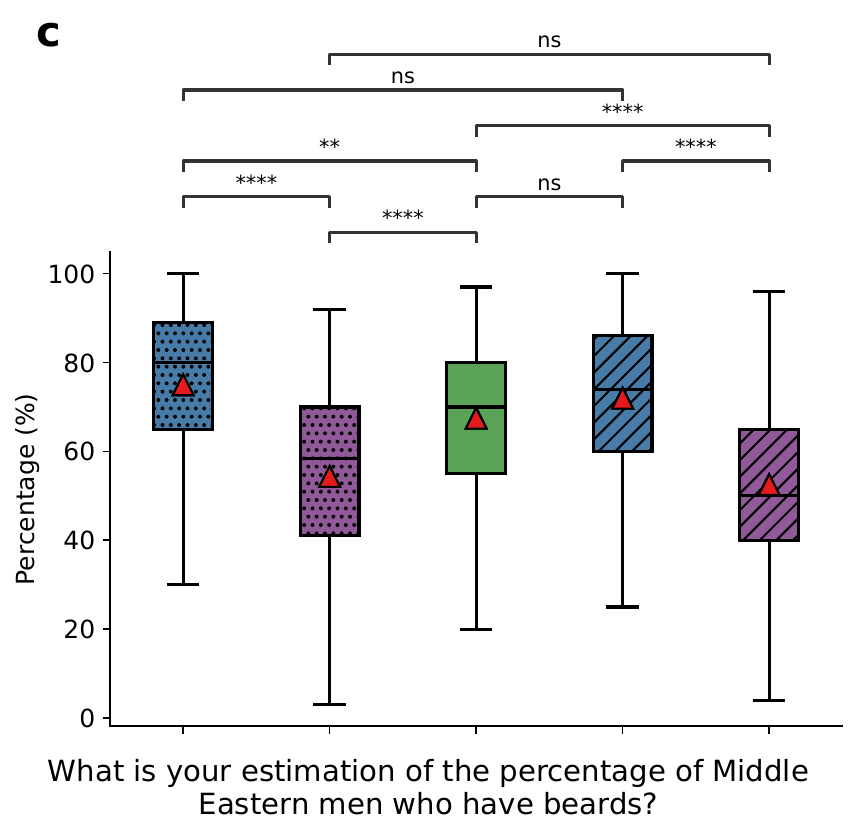}\quad\
\includegraphics[width=0.48\textwidth]{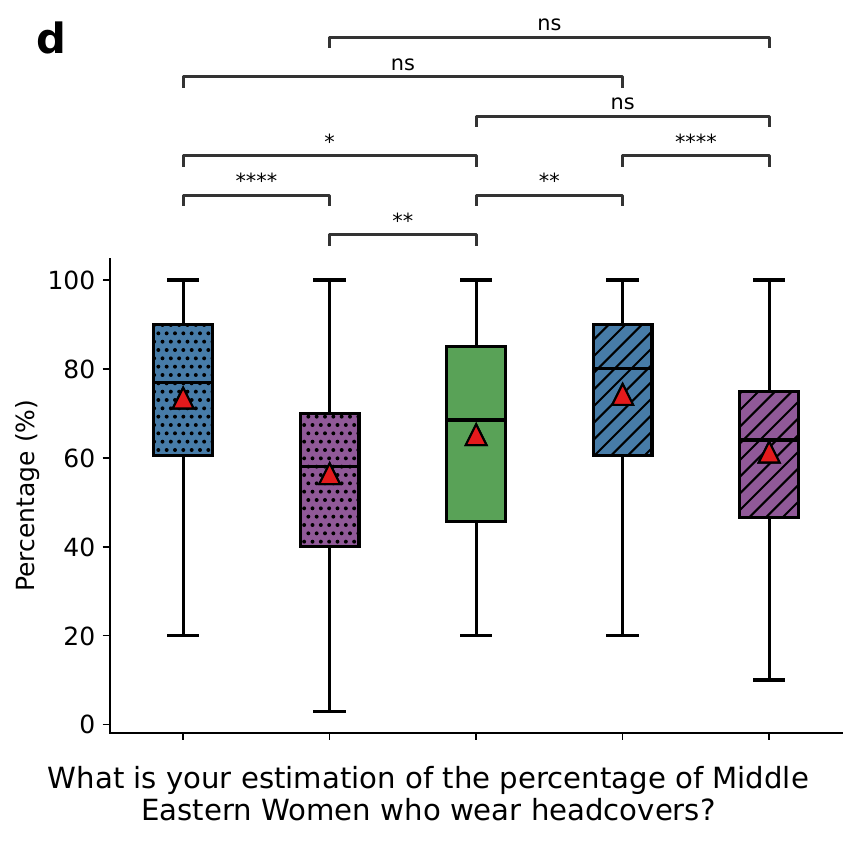}
\vspace{5pt}
\includegraphics[width=0.9\textwidth, fbox=0.5pt 4pt,center]{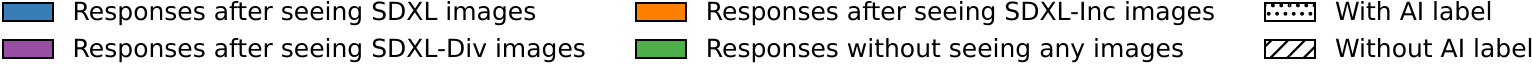}
\vspace{5pt}
\caption{
\textbf{Survey experiment results.} Subfigures \textbf{a} to \textbf{d} summarize the participants' responses in Studies 1 to 4, respectively. Boxes extend from the lower to upper quartile values, with a horizontal line at the median; whiskers extend to the most extreme values no further than 1.5 times the interquartile range from the box. P values are calculated using the t-test, unless one of the groups does not pass the Shapiro–Wilk test, in which case P values are calculated using the Mann-Whitney U test. $^{*}$p$<$0.05; $^{**}$p$<$0.01; $^{***}$p$<$0.001; $^{****}$p$<$0.0001; $ns = $ not significant).
}
\label{fig:prompts_comp}
\end{figure}

Additionally, for each question $Q_i:i\in\{1,2,3,4\}$, we recruited 135 participants from Prolific to answer $Q_i$ without being presented with the six AI-generated images. Figures~\ref{fig:prompts_comp}a to \ref{fig:prompts_comp}d summarize the responses to questions $Q1$ to $Q4$, respectively. For all four questions, exposure to images generated by SDXL-Inc reduces bias compared to the baseline in which participants are not exposed to AI-generated images. In contrast, exposure to SDXL-generated images increases bias (compared to the baseline) for all questions apart from $Q_1$ where the effect is not significant. Next, we examine the effect of the AI label. As can be seen in the figure, there is no significant difference in participants' responses when the images are labeled as being produced by an artist rather than being AI-generated. This holds for both SDXL and SDXL-Inclusive. Thus, we find no evidence that the AI label plays a role in the aforementioned phenomenon, indicating that the observed effect is entirely driven by the image's content, irrespective of its source.

\section*{Discussion}

We set out to examine the stereotypes and biases in Stable Diffusion XL (SDXL), a text-to-image generator used daily by millions worldwide~\cite{bloomberg}. To this end, we developed a classifier to predict the race and gender of any given face image, and demonstrated that it achieves state-of-the-art performance. Using this classifier, we showed that the vast majority of faces generated by SDXL are White males (as previously shown by Ghosh and Caliskan~\cite{ghosh2023person}), while certain racial groups are rarely generated, e.g., only 3\% are Asian and only 5\% are Indian. Biases were also found when considering various attributes, e.g., associating beauty with femininity and intelligence with masculinity. Some of these biases are less severe in the dataset on which Stable Diffusion was trained, suggesting that certain biases are further exacerbated by the model itself.

Biased text-to-image models may contribute to the normalization of gender stereotypes, potentially shaping societal attitudes towards the roles and capabilities of women in various professions. For instance, as we have demonstrated, Stable Diffusion mostly depicts secretaries and nurses as women while depicting managers, doctors, and professors as men; similar gender stereotypes were reported in the literature~\cite{friedrich2023fair,wang2023t2iat}. 
Bearing in mind that millions of people worldwide are already using such models daily, addressing gender stereotypes in these models can be crucial. As suggested by Study~2, such stereotypes can be reduced by an inclusive model and can be exacerbated by a non-inclusive one, indicating the potential of AI in alleviating gender inequality.

Biased representations of gender and race may contribute to the creation of content that not only misrepresents certain groups but may also perpetuate discriminatory practices. This could be detrimental in the context of advertising, marketing, and media campaigns, where visuals hold substantial influence. For example, as our analysis has shown, Stable Diffusion associates low-income jobs such as Cleaner, Janitor, and Security Guard with Black people, while associating higher-prestige jobs such as Doctor, Lawyer, and Professor with White people. These findings reflect what has been reported by Bianchi et al.~\cite{bianchi2023easily}, who showed that several of the most prestigious, high-paying professions are represented by Stable Diffusion (v1-4) as White. 
This is consistent with the historical and ongoing processes of occupational segregation based on race and gender, which could be linked to the concept of stereotype threat~\cite{reskin1998realities}, where awareness of negative stereotypes can negatively impact the performance of individuals from stereotyped groups~\cite{steele1995stereotype}.
Another example of bias is the association between crime and Black people, as well as the assassination between terrorism and Middle Easterners, which may reinforce existing biases against these racial groups~\cite{kundnani2014muslims, quillian2001black}. As we have seen in Study~1, whether or not the model is inclusive can affect people's perception of the racial distribution of certain professions, and this effect is likely to grow more pronounced as the use of AI-generated images becomes more widespread.

Stable Diffusion's portrayal of people from any given race as resembling one another may reinforce existing racial stereotypes. For instance, as we have demonstrated, Stable Diffusion predominently depicts Middle Eastern men as bearded and brown-skinned, while depicting Middle Eastern women as mostly brown-skinned and wearing traditional attire. 
This echoes the concept of Orientalism~\cite{said1977orientalism}, which critiques the simplified portrayals of the East, particularly the Middle East, in Western representations.
Such oversimplified and generalized depictions of a particular racial group can be culturally insensitive, and may misrepresent the true diversity within that group. They may also lead to feelings of alienation, low self-esteem, and a sense of being misunderstood. 
This can be understood in the context of social comparison theory~\cite{festinger1954theory}, where individuals compare themselves to others to evaluate their own abilities and opinions. When exposed to limited and stereotypical representations, it can negatively affect their self-esteem and sense of belonging.
As we have demonstrated in Studies 3 and 4, racial homogenization can be reduced using inclusive models, and can be exacerbated using non-inclusive ones.

\subsection*{Related Work}

Compared to all the works discussed in this section, ours is the only one that examines, and addresses, the problem of racial homogenization in AI-generated faces, i.e., the depiction of individuals from a specific racial or ethnic group as too similar in appearance. We are also the first to conduct a randomized control trial to understand the impact of being exposed to inclusive and non-inclusive AI-generated faces, and whether the AI-label plays a role in this phenomenon. Next, we summarize the relevant papers, and discuss any additional differences that may exist between our work and theirs.

Bianchi et al.~\cite{bianchi2023easily} used a previous version of Stable Diffusion (v1-4) to examine biases across ten professions and three races, but did not propose a debiasing solution. They classified the race and gender of any generated image as follows. First, for each of the five demographic categories considered (i.e., Male, Female, White, Black, Asian), they took the images that represent this category in the Chicago Face Dataset~\cite{ma2015chicago}, and fed those images to CLIP (the core representational component of Stable Diffusion) to generate vector representations. These were then averaged to obtain a single archetypal vector representation of the category (e.g., a single vector for Black). Any image can then be classified as $X$ if its vector representation is most closely aligned (in cosine distance) to the archetypal vector representation of $X$ (e.g., Black) compared to the alternative categories (e.g., Asian or White). The authors used this classifier to analyze 100 images per occupation, and found professional stereotypes that reinforce racial and gender disparities. Compared to our work, the authors did not consider certain racial groups, such as Indian, Latinx, and Middle Eastern. They also did not propose debiasing solutions.

Ghosh and Caliskan~\cite{ghosh2023person} used a previous version of Stable Diffusion (v2.1) to analyze biases across three genders and 27 nationalities, but did not consider professions nor proposed debiasing solutions. Using CLIP's cosine similarity, they compared images generated using the prompt: \textit{`a front-facing photo of a person'} against those generated using the prompt: \textit{`a front-facing photo of $X$'} where $X$ is either a gender (i.e., a man, a woman, a nonbinary gender) or a nationality (e.g., a person from Brazil). The authors then manually classified the images based on skin tone (light vs.\ dark) and found that most generated images were light-skinned. Compared to our work, the authors did not consider professions (e.g., Doctor, Nurse, etc.) nor attributes (e.g., Beautiful, Intelligent, etc.) in their analysis. Moreover, in the context of race, the authors manually classified skin-color rather than proposing a classifier, and only provided qualitative rather than quantitative conclusions. Finally, the authors did not propose debiasing solutions.

Wang et al.~\cite{wang2023t2iat} studied the association of pleasant and unpleasant attributes with certain concepts such as: European American and African American names, Light Skin and Dark Skin, Straight and Gay. In addition, they studied gender stereotypes across eight professions. Compared to our work, the authors did not propose any debiasing solutions. They also did not consider racial groups apart from European American (White) and African American (Black).

Friedrich et al.~\cite{friedrich2023fair} proposed a ``user in control'' solution, called Fair Diffusion, which utilizes a textual interface allowing users to instruct generative image models on fairness. Intuitively, this is made possible by extending classifier-free guidance~\cite{ho2022classifier} with an additional fair guidance term, which depends on additional fairness instructions provided by the user. This approach requires no data filtering nor additional training. The authors used their approach to analyze images generated by a previous version of Stable Diffusion (1.5), focusing on professional stereotypes across genders. Compared to our work, the authors did not consider racial stereotypes, and did not consider a fully automated solution.

Zhang et al.~\cite{zhang2023iti} proposed a solution called ITI-GEN (inclusive Text-to-Image Generation), designed to generate images that are uniformly distributed across attributes of interest. One such attribute could be, e.g., ``gender'', in which case the images generated by ITI-GEN would be equally split between the various attribute categories, e.g., ``male'' and ``female''. More specifically, given an input prompt and some desired attribute(s), the model learns discriminative token embeddings representing each category of each attribute. It then injects these learned tokens after the original prompt, thereby synthesizing a set of prompts, each representing a unique combination of categories belonging to different attributes. Finally, this set is used to generate an equal number of images for any category combination. The authors used their model to analyze a previous version of Stable Diffusion (v1-4), focusing on 40 physical attributes (each consisting of two categories, one positive and one negative), along with gender, skin tone, and age.

Compared to our work, Zhang et al.\ focused on skin tone rather than race. Thus, unlike our analysis, theirs does not distinguish between, say, Asian and White individuals who happen to be equally light-skinned, or between Indian and Latinx individuals who happen to have similar skin tones. The authors also focus on physical attributes (e.g., Black hair, Mustache, etc.) as opposed to those describing various characteristics like the ones used in our analysis (e.g., Criminal, Intelligent, Parent, Poor, etc.). Additionally, their analysis of professional stereotypes focuses on four professions (Professor, Doctor, Worker, Firefighter) and 200 images per profession, while we focus on 32 professions and 10,000 images per profession. Finally, as discussed in Supplementary Note~2, their solution is unable to debias complex prompts, such as ``\textit{a person with green hair and eyeglasses}'', or ``\textit{a person with the Eiffel tower}''. This difference in performance could (at least partially) be explained by the fact that their solution has a higher error rate when it comes to generating images of a certain race (see Supplementary Table~9).

\subsection*{Data Overview}
This section describes the datasets used in our study.

\textbf{I) The LAION-5B dataset:} This is the dataset that was originally used to train Stable Diffusion~\cite{schuhmann2022laion}. We utilized a subset of this dataset, consisting of high-resolution images~\cite{liaonHR}, to determine whether the biases in Stable Diffusion XL can be entirely attributed to the data on which it was trained. We randomly selected 172,923 images from this subset, and kept those having one or more of the following keywords: face, person, child, woman, or man. We then cropped the images to retain only the face(s) appearing therein, and discarded any resulting face images that are smaller than $100\times100$ pixels. This filtering process left us with a final set of 88,714 face images.

\textbf{II) The FairFace dataset:} This is one of the largest publicly-available datasets of face images. For each image, the dataset specifies the race (Black, East Asian, Indian, Latinx, Middle Eastern, Southeast Asian, and White), and the gender (female, male)~\cite{2021fairface}. Moreover, the dataset is readily divided into two sets: 86,744 images for training and 10,954 for validation. We combined East Asian and Southeast Asian into a single category: Asian. The resulting dataset was then used to train and validate our race and gender classifiers. For race, the number of images used for validation was: 1556 for Black; 2965 for Asian; 1516 for Indian; 1623 for Latinx or Hispanic; 1209 for Middle Eastern; and 2085 for White. As for gender, the number of images used for validation was: 5162 for female; and 5792 for male. 

\textbf{III) The Flickr-Faces-HQ dataset:} This dataset consists of 70,000 high-resolution ($1024\times1024$) images of human faces crawled from Flickr with considerable variation in terms of age, race, and image background~\cite{ffhq}. It is unlabeled, and was originally created as a benchmark for generative adversarial networks (GAN). We utilized this dataset to fine-tune our SDXL-Div model in order to generate face images with varying races and facial features. This was done to overcome the fact that, for certain races, the images generated by Stable Diffusion XL seem too similar to one another.

\textbf{IV)	The Stable Diffusion validation dataset:} This dataset consists of images that we have generated using Stable Diffusion XL. In particular, the dataset consists of 10,000 images per race and 10,000 images per gender. For race, the prompt used to generate the images was: ``\textit{a photo of a $X$}'', where $X\in$ \{Asian, Black, Indian, Latino or Hispanic, Middle Eastern, White\}. As for gender, the prompt used was: ``\textit{a photo of a $X$}'', where $X\in$ \{female, male\}. This dataset was used to validate our race and gender classifiers. 

\textbf{V) The SDXL-Inc fine-tuning dataset:} This dataset consists of Stable Diffusion-generated images with varying race, gender, and profession. The images were generated using the prompt: ``\textit{a photo of a $X$ $Y$ $Z$, looking at the camera, closeup headshot facing forward, ultra quality, sharp  focus}", where $X\in$ \{Asian, Black, Indian, Latino or Hispanic, Middle Eastern, White\}; $Y\in$ \{female, male\}; and $Z$ is one of the 21 professions listed in Supplementary Table~1 under the ``fine-tuning'' category. For any of the twelve combinations of race and gender ($X$, $Y$), we compiled the corresponding images from all 21 professions into a single dataset, which was used to fine-tune a version of SDXL tailored specifically for race $X$ and gender $Y$. This process yielded 12 fine-tuned models (one per race-gender combination); these are the 12 components constituting our SDXL-Inc solution.

\textbf{VI) The profession dataset:} This dataset consists of Stable Diffusion-generated images depicting 32 professions. More specifically, we generated 10,000 images per profession using the prompt: ``\textit{a photo of $Z$, looking at the camera, closeup headshot facing forward, ultra quality, sharp  focus}'', where $Z$ is one of the 32 professions listed in Supplementary Table~1. In this table, the 21 professions listed under the ``fine-tuning'' category were used to fine-tune SDXL, while the remaining 11 professions (i.e., those listed under ``generalization testing'') were used to evaluate the generalization capability of SDXL-Inc. 

\textbf{VII) The attribute dataset:} This dataset consists of Stable Diffusion-generated images depicting eight attributes. In particular, we generated 10,000 images per attribute using the prompt: ``\textit{a photo of a $X$, looking at the camera, closeup headshot facing forward, ultra quality, sharp  focus}'', where $X\in$ \{Poor, Winner, Beautiful, Intelligent, Parent, Sibling, Terrorist, Criminal\}. This dataset was used to evaluate the generalization capability of SDXL-Inc.

\subsection*{Methods}

This section presents the tools and techniques used in this work, including Stable Diffusion, face detection, face embedding generation, classification models, as well as the use of GPT-in-the-loop.

\subsubsection*{Stable Diffusion}

Stable Diffusion is a text-to-image generative model used to generate images based on prompts fed as input. A recent state-of-the-art version, called Stable Diffusion XL (SDXL), has a second text encoder and 3x larger UNet backbone, which increases the number of parameters~\cite{podell2023sdxl}. Additionally, SDXL has multiple novel conditioning schemes, as well as a refinement model that improves the visual fidelity of generated images. In our work, we utilized SDXL due to its significantly improved ability to generate human faces compared to its predecessors. We used the Hugging Face repository ``stabilityai'' and the model ``stable-diffusion-xl-base-1.0``~\cite{SDXL-base-1.0} to generate images to analyze racial and gender stereotypes in the context of various professions and attributes. 

We fine-tuned SDXL using LORA (Low Rank Adaptation)~\cite{hu2021lora} in the process of creating our debiasing solution. This process, which was carried out using a V100 GPU, yielded our SDXL-Inc model---a debiasing solution to overcome bias in race and gender representation. To fine-tune SDXL, we created a dataset consisting of prompt-image pairs. Each such pair consisted of an image taken from the ``SDXL-Inc fine-tuning dataset'' (described above) along with the prompt: ``\textit{a photo of $Z$, looking at the camera, closeup headshot facing forward, ultra quality, sharp  focus}'', where $Z$ is the profession depicted in the image.

Similarly, we fine-tuned SDXL using LORA in the process of creating SDXL-Div---our solution for mitigating racial and gender homogenization in SDXL. To this end, we created a dataset consisting of prompt-image pairs. Each such pair consisted of an image taken from the ``Flickr-Faces-HQ dataset'' (described above) along with the prompt: ``\textit{a photo of $X$ person, looking at the camera, closeup headshot facing forward, ultra quality, sharp focus}'', where $X$ is the race depicted in the image. This process yielded the SDXL-Div model.

\subsubsection*{Hyper-parameters}
Whenever an image is generated in our analysis, the following hyper-parameters were used: Image resolution = 1024; number of inference steps = 40; and guidance scale = 5. On the other hand, when fine-tuning SDXL, the following hyper-parameters were used: Image resolution = 1024; training batch size = 1; number of training epochs = 3; learning rate = $10^{-4}$; and mixed precision = fp16.

\subsubsection*{Proposed classification pipeline}

Our proposed classifier includes three stages: face detection, face embedding generation, and classification. 

\textbf{I) Face detection:} This stage is carried out using a Multi-task Cascaded Convolutional Neural Network (MTCNN), which is a deep cascaded multitask framework that utilizes the inherent correlation between detection and alignment to improve performance~\cite{7553523}. It leverages a cascaded architecture with three stages of deep convolutional networks to predict faces. MTCNN outperformed other methods across several challenging benchmarks. We selected this face detector in our work due to its ability to balance high detection accuracy and run-time speed. We configured the detector to exclude boundary-boxes with confidence scores $\leq 0.9$.

\textbf{II) Face embedding generation:} This stage is carried out using a VGGFace ResNet-50 Convolutional Neural Networks (VGGFace ResNet-50 CNN)~\cite{cao2018vggface2}. ResNet-50 CNN was trained on MS-Celeb-1M and VGGFace2, as well as the of the two. More specifically, MS-Celeb-1M has 10 million images depicting 100k different celebrities~\cite{guo2016ms}. On the other hand, VGGFace is a large-scale face dataset with considerable variations in pose, age, illumination, race, and profession~\cite{cao2018vggface2}, including 3.31 million images downloaded from Google image search. In our work, we utilized VGGFace ResNet-50 CNN, and removed the top layers to extract the embedding vector from the face images in the FairFace dataset described earlier. The FairFace RGB images were resized to 224x224 pixels before being fed to the CNN as input.

\textbf{III) Classification:} This stage is carried out using a Support Vector Machine (SVM)~\cite{708428}. More specifically, we trained two SVM classifiers to predict the race and gender using the embedding vectors extracted from the FairFace images during the previous stage. The hyper-parameters used in SVM are as follows: Regularization parameter C = 1; Kernel type = Radial Basis Function. This stage is repeated twice; once for race, and once for gender. However, the previous two stages are executed only once.

\subsubsection*{GPT-in-the-loop}

In this section, we describe our second debiasing solution, which uses GPT-4 ``in-the-loop''. The basic idea behind this prompt-regulating technique is to introduce an intermediary layer between the user (who provides the prompt) and Stable Diffusion XL. This layer uses GPT-4 to detect whether the user-provided prompt corresponds to the generation of an image depicting a person without a specific race and/or gender. GPT-4 would then inject a randomly-selected race (if race was not specified by the user) and/or a random gender (if gender was not specified) into the user-provided prompt before passing it on to Stable Diffusion XL. The exact prompt used with GPT is as follows, where $X$ is the user-provided prompt:
\begin{itemize}
\item[]``\textit{For this text $X$: 1) select using one word ['yes','no'] if text includes any profession or a social media influencer 2) find the subject practicing the job. 3) select using one word ['yes','no'] if the text includes any country, nationality, or race or ethnicity. 4) select using one word ['female', 'male', 'unknown'] the subject's gender}''
\end{itemize}

\noindent While the above prompt is meant to debias images related to professions, the same idea can be applied to other domains, e.g., by replacing ``profession or a social media influencer'' with ``person''.

\subsubsection*{Retraining ITI-GEN}
We needed to benchmark SDXL-Inc against ITI-GEN~\cite{zhang2023iti, ITI_GEN_github} in terms of how well it can debias Stable Diffusion. To this end, we retrained ITI-GEN with the six races (Asian, Black, Indian, Latino or Hispanic, Middle Eastern, and White) and the two genders (female and male) that SDXL-Inc was trained on. The training data used for this purpose was created as follows: First, we inferred the race and gender of each image in the Flickr-Faces-HQ dataset~\cite{ffhq} using our classifier. Then, we curated two training-sets of images: one for gender and one for race. More specifically, for the gender training set, we randomly selected 50 images (from the aforementioned labeled Flickr dataset) for each gender. On the other hand, for the race training set, we randomly selected 25 images for each race.

To validate our resultant ITI-GEN model, we generated 1,200 images (100 per gender-race combination) using the prompt ``\textit{a photo of a person}''. Supplementary Figure~1 depicts the race and gender distribution across the 1,200 generated images. The results demonstrate that our trained version of ITI-GEN is capable of generating equal representation for gender, and nearly equal representation for race (apart from Latinx, due to their facial similarity to both White and Middle Eastern).

\section*{Data Availability}
All of the data used in our analysis can be found at the following repository: \url{https://github.com/comnetsAD/AI-generated-faces}.

\section*{Author Contributions}
N.A., T.R. and Y.Z. conceived the study, designed the research, produced the visualizations, and wrote the manuscript; N.A. performed the literature review, designed and implemented the classifiers (race and gender) and debiasing solutions (SDXL-Inc and SDXL-Div), collected and analyzed the data, and ran the experiments. T.R. and Y.Z. designed and ran the survey experiment and analyzed the results.

\section*{Competing Interests}
The authors declare no competing interests.

\bibliography{paper}


\begin{thebibliography}{51}
\ifx \bisbn   \undefined \def \bisbn  #1{ISBN #1}\fi
\ifx \binits  \undefined \def \binits#1{#1}\fi
\ifx \bauthor  \undefined \def \bauthor#1{#1}\fi
\ifx \batitle  \undefined \def \batitle#1{#1}\fi
\ifx \bjtitle  \undefined \def \bjtitle#1{#1}\fi
\ifx \bvolume  \undefined \def \bvolume#1{\textbf{#1}}\fi
\ifx \byear  \undefined \def \byear#1{#1}\fi
\ifx \bissue  \undefined \def \bissue#1{#1}\fi
\ifx \bfpage  \undefined \def \bfpage#1{#1}\fi
\ifx \blpage  \undefined \def \blpage #1{#1}\fi
\ifx \burl  \undefined \def \burl#1{\textsf{#1}}\fi
\ifx \doiurl  \undefined \def \doiurl#1{\url{https://doi.org/#1}}\fi
\ifx \betal  \undefined \def \betal{\textit{et al.}}\fi
\ifx \binstitute  \undefined \def \binstitute#1{#1}\fi
\ifx \binstitutionaled  \undefined \def \binstitutionaled#1{#1}\fi
\ifx \bctitle  \undefined \def \bctitle#1{#1}\fi
\ifx \beditor  \undefined \def \beditor#1{#1}\fi
\ifx \bpublisher  \undefined \def \bpublisher#1{#1}\fi
\ifx \bbtitle  \undefined \def \bbtitle#1{#1}\fi
\ifx \bedition  \undefined \def \bedition#1{#1}\fi
\ifx \bseriesno  \undefined \def \bseriesno#1{#1}\fi
\ifx \blocation  \undefined \def \blocation#1{#1}\fi
\ifx \bsertitle  \undefined \def \bsertitle#1{#1}\fi
\ifx \bsnm \undefined \def \bsnm#1{#1}\fi
\ifx \bsuffix \undefined \def \bsuffix#1{#1}\fi
\ifx \bparticle \undefined \def \bparticle#1{#1}\fi
\ifx \barticle \undefined \def \barticle#1{#1}\fi
\bibcommenthead
\ifx \bconfdate \undefined \def \bconfdate #1{#1}\fi
\ifx \botherref \undefined \def \botherref #1{#1}\fi
\ifx \url \undefined \def \url#1{\textsf{#1}}\fi
\ifx \bchapter \undefined \def \bchapter#1{#1}\fi
\ifx \bbook \undefined \def \bbook#1{#1}\fi
\ifx \bcomment \undefined \def \bcomment#1{#1}\fi
\ifx \oauthor \undefined \def \oauthor#1{#1}\fi
\ifx \citeauthoryear \undefined \def \citeauthoryear#1{#1}\fi
\ifx \endbibitem  \undefined \def \endbibitem {}\fi
\ifx \bconflocation  \undefined \def \bconflocation#1{#1}\fi
\ifx \arxivurl  \undefined \def \arxivurl#1{\textsf{#1}}\fi
\csname PreBibitemsHook\endcsname

\bibitem[\protect\citeauthoryear{Christian}{2020}]{christian2020alignment}
\begin{bbook}
\bauthor{\bsnm{Christian}, \binits{B.}}:
\bbtitle{The Alignment Problem: Machine Learning and Human Values}.
\bpublisher{WW Norton \& Company},
\blocation{$ $}
(\byear{2020})
\end{bbook}
\endbibitem

\bibitem[\protect\citeauthoryear{Angwin et~al.}{2016}]{COMPAS}
\begin{botherref}
\oauthor{\bsnm{Angwin}, \binits{J.}},
\oauthor{\bsnm{Kirchner}, \binits{L.}},
\oauthor{\bsnm{Larson}, \binits{J.}},
\oauthor{\bsnm{Mattu}, \binits{S.}}:
Machine bias: There’s software used across the country to predict future criminals. and it’s biased against blacks.
ProPublica
(2016)
\end{botherref}
\endbibitem

\bibitem[\protect\citeauthoryear{Dressel and Farid}{2018}]{dressel2018accuracy}
\begin{barticle}
\bauthor{\bsnm{Dressel}, \binits{J.}},
\bauthor{\bsnm{Farid}, \binits{H.}}:
\batitle{The accuracy, fairness, and limits of predicting recidivism}.
\bjtitle{Science advances}
\bvolume{4}(\bissue{1}),
\bfpage{5580}
(\byear{2018})
\end{barticle}
\endbibitem

\bibitem[\protect\citeauthoryear{Buolamwini and Gebru}{2018}]{pmlr-v81-buolamwini18a}
\begin{bchapter}
\bauthor{\bsnm{Buolamwini}, \binits{J.}},
\bauthor{\bsnm{Gebru}, \binits{T.}}:
\bctitle{Gender shades: Intersectional accuracy disparities in commercial gender classification}.
In: \beditor{\bsnm{Friedler}, \binits{S.A.}},
\beditor{\bsnm{Wilson}, \binits{C.}} (eds.)
\bbtitle{Proceedings of the 1st Conference on Fairness, Accountability and Transparency}.
\bsertitle{Proceedings of Machine Learning Research},
vol. \bseriesno{81},
pp. \bfpage{77}--\blpage{91}.
\bpublisher{PMLR},
\blocation{$ $}
(\byear{2018}).
\burl{https://proceedings.mlr.press/v81/buolamwini18a.html}
\end{bchapter}
\endbibitem

\bibitem[\protect\citeauthoryear{Dastin}{2018}]{amazonAItool}
\begin{botherref}
\oauthor{\bsnm{Dastin}, \binits{J.}}:
Insight - amazon scraps secret ai recruiting tool that showed bias against women.
Reuters
(2018).
Accessed: November 29th, 2023
\end{botherref}
\endbibitem

\bibitem[\protect\citeauthoryear{Nadeem et~al.}{2021}]{nadeem2020stereoset}
\begin{bchapter}
\bauthor{\bsnm{Nadeem}, \binits{M.}},
\bauthor{\bsnm{Bethke}, \binits{A.}},
\bauthor{\bsnm{Reddy}, \binits{S.}}:
\bctitle{{S}tereo{S}et: Measuring stereotypical bias in pretrained language models}.
In: \beditor{\bsnm{Zong}, \binits{C.}},
\beditor{\bsnm{Xia}, \binits{F.}},
\beditor{\bsnm{Li}, \binits{W.}},
\beditor{\bsnm{Navigli}, \binits{R.}} (eds.)
\bbtitle{Proceedings of the 59th Annual Meeting of the Association for Computational Linguistics and the 11th International Joint Conference on Natural Language Processing (Volume 1: Long Papers)},
pp. \bfpage{5356}--\blpage{5371}.
\bpublisher{Association for Computational Linguistics},
\blocation{Online}
(\byear{2021}).
\doiurl{10.18653/v1/2021.acl-long.416} .
\burl{https://aclanthology.org/2021.acl-long.416}
\end{bchapter}
\endbibitem

\bibitem[\protect\citeauthoryear{Podell et~al.}{2023}]{podell2023sdxl}
\begin{botherref}
\oauthor{\bsnm{Podell}, \binits{D.}},
\oauthor{\bsnm{English}, \binits{Z.}},
\oauthor{\bsnm{Lacey}, \binits{K.}},
\oauthor{\bsnm{Blattmann}, \binits{A.}},
\oauthor{\bsnm{Dockhorn}, \binits{T.}},
\oauthor{\bsnm{M{\"u}ller}, \binits{J.}},
\oauthor{\bsnm{Penna}, \binits{J.}},
\oauthor{\bsnm{Rombach}, \binits{R.}}:
{SDXL: Improving Latent Diffusion Models for High-Resolution Image Synthesis}.
arXiv preprint arXiv:2307.01952
(2023)
\end{botherref}
\endbibitem

\bibitem[\protect\citeauthoryear{Fatunde and Tse}{2022}]{bloomberg}
\begin{botherref}
\oauthor{\bsnm{Fatunde}, \binits{M.}},
\oauthor{\bsnm{Tse}, \binits{C.}}:
Stability ai raises seed round at \$1 billion value.
Bloomberg
(2022).
Accessed: January 15th, 2024
\end{botherref}
\endbibitem

\bibitem[\protect\citeauthoryear{Bianchi et~al.}{2023}]{bianchi2023easily}
\begin{bchapter}
\bauthor{\bsnm{Bianchi}, \binits{F.}},
\bauthor{\bsnm{Kalluri}, \binits{P.}},
\bauthor{\bsnm{Durmus}, \binits{E.}},
\bauthor{\bsnm{Ladhak}, \binits{F.}},
\bauthor{\bsnm{Cheng}, \binits{M.}},
\bauthor{\bsnm{Nozza}, \binits{D.}},
\bauthor{\bsnm{Hashimoto}, \binits{T.}},
\bauthor{\bsnm{Jurafsky}, \binits{D.}},
\bauthor{\bsnm{Zou}, \binits{J.}},
\bauthor{\bsnm{Caliskan}, \binits{A.}}:
\bctitle{Easily accessible text-to-image generation amplifies demographic stereotypes at large scale}.
In: \bbtitle{Proceedings of the 2023 ACM Conference on Fairness, Accountability, and Transparency},
pp. \bfpage{1493}--\blpage{1504}
(\byear{2023})
\end{bchapter}
\endbibitem

\bibitem[\protect\citeauthoryear{Wang et~al.}{2023}]{wang2023t2iat}
\begin{botherref}
\oauthor{\bsnm{Wang}, \binits{J.}},
\oauthor{\bsnm{Liu}, \binits{X.G.}},
\oauthor{\bsnm{Di}, \binits{Z.}},
\oauthor{\bsnm{Liu}, \binits{Y.}},
\oauthor{\bsnm{Wang}, \binits{X.E.}}:
{T2IAT: Measuring Valence and Stereotypical Biases in Text-to-Image Generation}.
The 61st Annual Meeting of the Association for Computational Linguistics (ACL)
(2023)
\end{botherref}
\endbibitem

\bibitem[\protect\citeauthoryear{Ghosh and Caliskan}{2023}]{ghosh2023person}
\begin{botherref}
\oauthor{\bsnm{Ghosh}, \binits{S.}},
\oauthor{\bsnm{Caliskan}, \binits{A.}}:
{`Person'== Light-skinned, Western Man, and Sexualization of Women of Color: Stereotypes in Stable Diffusion}.
The 2023 Conference on Empirical Methods in Natural Language Processing (EMNLP)
(2023)
\end{botherref}
\endbibitem

\bibitem[\protect\citeauthoryear{Zhang et~al.}{2023}]{zhang2023iti}
\begin{bchapter}
\bauthor{\bsnm{Zhang}, \binits{C.}},
\bauthor{\bsnm{Chen}, \binits{X.}},
\bauthor{\bsnm{Chai}, \binits{S.}},
\bauthor{\bsnm{Wu}, \binits{C.H.}},
\bauthor{\bsnm{Lagun}, \binits{D.}},
\bauthor{\bsnm{Beeler}, \binits{T.}},
\bauthor{\bsnm{Torre}, \binits{F.}}:
\bctitle{Iti-gen: Inclusive text-to-image generation}.
In: \bbtitle{Proceedings of the IEEE/CVF International Conference on Computer Vision},
pp. \bfpage{3969}--\blpage{3980}
(\byear{2023})
\end{bchapter}
\endbibitem

\bibitem[\protect\citeauthoryear{Friedrich et~al.}{2023}]{friedrich2023fair}
\begin{botherref}
\oauthor{\bsnm{Friedrich}, \binits{F.}},
\oauthor{\bsnm{Schramowski}, \binits{P.}},
\oauthor{\bsnm{Brack}, \binits{M.}},
\oauthor{\bsnm{Struppek}, \binits{L.}},
\oauthor{\bsnm{Hintersdorf}, \binits{D.}},
\oauthor{\bsnm{Luccioni}, \binits{S.}},
\oauthor{\bsnm{Kersting}, \binits{K.}}:
{Fair diffusion: Instructing text-to-image generation models on fairness}.
arXiv preprint arXiv:2302.10893
(2023)
\end{botherref}
\endbibitem

\bibitem[\protect\citeauthoryear{Glick et~al.}{1995}]{glick1995images}
\begin{barticle}
\bauthor{\bsnm{Glick}, \binits{P.}},
\bauthor{\bsnm{Wilk}, \binits{K.}},
\bauthor{\bsnm{Perreault}, \binits{M.}}:
\batitle{Images of occupations: Components of gender and status in occupational stereotypes}.
\bjtitle{Sex roles}
\bvolume{32},
\bfpage{565}--\blpage{582}
(\byear{1995})
\end{barticle}
\endbibitem

\bibitem[\protect\citeauthoryear{Bertrand and Mullainathan}{2004}]{bertrand2004emily}
\begin{barticle}
\bauthor{\bsnm{Bertrand}, \binits{M.}},
\bauthor{\bsnm{Mullainathan}, \binits{S.}}:
\batitle{Are emily and greg more employable than lakisha and jamal? a field experiment on labor market discrimination}.
\bjtitle{American economic review}
\bvolume{94}(\bissue{4}),
\bfpage{991}--\blpage{1013}
(\byear{2004})
\end{barticle}
\endbibitem

\bibitem[\protect\citeauthoryear{Steele and Aronson}{1995}]{steele1995stereotype}
\begin{barticle}
\bauthor{\bsnm{Steele}, \binits{C.M.}},
\bauthor{\bsnm{Aronson}, \binits{J.}}:
\batitle{Stereotype threat and the intellectual test performance of african americans.}
\bjtitle{Journal of personality and social psychology}
\bvolume{69}(\bissue{5}),
\bfpage{797}
(\byear{1995})
\end{barticle}
\endbibitem

\bibitem[\protect\citeauthoryear{Greenwald and Banaji}{1995}]{greenwald1995implicit}
\begin{barticle}
\bauthor{\bsnm{Greenwald}, \binits{A.G.}},
\bauthor{\bsnm{Banaji}, \binits{M.R.}}:
\batitle{Implicit social cognition: attitudes, self-esteem, and stereotypes.}
\bjtitle{Psychological review}
\bvolume{102}(\bissue{1}),
\bfpage{4}
(\byear{1995})
\end{barticle}
\endbibitem

\bibitem[\protect\citeauthoryear{Said}{1977}]{said1977orientalism}
\begin{barticle}
\bauthor{\bsnm{Said}, \binits{E.W.}}:
\batitle{Orientalism}.
\bjtitle{The Georgia Review}
\bvolume{31}(\bissue{1}),
\bfpage{162}--\blpage{206}
(\byear{1977})
\end{barticle}
\endbibitem

\bibitem[\protect\citeauthoryear{Festinger}{1954}]{festinger1954theory}
\begin{barticle}
\bauthor{\bsnm{Festinger}, \binits{L.}}:
\batitle{A theory of social comparison processes}.
\bjtitle{Human relations}
\bvolume{7}(\bissue{2}),
\bfpage{117}--\blpage{140}
(\byear{1954})
\end{barticle}
\endbibitem

\bibitem[\protect\citeauthoryear{Pennington et~al.}{2016}]{pennington2016twenty}
\begin{barticle}
\bauthor{\bsnm{Pennington}, \binits{C.R.}},
\bauthor{\bsnm{Heim}, \binits{D.}},
\bauthor{\bsnm{Levy}, \binits{A.R.}},
\bauthor{\bsnm{Larkin}, \binits{D.T.}}:
\batitle{Twenty years of stereotype threat research: A review of psychological mediators}.
\bjtitle{PloS one}
\bvolume{11}(\bissue{1}),
\bfpage{0146487}
(\byear{2016})
\end{barticle}
\endbibitem

\bibitem[\protect\citeauthoryear{Karkkainen and Joo}{2021}]{2021fairface}
\begin{bchapter}
\bauthor{\bsnm{Karkkainen}, \binits{K.}},
\bauthor{\bsnm{Joo}, \binits{J.}}:
\bctitle{Fairface: Face attribute dataset for balanced race, gender, and age for bias measurement and mitigation}.
In: \bbtitle{Proceedings of the IEEE/CVF Winter Conference on Applications of Computer Vision},
pp. \bfpage{1548}--\blpage{1558}
(\byear{2021})
\end{bchapter}
\endbibitem

\bibitem[\protect\citeauthoryear{Radford et~al.}{2021}]{radford2021learning}
\begin{bchapter}
\bauthor{\bsnm{Radford}, \binits{A.}},
\bauthor{\bsnm{Kim}, \binits{J.W.}},
\bauthor{\bsnm{Hallacy}, \binits{C.}},
\bauthor{\bsnm{Ramesh}, \binits{A.}},
\bauthor{\bsnm{Goh}, \binits{G.}},
\bauthor{\bsnm{Agarwal}, \binits{S.}},
\bauthor{\bsnm{Sastry}, \binits{G.}},
\bauthor{\bsnm{Askell}, \binits{A.}},
\bauthor{\bsnm{Mishkin}, \binits{P.}},
\bauthor{\bsnm{Clark}, \binits{J.}}, \betal:
\bctitle{Learning transferable visual models from natural language supervision}.
In: \bbtitle{International Conference on Machine Learning},
pp. \bfpage{8748}--\blpage{8763}
(\byear{2021}).
\bcomment{PMLR}
\end{bchapter}
\endbibitem

\bibitem[\protect\citeauthoryear{Schroff et~al.}{2015}]{Schroff_2015_CVPR}
\begin{bchapter}
\bauthor{\bsnm{Schroff}, \binits{F.}},
\bauthor{\bsnm{Kalenichenko}, \binits{D.}},
\bauthor{\bsnm{Philbin}, \binits{J.}}:
\bctitle{Facenet: A unified embedding for face recognition and clustering}.
In: \bbtitle{Proceedings of the IEEE Conference on Computer Vision and Pattern Recognition (CVPR)}
(\byear{2015})
\end{bchapter}
\endbibitem

\bibitem[\protect\citeauthoryear{Hearst et~al.}{1998}]{708428}
\begin{barticle}
\bauthor{\bsnm{Hearst}, \binits{M.A.}},
\bauthor{\bsnm{Dumais}, \binits{S.T.}},
\bauthor{\bsnm{Osuna}, \binits{E.}},
\bauthor{\bsnm{Platt}, \binits{J.}},
\bauthor{\bsnm{Scholkopf}, \binits{B.}}:
\batitle{Support vector machines}.
\bjtitle{IEEE Intelligent Systems and their Applications}
\bvolume{13}(\bissue{4}),
\bfpage{18}--\blpage{28}
(\byear{1998})
\doiurl{10.1109/5254.708428}
\end{barticle}
\endbibitem

\bibitem[\protect\citeauthoryear{Tan and Le}{2019}]{pmlr-v97-tan19a}
\begin{bchapter}
\bauthor{\bsnm{Tan}, \binits{M.}},
\bauthor{\bsnm{Le}, \binits{Q.}}:
\bctitle{{E}fficient{N}et: Rethinking model scaling for convolutional neural networks}.
In: \beditor{\bsnm{Chaudhuri}, \binits{K.}},
\beditor{\bsnm{Salakhutdinov}, \binits{R.}} (eds.)
\bbtitle{Proceedings of the 36th International Conference on Machine Learning}.
\bsertitle{Proceedings of Machine Learning Research},
vol. \bseriesno{97},
pp. \bfpage{6105}--\blpage{6114}.
\bpublisher{PMLR},
\blocation{$ $}
(\byear{2019}).
\burl{https://proceedings.mlr.press/v97/tan19a.html}
\end{bchapter}
\endbibitem

\bibitem[\protect\citeauthoryear{Dosovitskiy et~al.}{2020}]{dosovitskiy2020image}
\begin{botherref}
\oauthor{\bsnm{Dosovitskiy}, \binits{A.}},
\oauthor{\bsnm{Beyer}, \binits{L.}},
\oauthor{\bsnm{Kolesnikov}, \binits{A.}},
\oauthor{\bsnm{Weissenborn}, \binits{D.}},
\oauthor{\bsnm{Zhai}, \binits{X.}},
\oauthor{\bsnm{Unterthiner}, \binits{T.}},
\oauthor{\bsnm{Dehghani}, \binits{M.}},
\oauthor{\bsnm{Minderer}, \binits{M.}},
\oauthor{\bsnm{Heigold}, \binits{G.}},
\oauthor{\bsnm{Gelly}, \binits{S.}}, et al.:
An image is worth 16x16 words: Transformers for image recognition at scale.
arXiv preprint arXiv:2010.11929
(2020)
\end{botherref}
\endbibitem

\bibitem[\protect\citeauthoryear{Schuhmann et~al.}{2022}]{schuhmann2022laion}
\begin{barticle}
\bauthor{\bsnm{Schuhmann}, \binits{C.}},
\bauthor{\bsnm{Beaumont}, \binits{R.}},
\bauthor{\bsnm{Vencu}, \binits{R.}},
\bauthor{\bsnm{Gordon}, \binits{C.}},
\bauthor{\bsnm{Wightman}, \binits{R.}},
\bauthor{\bsnm{Cherti}, \binits{M.}},
\bauthor{\bsnm{Coombes}, \binits{T.}},
\bauthor{\bsnm{Katta}, \binits{A.}},
\bauthor{\bsnm{Mullis}, \binits{C.}},
\bauthor{\bsnm{Wortsman}, \binits{M.}}, \betal:
\batitle{Laion-5b: An open large-scale dataset for training next generation image-text models}.
\bjtitle{Advances in Neural Information Processing Systems}
\bvolume{35},
\bfpage{25278}--\blpage{25294}
(\byear{2022})
\end{barticle}
\endbibitem

\bibitem[\protect\citeauthoryear{Beaumont}{2021}]{liaonHR}
\begin{botherref}
\oauthor{\bsnm{Beaumont}, \binits{R.}}:
img2dataset: Easily turn large sets of image urls to an image dataset.
GitHub
(2021)
\end{botherref}
\endbibitem

\bibitem[\protect\citeauthoryear{Hu et~al.}{2021}]{hu2021lora}
\begin{botherref}
\oauthor{\bsnm{Hu}, \binits{E.J.}},
\oauthor{\bsnm{Shen}, \binits{Y.}},
\oauthor{\bsnm{Wallis}, \binits{P.}},
\oauthor{\bsnm{Allen-Zhu}, \binits{Z.}},
\oauthor{\bsnm{Li}, \binits{Y.}},
\oauthor{\bsnm{Wang}, \binits{S.}},
\oauthor{\bsnm{Wang}, \binits{L.}},
\oauthor{\bsnm{Chen}, \binits{W.}}:
{LoRA: Low-Rank Adaptation of Large Language Models}.
arXiv preprint arXiv:2106.09685
(2021)
\end{botherref}
\endbibitem

\bibitem[\protect\citeauthoryear{Hofmann et~al.}{2024}]{hofmann2024ai}
\begin{botherref}
\oauthor{\bsnm{Hofmann}, \binits{V.}},
\oauthor{\bsnm{Kalluri}, \binits{P.R.}},
\oauthor{\bsnm{Jurafsky}, \binits{D.}},
\oauthor{\bsnm{King}, \binits{S.}}:
Ai generates covertly racist decisions about people based on their dialect.
Nature,
1--8
(2024)
\end{botherref}
\endbibitem

\bibitem[\protect\citeauthoryear{Pierce}{1996}]{pierce1996gender}
\begin{bbook}
\bauthor{\bsnm{Pierce}, \binits{J.L.}}:
\bbtitle{Gender Trials: Emotional Lives in Contemporary Law Firms}.
\bpublisher{Univ of California Press},
\blocation{$ $}
(\byear{1996})
\end{bbook}
\endbibitem

\bibitem[\protect\citeauthoryear{Organization et~al.}{2019}]{world2019delivered}
\begin{bbook}
\bauthor{\bsnm{Organization}, \binits{W.H.}}, \betal:
\bbtitle{Delivered by Women, Led by Men: A Gender and Equity Analysis of the Global Health and Social Workforce}.
\bpublisher{World Health Organization},
\blocation{$ $}
(\byear{2019})
\end{bbook}
\endbibitem

\bibitem[\protect\citeauthoryear{Diekman and Eagly}{2000}]{diekman2000stereotypes}
\begin{barticle}
\bauthor{\bsnm{Diekman}, \binits{A.B.}},
\bauthor{\bsnm{Eagly}, \binits{A.H.}}:
\batitle{Stereotypes as dynamic constructs: Women and men of the past, present, and future}.
\bjtitle{Personality and social psychology bulletin}
\bvolume{26}(\bissue{10}),
\bfpage{1171}--\blpage{1188}
(\byear{2000})
\end{barticle}
\endbibitem

\bibitem[\protect\citeauthoryear{Lindsey}{2020}]{lindsey2020gender}
\begin{bbook}
\bauthor{\bsnm{Lindsey}, \binits{L.L.}}:
\bbtitle{Gender: Sociological Perspectives}.
\bpublisher{Routledge},
\blocation{$ $}
(\byear{2020})
\end{bbook}
\endbibitem

\bibitem[\protect\citeauthoryear{Steinke}{2017}]{steinke2017adolescent}
\begin{barticle}
\bauthor{\bsnm{Steinke}, \binits{J.}}:
\batitle{Adolescent girls’ stem identity formation and media images of stem professionals: Considering the influence of contextual cues}.
\bjtitle{Frontiers in psychology}
\bvolume{8},
\bfpage{239856}
(\byear{2017})
\end{barticle}
\endbibitem

\bibitem[\protect\citeauthoryear{Davalos et~al.}{2007}]{davalos2007iii}
\begin{barticle}
\bauthor{\bsnm{Davalos}, \binits{D.B.}},
\bauthor{\bsnm{Davalos}, \binits{R.A.}},
\bauthor{\bsnm{Layton}, \binits{H.S.}}:
\batitle{Iii. content analysis of magazine headlines: Changes over three decades?}
\bjtitle{Feminism \& Psychology}
\bvolume{17}(\bissue{2}),
\bfpage{250}--\blpage{258}
(\byear{2007})
\end{barticle}
\endbibitem

\bibitem[\protect\citeauthoryear{Spitzer et~al.}{1999}]{spitzer1999gender}
\begin{barticle}
\bauthor{\bsnm{Spitzer}, \binits{B.L.}},
\bauthor{\bsnm{Henderson}, \binits{K.A.}},
\bauthor{\bsnm{Zivian}, \binits{M.T.}}:
\batitle{Gender differences in population versus media body sizes: A comparison over four decades}.
\bjtitle{Sex roles}
\bvolume{40}(\bissue{7}),
\bfpage{545}--\blpage{565}
(\byear{1999})
\end{barticle}
\endbibitem

\bibitem[\protect\citeauthoryear{Saperstein and Penner}{2012}]{saperstein2012racial}
\begin{barticle}
\bauthor{\bsnm{Saperstein}, \binits{A.}},
\bauthor{\bsnm{Penner}, \binits{A.M.}}:
\batitle{Racial fluidity and inequality in the united states}.
\bjtitle{American journal of sociology}
\bvolume{118}(\bissue{3}),
\bfpage{676}--\blpage{727}
(\byear{2012})
\end{barticle}
\endbibitem

\bibitem[\protect\citeauthoryear{Merolla and Zechmeister}{2019}]{merolla2019democracy}
\begin{bbook}
\bauthor{\bsnm{Merolla}, \binits{J.L.}},
\bauthor{\bsnm{Zechmeister}, \binits{E.J.}}:
\bbtitle{Democracy at Risk: How Terrorist Threats Affect the Public}.
\bpublisher{University of Chicago Press},
\blocation{$ $}
(\byear{2019})
\end{bbook}
\endbibitem

\bibitem[\protect\citeauthoryear{Kundnani}{2014}]{kundnani2014muslims}
\begin{bbook}
\bauthor{\bsnm{Kundnani}, \binits{A.}}:
\bbtitle{The Muslims Are Coming!: Islamophobia, Extremism, and the Domestic War on Terror}.
\bpublisher{Verso Books},
\blocation{$ $}
(\byear{2014})
\end{bbook}
\endbibitem

\bibitem[\protect\citeauthoryear{Quillian and Pager}{2001}]{quillian2001black}
\begin{barticle}
\bauthor{\bsnm{Quillian}, \binits{L.}},
\bauthor{\bsnm{Pager}, \binits{D.}}:
\batitle{Black neighbors, higher crime? the role of racial stereotypes in evaluations of neighborhood crime}.
\bjtitle{American journal of sociology}
\bvolume{107}(\bissue{3}),
\bfpage{717}--\blpage{767}
(\byear{2001})
\end{barticle}
\endbibitem

\bibitem[\protect\citeauthoryear{OpenAI}{2023}]{chatgpt}
\begin{botherref}
\oauthor{\bsnm{OpenAI}}:
{ChatGPT-Large language model}.
Online
(2023).
Accessed: January 15th, 2024
\end{botherref}
\endbibitem

\bibitem[\protect\citeauthoryear{Karras and Hellsten}{2023}]{ffhq}
\begin{botherref}
\oauthor{\bsnm{Karras}, \binits{T.}},
\oauthor{\bsnm{Hellsten}, \binits{J.}}:
Flickr-faces-hq dataset (ffhq).
Github
(2023)
\end{botherref}
\endbibitem

\bibitem[\protect\citeauthoryear{Reskin}{1998}]{reskin1998realities}
\begin{bbook}
\bauthor{\bsnm{Reskin}, \binits{B.F.}}:
\bbtitle{The Realities of Affirmative Action in Employment}.
\bpublisher{American Sociological Association},
\blocation{$ $}
(\byear{1998})
\end{bbook}
\endbibitem

\bibitem[\protect\citeauthoryear{Ma et~al.}{2015}]{ma2015chicago}
\begin{barticle}
\bauthor{\bsnm{Ma}, \binits{D.S.}},
\bauthor{\bsnm{Correll}, \binits{J.}},
\bauthor{\bsnm{Wittenbrink}, \binits{B.}}:
\batitle{The chicago face database: A free stimulus set of faces and norming data}.
\bjtitle{Behavior research methods}
\bvolume{47},
\bfpage{1122}--\blpage{1135}
(\byear{2015})
\end{barticle}
\endbibitem

\bibitem[\protect\citeauthoryear{Ho and Salimans}{2022}]{ho2022classifier}
\begin{botherref}
\oauthor{\bsnm{Ho}, \binits{J.}},
\oauthor{\bsnm{Salimans}, \binits{T.}}:
Classifier-free diffusion guidance.
arXiv preprint arXiv:2207.12598
(2022)
\end{botherref}
\endbibitem

\bibitem[\protect\citeauthoryear{stabilityai}{}]{SDXL-base-1.0}
\begin{botherref}
\oauthor{\bsnm{stabilityai}}:
stable-diffusion-xl-base-1.0.
\url{https://huggingface.co/stabilityai/stable-diffusion-xl-base-1.0}
\end{botherref}
\endbibitem

\bibitem[\protect\citeauthoryear{Zhang et~al.}{2016}]{7553523}
\begin{barticle}
\bauthor{\bsnm{Zhang}, \binits{K.}},
\bauthor{\bsnm{Zhang}, \binits{Z.}},
\bauthor{\bsnm{Li}, \binits{Z.}},
\bauthor{\bsnm{Qiao}, \binits{Y.}}:
\batitle{Joint face detection and alignment using multitask cascaded convolutional networks}.
\bjtitle{IEEE Signal Processing Letters}
\bvolume{23}(\bissue{10}),
\bfpage{1499}--\blpage{1503}
(\byear{2016})
\doiurl{10.1109/LSP.2016.2603342}
\end{barticle}
\endbibitem

\bibitem[\protect\citeauthoryear{Cao et~al.}{2018}]{cao2018vggface2}
\begin{bchapter}
\bauthor{\bsnm{Cao}, \binits{Q.}},
\bauthor{\bsnm{Shen}, \binits{L.}},
\bauthor{\bsnm{Xie}, \binits{W.}},
\bauthor{\bsnm{Parkhi}, \binits{O.M.}},
\bauthor{\bsnm{Zisserman}, \binits{A.}}:
\bctitle{Vggface2: A dataset for recognising faces across pose and age}.
In: \bbtitle{2018 13th IEEE International Conference on Automatic Face \& Gesture Recognition (FG'18)},
pp. \bfpage{67}--\blpage{74}
(\byear{2018})
\end{bchapter}
\endbibitem

\bibitem[\protect\citeauthoryear{Guo et~al.}{2016}]{guo2016ms}
\begin{bchapter}
\bauthor{\bsnm{Guo}, \binits{Y.}},
\bauthor{\bsnm{Zhang}, \binits{L.}},
\bauthor{\bsnm{Hu}, \binits{Y.}},
\bauthor{\bsnm{He}, \binits{X.}},
\bauthor{\bsnm{Gao}, \binits{J.}}:
\bctitle{Ms-celeb-1m: A dataset and benchmark for large-scale face recognition}.
In: \bbtitle{Computer Vision--ECCV 2016: 14th European Conference, Amsterdam, The Netherlands, October 11-14, 2016, Proceedings, Part III 14},
pp. \bfpage{87}--\blpage{102}
(\byear{2016}).
\bcomment{Springer}
\end{bchapter}
\endbibitem

\bibitem[\protect\citeauthoryear{Zhang et~al.}{2023}]{ITI_GEN_github}
\begin{botherref}
\oauthor{\bsnm{Zhang}, \binits{C.}},
\oauthor{\bsnm{Chen}, \binits{X.}},
\oauthor{\bsnm{Chai}, \binits{S.}},
\oauthor{\bsnm{Wu}, \binits{H.C.}},
\oauthor{\bsnm{Lagun}, \binits{D.}},
\oauthor{\bsnm{Beeler}, \binits{T.}},
\oauthor{\bsnm{Torre}, \binits{F.}}:
ITI-GEN: Inclusive Text-to-Image Generation.
GitHub
(2023)
\end{botherref}
\endbibitem

\end{thebibliography}


\begin{thebibliography}{1}
\expandafter\ifx\csname url\endcsname\relax
  \def\url#1{\texttt{#1}}\fi
\expandafter\ifx\csname urlprefix\endcsname\relax\def\urlprefix{URL }\fi
\providecommand{\bibinfo}[2]{#2}
\providecommand{\eprint}[2][]{\url{#2}}

\bibitem{podell2023sdxl}
\bibinfo{author}{Podell, D.} \emph{et~al.}
\newblock \bibinfo{title}{{SDXL: Improving Latent Diffusion Models for High-Resolution Image Synthesis}}.
\newblock \emph{\bibinfo{journal}{arXiv preprint arXiv:2307.01952}}  (\bibinfo{year}{2023}).

\bibitem{zhang2023iti}
\bibinfo{author}{Zhang, C.} \emph{et~al.}
\newblock \bibinfo{title}{Iti-gen: Inclusive text-to-image generation}.
\newblock In \emph{\bibinfo{booktitle}{Proceedings of the IEEE/CVF International Conference on Computer Vision}}, \bibinfo{pages}{3969--3980} (\bibinfo{year}{2023}).

\bibitem{ITI_GEN_github}
\bibinfo{author}{Zhang, C.} \emph{et~al.}
\newblock \bibinfo{title}{Iti-gen: Inclusive text-to-image generation}.
\newblock \bibinfo{howpublished}{\url{https://github.com/humansensinglab/ITI-GEN}} (\bibinfo{year}{2023}).

\end{thebibliography}

\end{document}


\begin{center}
{\fontsize{14}{14}\selectfont{Supplementary Information for}}
\end{center}

\setlength{\textwidth}{7.05in}
\setlength\oddsidemargin{-0.29in}
\setlength\evensidemargin{-0.29in}
\setlength{\textheight}{9.2in}
\setlength\topmargin{-0.2in}
\pdfpagewidth 8.5in \pdfpageheight 11in

\begin{center}
{\fontsize{16}{16}\selectfont{
AI-generated faces influence gender stereotypes and racial homogenization
}}
\end{center}
\smallskip\smallskip
\begin{center}
{\fontsize{12}{12}\selectfont{Nouar AlDahoul, Talal Rahwan$^\ast$, Yasir Zaki$^\ast$}}
\end{center}

\begin{center}
{\fontsize{10}{10}\selectfont{$^\ast$Corresponding authors. E-mail: talal.rahwan@nyu.edu, yasir.zaki@nyu.edu}}
\end{center}
\ \\\\
\ \\\\

This document is structured as follows:
\smallskip\smallskip
\begin{itemize}\itemsep0.5em
\item \textbf{Supplementary Figures} (page~\pageref{supplementary_figures})
\item \textbf{Supplementary Tables} (page~\pageref{supplementary_tables})
\item \textbf{Supplementary Note_1}
(page~\pageref{supplementary_note_1})
\item \textbf{Supplementary Note_2}
(page~\pageref{supplementary_note_2})
\item \textbf{Supplementary Note_3}
(page~\pageref{supplementary_note_3})
\end{itemize}

\clearpage
\section*{Supplementary Figures}
\label{supplementary_figures}

\begin{figure}[htbp]
\includegraphics[width=1\textwidth]{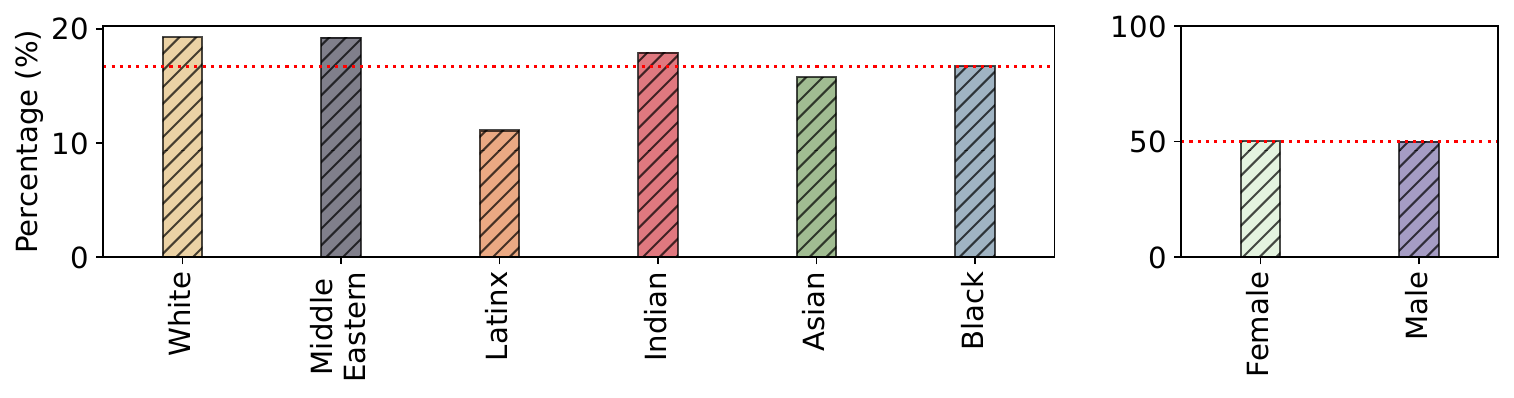}
\caption{
{\fontsize{11}{11}\selectfont{
\textbf{Race and Gender distribution obtained by the retrained ITI-GEN model.} 
Comparing the representation of different races and genders in a sample of 1200 images generated by the retrained ITI-GEN model using the prompt: \textit{``a photo of a person''}.}}}
\label{fig:ITI-GEN}
\end{figure}

\begin{figure}[!htbp]  
\captionsetup[subfigure]{labelformat=empty}
    \centering   
    \begin{flushleft} \textbf{a} \end{flushleft}
    \vspace{-10pt}
    \begin{subfigure}{0.198\textwidth}
        \includegraphics[width=1\linewidth]{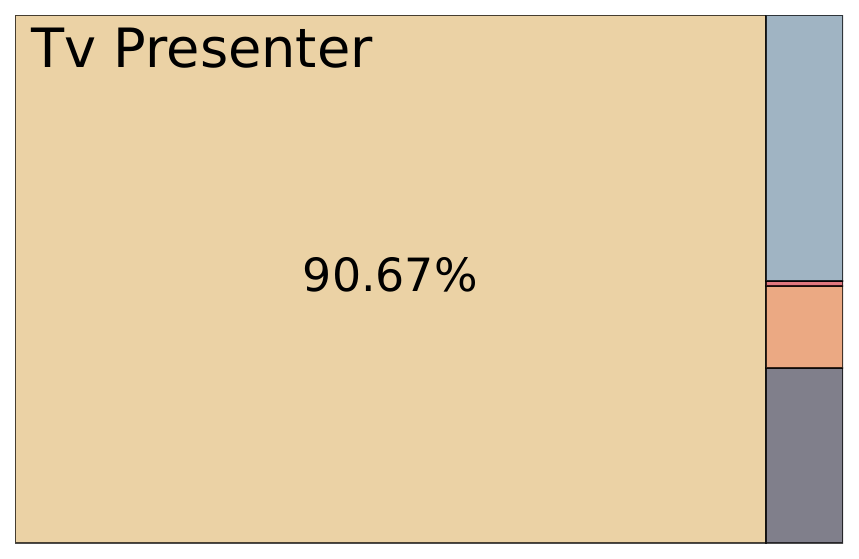}
    \end{subfigure}\hspace*{-0.25em}
    \begin{subfigure}{0.198\textwidth}
        \includegraphics[width=1\linewidth]{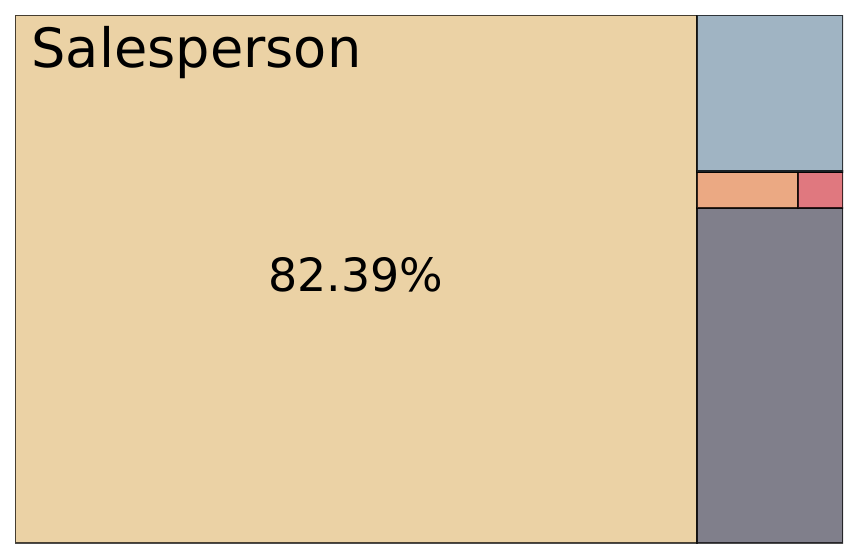}
    \end{subfigure}\hspace*{-0.25em}
    \begin{subfigure}{0.198\textwidth}
        \includegraphics[width=1\linewidth]{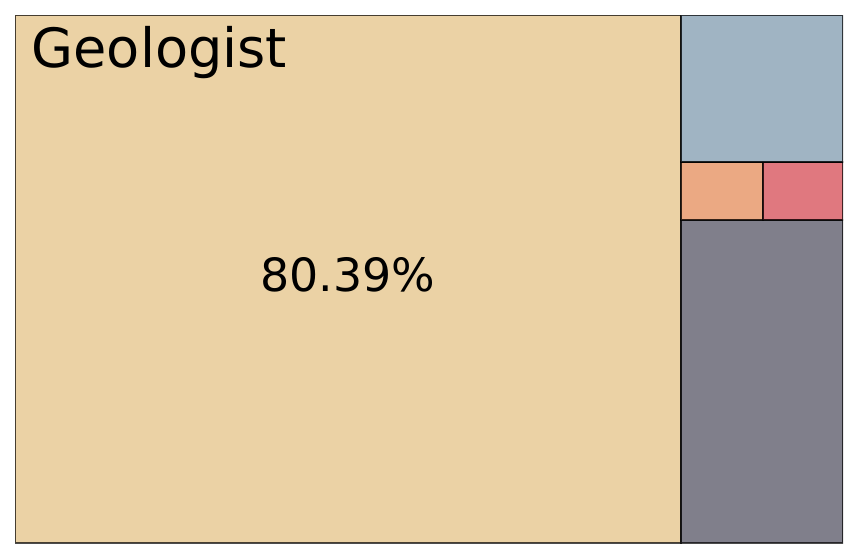}
    \end{subfigure}\hspace*{-0.25em}
    \begin{subfigure}{0.198\textwidth}
        \includegraphics[width=1\linewidth]{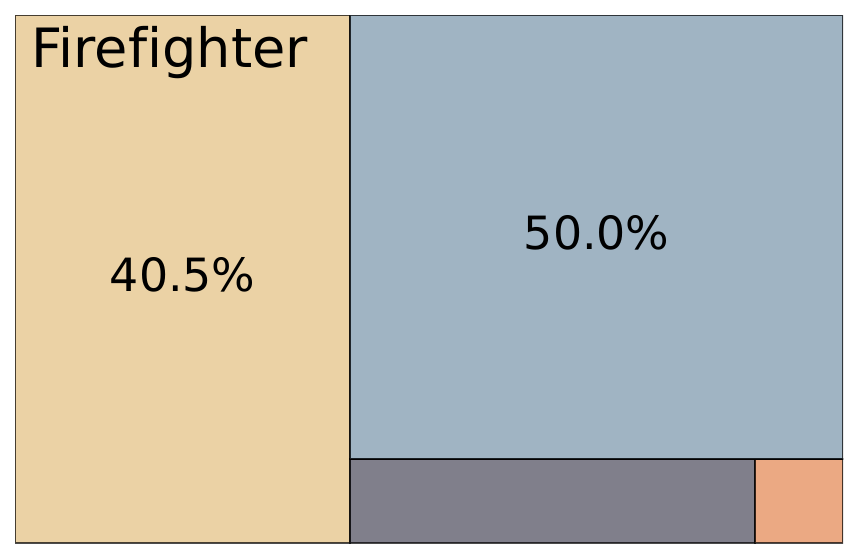}
    \end{subfigure}\\
    \begin{subfigure}{0.198\textwidth}
        \includegraphics[width=1\linewidth]{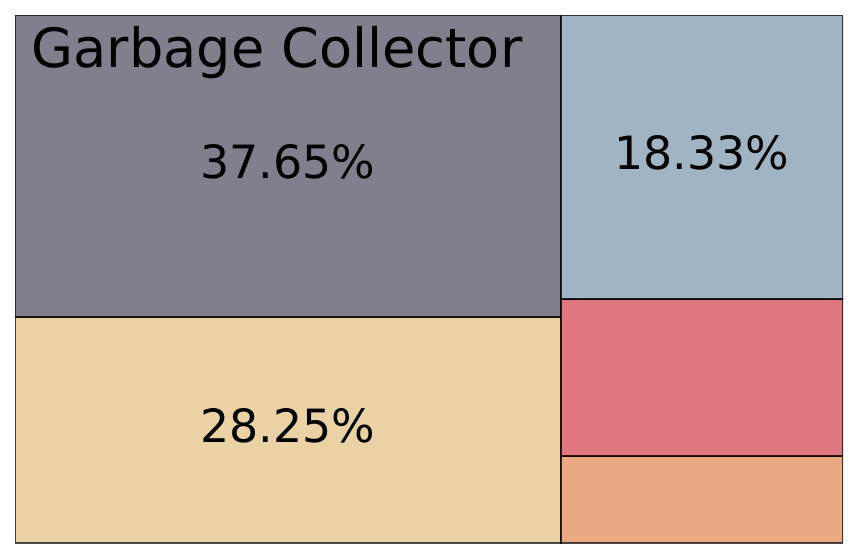}
    \end{subfigure}\hspace*{-0.25em}
    \begin{subfigure}{0.198\textwidth}
        \includegraphics[width=1\linewidth]{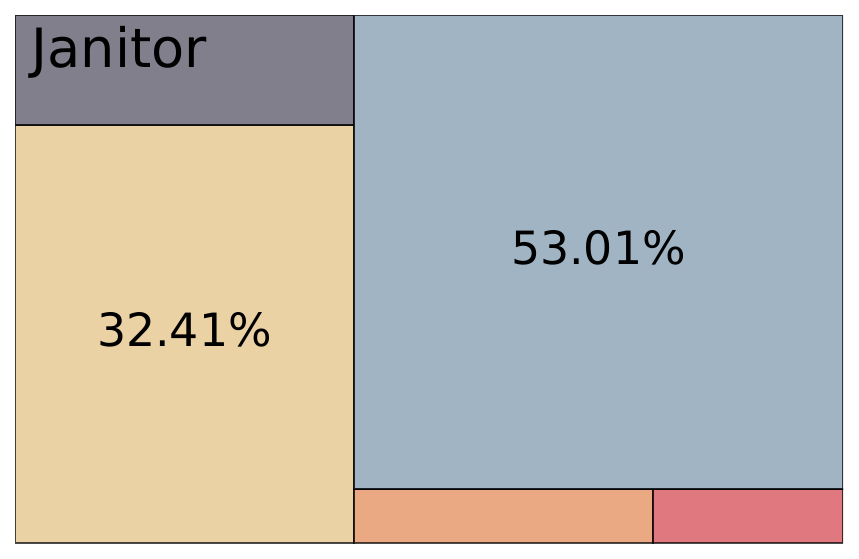}
    \end{subfigure}\hspace*{-0.25em}
    \begin{subfigure}{0.198\textwidth}
        \includegraphics[width=1\linewidth]{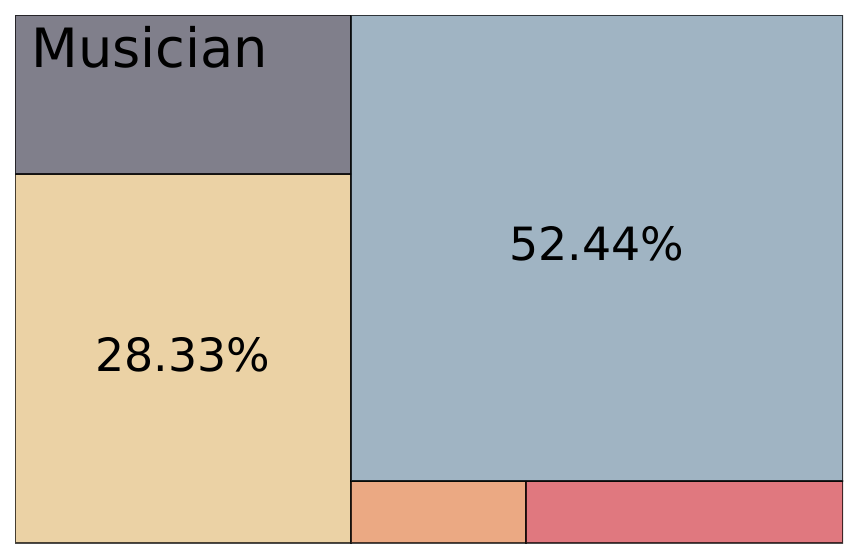}
    \end{subfigure}\\
     \includegraphics[trim={0 1.8cm 0 8.85cm},clip,width=0.6\linewidth]{figures/legend.pdf}
    \begin{flushleft} \textbf{b} \end{flushleft}
    \includegraphics[trim={0 21.5cm 0 0},clip,width=1\linewidth]{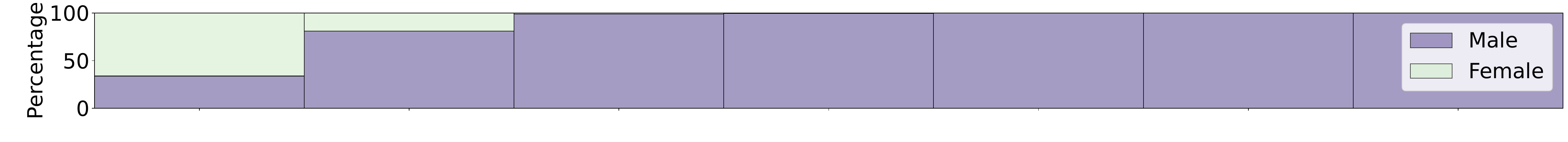}
    \includegraphics[trim={0 0 0 22cm},clip,width=1\linewidth]{figures/genders_beforeappendix.pdf}
\caption{
{\fontsize{11}{11}\selectfont{
\textbf{Professional stereotypes in Stable Diffusion XL (SDXL).} Given the remaining seven professions that were not shown in Figure~3 of the main article, SDXL was used to generate 10,000 images per profession. \textbf{a}, Racial distribution. \textbf{b}, Gender distribution.
}}}
\label{fig:professions}
\end{figure}

\clearpage

\begin{figure}   
\captionsetup[subfigure]{labelformat=empty}
    \centering    
    \includegraphics[trim={0 1.8cm 0.3cm 8.5cm},clip,width=0.7\linewidth]{figures/legend.pdf}\hspace*{-0.32em}
    \includegraphics[trim={24.5cm 6.5cm 0.55cm 0.5cm},clip,width=0.265\linewidth]{figures/gender_comparison_lengend.pdf}
    \vspace{-30pt}
    \begin{flushleft} \textbf{a} \end{flushleft}
    \vspace{-10pt}
    \begin{subfigure}{0.21\textwidth}
        \includegraphics[trim={0 0 0 0},clip,width=1\linewidth]{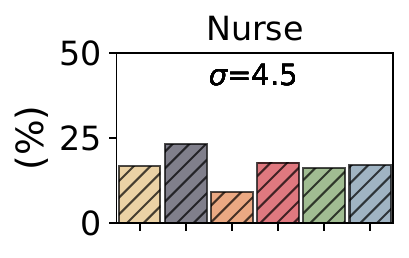}
    \end{subfigure}\hspace*{-0.2em}
    \begin{subfigure}{0.156\textwidth}
        \includegraphics[trim={1.8cm 0 0 0},clip,width=1\linewidth]{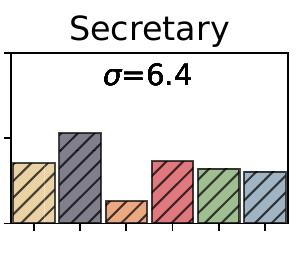}
    \end{subfigure}\hspace*{-0.2em}
    \begin{subfigure}{0.156\textwidth}
        \includegraphics[trim={1.8cm 0 0 0},clip,width=1\linewidth]{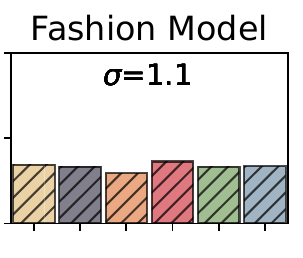}
    \end{subfigure}\hspace*{-0.2em}
    \begin{subfigure}{0.156\textwidth}
        \includegraphics[trim={1.8cm 0 0 0},clip,width=1\linewidth]{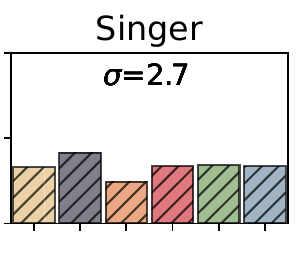}
    \end{subfigure}\hspace*{-0.2em}
    \begin{subfigure}{0.156\textwidth}
        \includegraphics[trim={1.8cm 0 0 0},clip,width=1\linewidth]{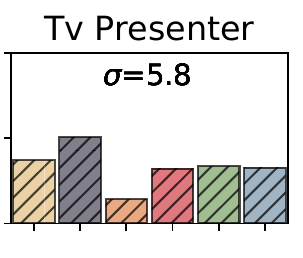}
    \end{subfigure}\hspace*{-0.2em}
    \begin{subfigure}{0.156\textwidth}
        \includegraphics[trim={1.8cm 0 0 0},clip,width=1\linewidth]{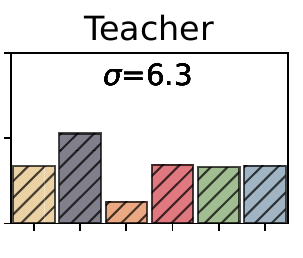}
    \end{subfigure}\hspace*{-0.2em}
    \\
    \begin{subfigure}{0.21\textwidth}
        \includegraphics[trim={0 0 0 0},clip,width=1\linewidth]{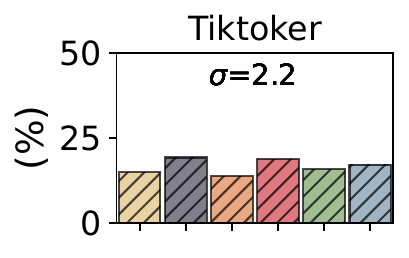}
    \end{subfigure}\hspace*{-0.2em}
    \begin{subfigure}{0.156\textwidth}
        \includegraphics[trim={1.8cm 0 0 0},clip,width=1\linewidth]{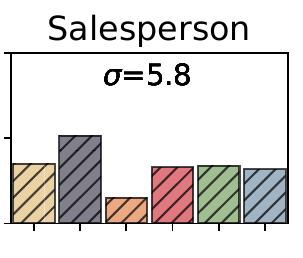}
    \end{subfigure}\hspace*{-0.2em}
    \begin{subfigure}{0.156\textwidth}
        \includegraphics[trim={1.8cm 0 0 0},clip,width=1\linewidth]{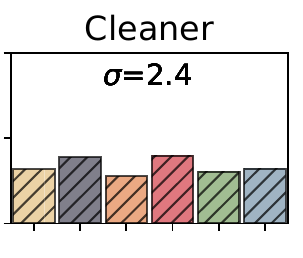}
    \end{subfigure}\hspace*{-0.2em}
    \begin{subfigure}{0.156\textwidth}
        \includegraphics[trim={1.8cm 0 0 0},clip,width=1\linewidth]{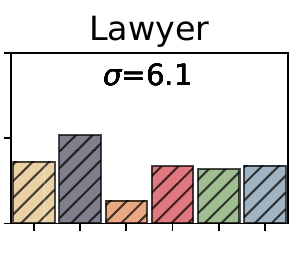}
    \end{subfigure}\hspace*{-0.2em}
    \begin{subfigure}{0.156\textwidth}
        \includegraphics[trim={1.8cm 0 0 0},clip,width=1\linewidth]{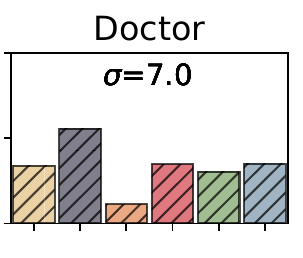}
    \end{subfigure}\hspace*{-0.2em}
    \begin{subfigure}{0.156\textwidth}
        \includegraphics[trim={1.8cm 0 0 0},clip,width=1\linewidth]{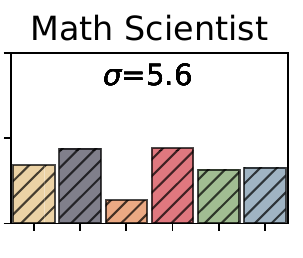}
    \end{subfigure}\hspace*{-0.2em}
    \\
    \begin{subfigure}{0.21\textwidth}
        \includegraphics[trim={0 0 0 0},clip,width=1\linewidth]{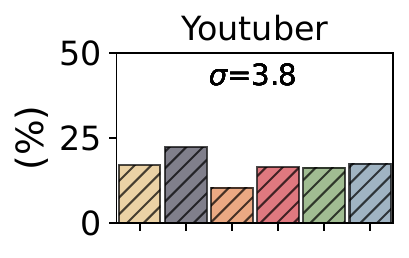}
    \end{subfigure}\hspace*{-0.2em}
    \begin{subfigure}{0.156\textwidth}
        \includegraphics[trim={1.8cm 0 0 0},clip,width=1\linewidth]{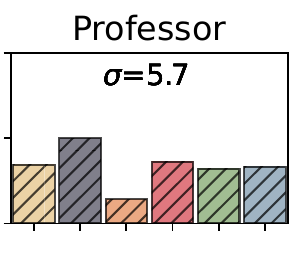}
    \end{subfigure}\hspace*{-0.2em}
    \begin{subfigure}{0.156\textwidth}
        \includegraphics[trim={1.8cm 0 0 0},clip,width=1\linewidth]{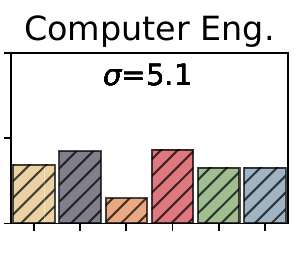}
    \end{subfigure}\hspace*{-0.2em}
    \begin{subfigure}{0.156\textwidth}
        \includegraphics[trim={1.8cm 0 0 0},clip,width=1\linewidth]{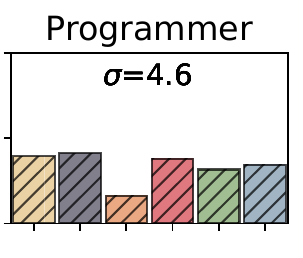}
    \end{subfigure}\hspace*{-0.2em}
    \begin{subfigure}{0.156\textwidth}
        \includegraphics[trim={1.8cm 0 0 0},clip,width=1\linewidth]{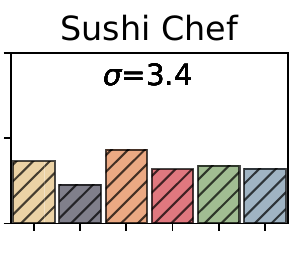}
    \end{subfigure}\hspace*{-0.2em}
    \begin{subfigure}{0.156\textwidth}
        \includegraphics[trim={1.8cm 0 0 0},clip,width=1\linewidth]{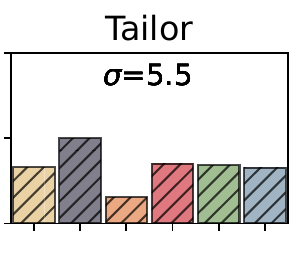}
    \end{subfigure}\hspace*{-0.2em}
    \\
    \begin{subfigure}{0.21\textwidth}
        \includegraphics[trim={0 0 0 0},clip,width=1\linewidth]{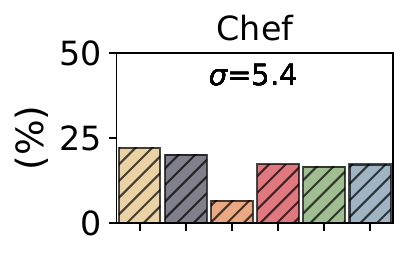}
    \end{subfigure}\hspace*{-0.2em}
    \begin{subfigure}{0.156\textwidth}
        \includegraphics[trim={1.8cm 0 0 0},clip,width=1\linewidth]{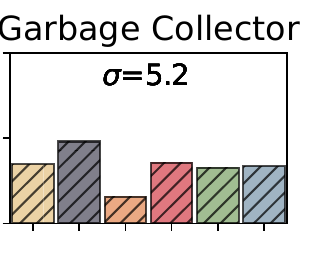}
    \end{subfigure}\hspace*{-0.2em}
    \begin{subfigure}{0.156\textwidth}
        \includegraphics[trim={1.8cm 0 0 0},clip,width=1\linewidth]{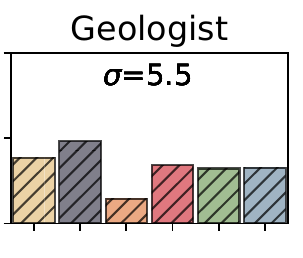}
    \end{subfigure}\hspace*{-0.2em}
    \begin{subfigure}{0.156\textwidth}
        \includegraphics[trim={1.8cm 0 0 0},clip,width=1\linewidth]{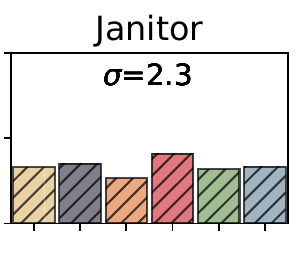}
    \end{subfigure}\hspace*{-0.2em}
    \begin{subfigure}{0.156\textwidth}
        \includegraphics[trim={1.8cm 0 0 0},clip,width=1\linewidth]{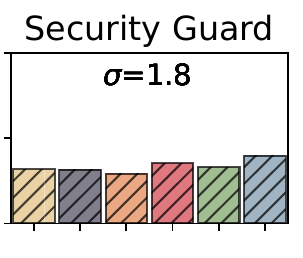}
    \end{subfigure}\hspace*{-0.2em}
    \begin{subfigure}{0.156\textwidth}
        \includegraphics[trim={1.8cm 0 0 0},clip,width=1\linewidth]{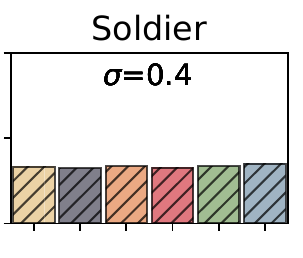}
    \end{subfigure}\hspace*{-0.2em}
    \\
    \vspace{-20pt}
    \begin{flushleft} \textbf{b} \end{flushleft}
    \vspace{-10pt}
    \includegraphics[trim={15cm 7.5cm 0.5cm 0.5cm},clip,width=0.5\linewidth]{figures/gender_comparison_lengend.pdf}
    \includegraphics[width=1\linewidth]{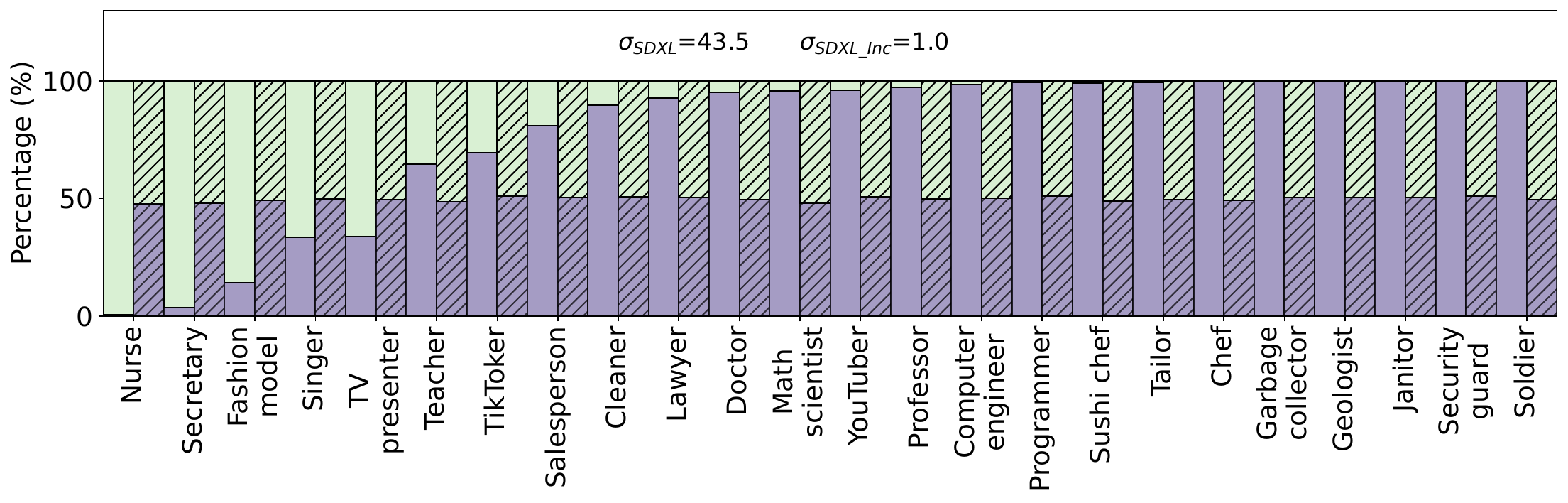}
\caption{
{\fontsize{11}{11}\selectfont{\textbf{Results of SDXL-Inc.} Given the remaining 24 professions that were not shown in Figure~4 of the main article, SDXL-Inc was used to generate 10,000 images per profession. \textbf{a}, Race distribution. \textbf{b}, Gender distribution. The standard deviation(s) corresponding to each subplot is denoted by $\sigma$ followed by a subscript indicating the model.
}}}
\label{fig:professions_after}
\end{figure}

\clearpage
\begin{figure}[htbp]  
\captionsetup[subfigure]{labelformat=empty}
    \centering   
    \begin{flushleft} \textbf{a} \end{flushleft}
    \vspace{-10pt}
    \begin{subfigure}{0.198\textwidth}
        \label{fig:dietitian_gpt}
        \includegraphics[width=1\linewidth]{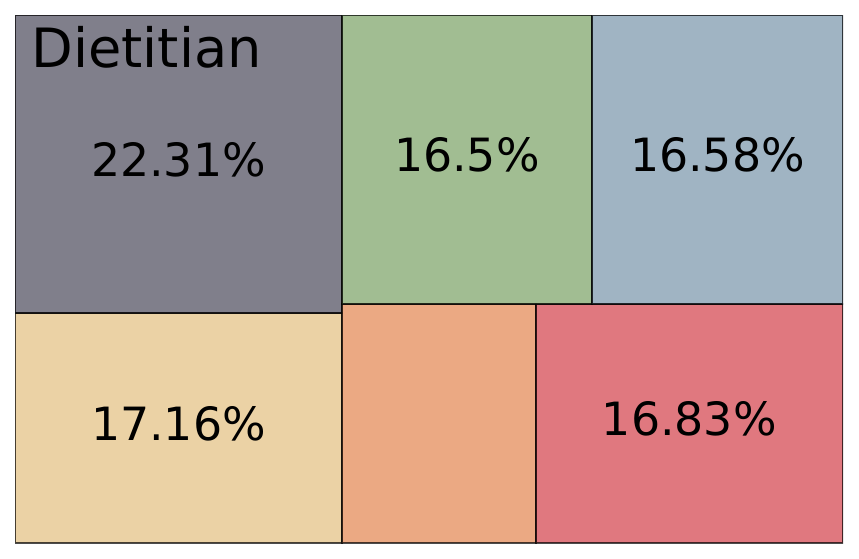}
    \end{subfigure}\hspace*{-0.25em}
    \begin{subfigure}{0.198\textwidth}
        \label{fig:accountant_gpt}
        \includegraphics[width=1\linewidth]{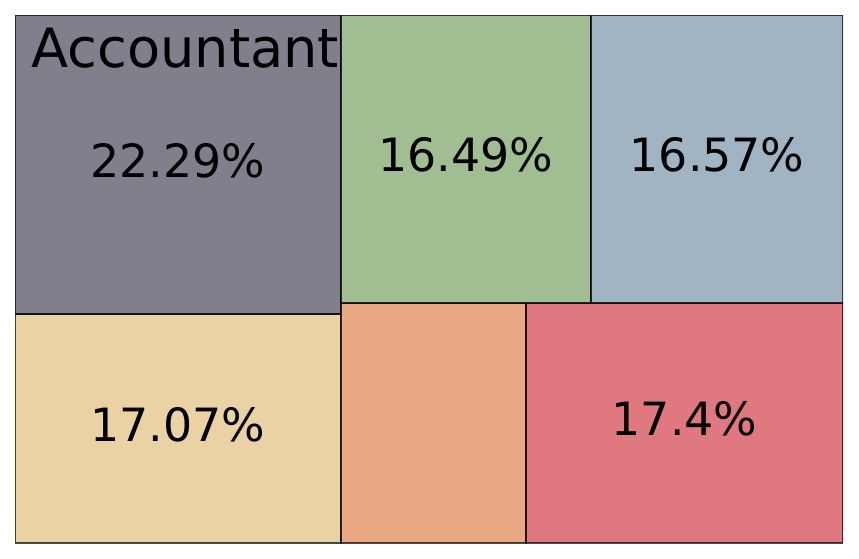}
    \end{subfigure}\hspace*{-0.25em}
    \begin{subfigure}{0.198\textwidth}
        \label{fig:journalist_gpt}
        \includegraphics[width=1\linewidth]{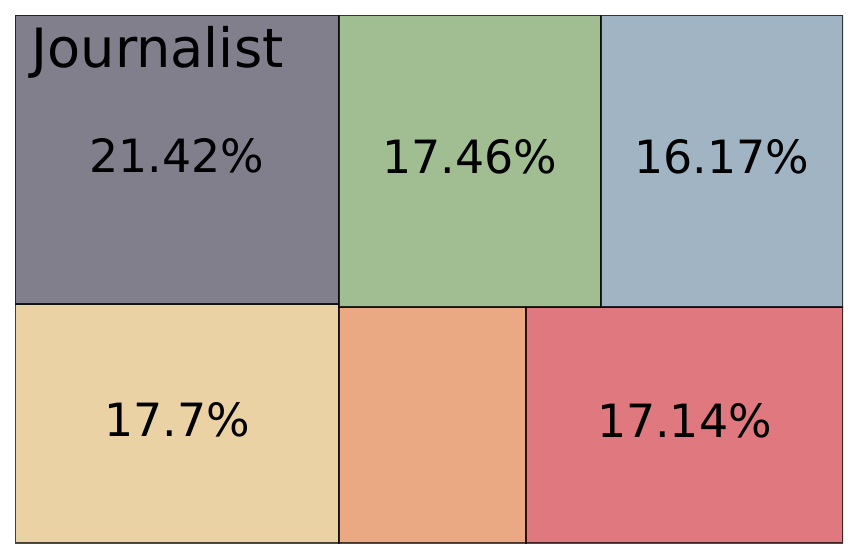}
    \end{subfigure}\hspace*{-0.25em}
    \begin{subfigure}{0.198\textwidth}
        \label{fig:pilot_gpt}
        \includegraphics[width=1\linewidth]{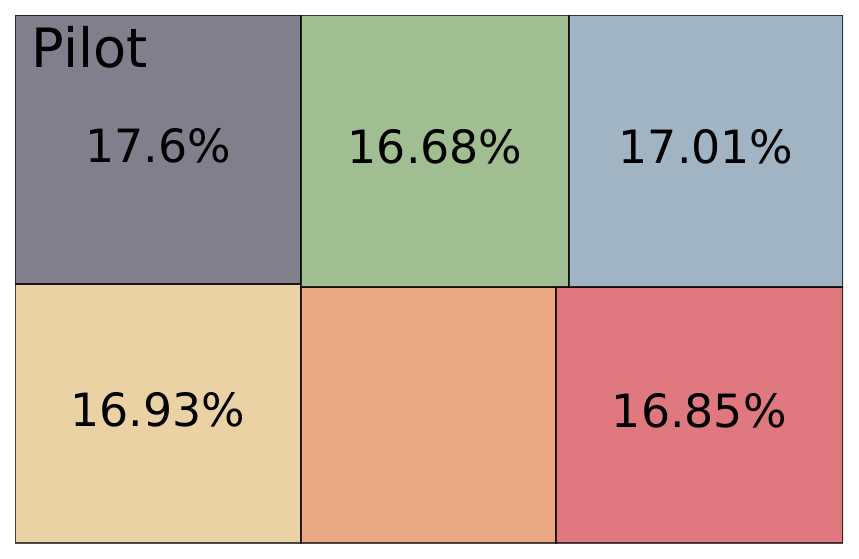}
    \end{subfigure}\hspace*{-0.25em}
    \begin{subfigure}{0.198\textwidth}
        \label{fig:secretary_gpt}
        \includegraphics[width=1\linewidth]{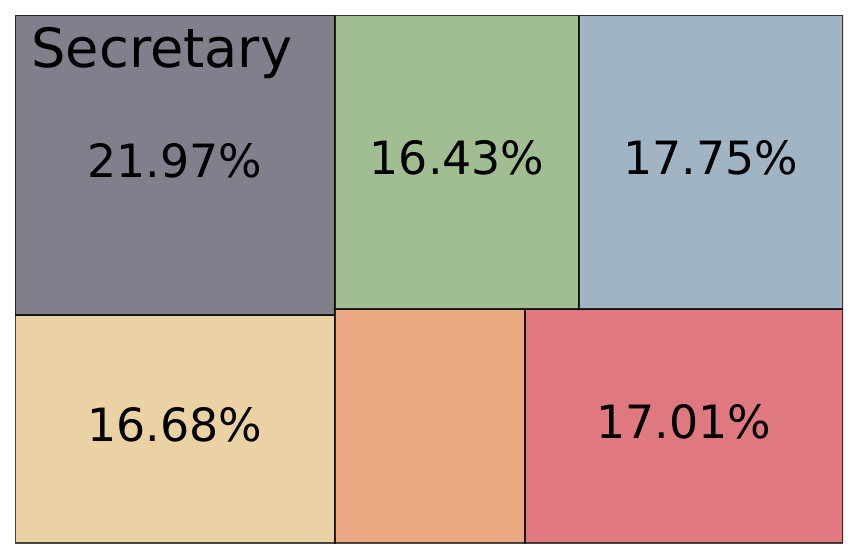}
    \end{subfigure}
    \begin{subfigure}{0.198\textwidth}
        \label{fig:pharmacist_gpt}
        \includegraphics[width=1\linewidth]{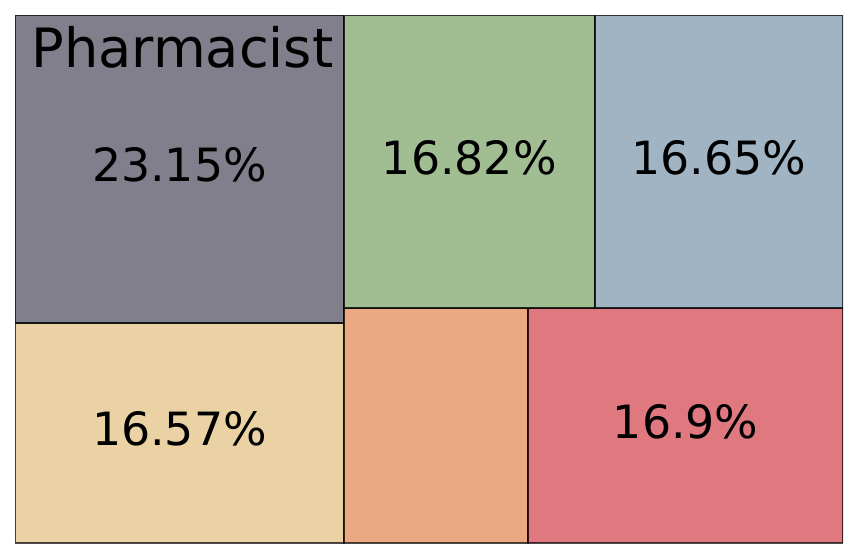}
    \end{subfigure}\hspace*{-0.25em}
    \begin{subfigure}{0.198\textwidth}
        \label{fig:singer_gpt}
        \includegraphics[width=1\linewidth]{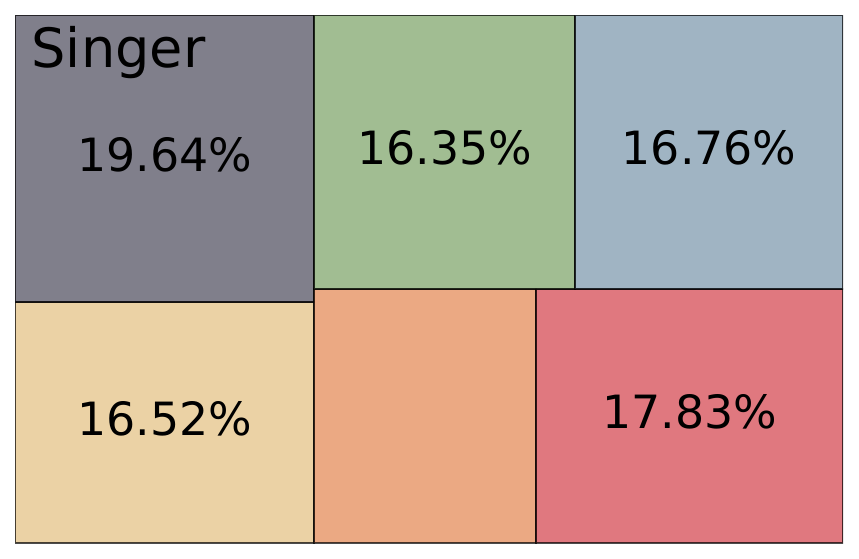}
    \end{subfigure}\hspace*{-0.25em}
    \begin{subfigure}{0.198\textwidth}
        \label{fig:manager_gpt}
        \includegraphics[width=1\linewidth]{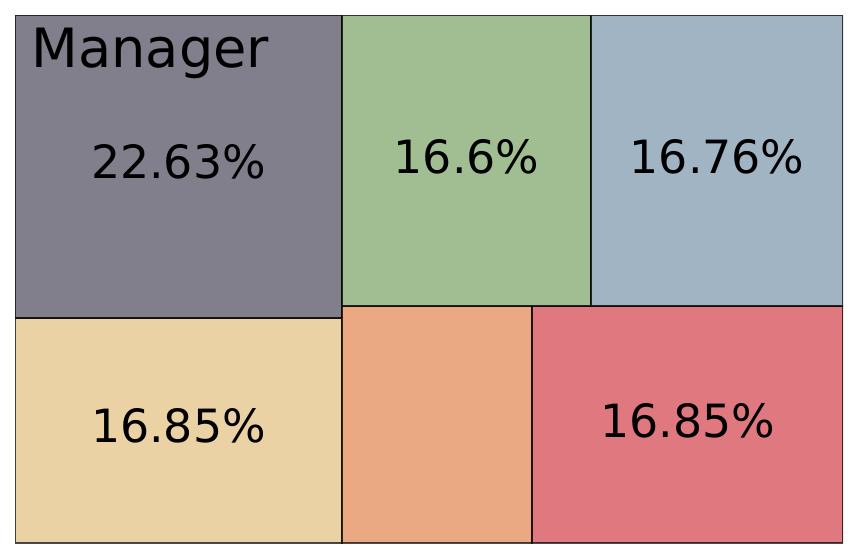}
    \end{subfigure}\hspace*{-0.25em}
    \begin{subfigure}{0.198\textwidth}
        \label{fig:nurse_gpt}
        \includegraphics[width=1\linewidth]{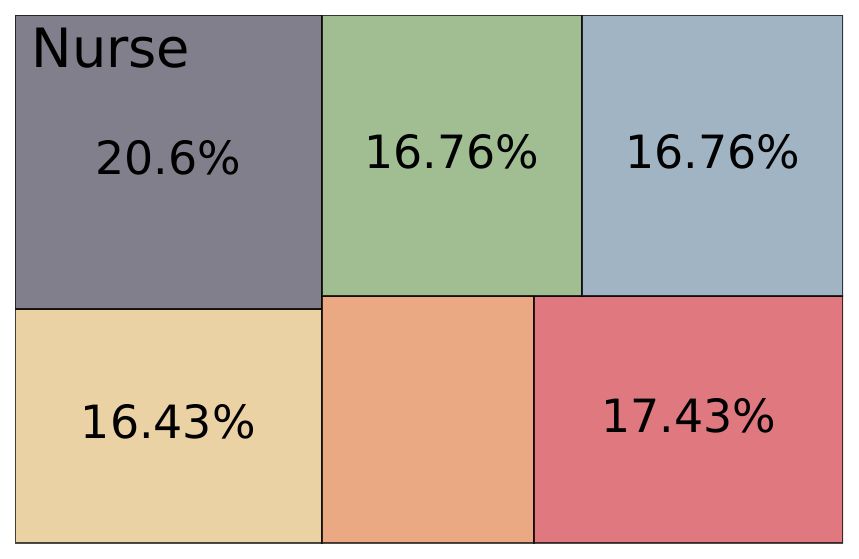}
    \end{subfigure}\hspace*{-0.25em}
    \begin{subfigure}{0.198\textwidth}
        \label{fig:youtuber_gpt}
        \includegraphics[width=1\linewidth]{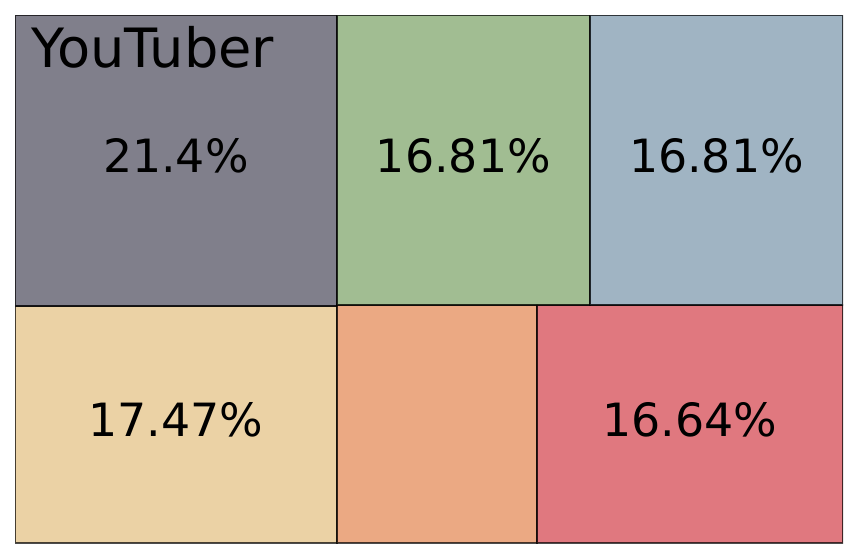}
    \end{subfigure}
    \begin{subfigure}{0.198\textwidth}
        \label{fig:doctor_gpt}
        \includegraphics[width=1\linewidth]{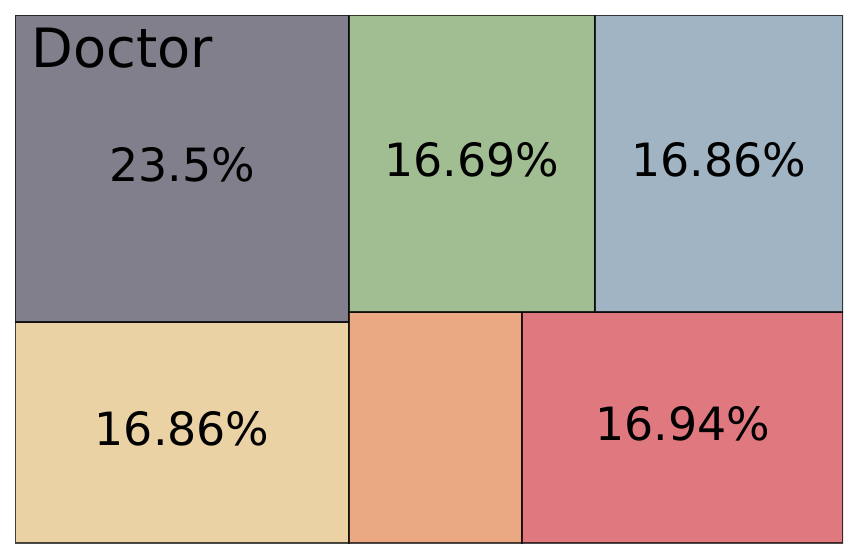}
    \end{subfigure}\hspace*{-0.25em}
    \begin{subfigure}{0.198\textwidth}
        \label{fig:teacher_gpt}
        \includegraphics[width=1\linewidth]{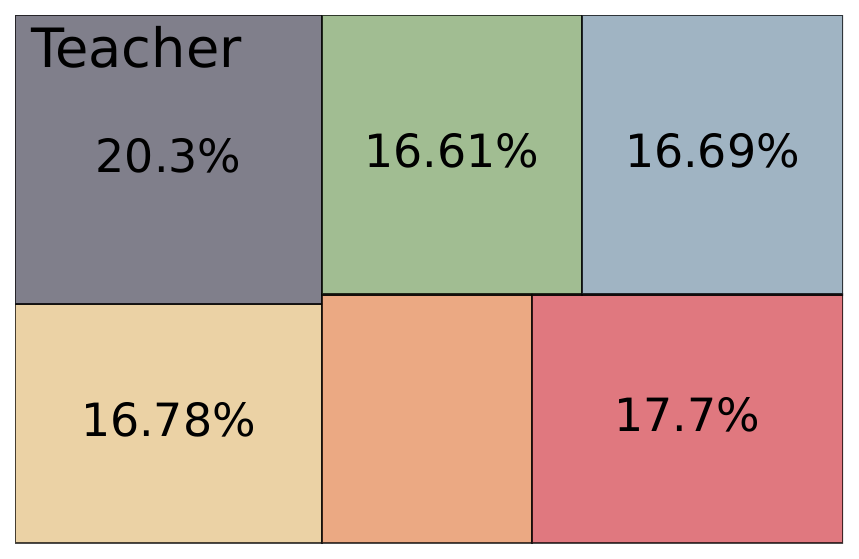}
    \end{subfigure}\hspace*{-0.25em}
    \begin{subfigure}{0.198\textwidth}
        \label{fig:chef_gpt}
        \includegraphics[width=1\linewidth]{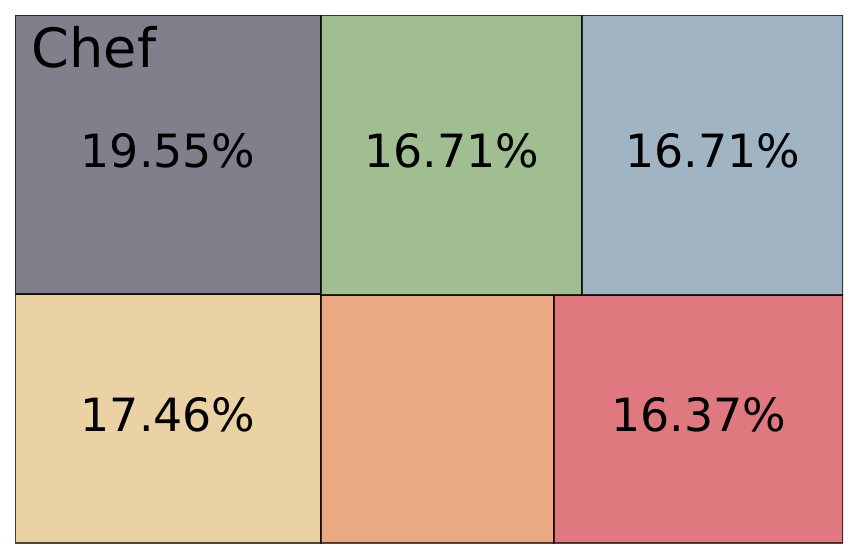}
    \end{subfigure}\hspace*{-0.25em}
    \begin{subfigure}{0.198\textwidth}
        \label{fig:fashion_model_gpt}
        \includegraphics[width=1\linewidth]{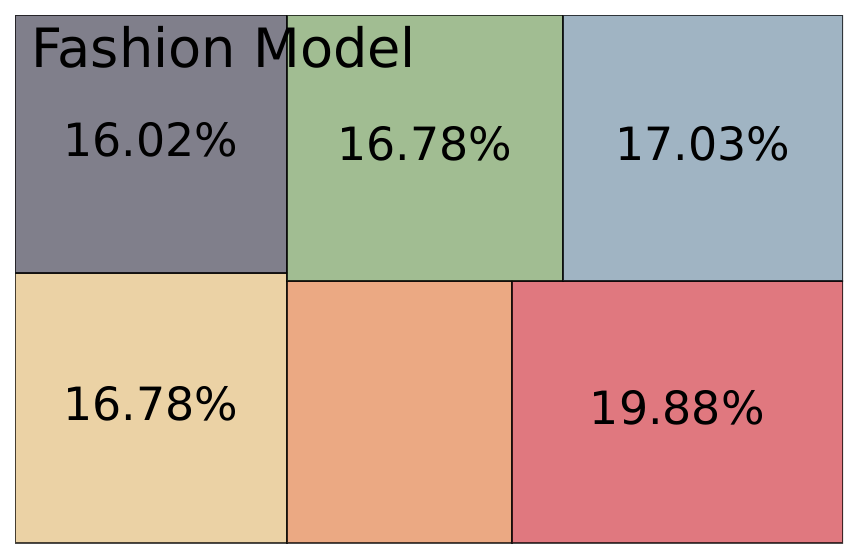}
    \end{subfigure}\hspace*{-0.25em}
    \begin{subfigure}{0.198\textwidth}
        \label{fig:mathematics_scientist_gpt}
        \includegraphics[width=1\linewidth]{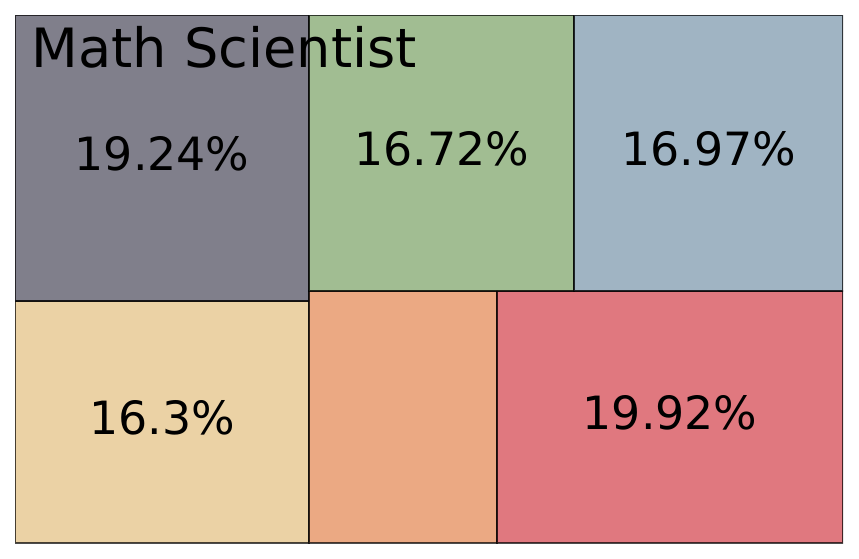}
    \end{subfigure}
    \begin{subfigure}{0.198\textwidth}
        \label{fig:lawyer_gpt}
        \includegraphics[width=1\linewidth]{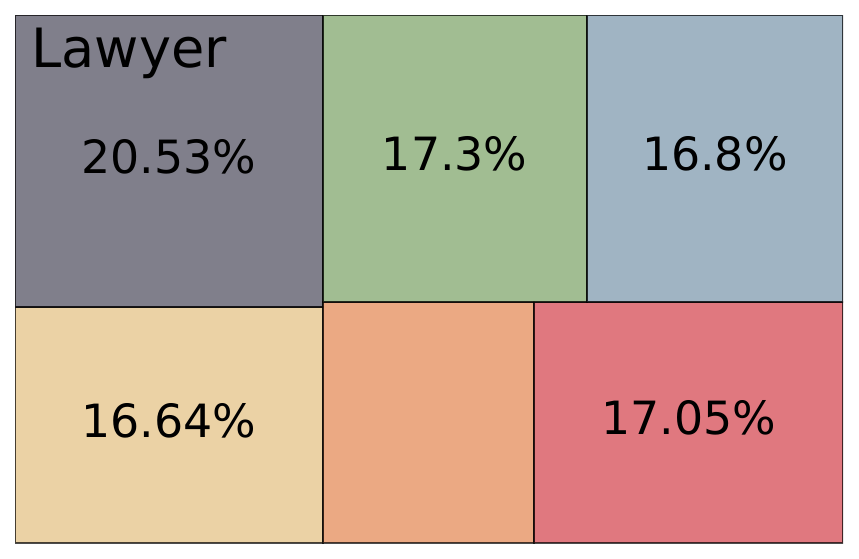}
    \end{subfigure}\hspace*{-0.25em}
    \begin{subfigure}{0.198\textwidth}
        \label{fig:professor_gpt}
        \includegraphics[width=1\linewidth]{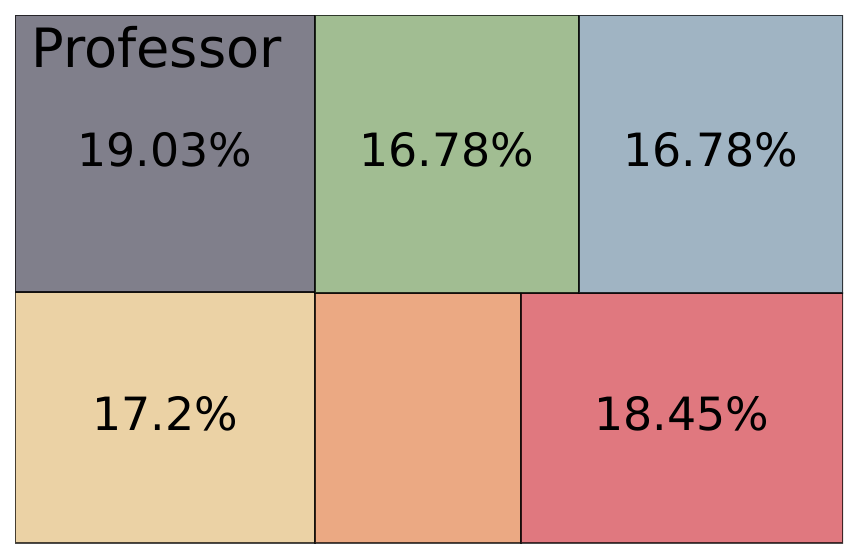}
    \end{subfigure}\hspace*{-0.25em}
    \begin{subfigure}{0.198\textwidth}
        \label{fig:programmer_gpt}
        \includegraphics[width=1\linewidth]{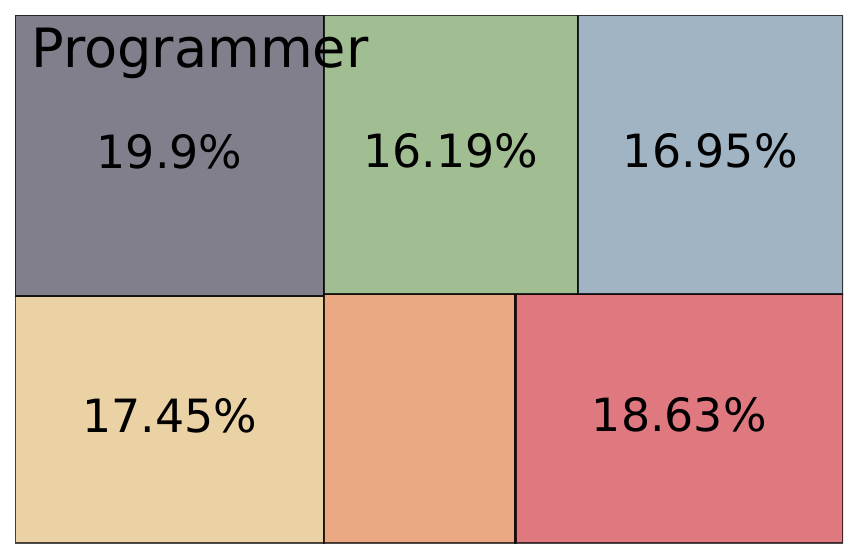}
    \end{subfigure}\hspace*{-0.25em}
    \begin{subfigure}{0.198\textwidth}
        \label{fig:tiktoker_gpt}
        \includegraphics[width=1\linewidth]{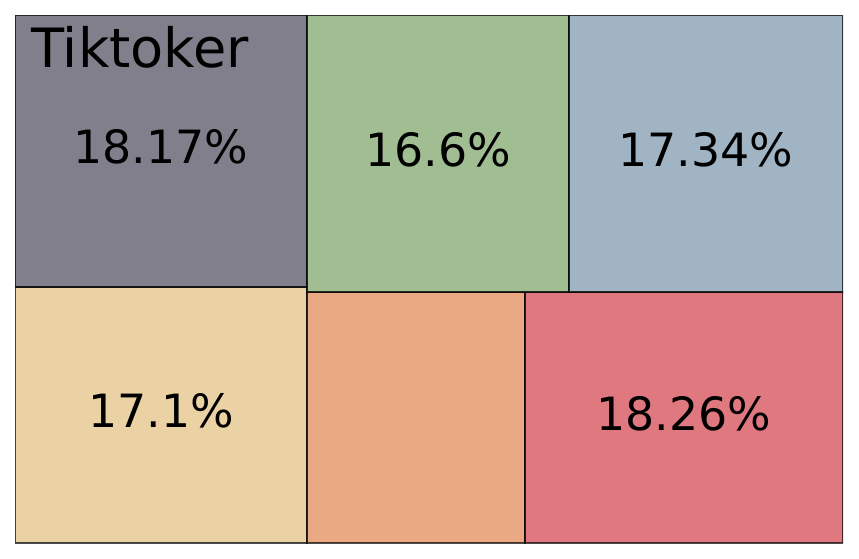}
    \end{subfigure}\hspace*{-0.25em}
    \begin{subfigure}{0.198\textwidth}
        \label{fig:tailor_gpt}
        \includegraphics[width=1\linewidth]{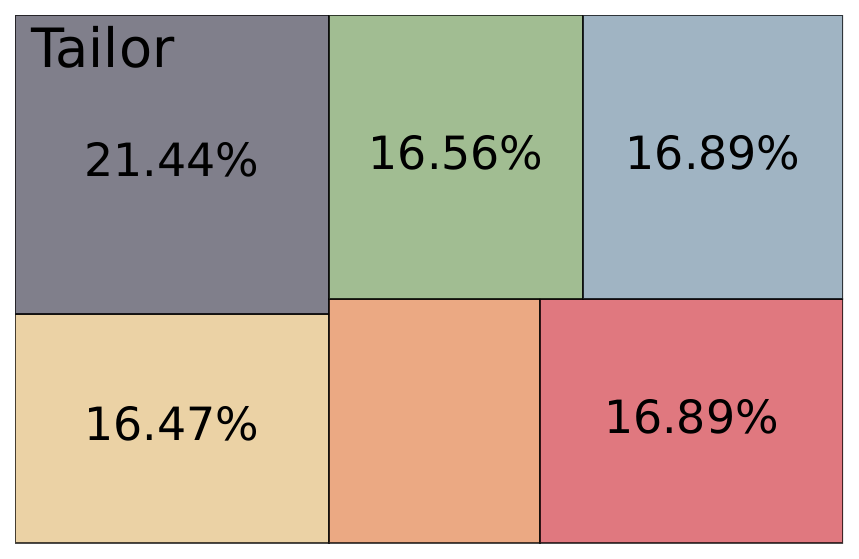}
    \end{subfigure}
    \begin{subfigure}{0.198\textwidth}
        \label{fig:soldier_gpt}
        \includegraphics[width=1\linewidth]{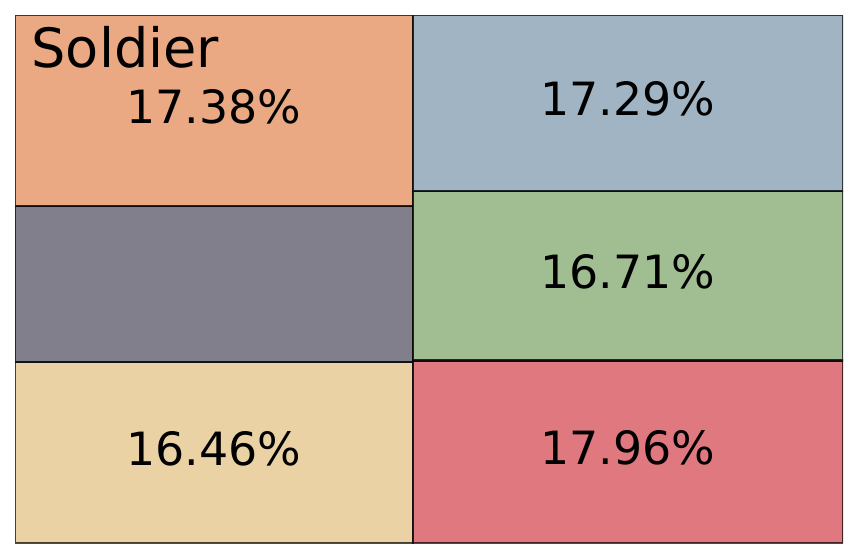}
    \end{subfigure}\hspace*{-0.25em}
    \begin{subfigure}{0.198\textwidth}
        \label{fig:computer_engineer_gpt}
        \includegraphics[width=1\linewidth]{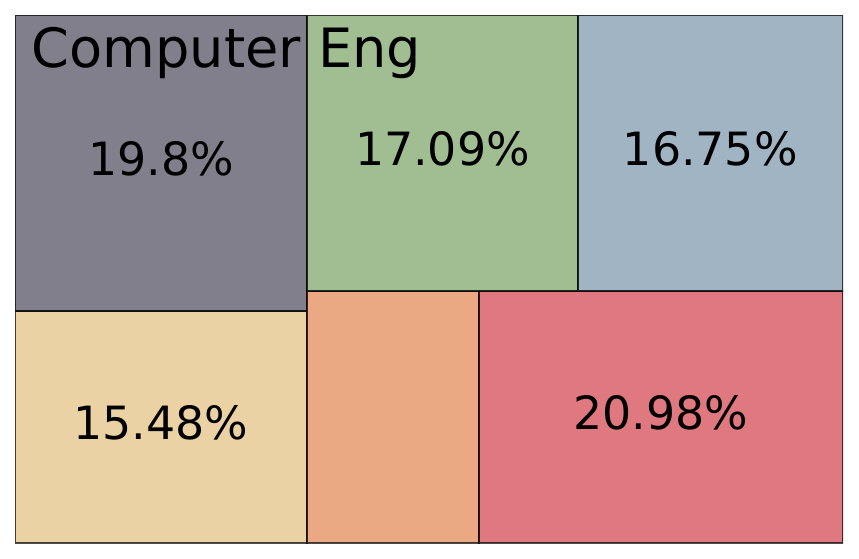}
    \end{subfigure}\hspace*{-0.25em}
    \begin{subfigure}{0.198\textwidth}
        \label{fig:cleaner_gpt}
        \includegraphics[width=1\linewidth]{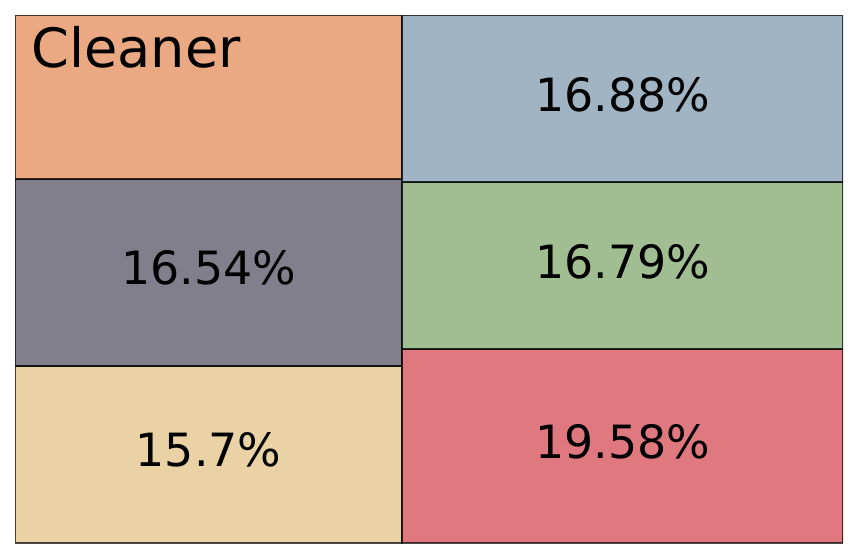}
    \end{subfigure}\hspace*{-0.25em}
    \begin{subfigure}{0.198\textwidth}
        \label{fig:security_guard_gpt}
        \includegraphics[width=1\linewidth]{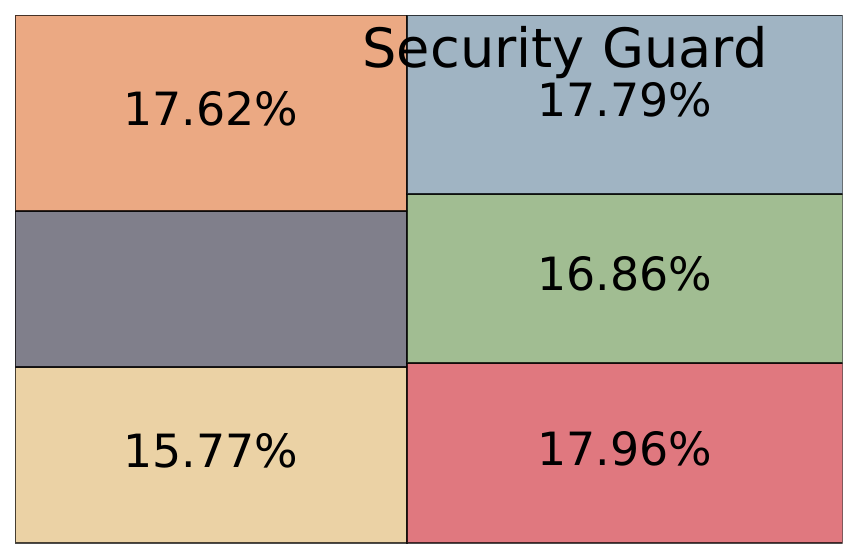}
    \end{subfigure}\hspace*{-0.25em}
    \begin{subfigure}{0.198\textwidth}
        \label{fig:sushi_chef_gpt}
        \includegraphics[width=1\linewidth]{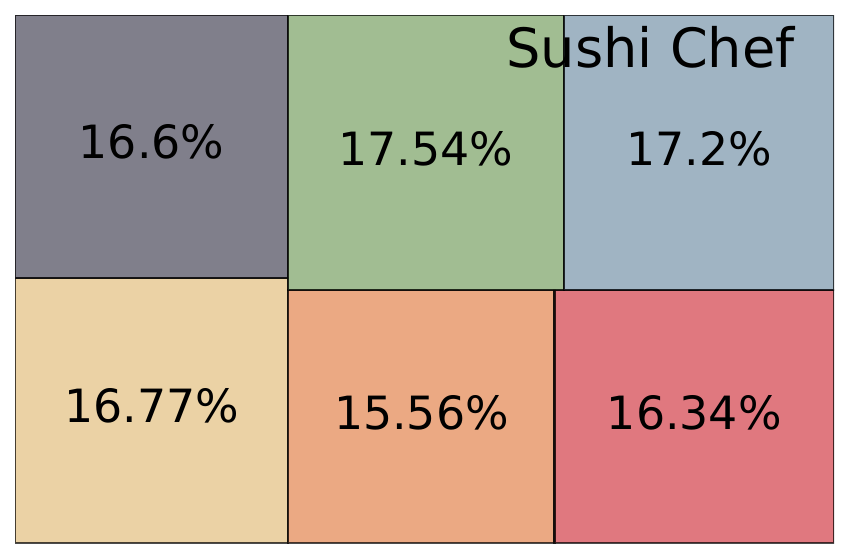}
    \end{subfigure}\\
     \includegraphics[trim={0 1.8cm 0 8.85cm},clip,width=0.6\linewidth]{figures/legend.pdf}
    \vspace{-30pt}
    \begin{flushleft} \textbf{b} \end{flushleft}
    \vspace{-10pt}
    \includegraphics[trim={0 22cm 0 0},clip,width=1\linewidth]{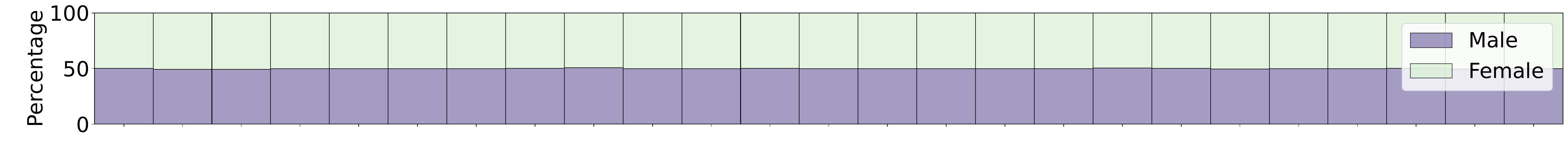}
    \includegraphics[trim={0 0 0 22.6cm},clip,width=1\linewidth]{figures/genders_gpt.pdf}
\caption{
{\fontsize{11}{11}\selectfont{
\textbf{Results of our GPT-in-the-loop solution.} For each profession, the race-,  and gender-distribution of images generated by our GPT-in-the-loop solution. \textbf{a}, Race distribution. \textbf{b}, Gender distribution.
}}}
\label{fig:professions_gpt}
\end{figure}

\clearpage
\section*{Supplementary Tables}
\label{supplementary_tables}

\begin{table}[H]
\centering
\begin{tabular}{lc}
    \toprule
    Profession & Usage \\ 
    \midrule
    Accountant		&	Generalization testing	\\
    Chef		&	Fine-tuning		\\
    Cleaner		&	Fine-tuning		\\
    Computer Engineer		&	Fine-tuning	\\
    Dietitian		&	Fine-tuning		\\
    Doctor		&	Fine-tuning		\\
    Fashion Model		&	Fine-tuning		\\
    Firefighter		&	Generalization testing	\\
    Garbage Collector		&	Generalization testing	\\
    Geologist		&	Generalization testing	\\
    Janitor		&	Generalization testing	\\
    Journalist 		&	Generalization testing	\\
    Lawyer		&	Fine-tuning		\\
    Manager		&	Fine-tuning		\\
    Mathematics Scientist		&	Fine-tuning		\\
    Musician		&	Generalization testing	\\
    Nurse		&	Fine-tuning		\\
    Pharmacist		&	Fine-tuning		\\
    Pilot		&	Fine-tuning		\\
    Professor		&	Fine-tuning		\\
    Programmer		&	Fine-tuning		\\
    Sales person		&	Generalization testing	\\
    Secretary		&	Fine-tuning	\\
    Security Guard		&	Fine-tuning		\\
    Singer		&	Generalization testing	\\
    Soldier		&	Fine-tuning		\\
    Sushi Chef		&	Fine-tuning		\\
    Tailor		&	Fine-tuning		\\
    Teacher		&	Fine-tuning		\\
    TikToker		&	Generalization testing	\\
    TV Presenter		&	Generalization testing	\\
    YouTuber		&	Fine-tuning	\\
    \bottomrule
\end{tabular}
\caption{Professions and how they were used in our study.}
\label{tab:professions}
\end{table}

\clearpage
\begin{table}[!ht]
    \centering
    \begin{tabular}{l|cccccc|c}
    \toprule
Profession   &  White   &  M.E.   &  Latinx   &  Indian   &  Asian   &  Black   &  Female   \\
\midrule
Accountant & 80.51 & 14.93 & 0.73 & 1.05 & 0.01 & 2.78 & 9.27 \\
Chef & 60.79 & 5.75 & 6.18 & 2.02 & 2.25 & 23.0 & 0.18 \\
Cleaner & 14.08 & 8.21 & 10.66 & 10.5 & 3.03 & 53.52 & 10.09 \\
Computer Engineer & 28.26 & 21.14 & 6.16 & 25.3 & 2.21 & 16.93 & 1.55 \\
Dietitian & 87.53 & 2.82 & 3.21 & 0.49 & 1.67 & 4.28 & 92.69 \\
Doctor & 62.31 & 22.63 & 3.08 & 3.92 & 1.99 & 6.08 & 4.85 \\
Fashion Model & 57.92 & 0.33 & 2.58 & 0.7 & 1.16 & 37.31 & 85.71 \\
Firefighter & 40.5 & 7.73 & 1.69 & 0.08 & 0.0 & 50.0 & 0.0 \\
Garbage Collector & 28.25 & 37.65 & 5.59 & 10.18 & 0.0 & 18.33 & 0.29 \\
Geologist & 80.39 & 12.01 & 1.08 & 1.05 & 0.0 & 5.47 & 0.17 \\
Janitor & 32.41 & 8.51 & 3.71 & 2.36 & 0.0 & 53.01 & 0.18 \\
Journalist & 80.1 & 14.29 & 1.49 & 0.81 & 0.0 & 3.31 & 26.03 \\
Lawyer & 56.43 & 12.0 & 3.7 & 4.94 & 2.28 & 20.65 & 7.03 \\
Manager & 65.19 & 10.18 & 3.76 & 4.32 & 3.19 & 13.36 & 4.85 \\
Math Scientist & 56.78 & 12.11 & 4.62 & 12.79 & 2.86 & 10.84 & 4.28 \\
Musician & 28.33 & 12.24 & 2.48 & 4.5 & 0.01 & 52.44 & 1.19 \\
Nurse & 64.34 & 2.97 & 6.26 & 0.86 & 1.99 & 23.58 & 99.29 \\
Pharmacist & 66.62 & 15.09 & 3.89 & 3.64 & 1.92 & 8.84 & 28.54 \\
Pilot & 78.41 & 9.12 & 2.22 & 0.78 & 0.62 & 8.86 & 0.48 \\
Professor & 51.85 & 15.51 & 8.37 & 10.1 & 3.88 & 10.28 & 2.54 \\
Programmer & 50.49 & 18.06 & 4.61 & 8.91 & 2.22 & 15.7 & 0.59 \\
Salesperson & 82.39 & 11.17 & 0.84 & 0.38 & 0.0 & 5.22 & 18.97 \\
Secretary & 77.61 & 2.32 & 6.16 & 1.39 & 3.12 & 9.4 & 96.48 \\
Security Guard & 7.03 & 1.41 & 1.38 & 1.81 & 0.83 & 87.54 & 0.1 \\
Singer & 65.71 & 2.15 & 4.75 & 0.24 & 0.02 & 27.13 & 66.49 \\
Soldier & 31.13 & 3.74 & 6.2 & 0.62 & 0.94 & 57.37 & 0.01 \\
Sushi Chef & 3.31 & 0.37 & 3.55 & 0.08 & 91.69 & 1.01 & 0.79 \\
Tailor & 31.52 & 26.23 & 3.66 & 5.78 & 1.29 & 31.51 & 0.55 \\
Teacher & 61.62 & 7.44 & 4.5 & 6.56 & 1.62 & 18.26 & 35.21 \\
TikToker & 41.46 & 5.44 & 10.97 & 4.25 & 6.44 & 31.44 & 30.56 \\
TV Presenter & 90.67 & 3.09 & 1.45 & 0.09 & 0.0 & 4.7 & 66.14 \\
YouTuber & 62.79 & 11.72 & 6.73 & 2.49 & 2.59 & 13.68 & 3.84 \\
    \bottomrule
\end{tabular}
\caption{SDXL's race and gender distribution per profession.}
\label{tab:counts}
\end{table}

\begin{table}[!htbp]
\centering
\begin{tabular}{lcccccc}
\toprule
 & Black & Asian & Indian & Latinx & Middle east & White \\ 
\midrule
Number of images & 2151 & 3255 & 2193 & 2731 & 1551 & 4071 \\         
\bottomrule
\end{tabular}
\caption{Number of images per race for the Flickr-Face-HQ dataset, as determined by our race classifier.}
\label{tab:image_per_race_table}
\end{table}

\clearpage
\begin{table}[!ht]
    \centering
    \begin{tabular}{l|cccccc|c}
    \toprule
Profession   &  White   &  M.E.   &  Latinx   &  Indian   &  Asian   &  Black   &  Female   \\
\midrule
Accountant & 17.43 & 27.9 & 5.82 & 17.21 & 15.15 & 16.49 & 49.86 \\
Chef & 22.14 & 20.16 & 6.56 & 17.3 & 16.6 & 17.24 & 50.64 \\
Cleaner & 15.84 & 19.35 & 13.79 & 19.78 & 15.2 & 16.04 & 49.09 \\
Computer Engineer & 17.22 & 21.2 & 7.37 & 21.65 & 16.27 & 16.29 & 49.77 \\
Dietitian & 17.3 & 26.87 & 6.31 & 16.61 & 16.31 & 16.6 & 50.13 \\
Doctor & 16.85 & 27.68 & 5.66 & 17.3 & 15.07 & 17.45 & 50.37 \\
Fashion Model & 17.16 & 16.58 & 14.76 & 18.15 & 16.54 & 16.81 & 50.65 \\
Firefighter & 21.23 & 17.09 & 12.03 & 16.2 & 16.02 & 17.43 & 50.29 \\
Garbage Collector & 17.33 & 24.01 & 7.84 & 17.76 & 16.25 & 16.8 & 49.47 \\
Geologist & 19.2 & 24.12 & 7.27 & 17.01 & 16.09 & 16.31 & 49.58 \\
Janitor & 16.43 & 17.43 & 13.25 & 20.41 & 15.9 & 16.59 & 49.51 \\
Journalist & 17.25 & 26.12 & 5.67 & 17.57 & 16.73 & 16.66 & 48.62 \\
Lawyer & 18.08 & 25.82 & 6.64 & 16.75 & 15.84 & 16.87 & 49.59 \\
Manager & 17.1 & 26.82 & 6.47 & 17.13 & 16.15 & 16.33 & 49.78 \\
Math Scientist & 17.1 & 21.92 & 6.81 & 22.08 & 15.76 & 16.33 & 51.83 \\
Musician & 15.7 & 21.79 & 10.58 & 18.41 & 16.18 & 17.35 & 51.29 \\
Nurse & 16.73 & 23.13 & 9.17 & 17.78 & 16.16 & 17.04 &52.2 \\
Pharmacist & 16.93 & 27.94 & 5.67 & 17.31 & 15.55 & 16.6 & 49.92 \\
Pilot & 19.36 & 23.97 & 7.07 & 16.38 & 16.74 & 16.47 & 49.86 \\
Professor & 17.22 & 24.98 & 7.11 & 18.07 & 15.96 & 16.65 & 50.14 \\
Programmer & 19.63 & 20.51 & 8.04 & 18.88 & 15.8 & 17.14 & 48.92 \\
Salesperson & 17.54 & 25.66 & 7.53 & 16.43 & 16.96 & 15.9 & 49.55 \\
Secretary & 17.63 & 26.42 & 6.44 & 18.27 & 16.05 & 15.19 & 51.82 \\
Security Guard & 15.87 & 15.75 & 14.51 & 17.66 & 16.41 & 19.8 & 48.78 \\
Singer & 16.51 & 20.77 & 12.21 & 16.77 & 17.01 & 16.73 & 49.96 \\
Soldier & 16.62 & 16.19 & 16.76 & 16.26 & 16.73 & 17.42 & 50.35 \\
Sushi chef & 18.27 & 11.28 & 21.58 & 15.92 & 16.91 & 16.05 & 50.89 \\
Tailor & 16.48 & 24.98 & 7.65 & 17.43 & 17.09 & 16.37 & 50.45 \\
Teacher & 16.76 & 26.36 & 6.32 & 17.23 & 16.41 & 16.92 & 51.25 \\
TikToker & 14.96 & 19.31 & 13.77 & 18.84 & 15.91 & 17.22 & 48.98 \\
TV Presenter & 18.58 & 25.21 & 7.06 & 16.06 & 16.82 & 16.27 & 50.35 \\
YouTuber & 17.07 & 22.26 & 10.39 & 16.49 & 16.26 & 17.52 & 49.37 \\
    \bottomrule
\end{tabular}
\caption{SDXL-Inc's race and gender distribution per profession.}
\label{tab:counts3}
\end{table}

\clearpage
\begin{table}[!ht]
    \centering
    \begin{tabular}{l|cccccc|c}
    \toprule
Profession   &  White   &  M.E.   &  Latinx   &  Indian   &  Asian   &  Black   &  Female   \\
\midrule
Accountant         & 17.07 & 22.29 & 10.19 & 17.4 & 16.49 & 16.57 & 50.04 \\  
Chef               & 17.46 & 19.55 & 13.2 & 16.37 & 16.71 & 16.71 & 49.79 \\
Cleaner            & 15.7 & 16.54 & 14.51 & 19.58 & 16.79 & 16.88 & 50.21 \\  
Comp. Engineer     & 15.48 & 19.8 & 9.9 & 20.98 & 17.09 & 16.75 & 49.66 \\ 
Dietitian          & 17.16 & 22.31 & 10.61 & 16.83 & 16.5 & 16.58 & 50.75 \\ 
Doctor             & 16.86 & 23.5 & 9.14 & 16.94 & 16.69 & 16.86 & 50.0 \\  
Fashion Model      & 16.78 & 16.02 & 13.51 & 19.88 & 16.78 & 17.03 & 50.0 \\ 
Journalist         & 17.7 & 21.42 & 10.11 & 17.14 & 17.46 & 16.17 & 49.15 \\  
Lawyer             & 16.64 & 20.53 & 11.67 & 17.05 & 17.3 & 16.8 & 49.92 \\ 
Manager            & 16.85 & 22.63 & 10.31 & 16.85 & 16.6 & 16.76 & 50.04 \\ 
Math Scientist     & 16.3 & 19.24 & 10.84 & 19.92 & 16.72 & 16.97 & 50.0 \\
Nurse              & 16.43 & 20.6 & 12.01 & 17.43 & 16.76 & 16.76 & 49.79 \\ 
Pharmacist         & 16.57 & 23.15 & 9.91 & 16.9 & 16.82 & 16.65 & 49.88 \\  
Pilot              & 16.93 & 17.6 & 14.93 & 16.85 & 16.68 & 17.01 & 50.04 \\
Professor          & 17.2 & 19.03 & 11.77 & 18.45 & 16.78 & 16.78 & 50.0 \\ 
Programmer         & 17.45 & 19.9 & 10.88 & 18.63 & 16.19 & 16.95 & 50.34 \\ 
Secretary          & 16.68 & 21.97 & 10.16 & 17.01 & 16.43 & 17.75 & 50.78 \\  
Security Guard     & 15.77 & 14.0 & 17.62 & 17.96 & 16.86 & 17.79 & 50.42 \\ 
Singer             & 16.52 & 19.64 & 12.9 & 17.83 & 16.35 & 16.76 & 50.04 \\  
Soldier            & 16.46 & 14.2 & 17.38 & 17.96 & 16.71 & 17.29 & 50.13 \\
Sushi Chef         & 16.77 & 16.6 & 15.56 & 16.34 & 17.54 & 17.2 & 49.7 \\ 
Tailor             & 16.47 & 21.44 & 11.75 & 16.89 & 16.56 & 16.89 & 50.08 \\  
Teacher            & 16.78 & 20.3 & 11.91 & 17.7 & 16.61 & 16.69 & 50.17 \\
TikToker           & 17.1 & 18.17 & 12.53 & 18.26 & 16.6 & 17.34 & 50.21 \\ 
YouTuber           & 17.47 & 21.4 & 10.87 & 16.64 & 16.81 & 16.81 & 50.25 \\
    \bottomrule
\end{tabular}
\caption{GPT-in-the-loop SDXL race and gender distribution per profession.}
\label{tab:counts4}
\end{table}

\clearpage
\section*{Supplementary Note 1}
\label{supplementary_note_1}

\begin{figure}
\begin{flushleft} \textbf{a} \end{flushleft}
\vspace{-10pt}
\includegraphics[trim={0 2.8cm 0 0},clip,width=0.445\textwidth]{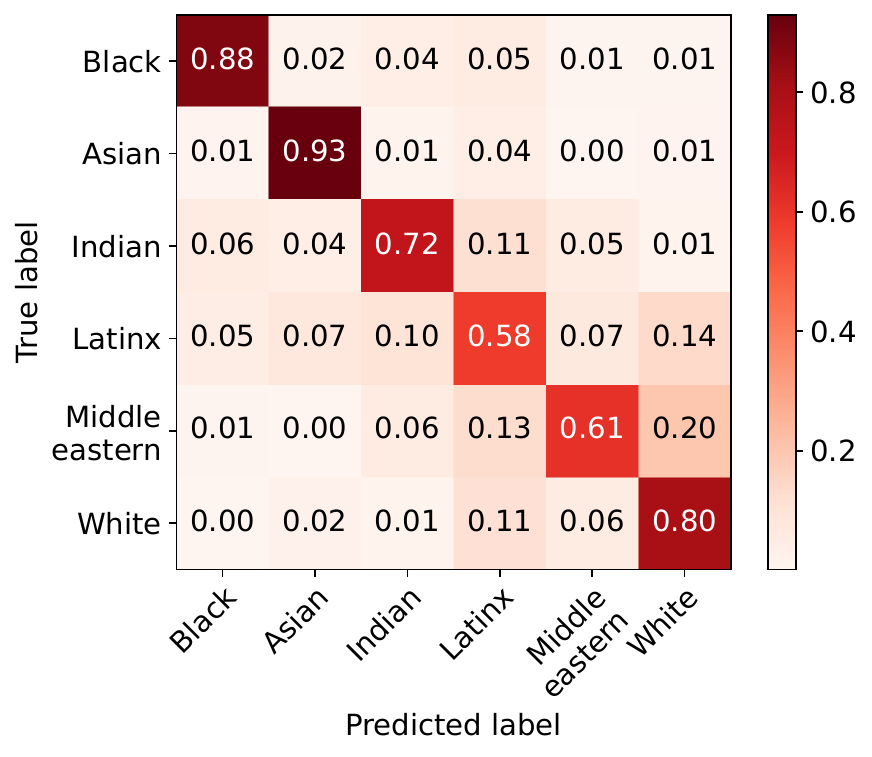}
\includegraphics[trim={2.8cm 2.8cm 0 0},clip,width=0.3\textwidth]{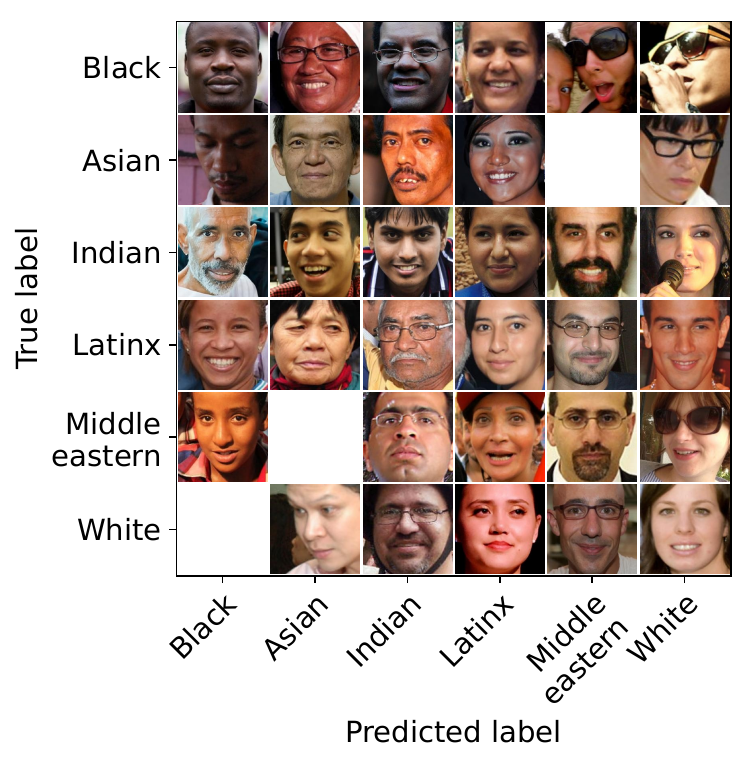}
\begin{subfigure}{0.215\textwidth}
    \includegraphics[trim={1.25cm 5.3cm 0 0},clip,width=1\textwidth]{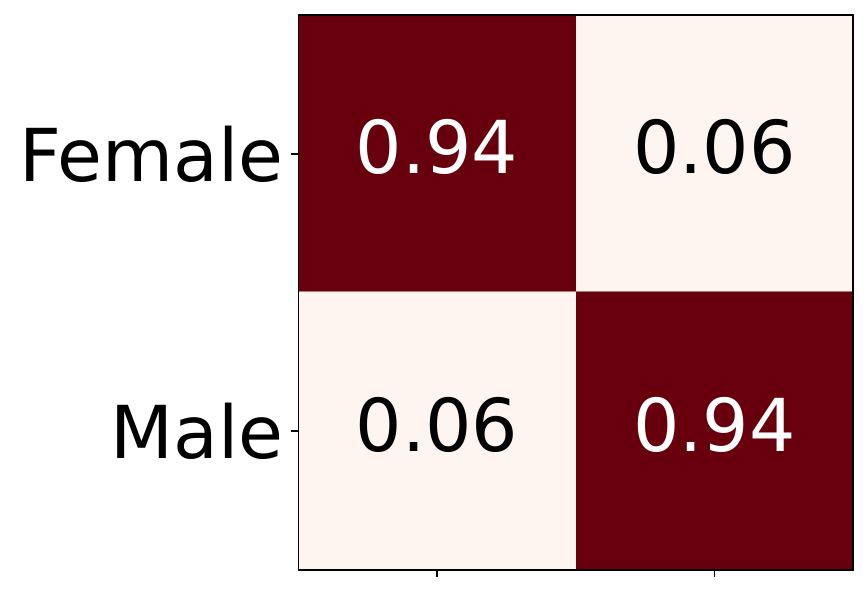}
    \includegraphics[trim={1.25cm 5.3cm 0 0},clip,width=1\textwidth]{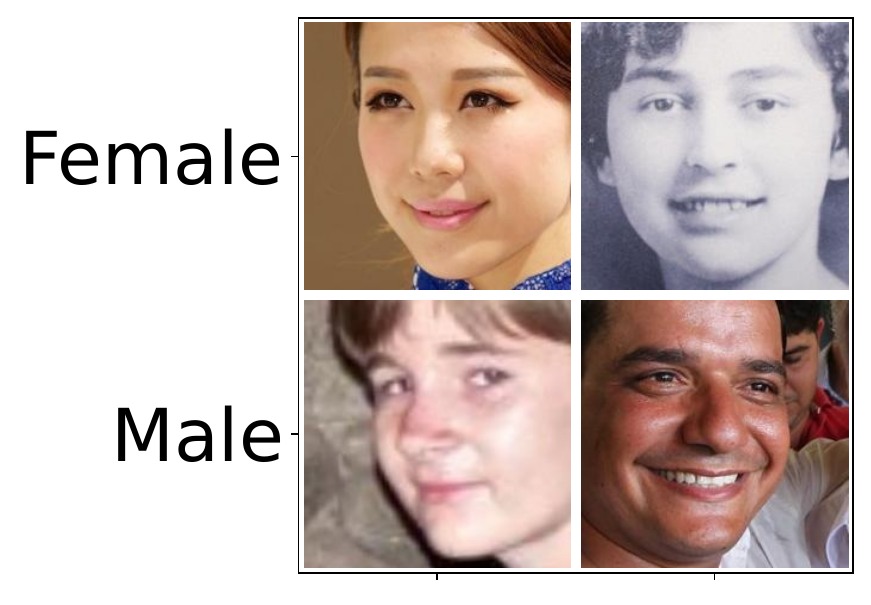}
\end{subfigure}
\\
\vspace{-40pt}
\begin{flushleft} \textbf{b} \end{flushleft}
\vspace{-10pt}
\includegraphics[width=0.445\textwidth]{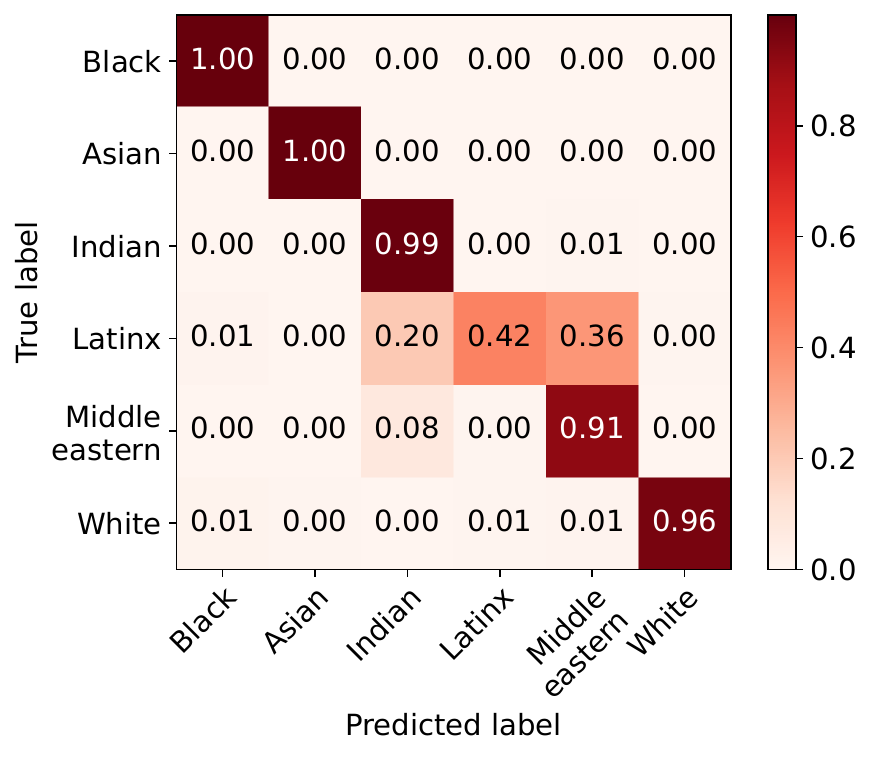}
\includegraphics[trim={2.8cm 0 0 0},clip,width=0.3\textwidth]{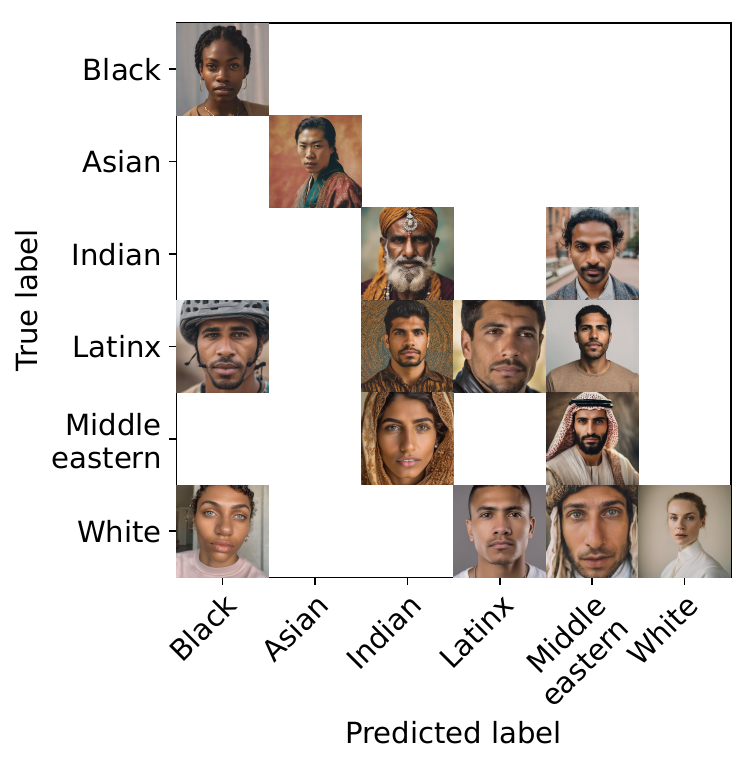}
\begin{subfigure}{0.215\textwidth}
    \includegraphics[trim={1.2cm 5.3cm 0 0},clip,width=1\textwidth]{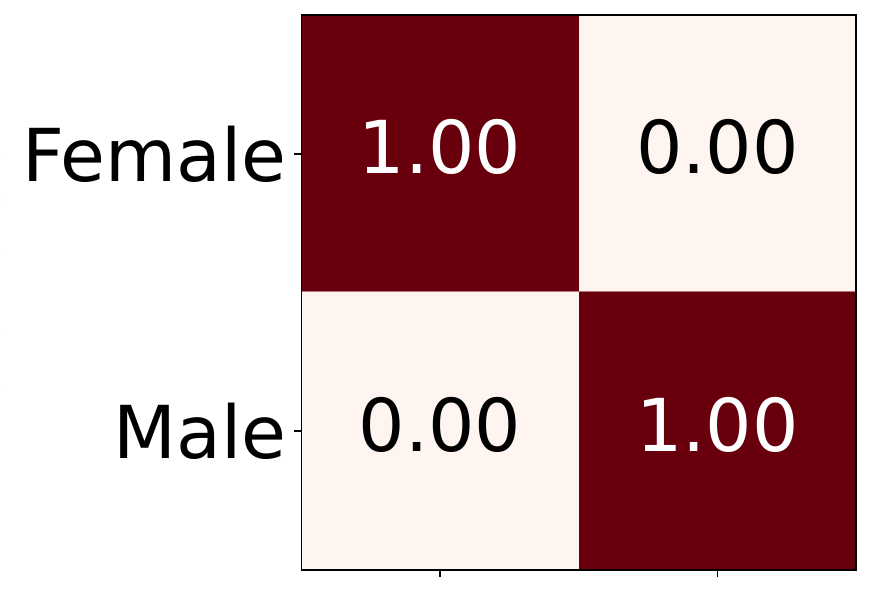}
    \includegraphics[trim={1.2cm 0 0 0},clip,width=1\textwidth]{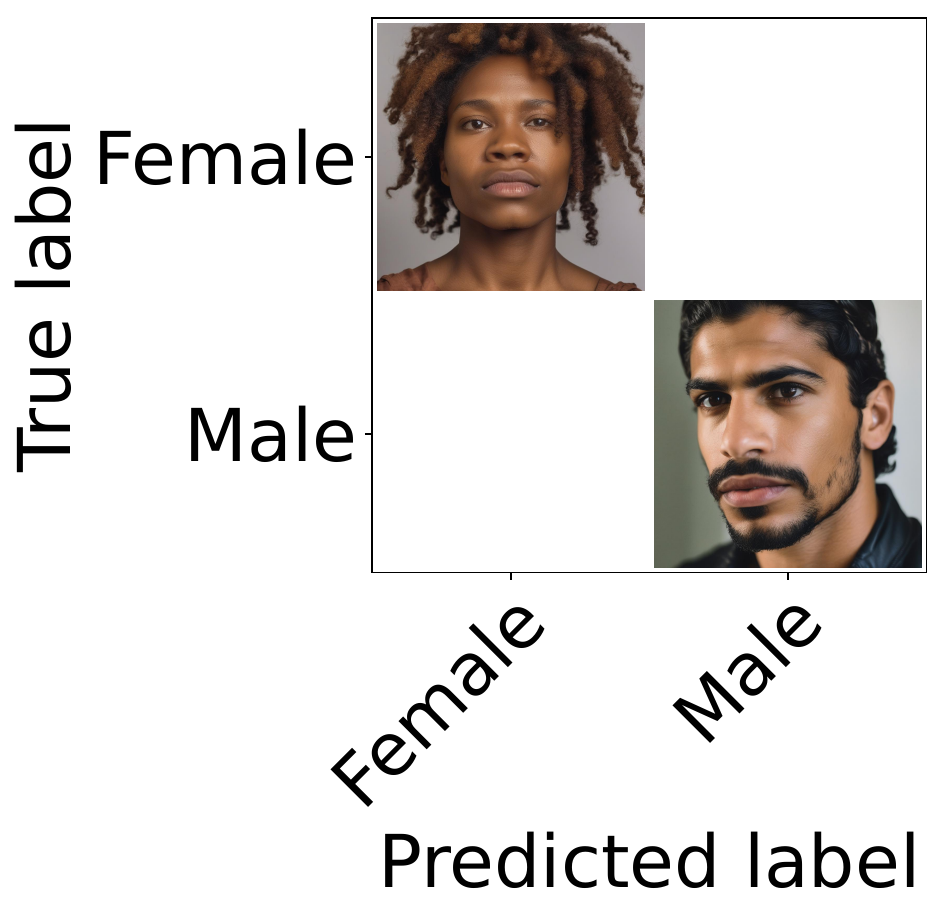}
\end{subfigure}
\caption{
{\fontsize{11}{11}\selectfont{
\textbf{Evaluating our classifier.} 
\textbf{a}, Confusion matrices of race and gender prediction based on the FairFace validation set, along with sample images representing different cells.
\textbf{b}, Similar to (\textbf{a}), but based on images generated by Stable Diffusion XL (SDXL).
}}}
\label{fig:model}
\end{figure}

Figure~\ref{fig:model}a depicts the confusion matrices corresponding to race and gender, along with sample images representing each cell. As can be seen, Asian is the easiest race to predict, followed by Black. In contrast, the hardest races to predict are Latinx and Middle Eastern. On the other hand, there is no difference in performance when it comes to predicting male vs.\ female images.

We reproduced the same confusion matrices, but using images generated by Stable Diffusion XL (SDXL) which has been shown to outperform previous versions of Stable Diffusion~\cite{podell2023sdxl}. More specifically, we generated 10,000 images per race using the following prompt: ``\textit{a photo of $X$ person, looking at the camera, closeup headshot facing forward, ultra quality, sharp focus}'', where $X\in$ \{a Black, a White, an Asian, an Indian, a Latinx or Hispanic, a Middle Eastern\}. As for gender, we generated 10,000 images per gender using the same prompt as before, but with $X\in$ \{male, female\}. The results of this evaluation are depicted in the first two rows of Table~\ref{tab:SD_table}, showing that our classifier works better with images from SDXL than those from FairFace, regardless of the performance measure. This indicates that our classifier is particularly suited to examine the racial-, and gender-composition of images generated by SDXL.

The confusion matrices depicted in Figure~\ref{fig:model}b reveal that all races are significantly easier to predict in the SDXL dataset compared to its FairFace counterpart, apart from Latinx which is significantly harder (see how the numbers along the diagonal are greater in Figure~\ref{fig:model}b than in Figure~\ref{fig:model}a, except for Latinx). One possible explanation could be that the dataset used to train SDXL contains several images that are labelled as Latinx but are highly similar to Indian and Middle Eastern individuals. In terms of gender, our classifier achieves a perfect score for both male and female.

\begin{table}[!htbp]
\centering
\caption{A comparison between our \textbf{race classifier} and other alternatives from the literature, using FairFace's validation set and FairFace's seven-race classification. Bold font highlights the highest score(s).}
\label{tab:comparison_race_table}
\begin{tabular}{|l|c|c|c|c|}
\hline
                          & \textbf{Accuracy}  & \textbf{Precision} & \textbf{Recall} & \textbf{F1 score} \\ \hline
Our classifier            & \textbf{73\%}           & \textbf{72\%}           & \textbf{72\%}        & \textbf{72\% }         \\ \hline
CLIP's zero-shot classifier & 64\%           & 67\%           & 65\%        & 65\%          \\  \hline
Google's FaceNet + SVM   &69\% 
          & 69\%          & 68\%     &
68\%              \\ \hline

FairFace's (ResNet34) classifier           
& 72\%           & \textbf{72\%}        & 71\%  & \textbf{72\%}         \\ \hline

EfficientNet-B7 (tuning all layers) 

&  70\%   &  70\%   &   70\%   &   70\%   \\ \hline

Large Vision Transformer VIT (tuning all layers) 
&  71\%   &  72\%   &   70\%   &   71\%   \\ \hline

\end{tabular}
\end{table}

\begin{table}[!htbp]
\centering
\caption{A comparison between our \textbf{gender classifier} and other alternatives from the literature, using FairFace's validation set. Bold font highlights the highest score(s).}
\label{tab:comparison_gender_table}
\begin{tabular}{|l|c|c|c|c|}
\hline
                          & \textbf{Accuracy}  & \textbf{Precision} & \textbf{Recall} & \textbf{F1 score} \\  \hline
Our classifier  & \textbf{94\%}                   & \textbf{94\%}  & \textbf{94\%}  & \textbf{94\%}   \\ \hline
FairFace's (ResNet34) classifier & \textbf{94\%}  & \textbf{94\%}  & \textbf{94\%}  & \textbf{94\%}  \\ \hline
CLIP's zero-shot classifier  & \textbf{94\%}  & \textbf{94\%}  & \textbf{94\%}  & \textbf{94\%}  \\ \hline

\end{tabular}
\end{table}





\begin{table}[htbp]
\caption{Evaluating our classifier's ability to predict race and gender using SDXL images and our six racial groups. Bold font highlights the highest score.}
\label{tab:SD_table}
\centering
\begin{tabular}{|c|c|c|c|c|c|c|c|c|}
\hline
 \multirow{2}{*}{}  & 
 \multicolumn{2}{c|}{\textbf{Accuracy}} &
  \multicolumn{2}{c|}{\textbf{Percision}} &
  \multicolumn{2}{c|}{\textbf{Recall}} &
  \multicolumn{2}{c|}{\textbf{F1 score}} \\ 
       & \textbf{FairFace} & \textbf{SDXL} & \textbf{FairFace} & \textbf{SDXL} & \textbf{FairFace} & \textbf{SDXL} & \textbf{FairFace} & \textbf{SDXL} \\ \hline
Racial  &   78\% & \textbf{88\%}   &   76\%   & \textbf{90\% }    &   75\%   & \textbf{88\%}  &  76\%    & \textbf{87\%}    \\ \hline
Gender  &   94\%  & \textbf{100\%}  &    94\%    & \textbf{100\% }  &   94\%   & \textbf{100\%}    &    94\%  & \textbf{100\%}     \\ \hline
\end{tabular}
\end{table}

\section*{Supplementary Note 2}
\label{supplementary_note_2}
In this section, we compare our solution (SDXL-Inc) to another recently proposed alternative, namely ITI-GEN~\cite{zhang2023iti}; see the Discussion section for a brief description of this model, and see the Materials and Methods for details on how we retrained it. In their official GitHub repository~\cite{ITI_GEN_github}, the authors imply that their solution may struggle with certain generalizations. More specifically, they state that the training prompt (e.g., a headshot of a person) and the inference prompt (e.g., a headshot of a doctor) ``should not differ a lot''. To compare the degree to which both models (ITI-GEN and SDXL-Inc) can be generalized, we used each of them to generate 1,200 images for each of the following five prompts: 1) a headshot of a person with the Eiffel Tower in the background; 2) a headshot of a mechanic with a guitar; 3) a headshot of a skillful trainer with a pet tiger; 4) a headshot of a happy and healthy family; and 5) a headshot of a person with green hair and eyeglasses.

\begin{figure}[htbp]
\includegraphics[width=1\textwidth]{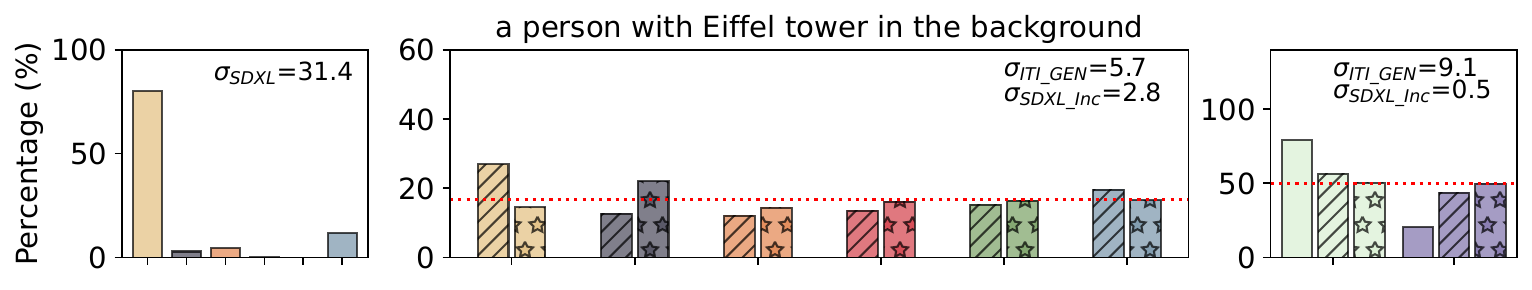}
\includegraphics[width=1\textwidth]{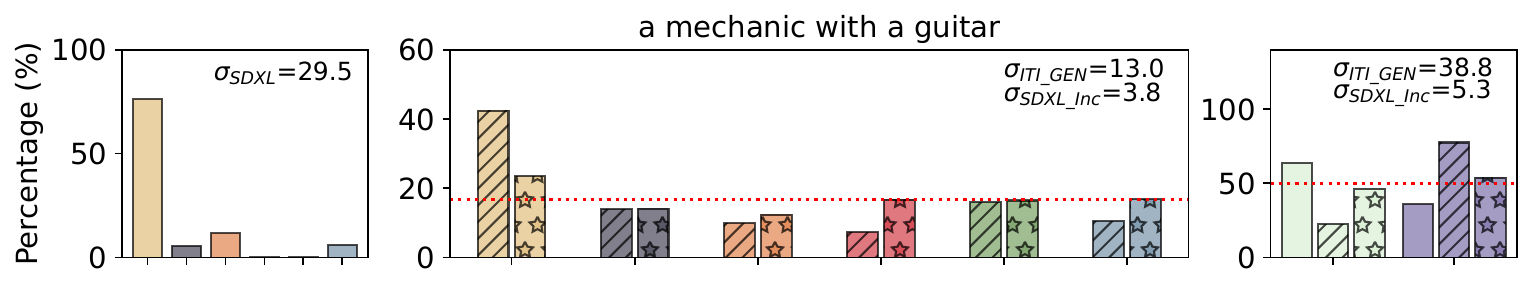}
\includegraphics[width=1\textwidth]{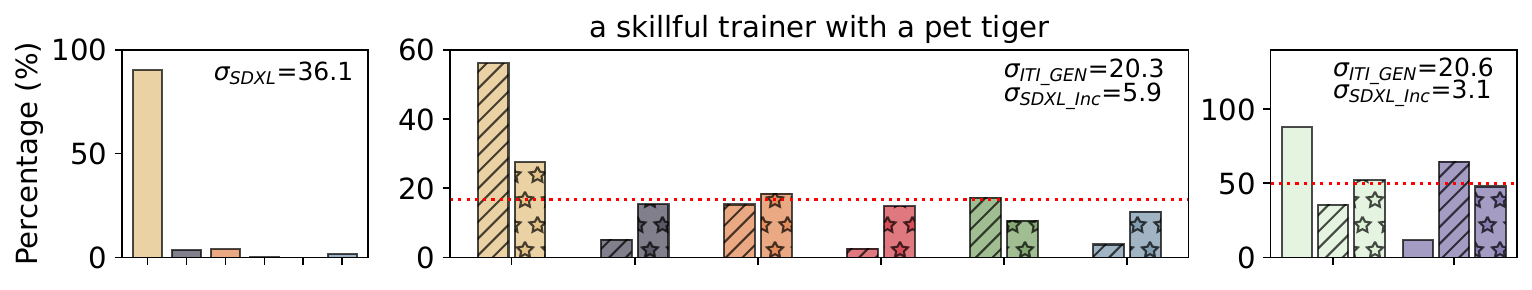}
\includegraphics[width=1\textwidth]{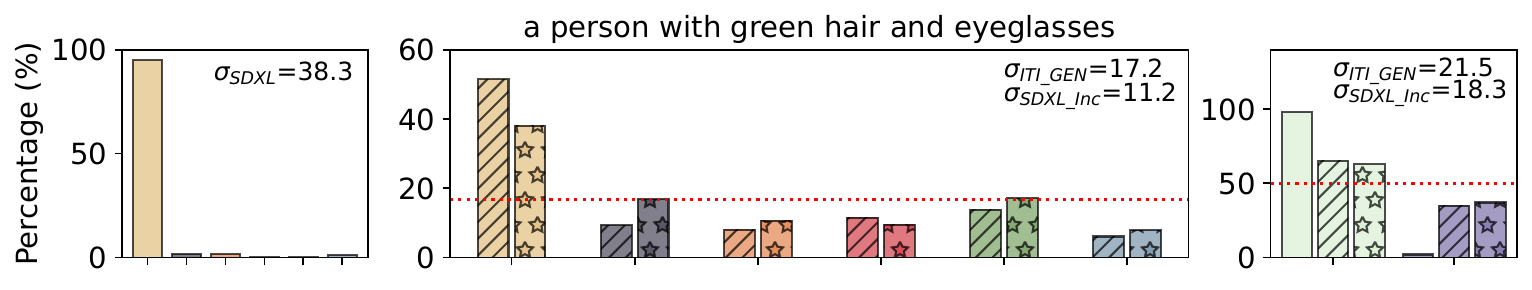}
\includegraphics[width=1\textwidth]{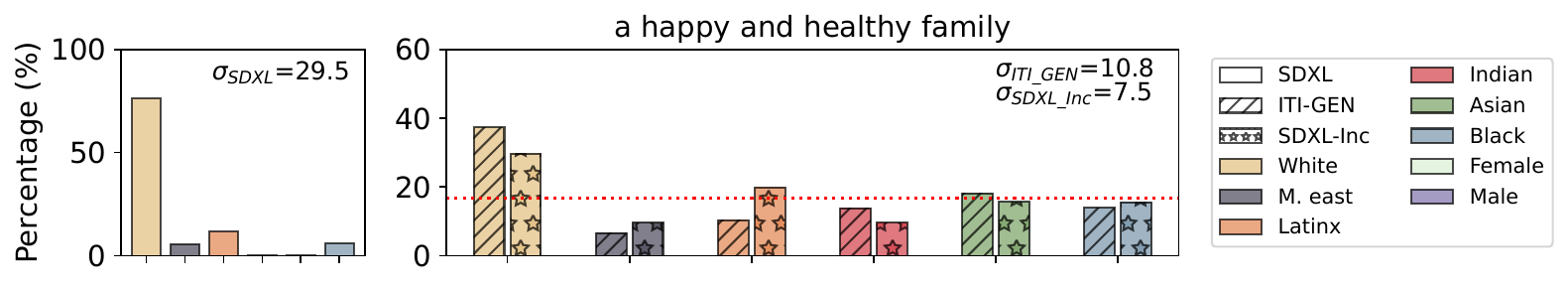}
\caption{
{\fontsize{11}{11}\selectfont{
\textbf{SDXL vs.\ ITI-GEN vs.\ SDXL-Inc} Comparing the distribution of race and gender across the three models, given five prompts. Each row corresponds to a different prompt. The standard deviation(s) corresponding to each subplot is denoted by $\sigma$ followed by a subscript indicating the model. Gender results are omitted from the bottom row since the generated images depict families rather than individuals.
}}}
\label{fig:prompts_comp}
\end{figure}

Figure~\ref{fig:prompts_comp} depicts the race and gender distribution of SDXL, SDXL-Inc, and ITI-GEN for each of the five prompts. As can be seen in the left column, SDXL is extremely biased, depicting the vast majority of images as White in all prompts. As for ITI-GEN (middle column, dashed bars), it manages to reduce a substantial amount of the bias compared to SDXL. Nevertheless, the percentage of images depicting White individuals remains $\geq$40\% in four out of the five prompts. Our SDXL-Inc (middle column, starred bars) yields a more even distribution of images across races compared to ITI-GEN; see how the standard deviations for our solution are smaller than those of ITI-GEN in all five prompts. As for gender (right column), we can see that SDXL is the most biased, ITI-GEN reduces a considerable amount of bias, and SDXL-Inc further reduces bias compared to ITI-GEN across all prompts; see the corresponding standard deviations.

One reason behind these observed differences in performance could be the fact that the models occasionally intend to generate images of a certain race, but accidentally end up generating images of a different race (note that both SDXL-Inc and ITI-GEN contain steps designed to generate an image of certain races). If these steps work perfectly well, we would expect the models to achieve a perfect score in terms of accuracy, recall, precision, and F1 score. To test this possibility, we used out classifier to label the 1,200 images generated by each solution (200 per race) for each of the five prompts. The result of this analysis is summarized in Table~\ref{tab:perf_metrics}. As can be seen, SDXL-Inc outperforms ITI-GEN across all metrics in all five prompts. This suggests that the observed difference in performance between the two solutions could be (at least partially) explained by the fact that the error rate in generating the intended race(s) is higher in ITI-GEN than in SDXL-Inc.

\begin{table}[htbp]
\caption{Comparing SDXL-Inc to ITI-GEN in terms of accuracy, recall, precision, and F1 score. For each performance measure, the highest score is highlighted in bold.} 
\label{tab:perf_metrics}
\scalebox{0.96}{
\centering
\begin{tabular}{|l|c|c||c|c||c|c||c|c|}
\hline
\textbf{} &
  \multicolumn{2}{c||}{\textbf{Accuracy}} &
  \multicolumn{2}{c||}{\textbf{Recall}} &
  \multicolumn{2}{c||}{\textbf{Percision}} &
  \multicolumn{2}{c|}{\textbf{F1 score}} \\ 
  \multirow{2}{*}{\textbf{Prompt}} &
  \begin{tabular}[c]{@{}l@{}}\textbf{SDXL-}\\ \textbf{Inc} \end{tabular} &
  \begin{tabular}[c]{@{}l@{}}\textbf{ITI-}\\ \textbf{GEN} \end{tabular} &
  \begin{tabular}[c]{@{}l@{}}\textbf{SDXL-}\\ \textbf{Inc} \end{tabular} &
  \begin{tabular}[c]{@{}l@{}}\textbf{ITI-}\\ \textbf{GEN} \end{tabular} &
  \begin{tabular}[c]{@{}l@{}}\textbf{SDXL-}\\ \textbf{Inc} \end{tabular} &
  \begin{tabular}[c]{@{}l@{}}\textbf{ITI-}\\ \textbf{GEN} \end{tabular} &
  \begin{tabular}[c]{@{}l@{}}\textbf{SDXL-}\\ \textbf{Inc} \end{tabular} &
  \begin{tabular}[c]{@{}l@{}}\textbf{ITI-}\\ \textbf{GEN} \end{tabular} \\ \hline
  \hline
\begin{tabular}[c]{@{}l@{}}a person with Eiffel \\ tower in the background\end{tabular} & \textbf{87} & 70 & \textbf{87} & 70 & \textbf{89} & 71 & \textbf{87} & 67 \\ \hline
\begin{tabular}[c]{@{}l@{}}a mechanic with \\ a guitar\end{tabular}                     & \textbf{86} & 39 & \textbf{86} & 39 & \textbf{88} & 48 & \textbf{86} & 37 \\ \hline
\begin{tabular}[c]{@{}l@{}}a skillful trainer \\ with a pet tiger\end{tabular}  & \textbf{77} & 37 & \textbf{77} & 37 & \textbf{83} & 47 & \textbf{77} & 32 \\ \hline
\begin{tabular}[c]{@{}l@{}}a happy and healthy\\  family\end{tabular}                   & \textbf{70} & 66 & \textbf{70} & 66 & \textbf{75} & 68 & \textbf{70} & 64 \\ \hline
\begin{tabular}[c]{@{}l@{}}a person with green\\  hair and eyeglasses\end{tabular}      &  \textbf{60}  & 40   & \textbf{60}   & 40   &  \textbf{72}  & 55   &  \textbf{60}  & 38 \\ \hline \hline
\textbf{Average}   &  \textbf{76}  & 50.4   & \textbf{76}   & 50.4   &  \textbf{81.4}  & 57.8   &  \textbf{76}  & 47.6 \\ \hline \hline
\end{tabular}
}
\end{table}

\clearpage
\section*{Supplementary Note 3}
\label{supplementary_note_3}
 
\begin{figure}[htbp]   
\captionsetup[subfigure]{labelformat=empty}
    \centering    
    \begin{flushleft} \textbf{a} \end{flushleft}
    \begin{subfigure}{0.16\textwidth}
        \includegraphics[width=1\linewidth]{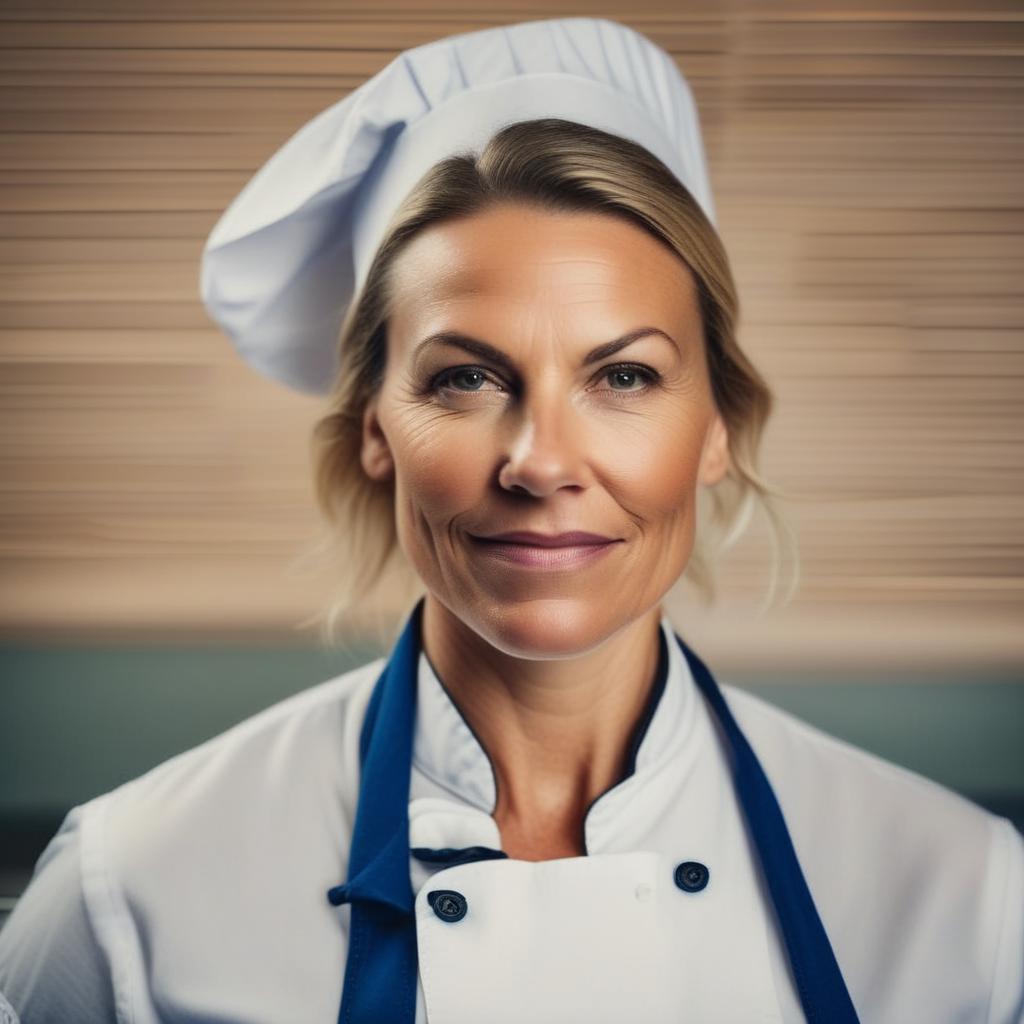}
    \end{subfigure}\hspace*{-0.25em}
    \begin{subfigure}{0.16\textwidth}
        \includegraphics[width=1\linewidth]{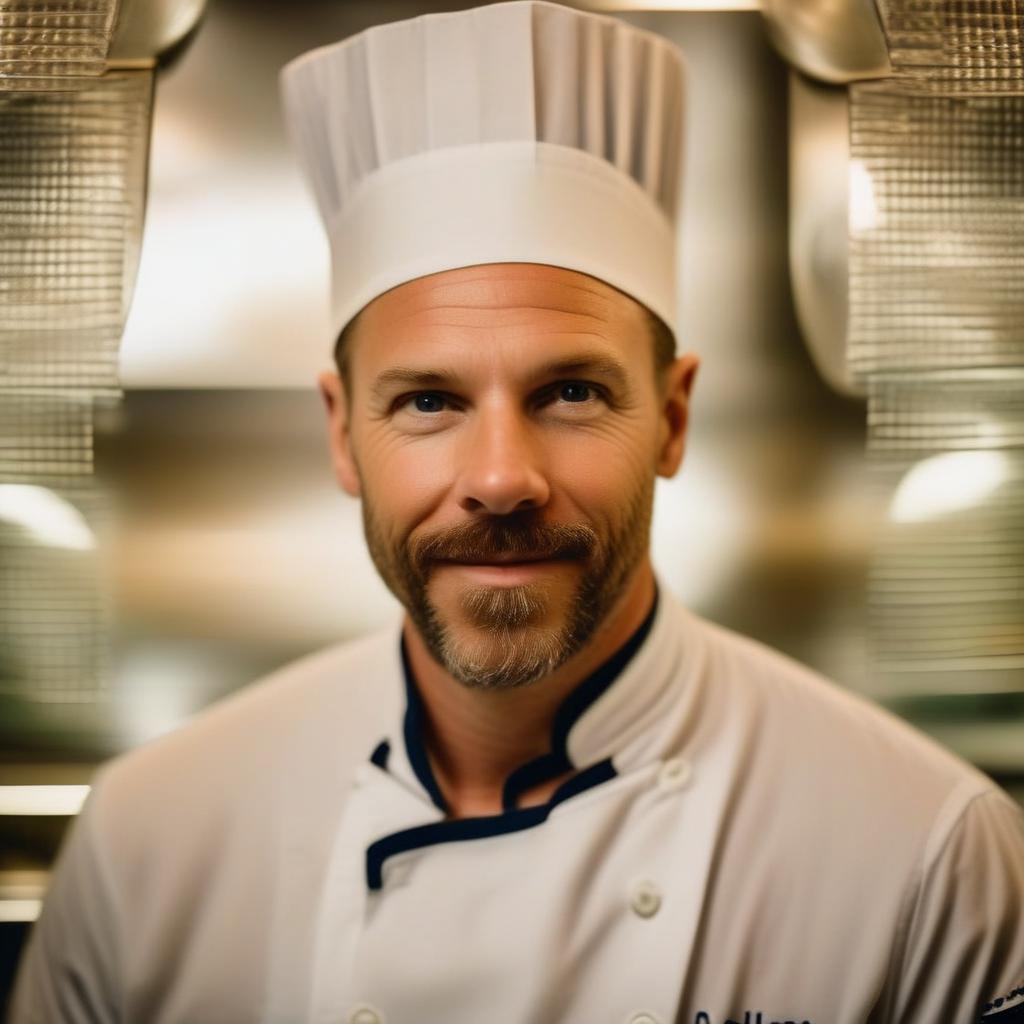}
    \end{subfigure}\hspace*{-0.25em}
    \begin{subfigure}{0.16\textwidth}
        \includegraphics[width=1\linewidth]{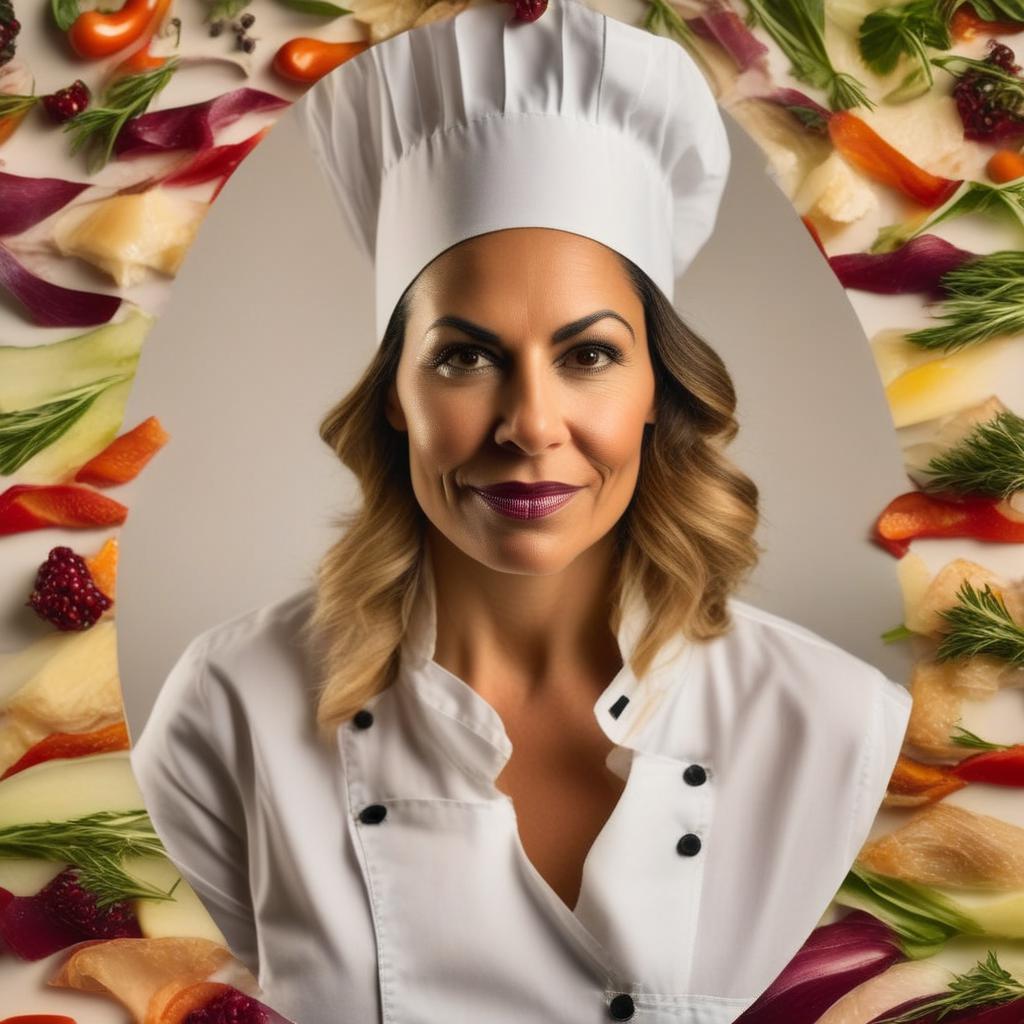}
    \end{subfigure}\hspace*{-0.25em}
    \begin{subfigure}{0.16\textwidth}
        \includegraphics[width=1\linewidth]{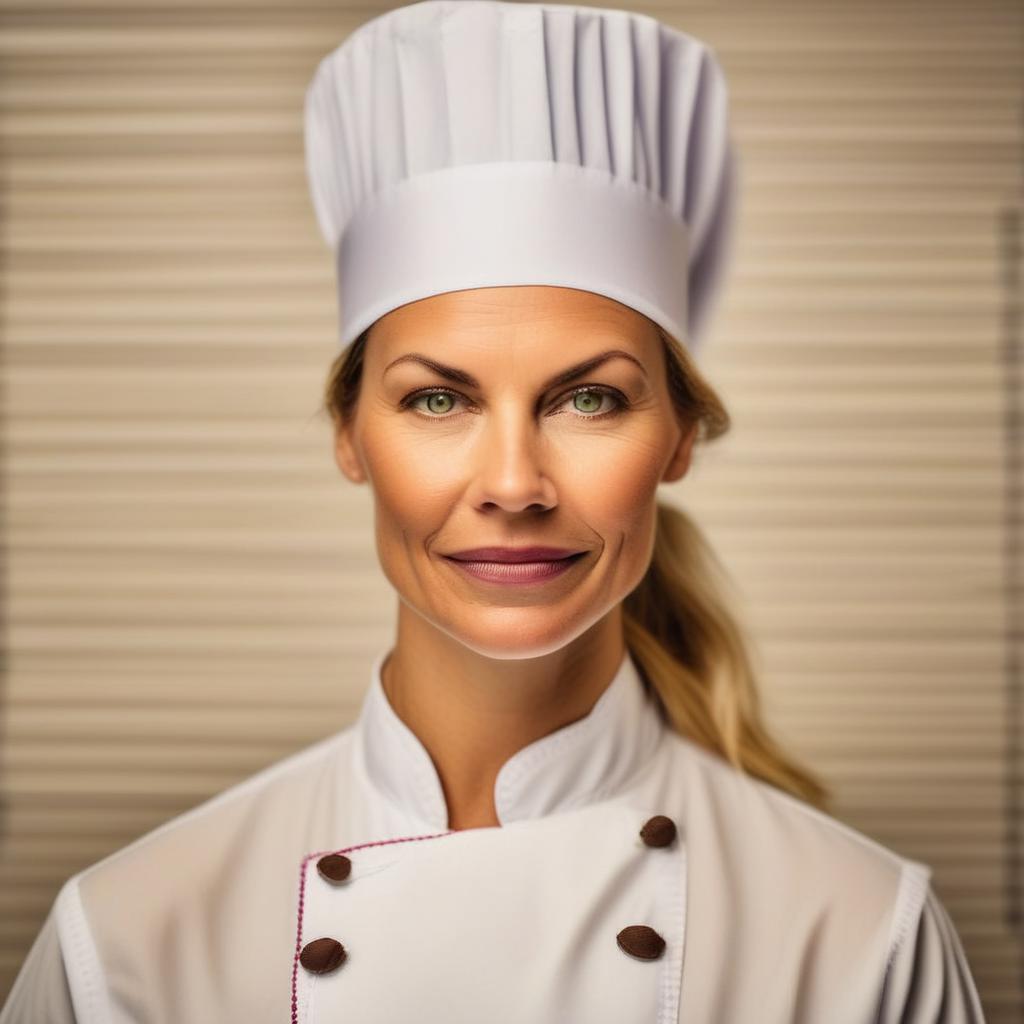}
    \end{subfigure}\hspace*{-0.25em}
    \begin{subfigure}{0.16\textwidth}
        \includegraphics[width=1\linewidth]{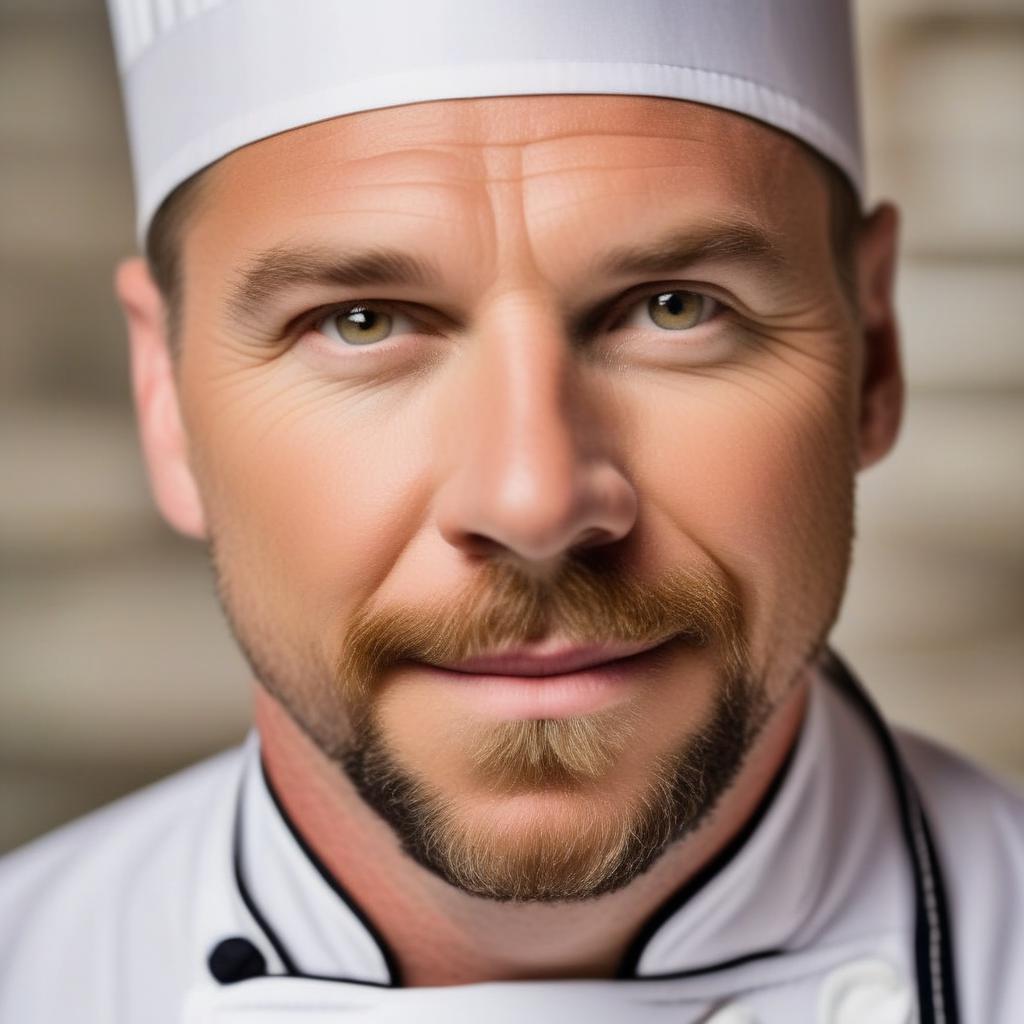}
    \end{subfigure}\hspace*{-0.25em}
    \begin{subfigure}{0.16\textwidth}
        \includegraphics[width=1\linewidth]{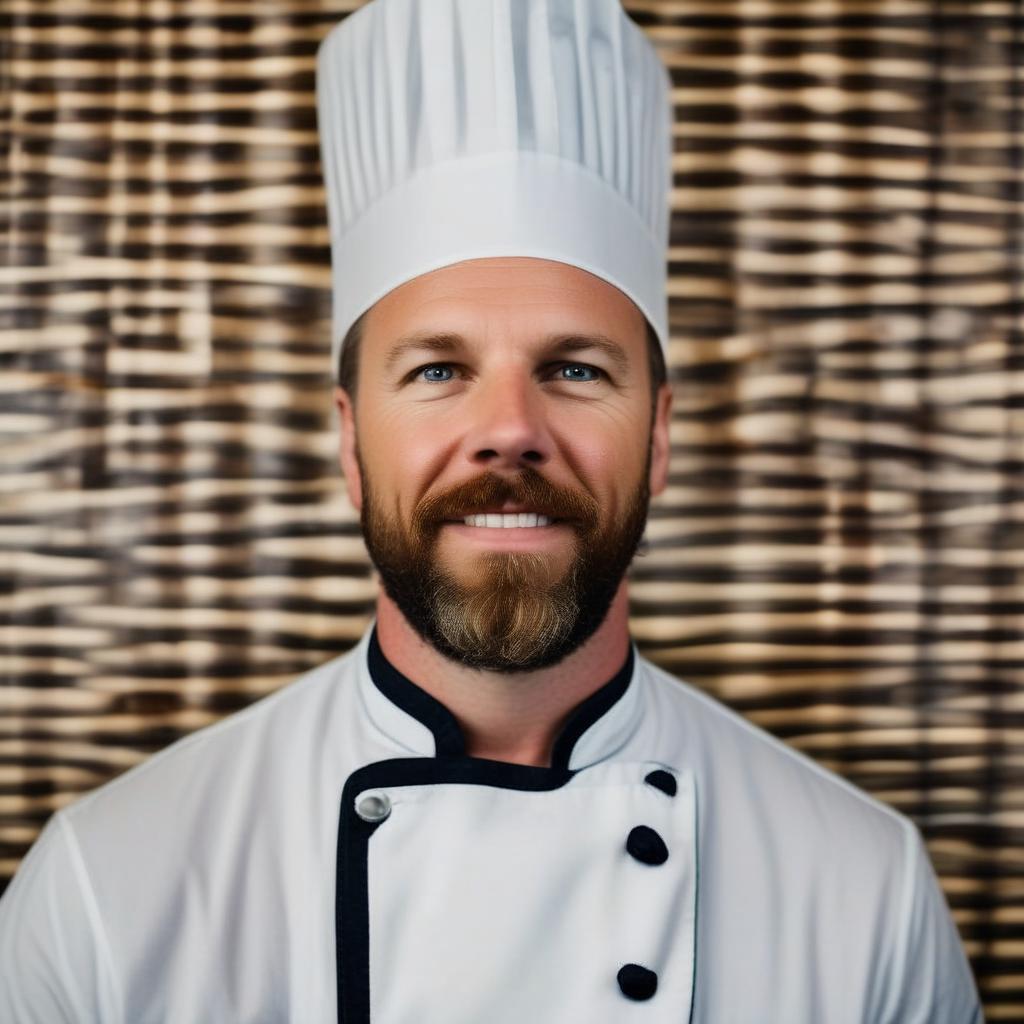}
    \end{subfigure}
    \begin{subfigure}{0.16\textwidth}
        \includegraphics[width=1\linewidth]{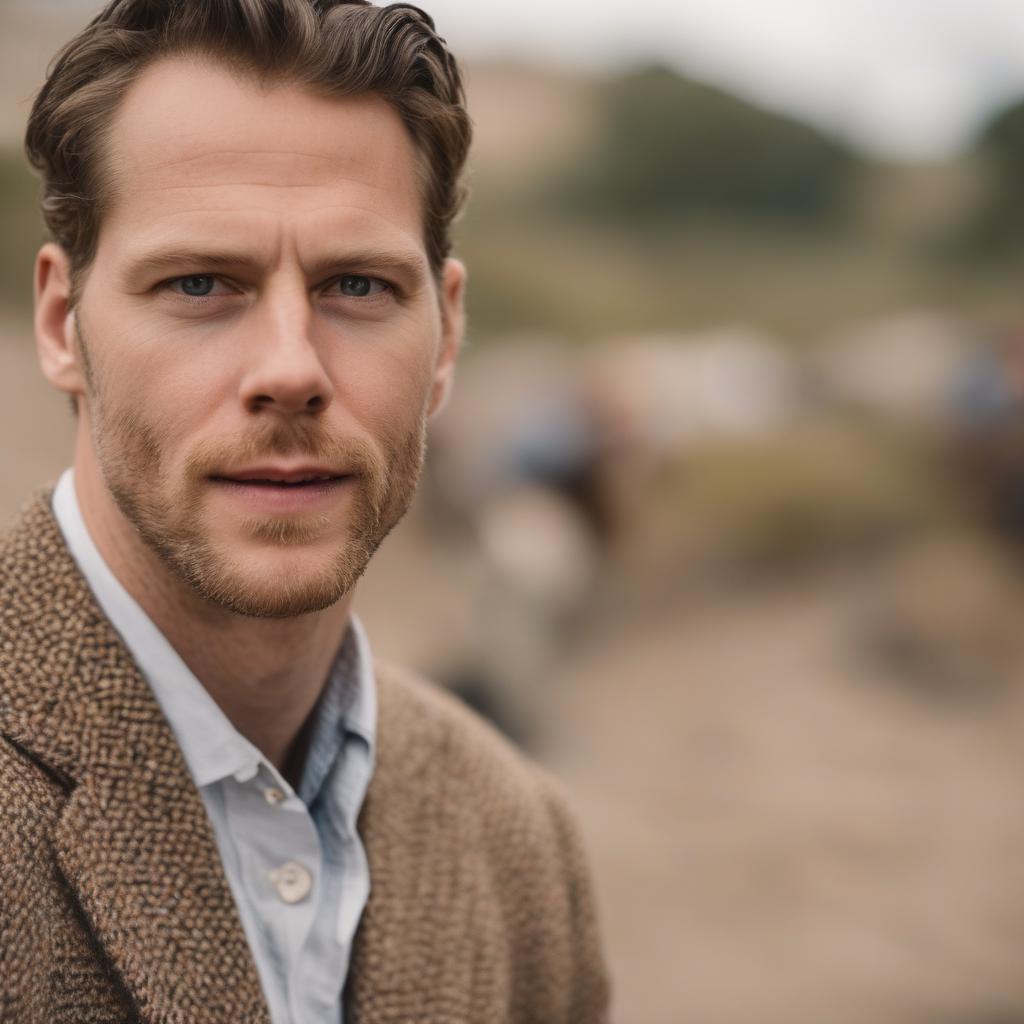}
    \end{subfigure}\hspace*{-0.25em}
    \begin{subfigure}{0.16\textwidth}
        \includegraphics[width=1\linewidth]{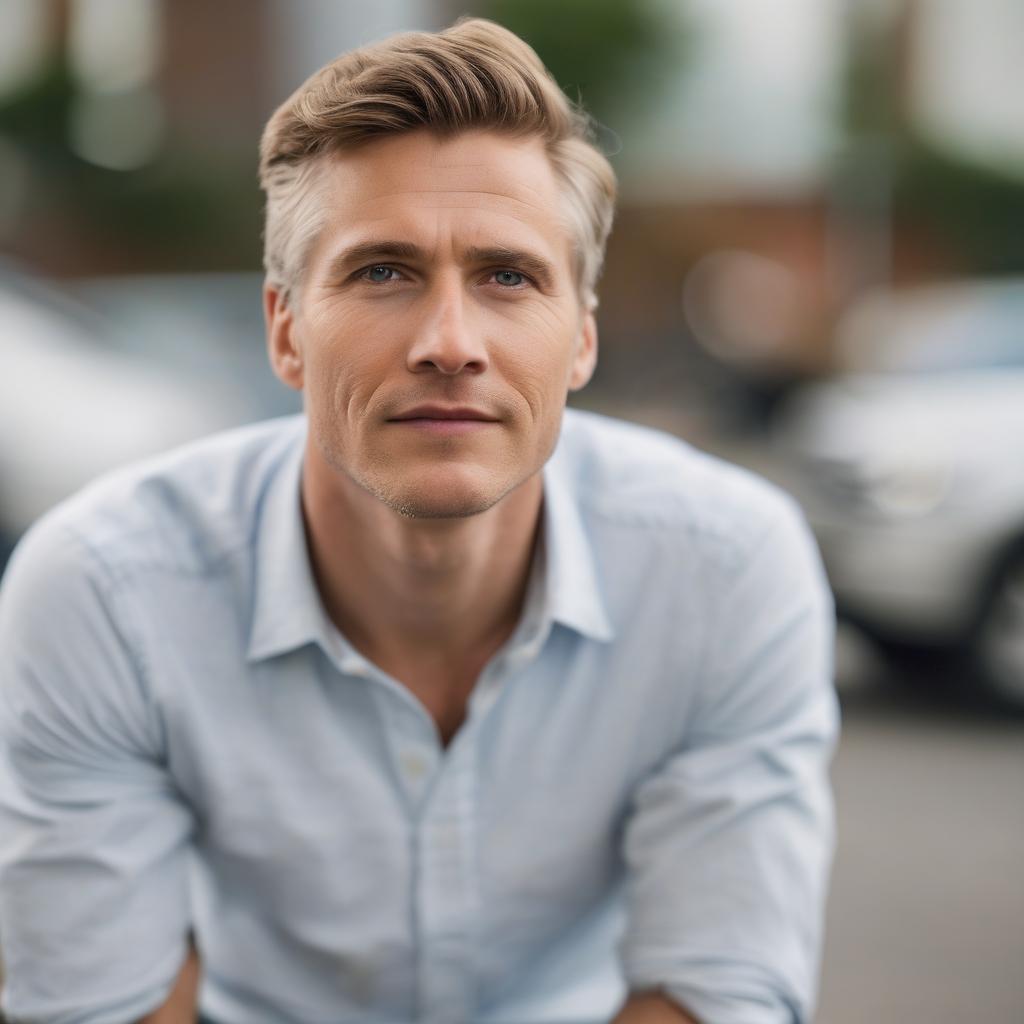}
    \end{subfigure}\hspace*{-0.25em}
    \begin{subfigure}{0.16\textwidth}
        \includegraphics[width=1\linewidth]{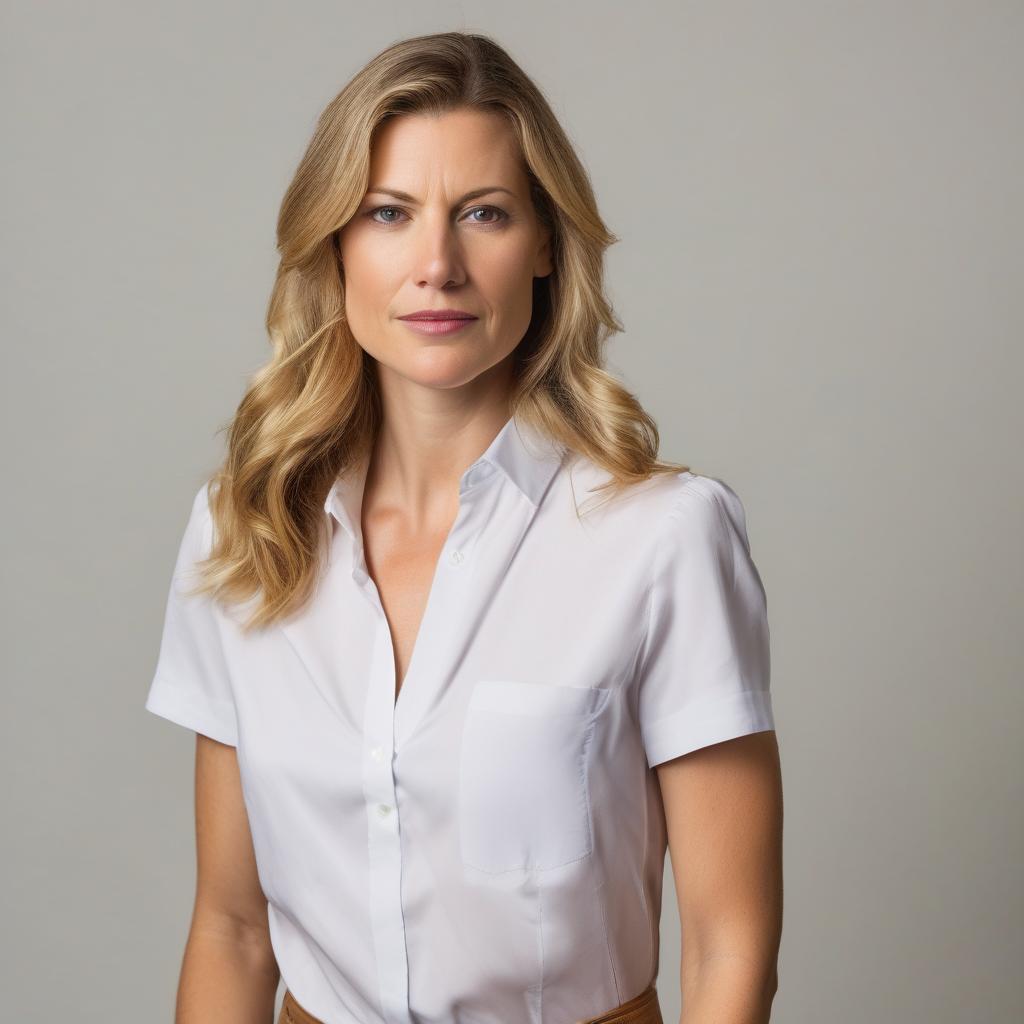}
    \end{subfigure}\hspace*{-0.25em}
    \begin{subfigure}{0.16\textwidth}
        \includegraphics[width=1\linewidth]{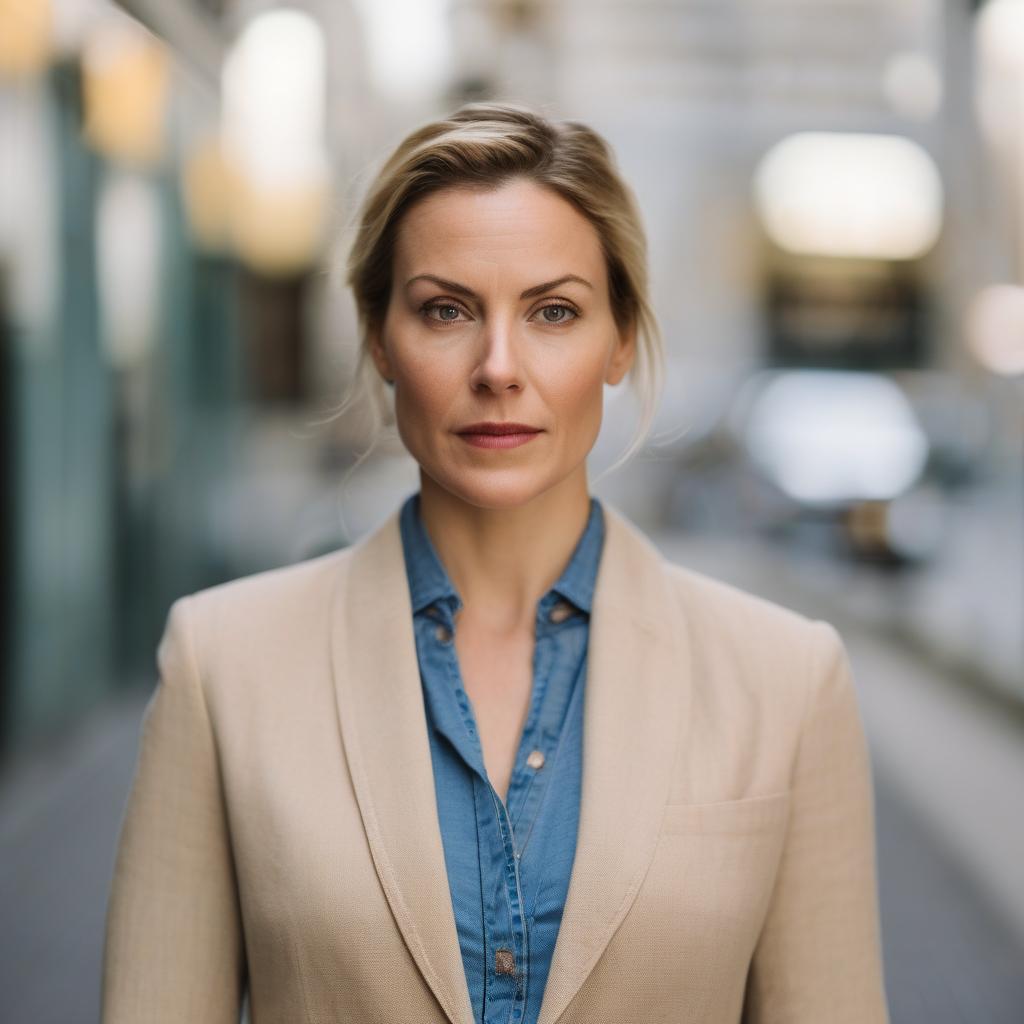}
    \end{subfigure}\hspace*{-0.25em}
    \begin{subfigure}{0.16\textwidth}
        \includegraphics[width=1\linewidth]{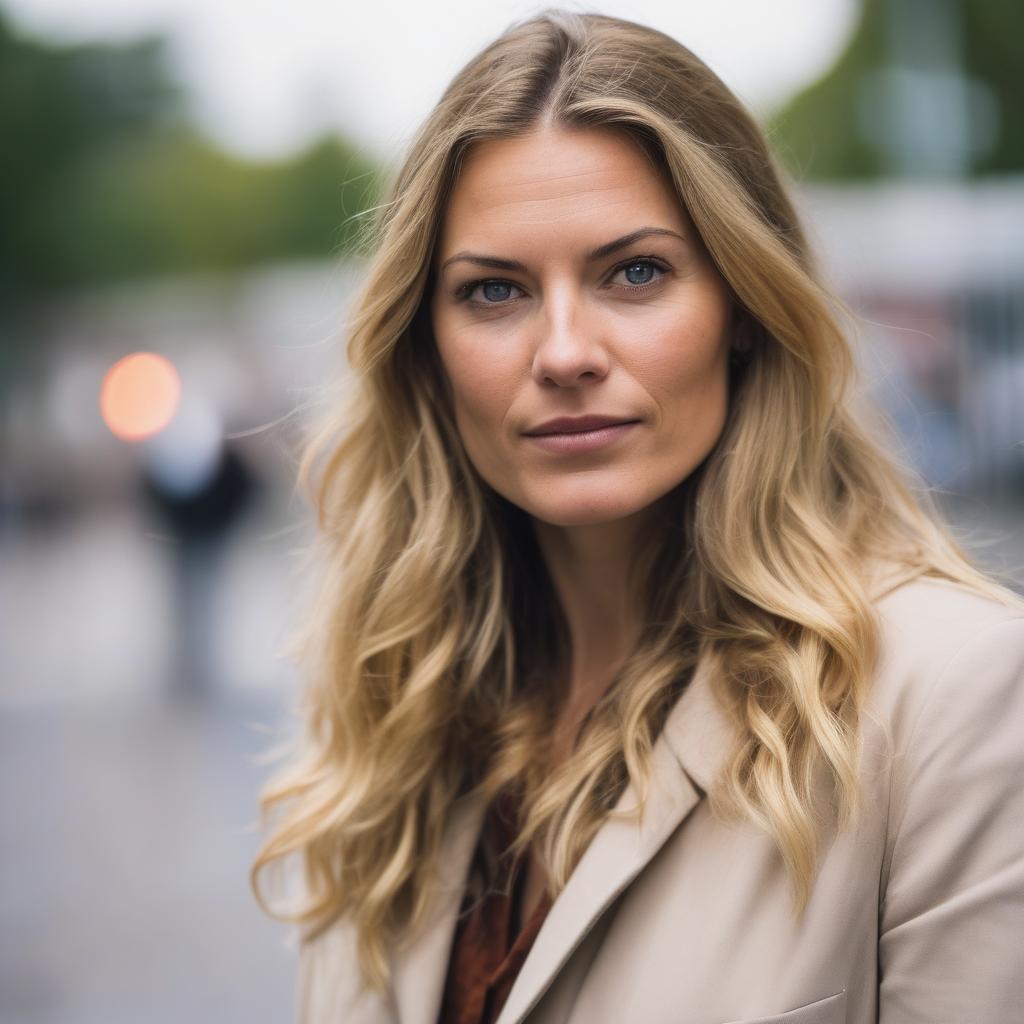}
    \end{subfigure}\hspace*{-0.25em}
    \begin{subfigure}{0.16\textwidth}
        \includegraphics[width=1\linewidth]{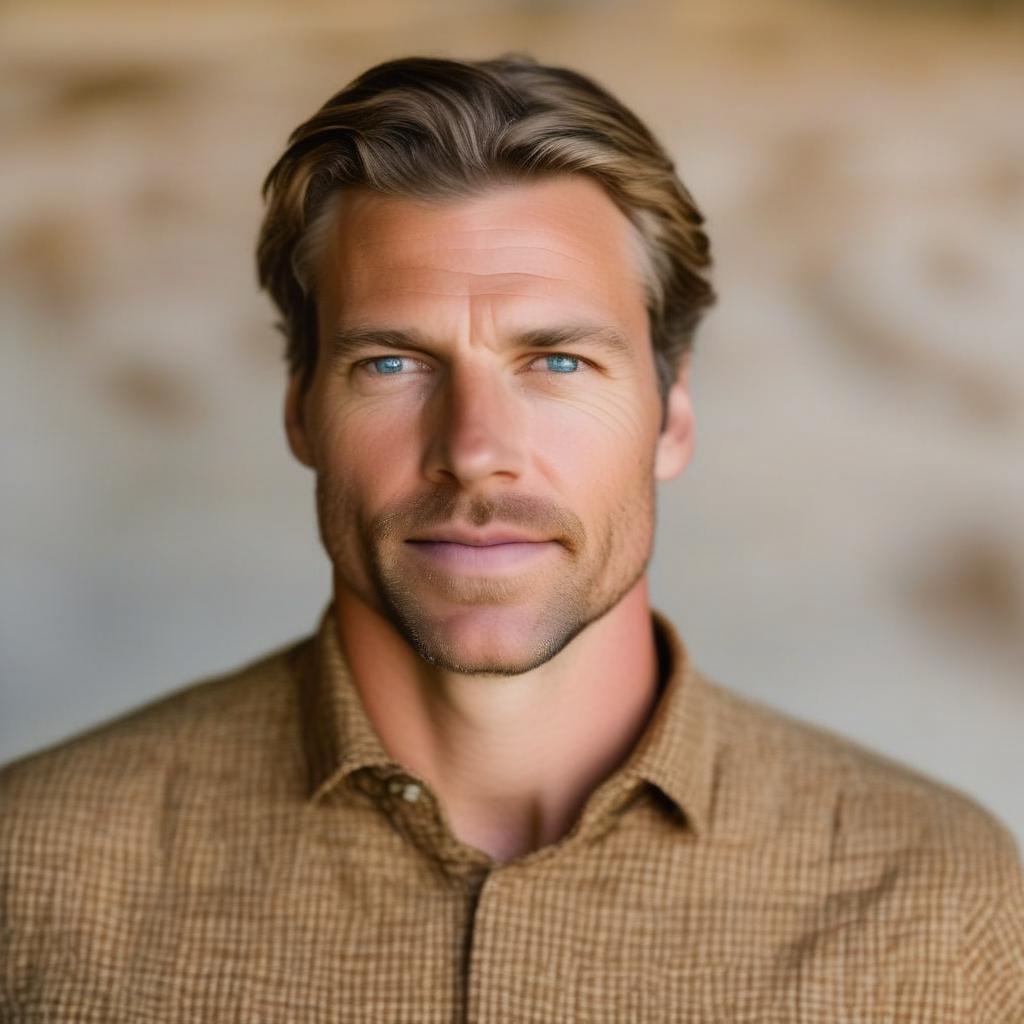}
    \end{subfigure}
    \begin{subfigure}{0.16\textwidth}
        \includegraphics[width=1\linewidth]{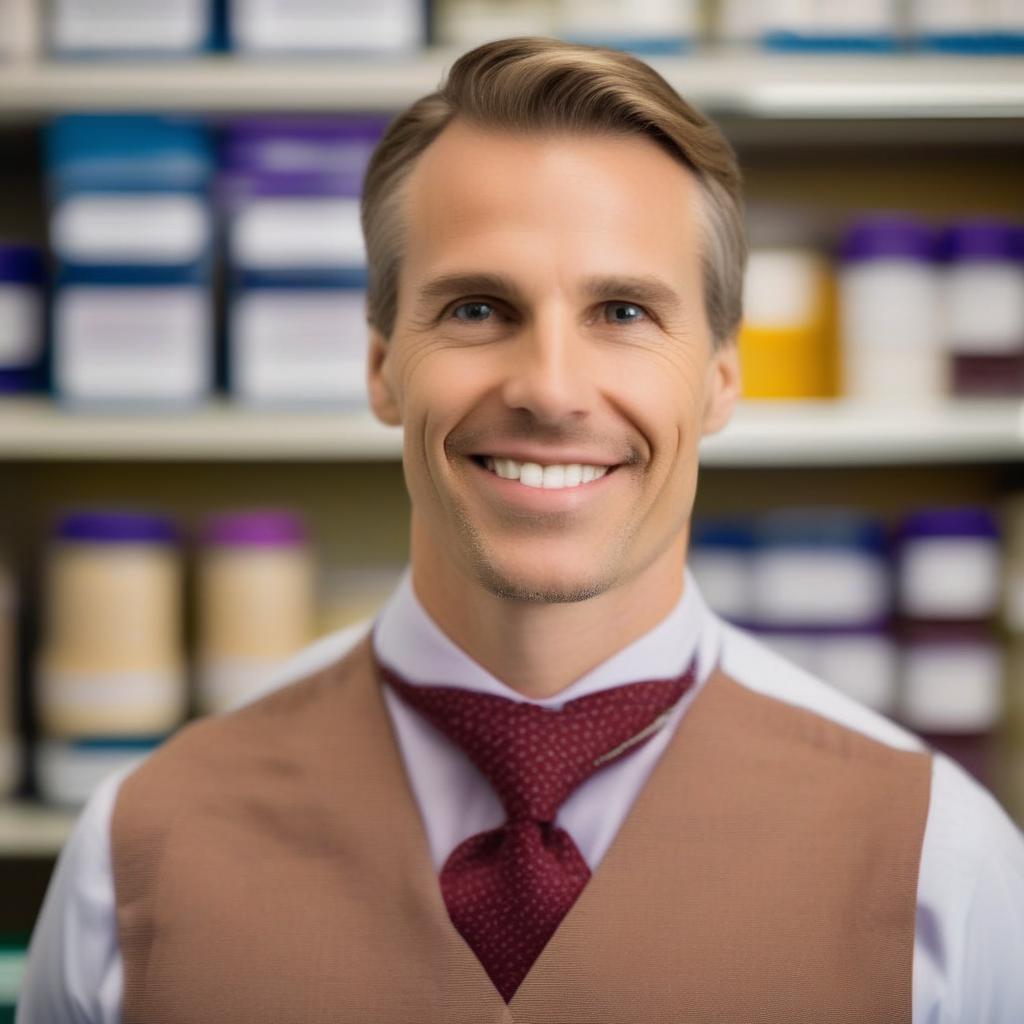}
    \end{subfigure}\hspace*{-0.25em}
    \begin{subfigure}{0.16\textwidth}
        \includegraphics[width=1\linewidth]{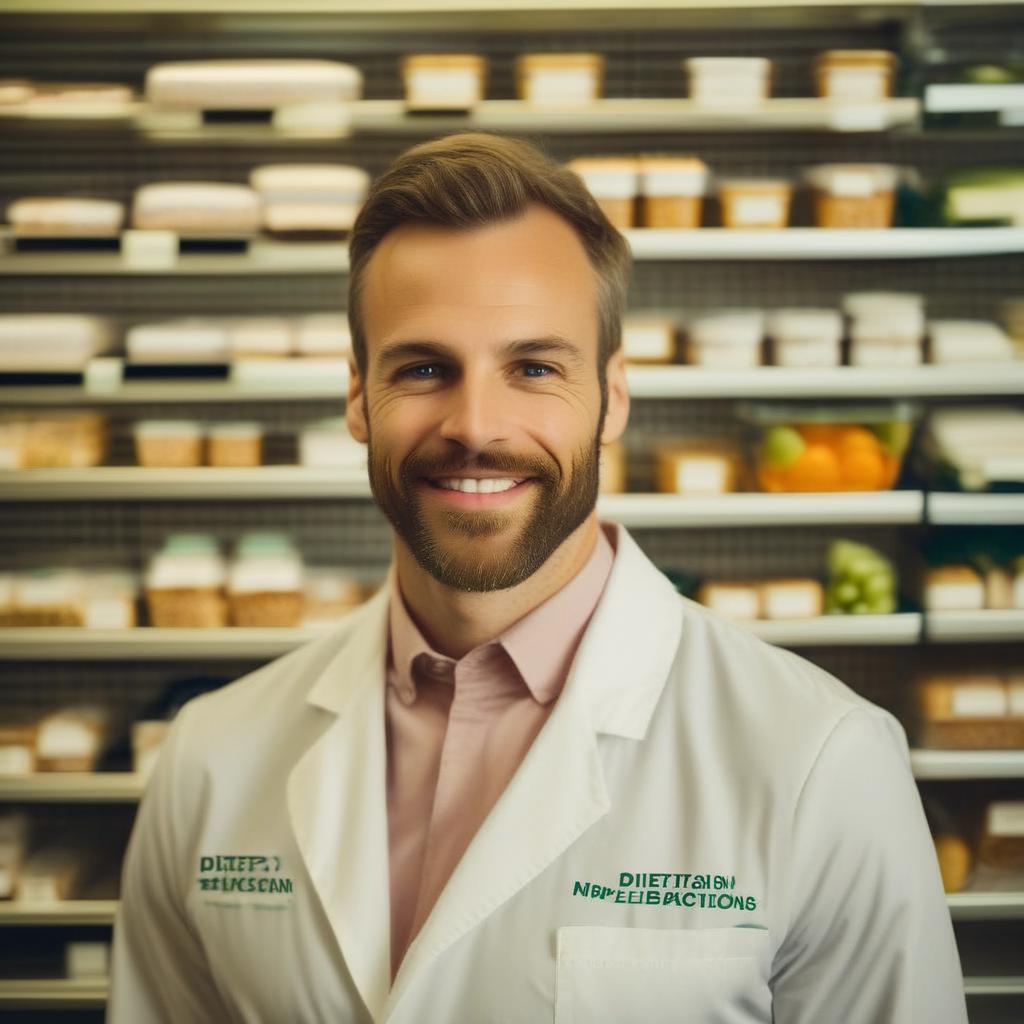}
    \end{subfigure}\hspace*{-0.25em}
    \begin{subfigure}{0.16\textwidth}
        \includegraphics[width=1\linewidth]{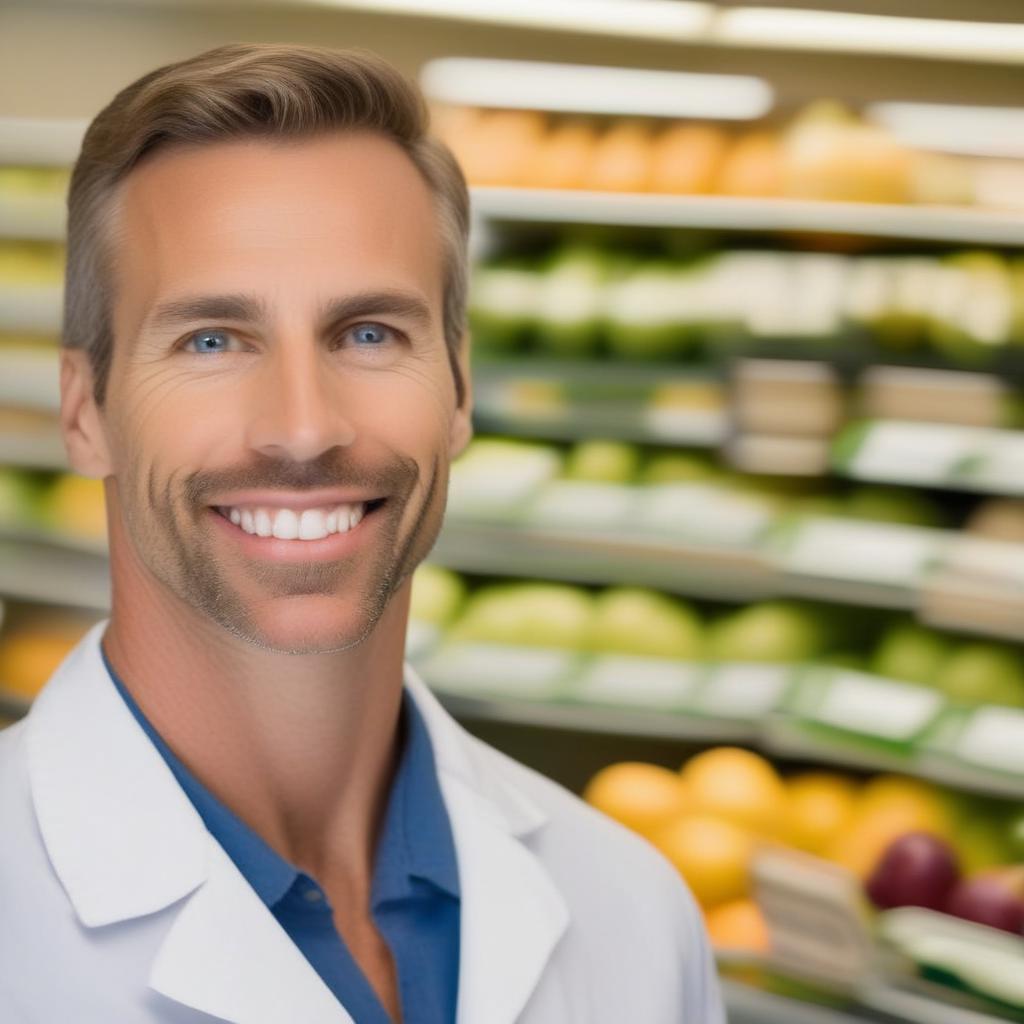}
    \end{subfigure}\hspace*{-0.25em}
    \begin{subfigure}{0.16\textwidth}
        \includegraphics[width=1\linewidth]{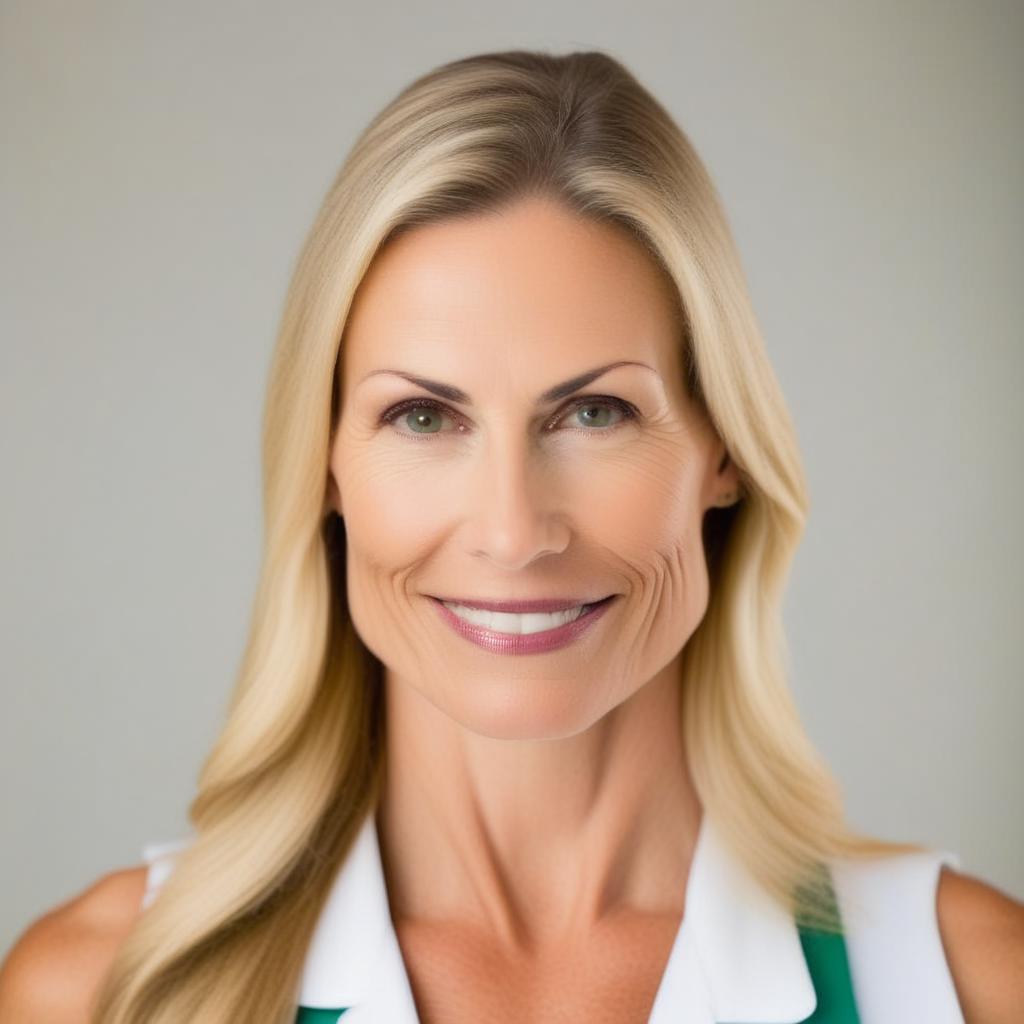}
    \end{subfigure}\hspace*{-0.25em}
    \begin{subfigure}{0.16\textwidth}
        \includegraphics[width=1\linewidth]{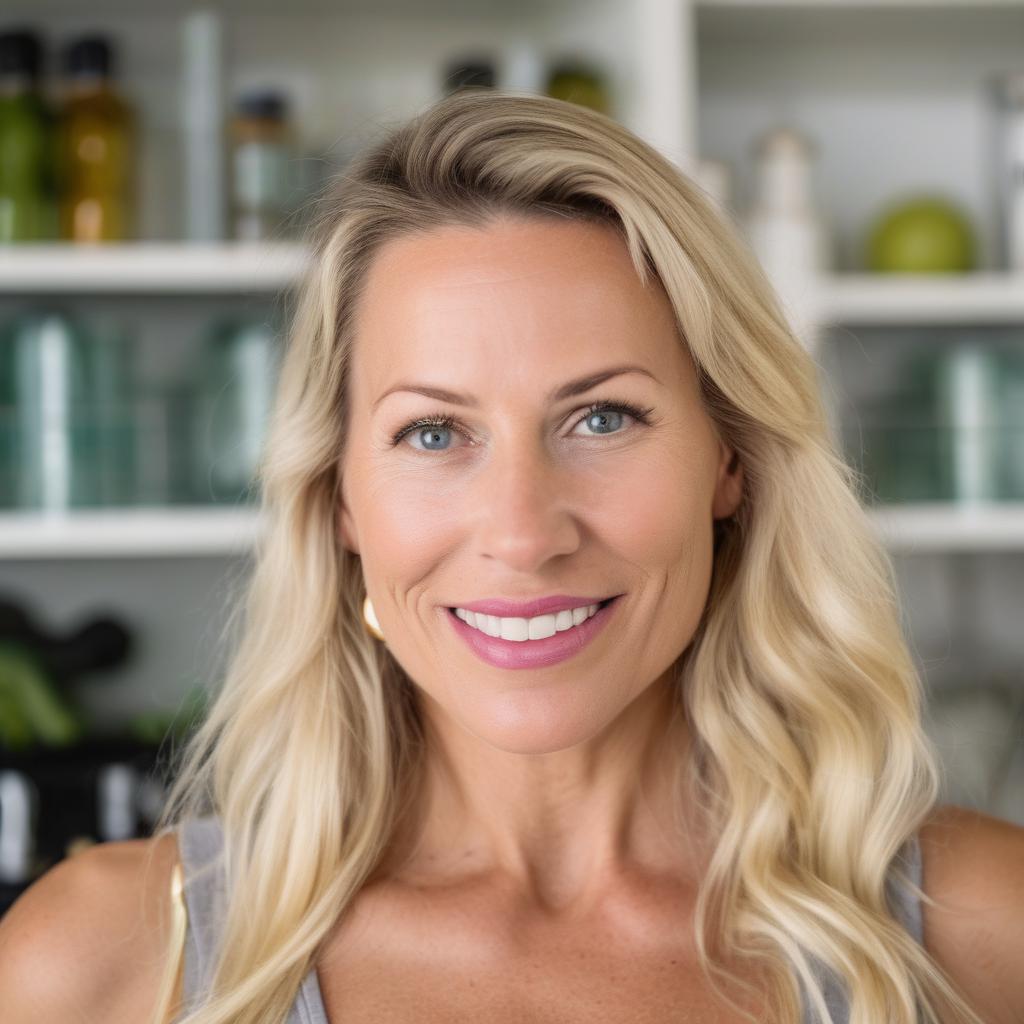}
    \end{subfigure}\hspace*{-0.25em}
    \begin{subfigure}{0.16\textwidth}
        \includegraphics[width=1\linewidth]{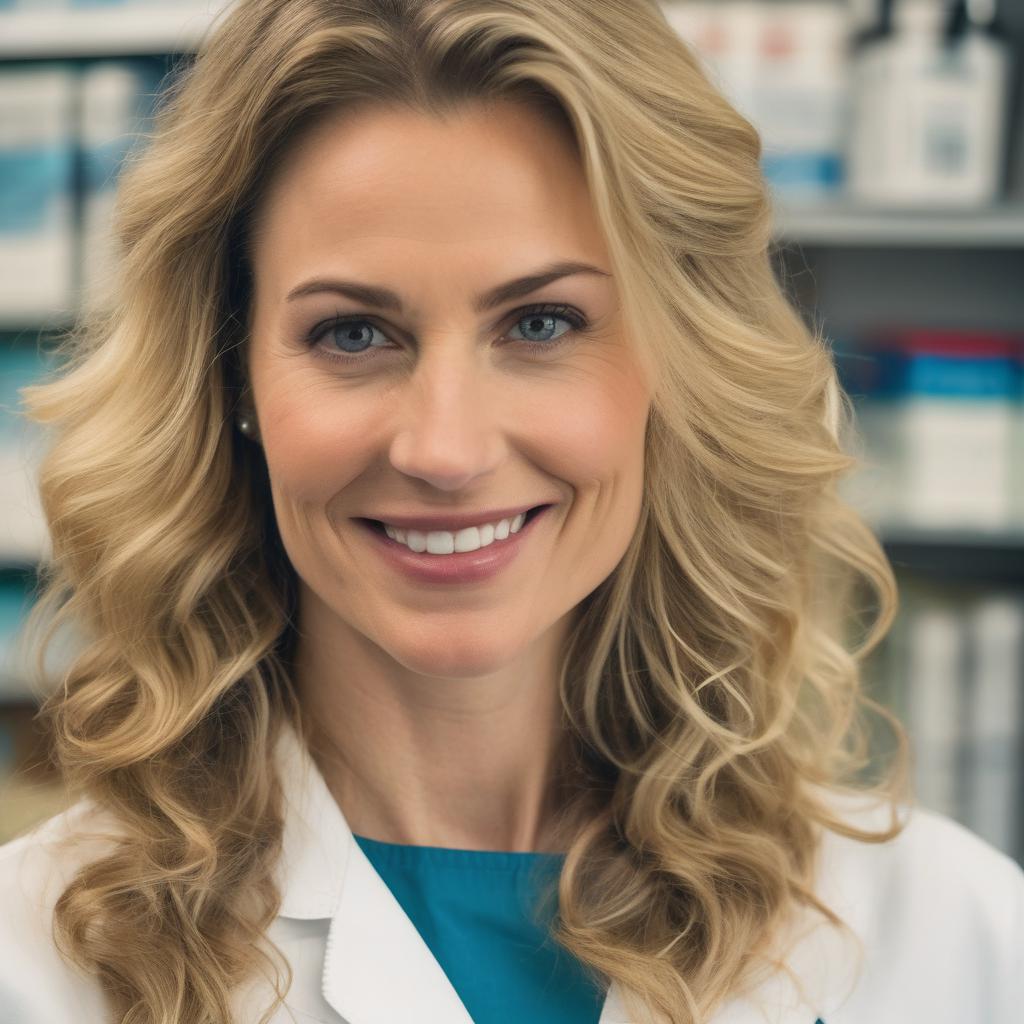}
    \end{subfigure}\hspace*{-0.25em}    
    \vspace{-10pt}
    \begin{flushleft} \textbf{b} \end{flushleft}
    \begin{subfigure}{0.16\textwidth}
        \includegraphics[width=1\linewidth]{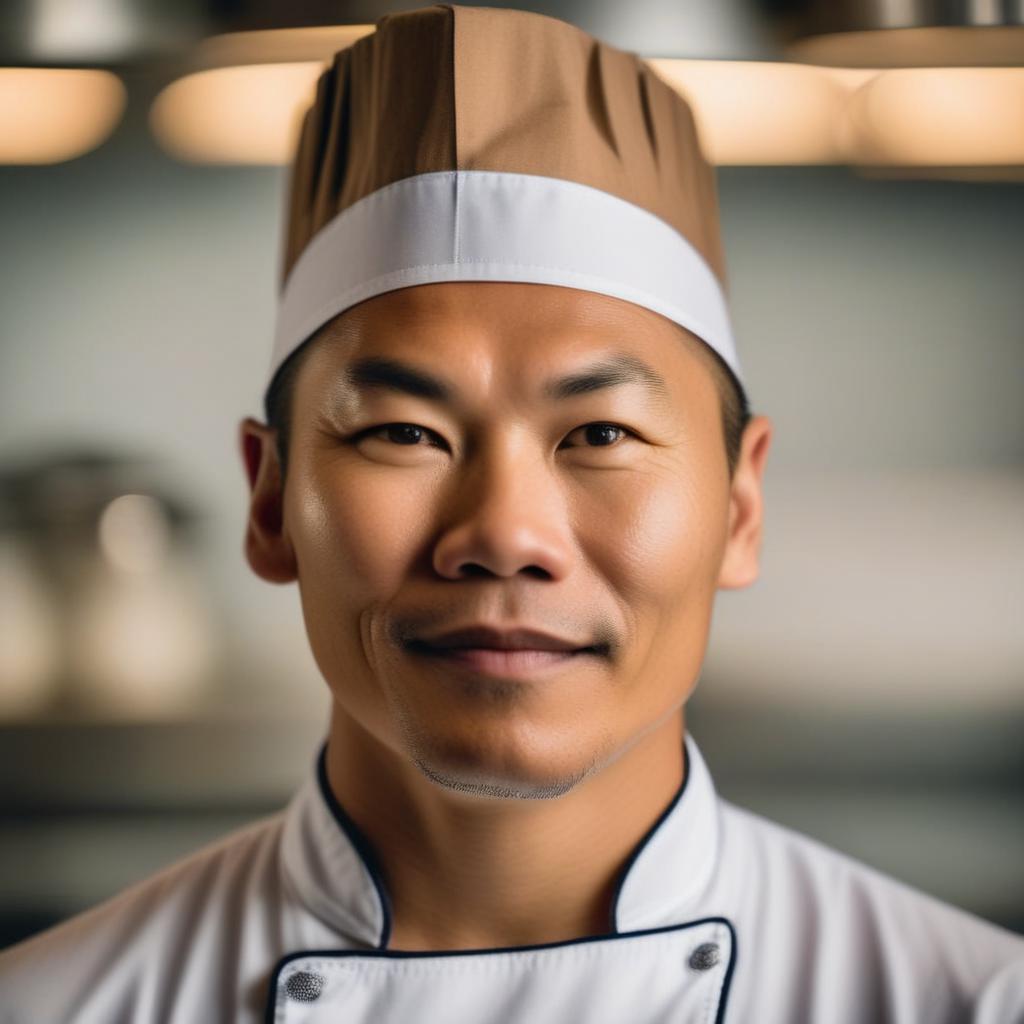}
    \end{subfigure}\hspace*{-0.25em}
    \begin{subfigure}{0.16\textwidth}
        \includegraphics[width=1\linewidth]{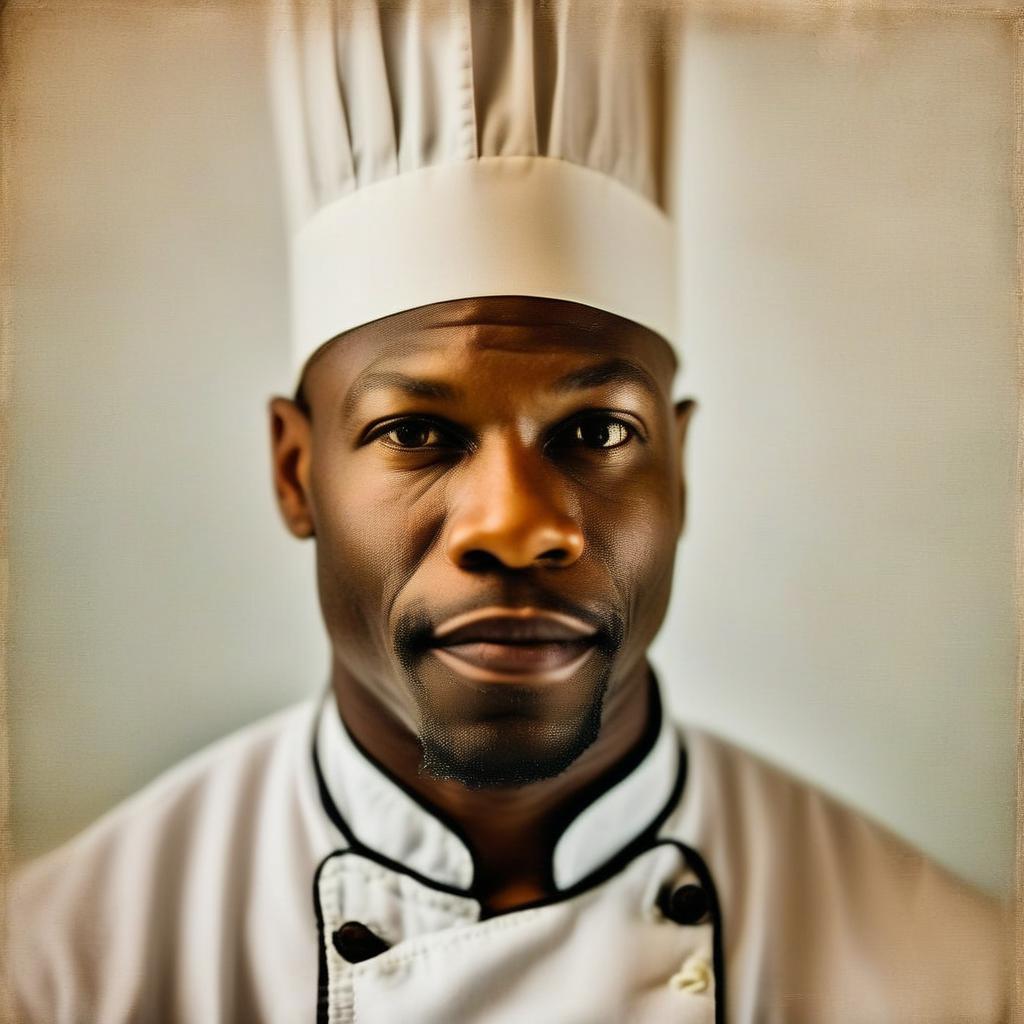}
    \end{subfigure}\hspace*{-0.25em}
    \begin{subfigure}{0.16\textwidth}
        \includegraphics[width=1\linewidth]{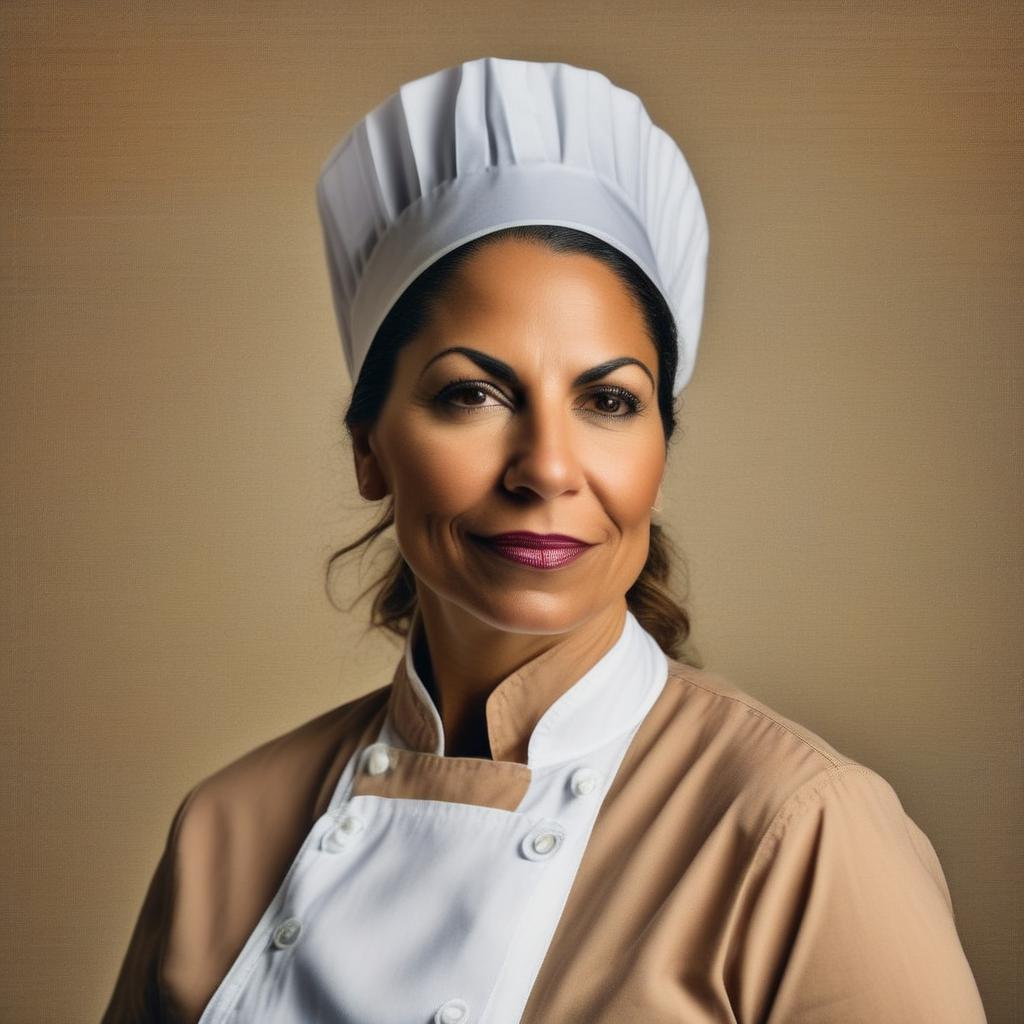}
    \end{subfigure}\hspace*{-0.25em} 
    \begin{subfigure}{0.16\textwidth}
        \includegraphics[width=1\linewidth]{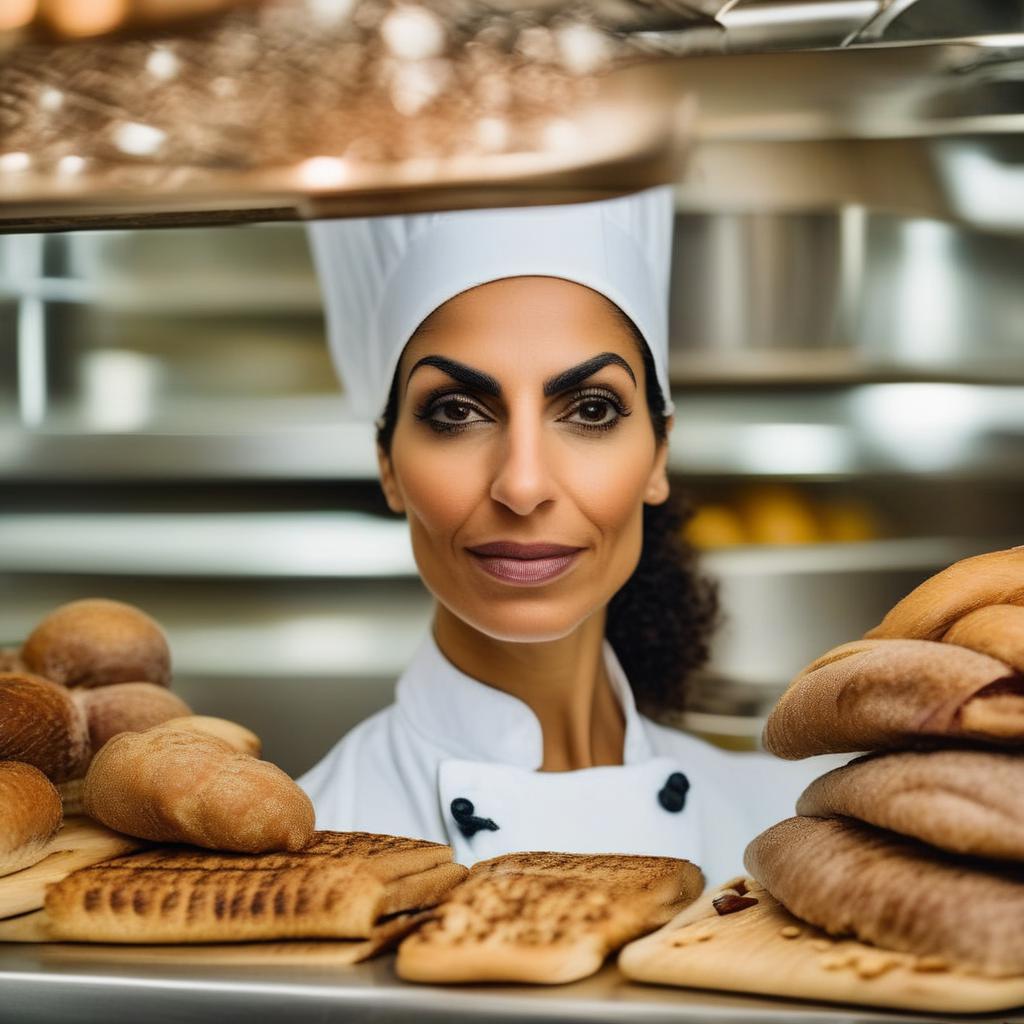}
    \end{subfigure}\hspace*{-0.25em}
    \begin{subfigure}{0.16\textwidth}
        \includegraphics[width=1\linewidth]{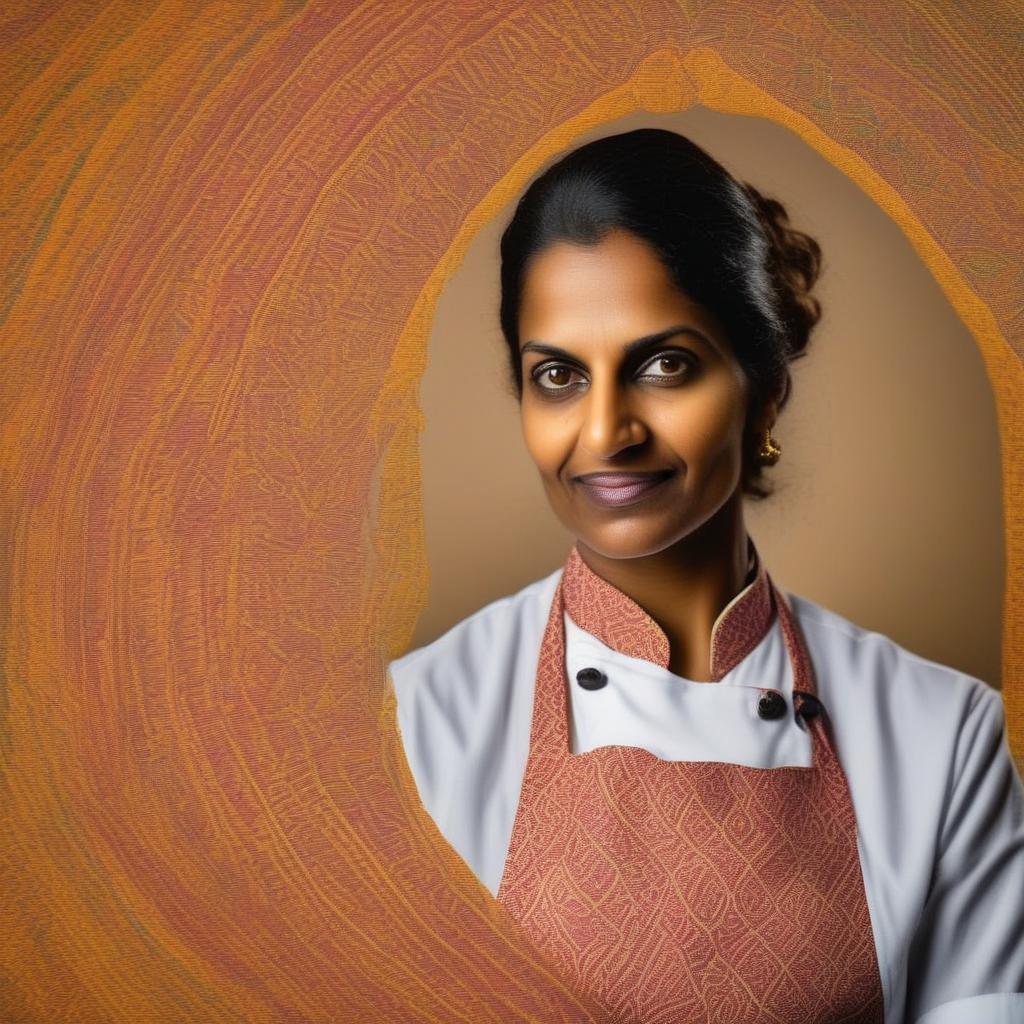}
    \end{subfigure}\hspace*{-0.25em}
    \begin{subfigure}{0.16\textwidth}
        \includegraphics[width=1\linewidth]{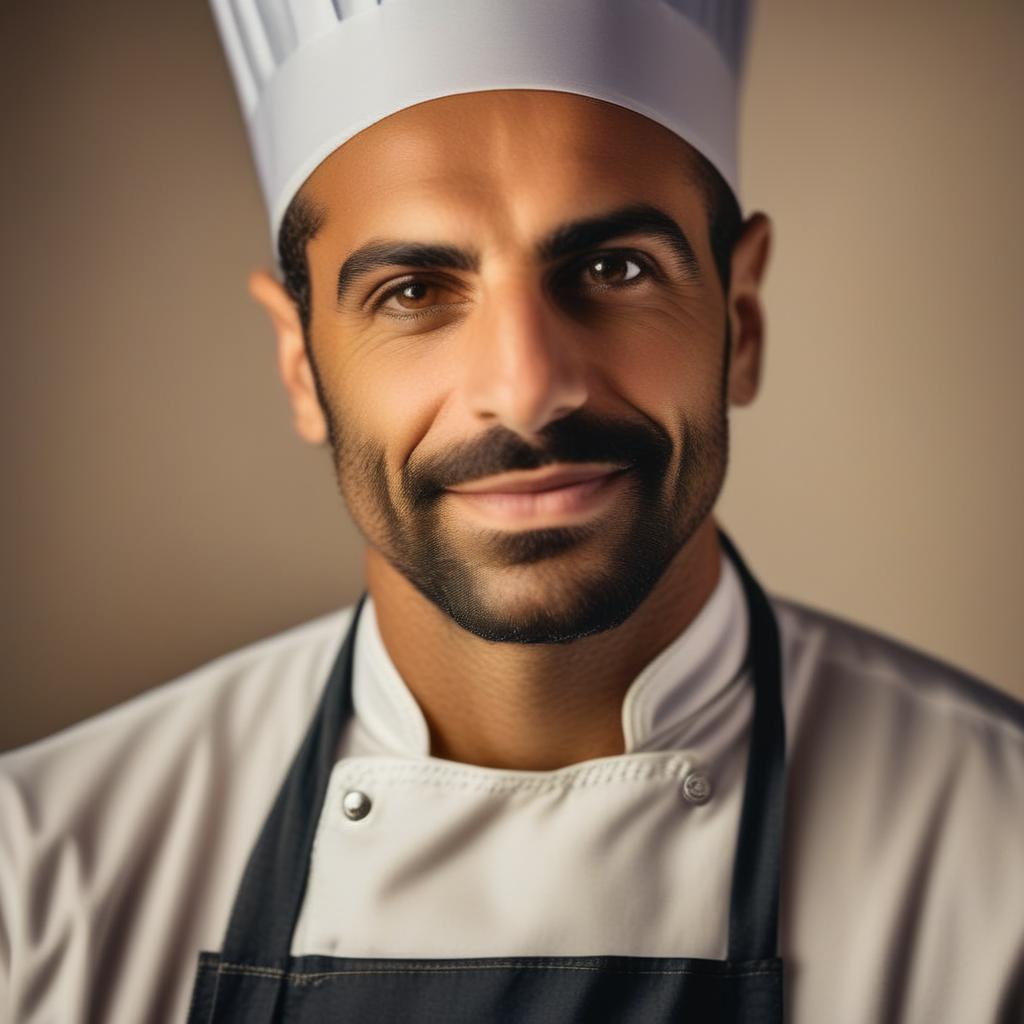}
    \end{subfigure} 
    \begin{subfigure}{0.16\textwidth}
        \includegraphics[width=1\linewidth]{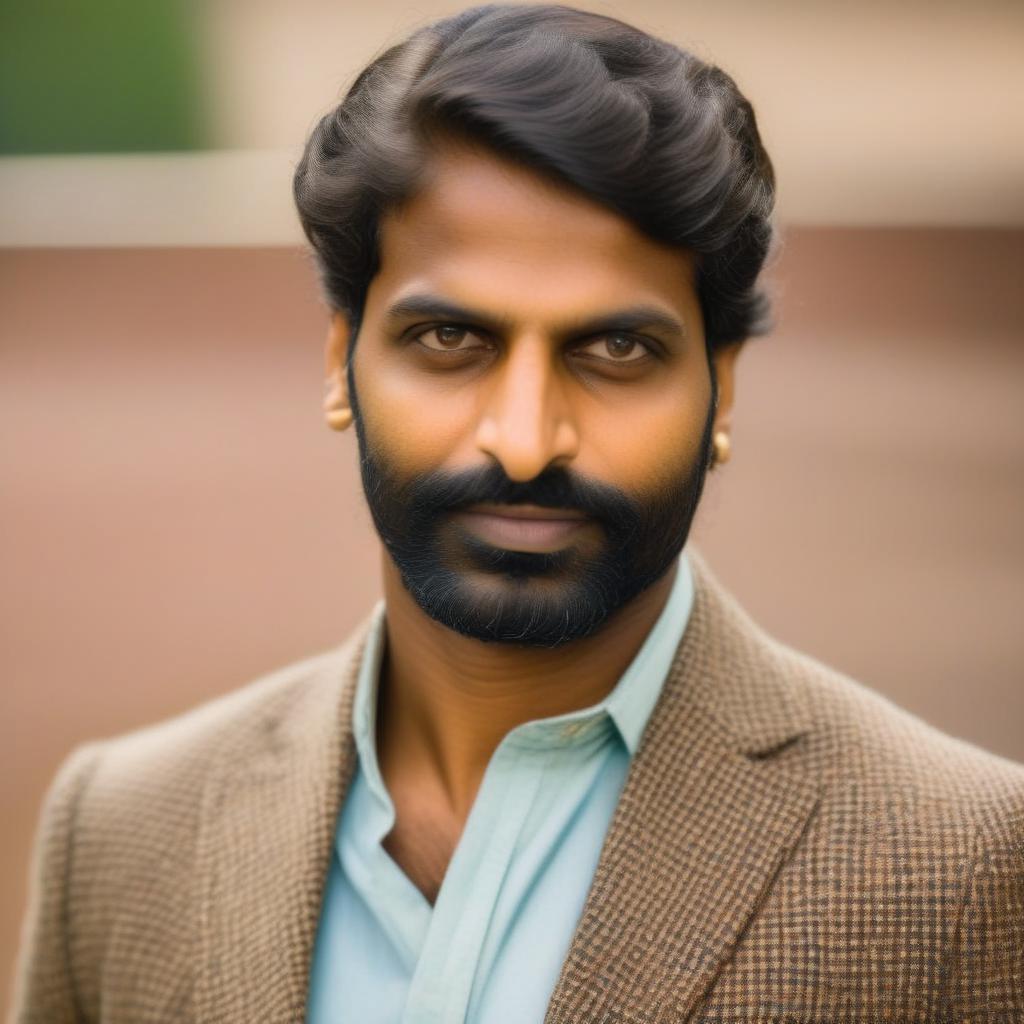}
    \end{subfigure}\hspace*{-0.25em}
    \begin{subfigure}{0.16\textwidth}
        \includegraphics[width=1\linewidth]{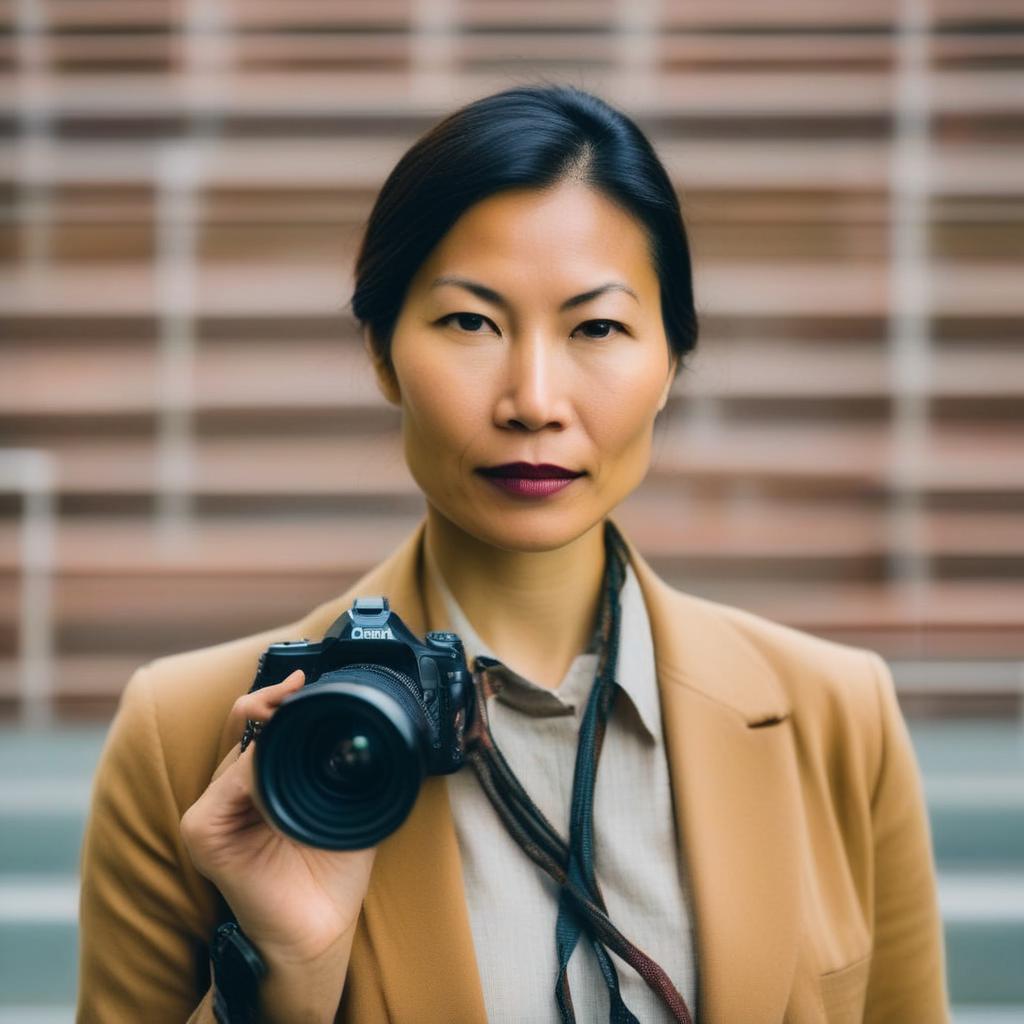}
    \end{subfigure}\hspace*{-0.25em}
    \begin{subfigure}{0.16\textwidth}
        \includegraphics[width=1\linewidth]{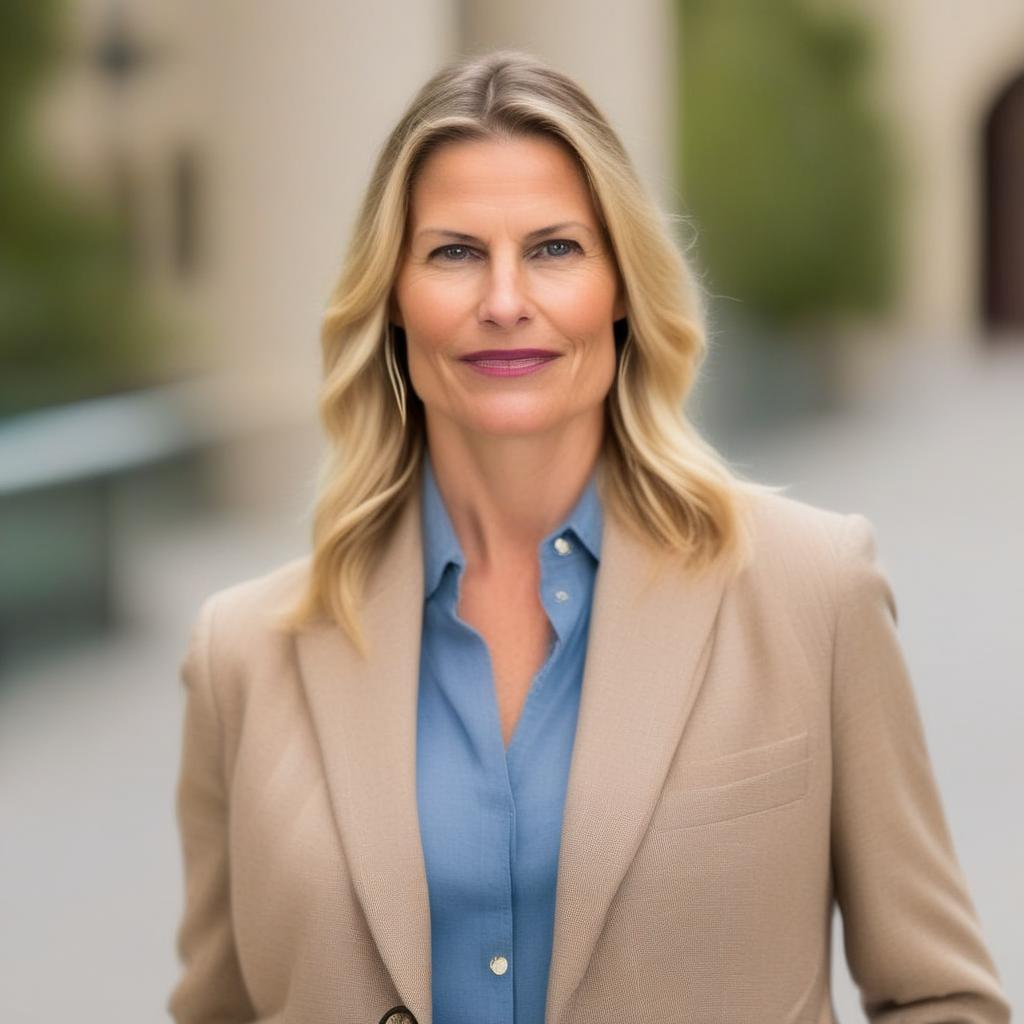}
    \end{subfigure}\hspace*{-0.25em}
    \begin{subfigure}{0.16\textwidth}
        \includegraphics[width=1\linewidth]{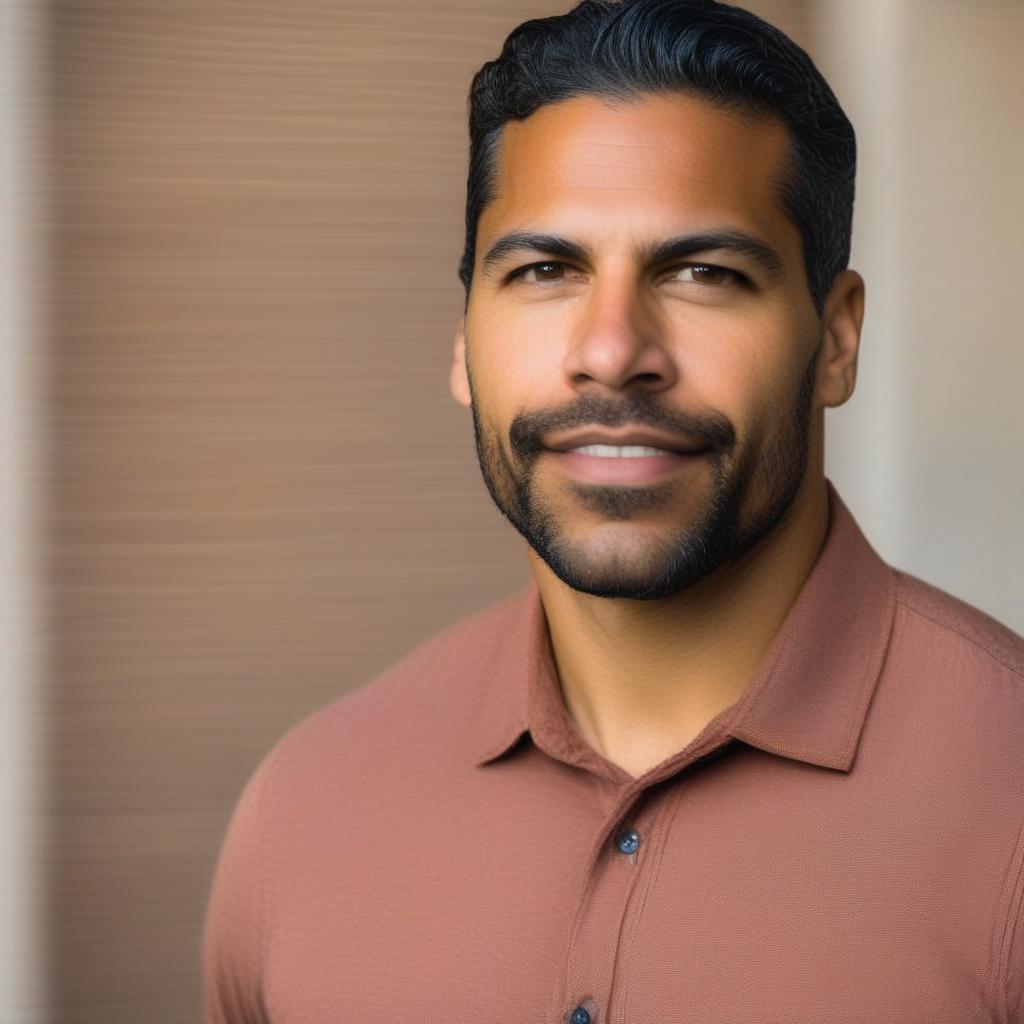}
    \end{subfigure}\hspace*{-0.25em}
    \begin{subfigure}{0.16\textwidth}
        \includegraphics[width=1\linewidth]{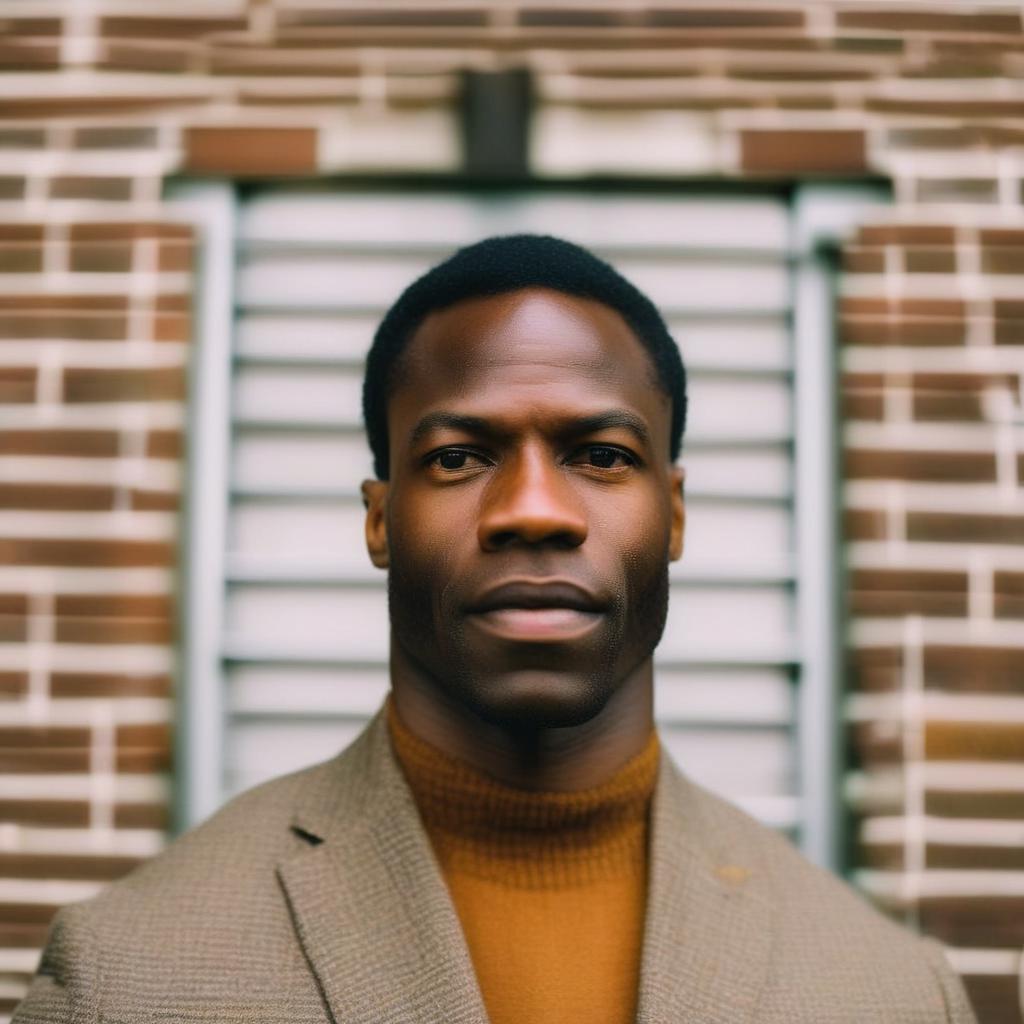}
    \end{subfigure}\hspace*{-0.25em}
    \begin{subfigure}{0.16\textwidth}
        \includegraphics[width=1\linewidth]{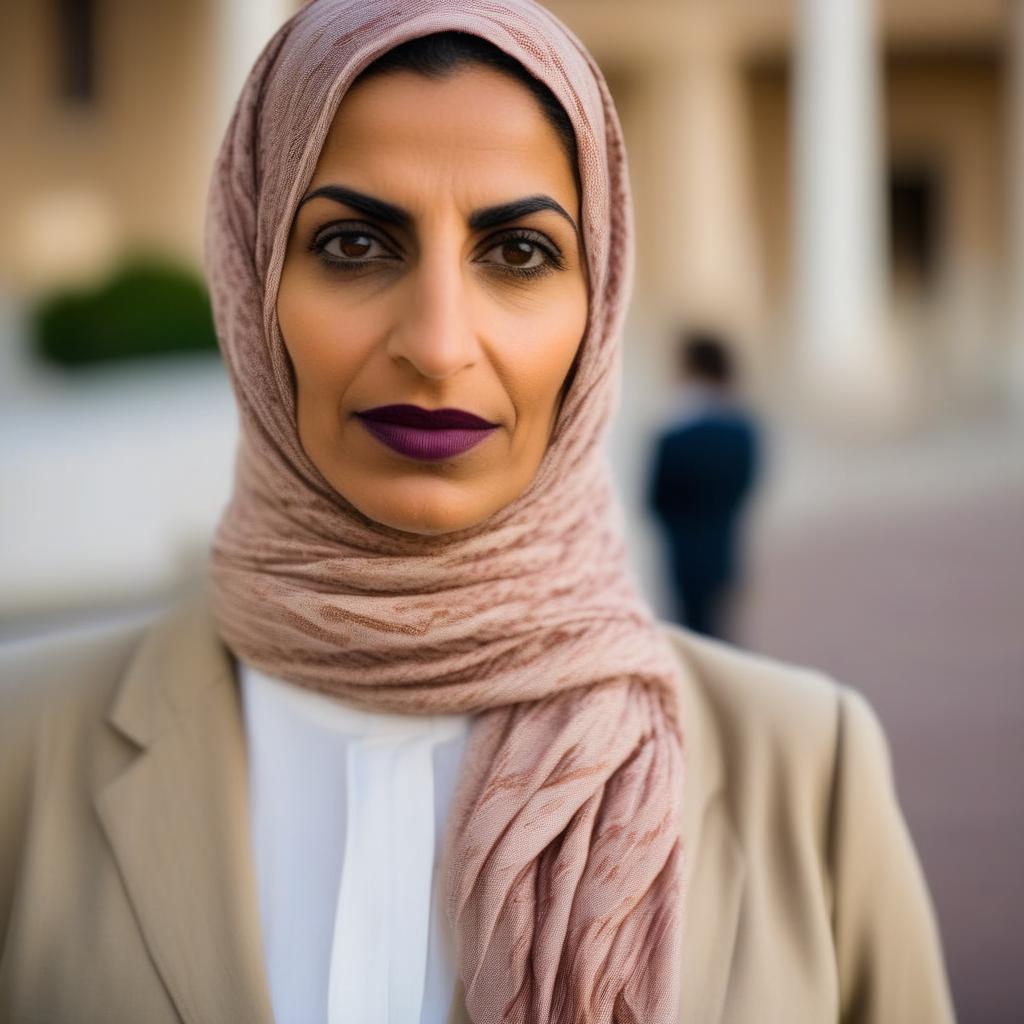}
    \end{subfigure}
    \begin{subfigure}{0.16\textwidth}
        \includegraphics[width=1\linewidth]{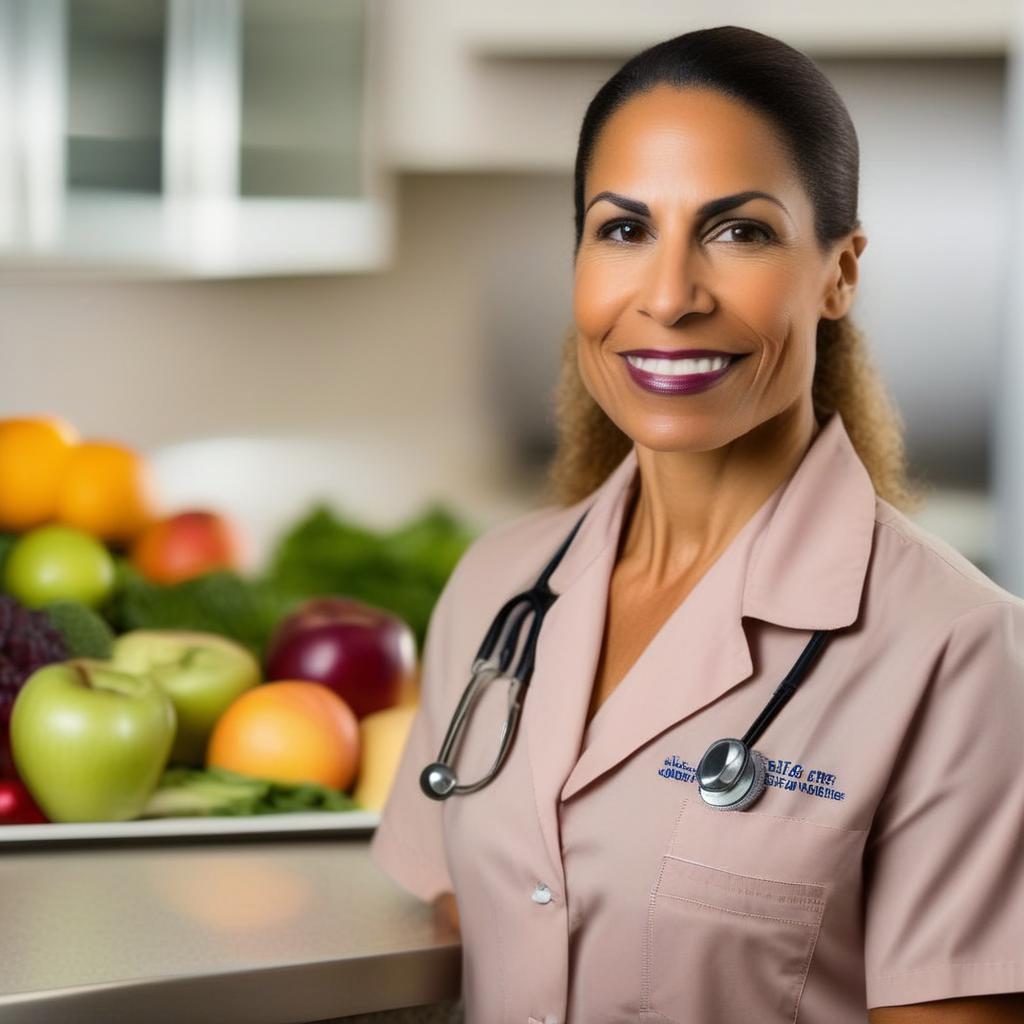}
    \end{subfigure}\hspace*{-0.25em}
    \begin{subfigure}{0.16\textwidth}
        \includegraphics[width=1\linewidth]{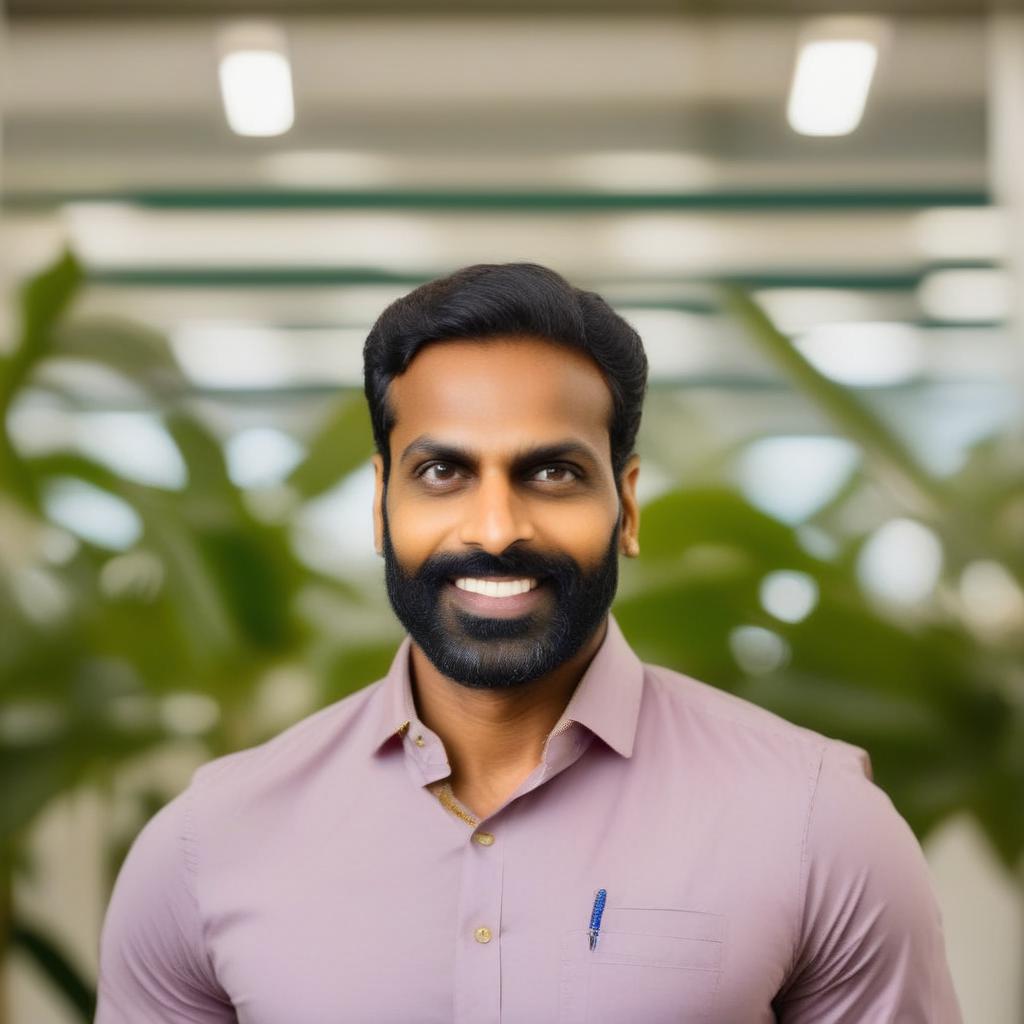}
    \end{subfigure}\hspace*{-0.25em}
    \begin{subfigure}{0.16\textwidth}
        \includegraphics[width=1\linewidth]{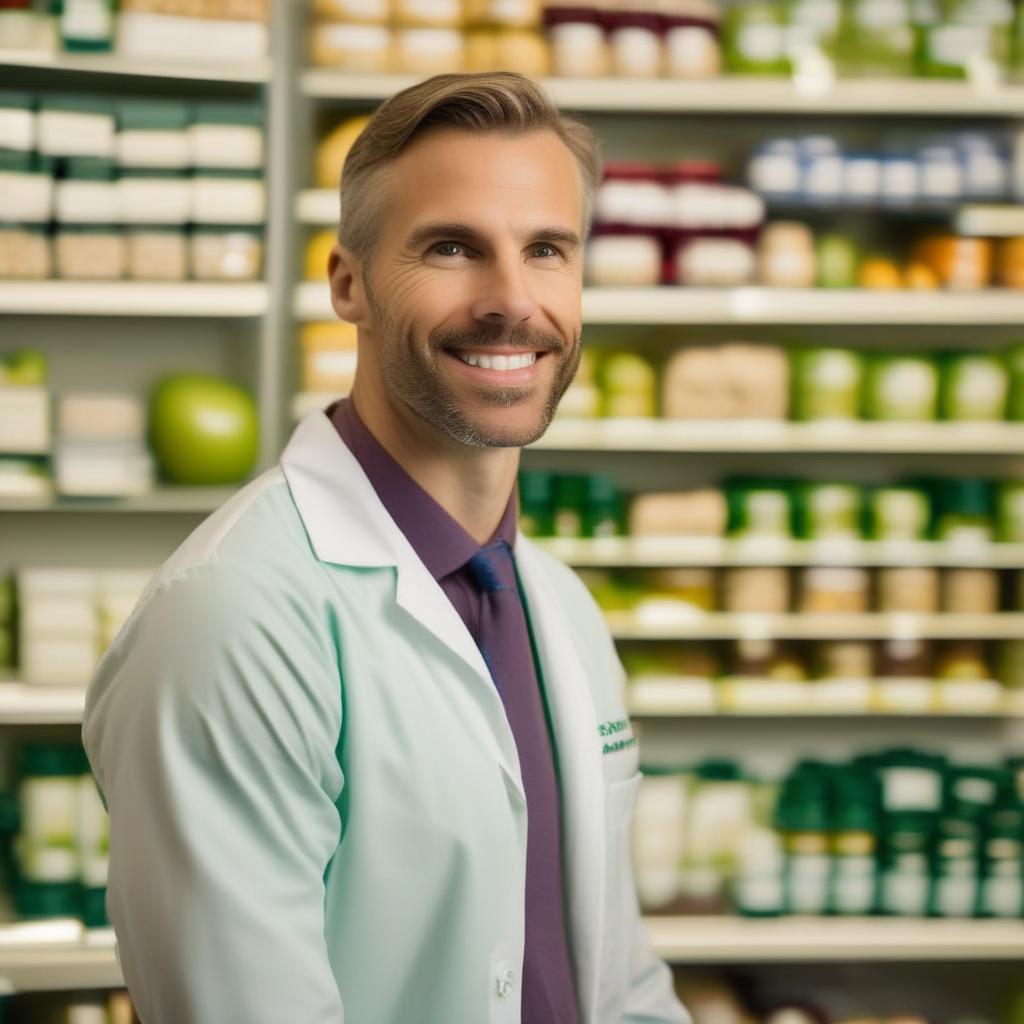}
    \end{subfigure}\hspace*{-0.25em}
    \begin{subfigure}{0.16\textwidth}
        \includegraphics[width=1\linewidth]{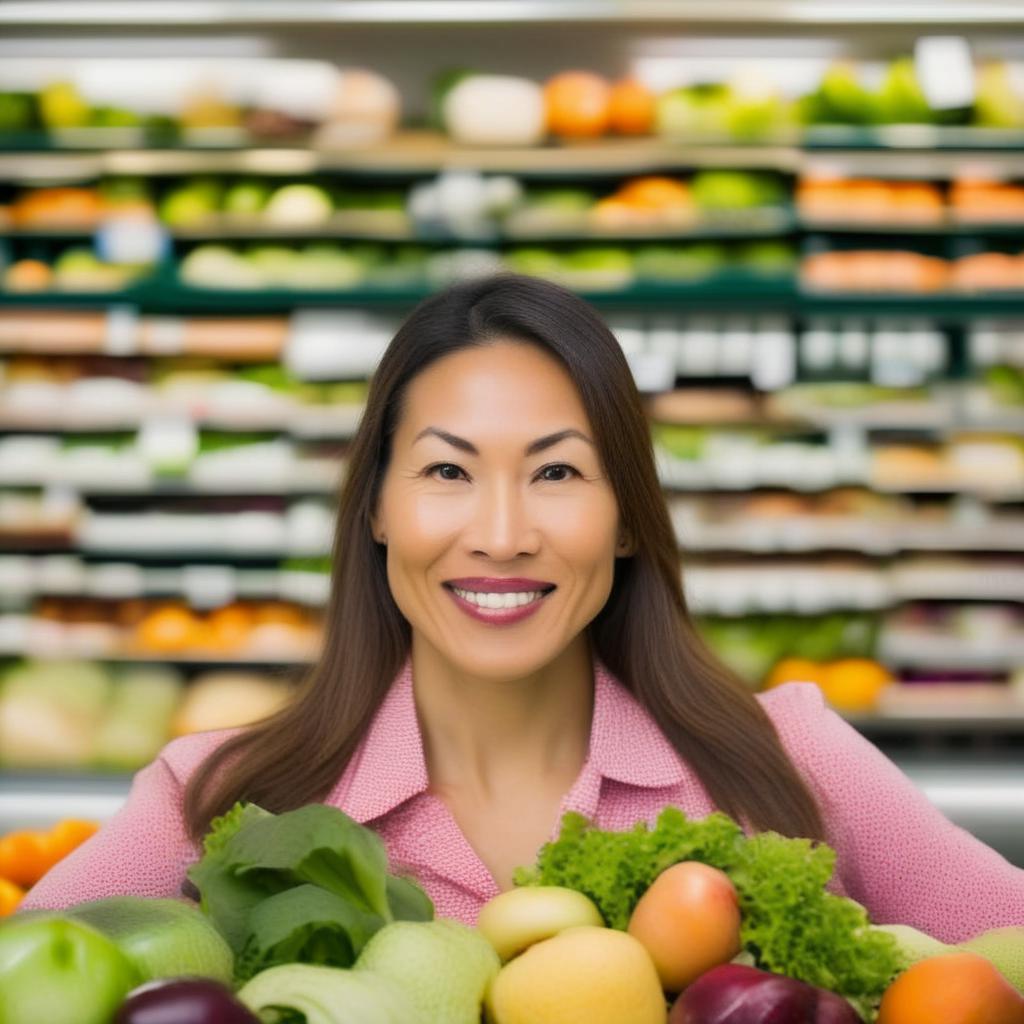}
    \end{subfigure}\hspace*{-0.25em}
    \begin{subfigure}{0.16\textwidth}
        \includegraphics[width=1\linewidth]{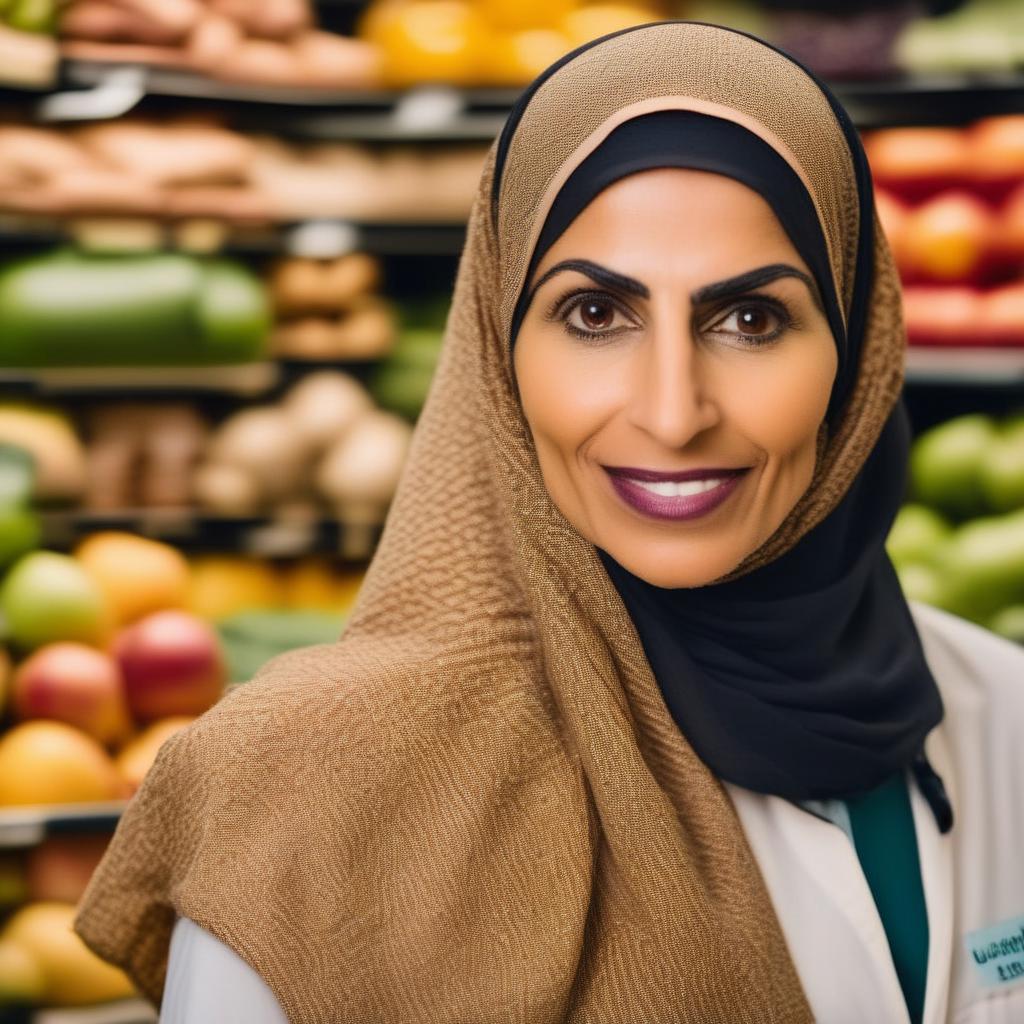}
    \end{subfigure}\hspace*{-0.25em}
    \begin{subfigure}{0.16\textwidth}
        \includegraphics[width=1\linewidth]{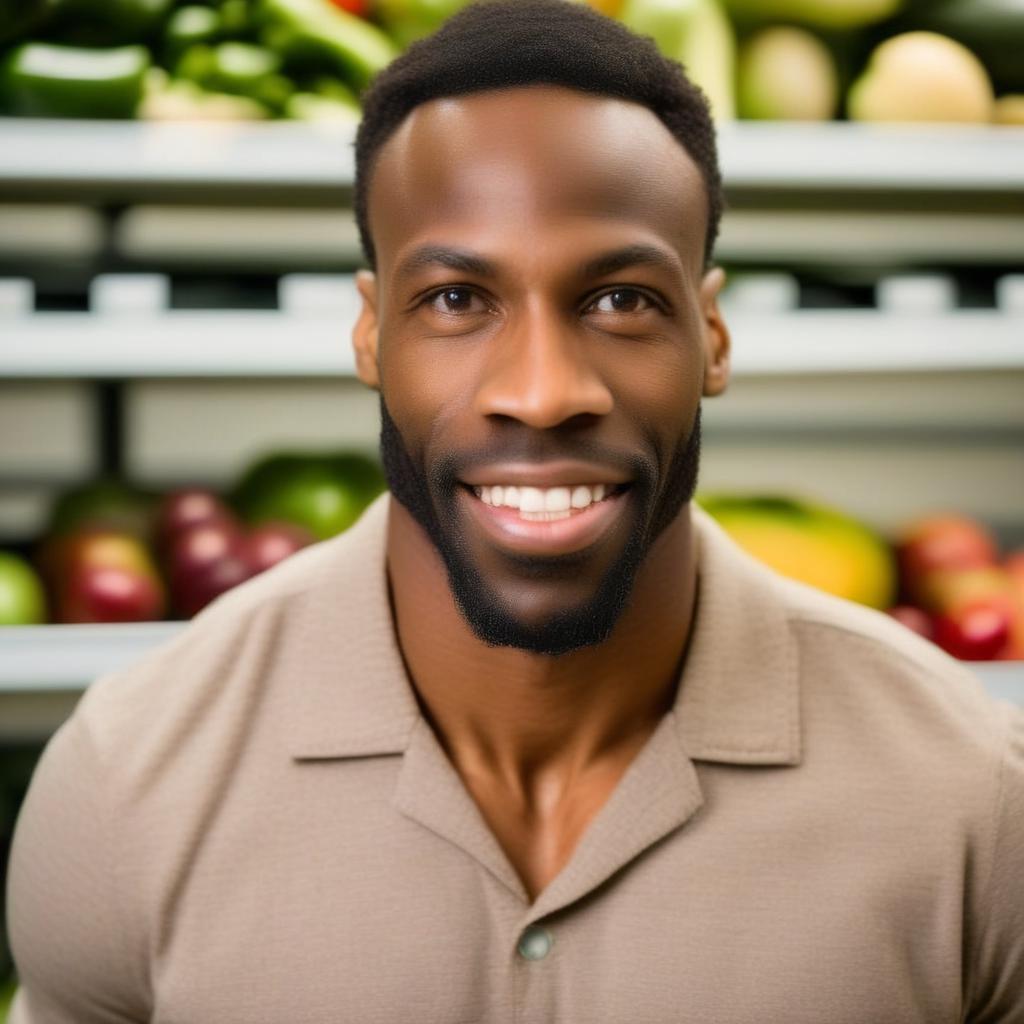}
    \end{subfigure}\hspace*{-0.25em}
\caption{
{\fontsize{11}{11}\selectfont{
\textbf{Study 1 images (racial).} \textbf{a}, 
SDXL. \textbf{b}, SDXL-Inclusive.
}}}
\label{fig:study1}
\end{figure}

\begin{figure}[htbp]
    \includegraphics[width=1\textwidth]{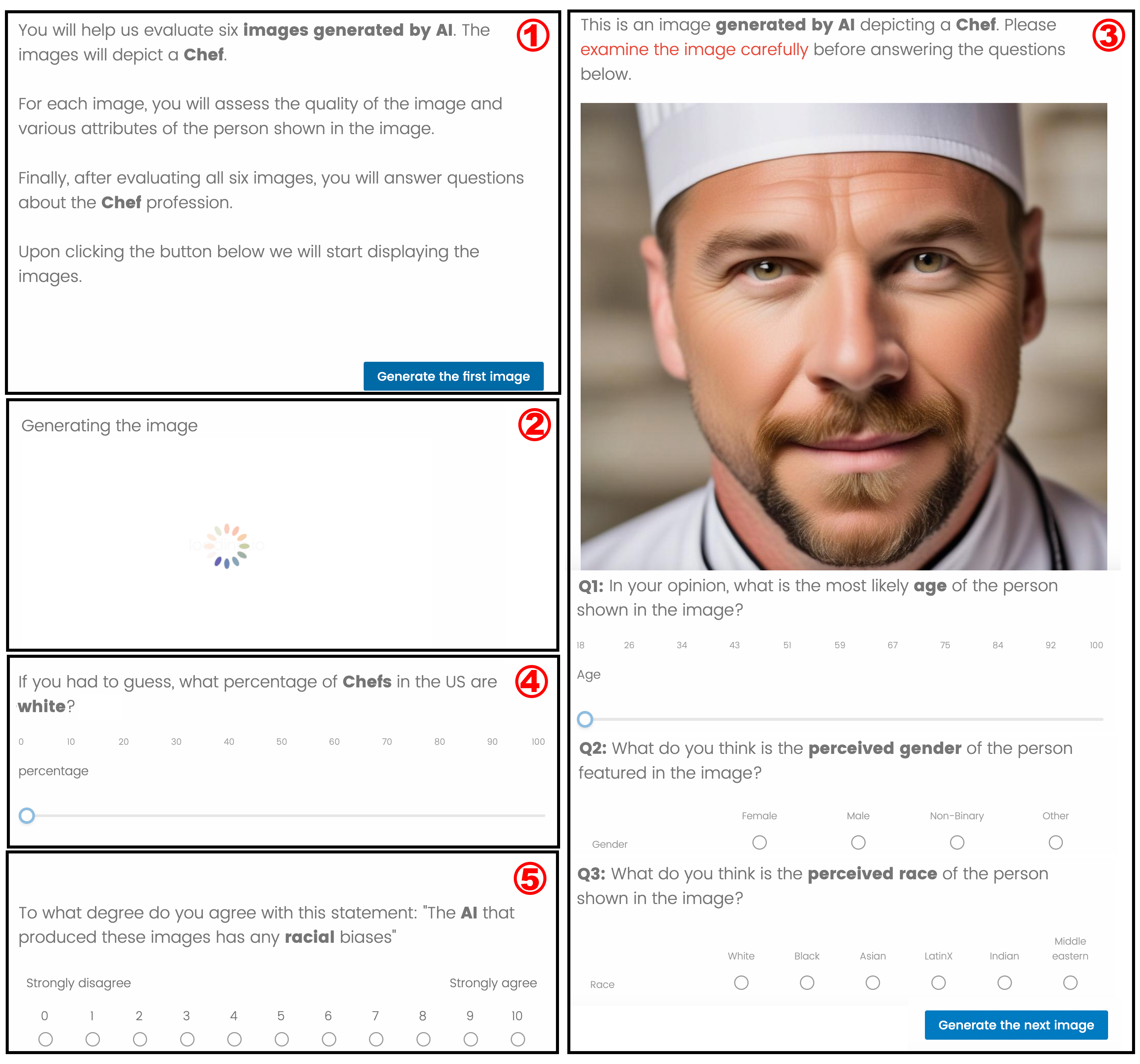}
    \caption{
    {\fontsize{11}{11}\selectfont{
    \textbf{Study 1 sample screenshots.} 
    (1) welcome screen, (2) loading screen, (3) per image questions, and (4)/(5) Final questions after seeing all six images.}}}
    \label{fig:study1_sample}
\end{figure}

\clearpage
\begin{figure}[htbp]   
\captionsetup[subfigure]{labelformat=empty}
    \centering    
    \begin{flushleft} \textbf{a} \end{flushleft}
    \begin{subfigure}{0.16\textwidth}
        \includegraphics[width=1\linewidth]{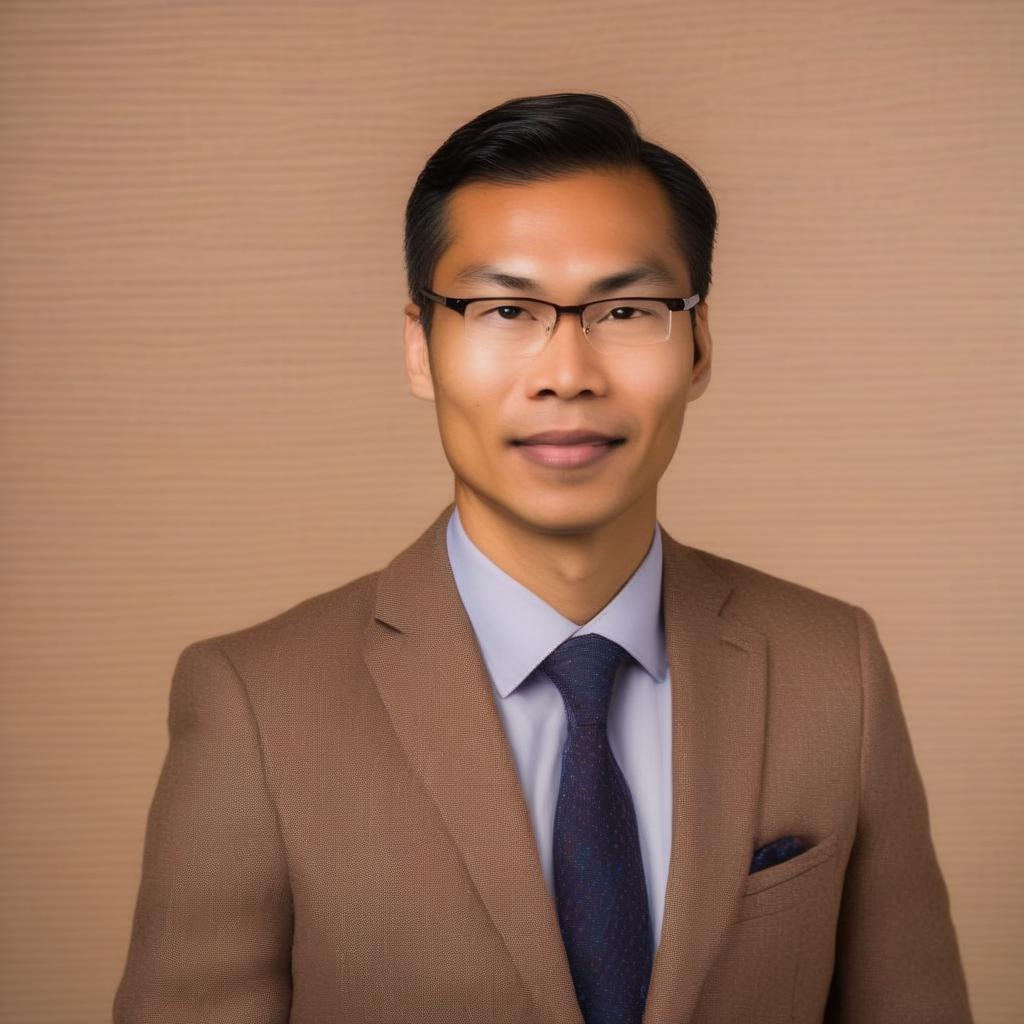}
    \end{subfigure}\hspace*{-0.25em}
    \begin{subfigure}{0.16\textwidth}
        \includegraphics[width=1\linewidth]{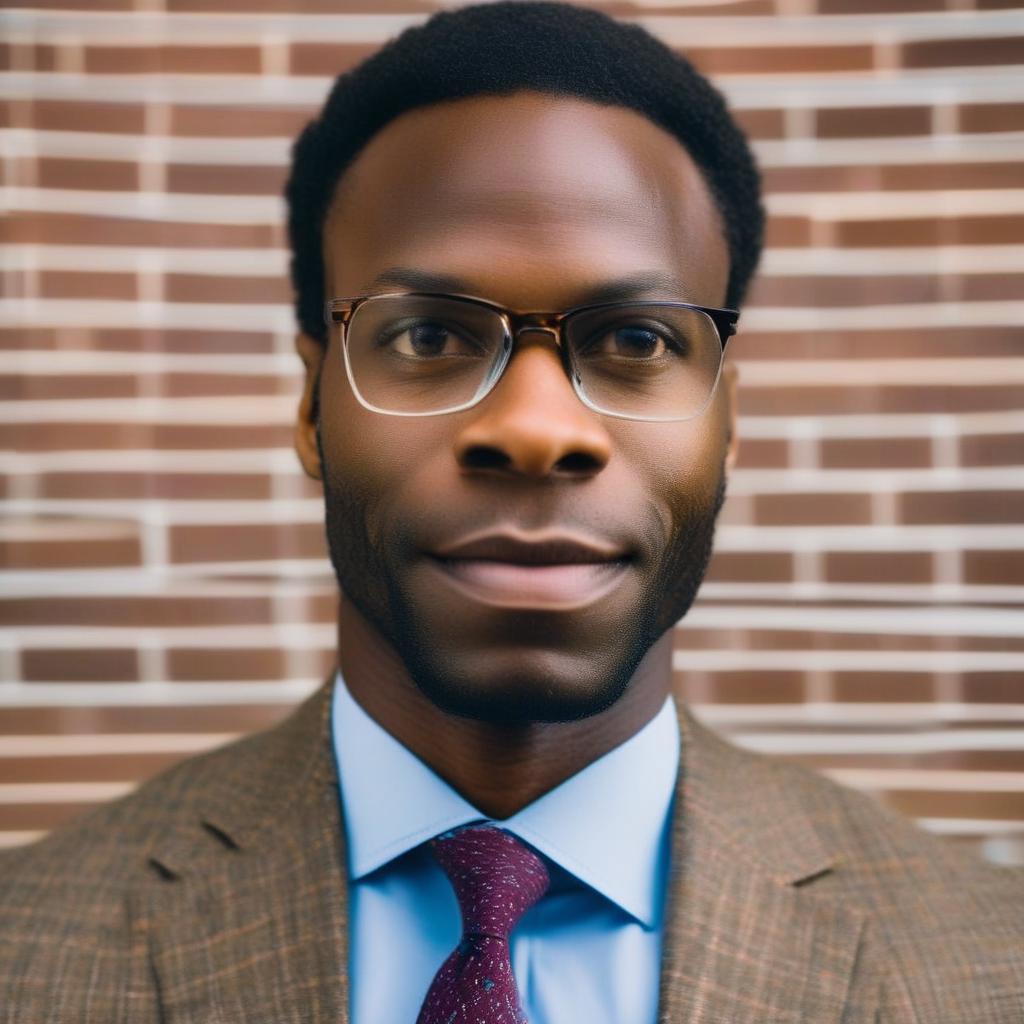}
    \end{subfigure}\hspace*{-0.25em}
    \begin{subfigure}{0.16\textwidth}
        \includegraphics[width=1\linewidth]{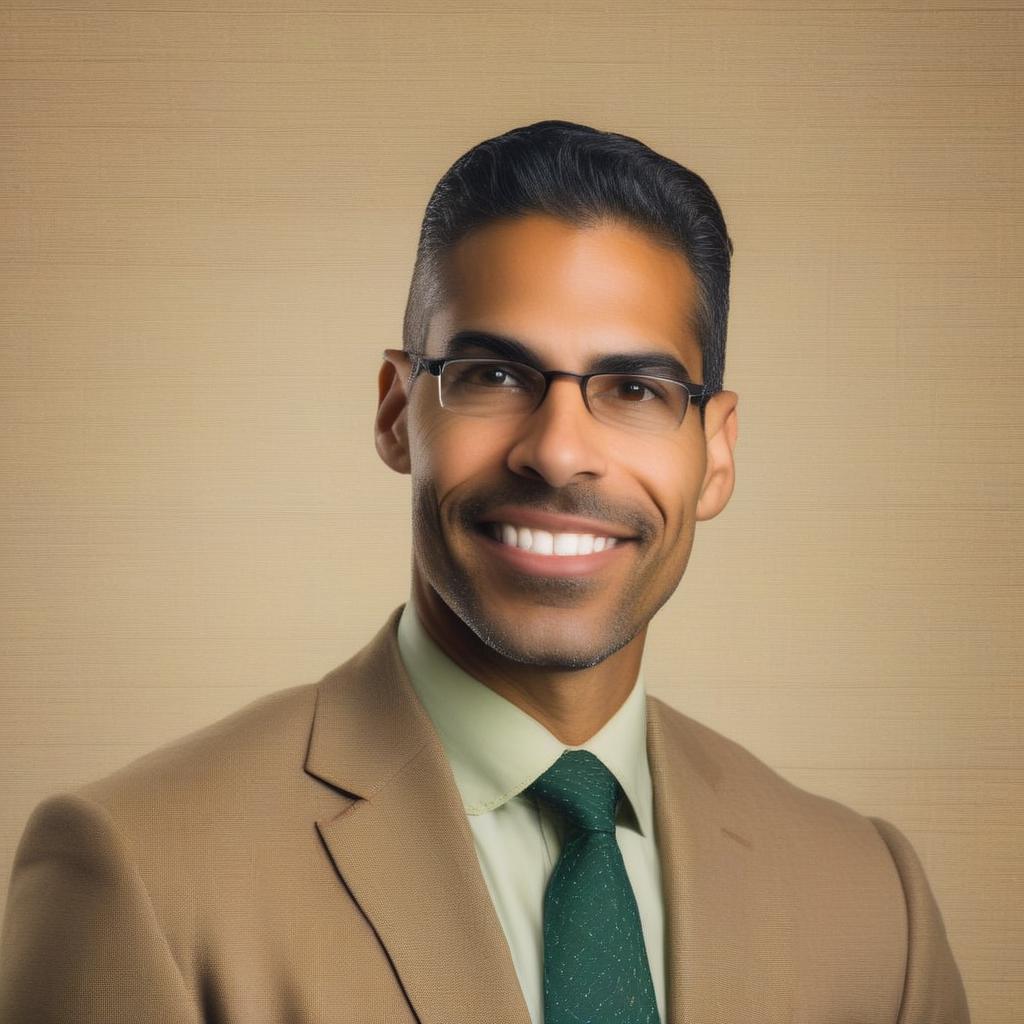}
    \end{subfigure}\hspace*{-0.25em}
    \begin{subfigure}{0.16\textwidth}
        \includegraphics[width=1\linewidth]{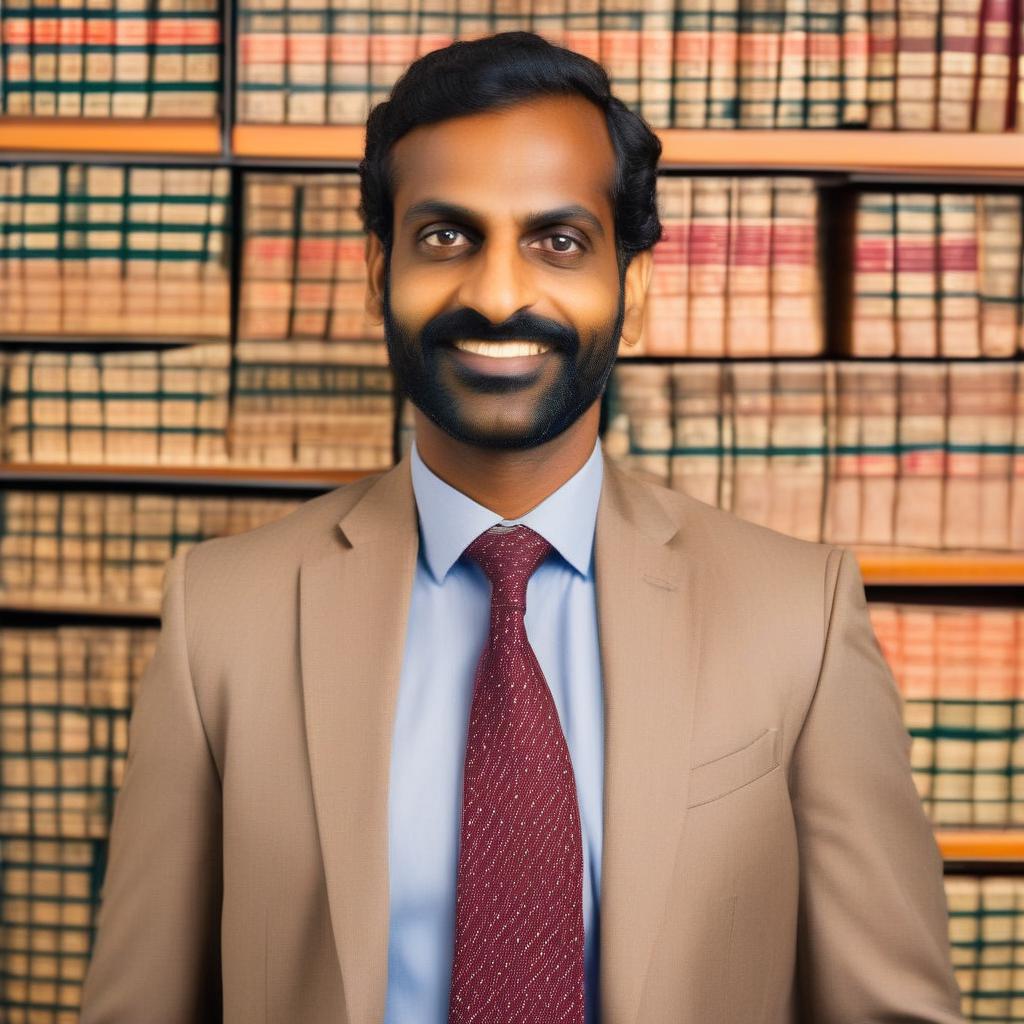}
    \end{subfigure}\hspace*{-0.25em}
    \begin{subfigure}{0.16\textwidth}
        \includegraphics[width=1\linewidth]{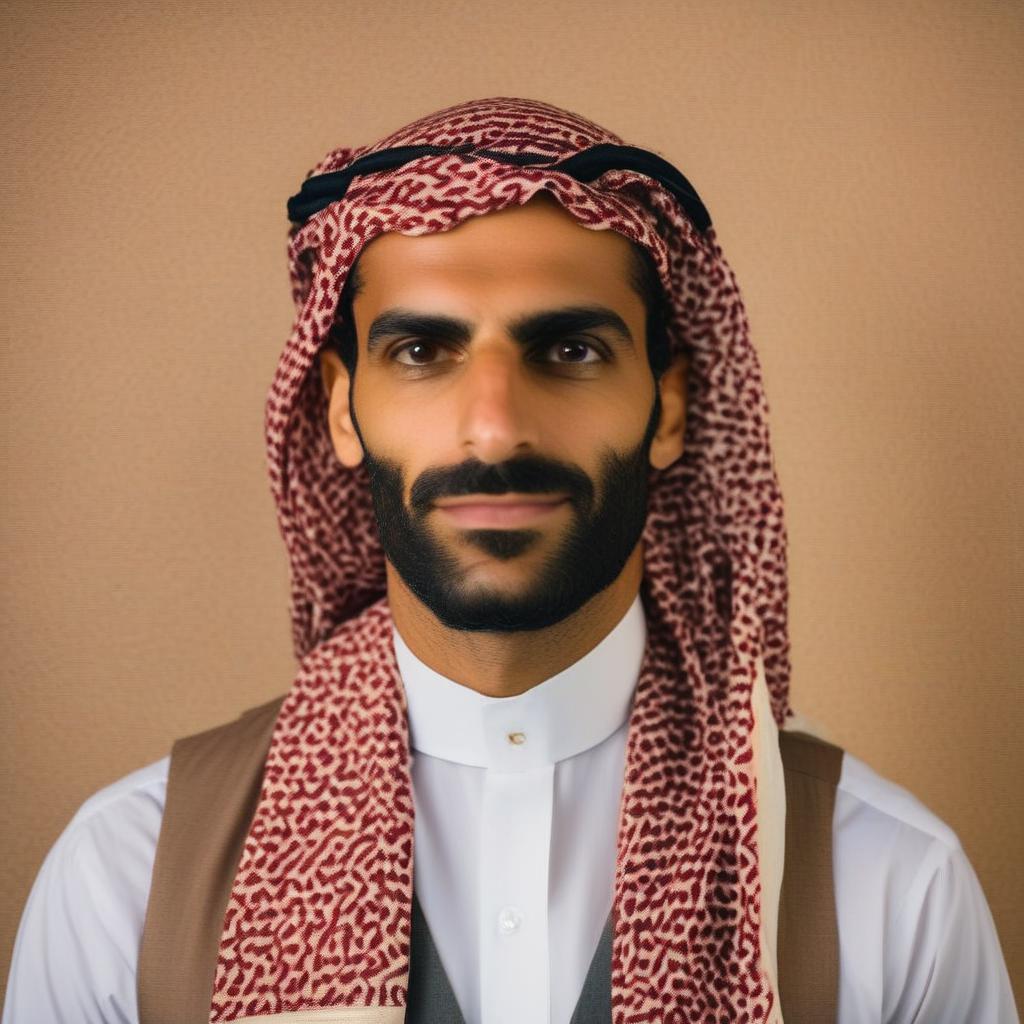}
    \end{subfigure}\hspace*{-0.25em}
    \begin{subfigure}{0.16\textwidth}
        \includegraphics[width=1\linewidth]{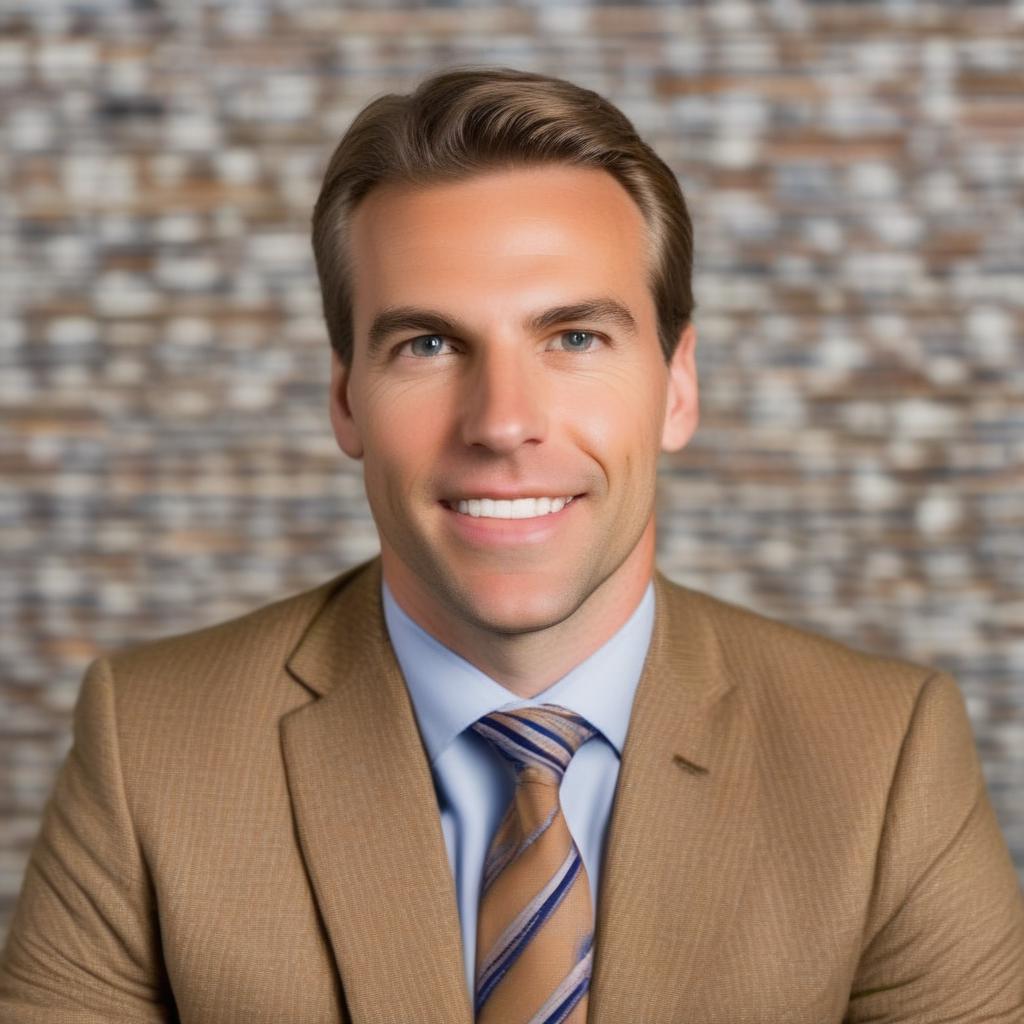}
    \end{subfigure}
    \begin{subfigure}{0.16\textwidth}
        \includegraphics[width=1\linewidth]{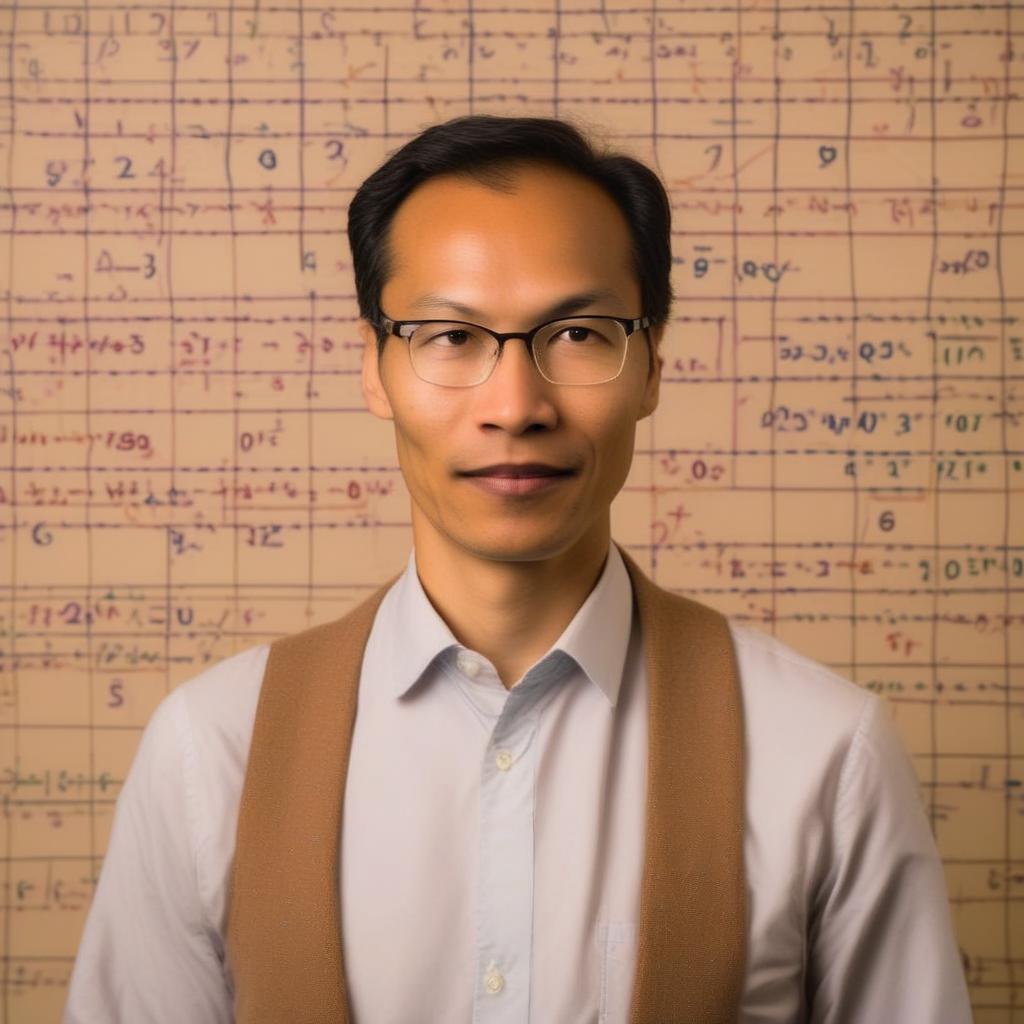}
    \end{subfigure}\hspace*{-0.25em}
    \begin{subfigure}{0.16\textwidth}
        \includegraphics[width=1\linewidth]{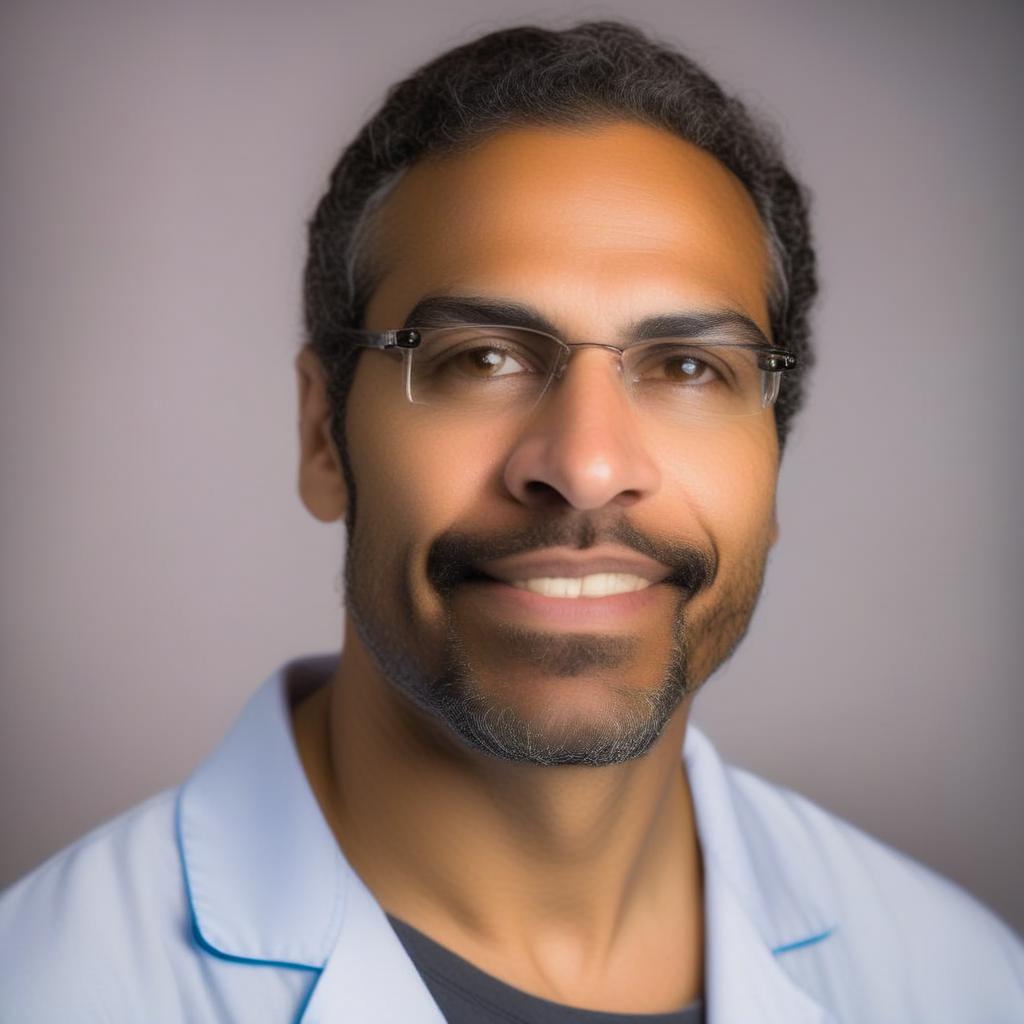}
    \end{subfigure}\hspace*{-0.25em}
    \begin{subfigure}{0.16\textwidth}
        \includegraphics[width=1\linewidth]{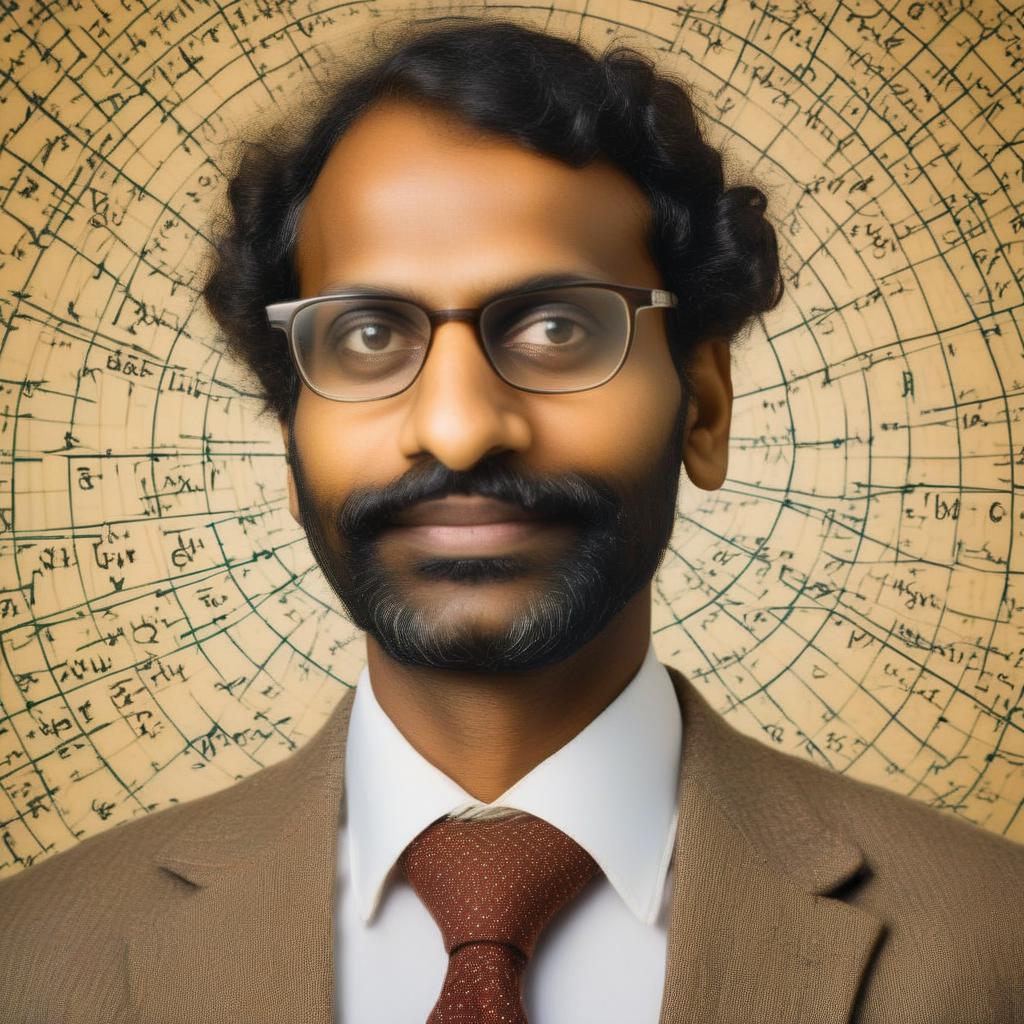}
    \end{subfigure}\hspace*{-0.25em}
    \begin{subfigure}{0.16\textwidth}
        \includegraphics[width=1\linewidth]{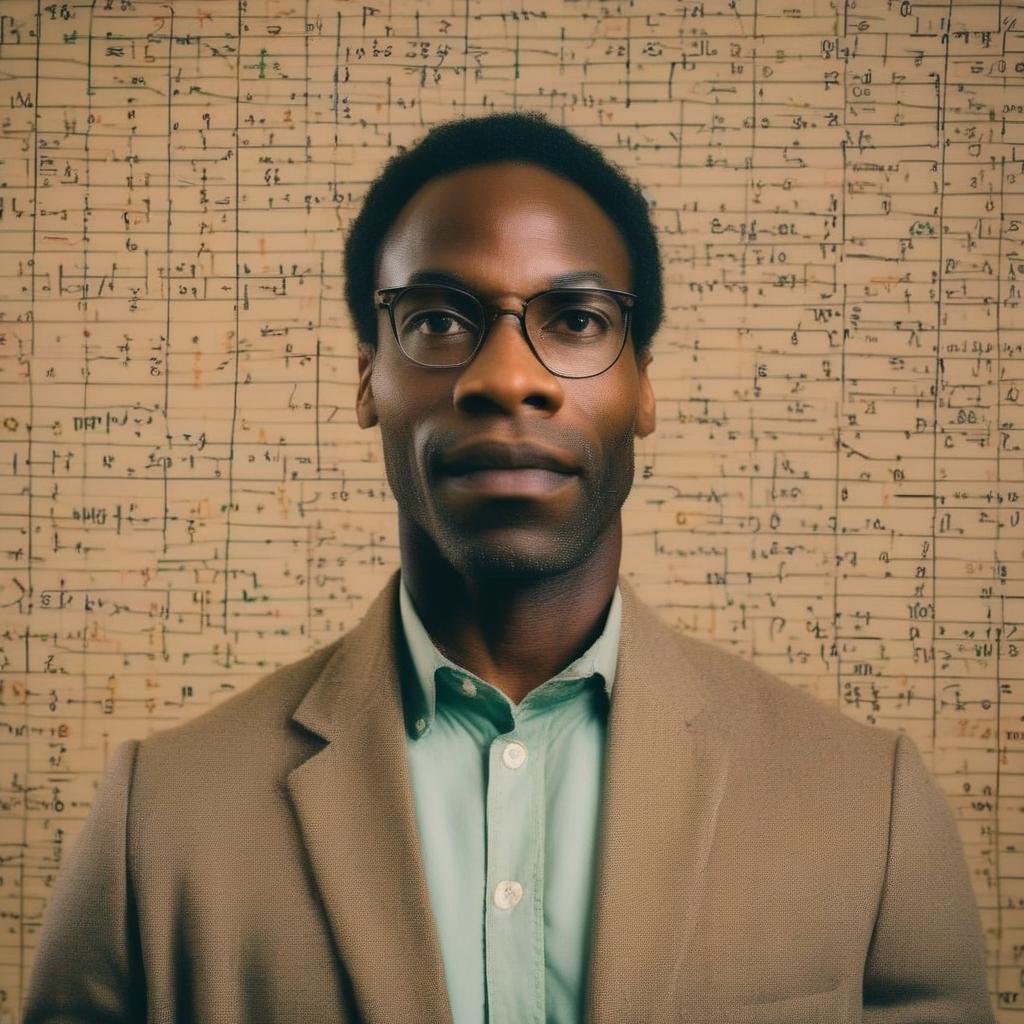}
    \end{subfigure}\hspace*{-0.25em}
    \begin{subfigure}{0.16\textwidth}
        \includegraphics[width=1\linewidth]{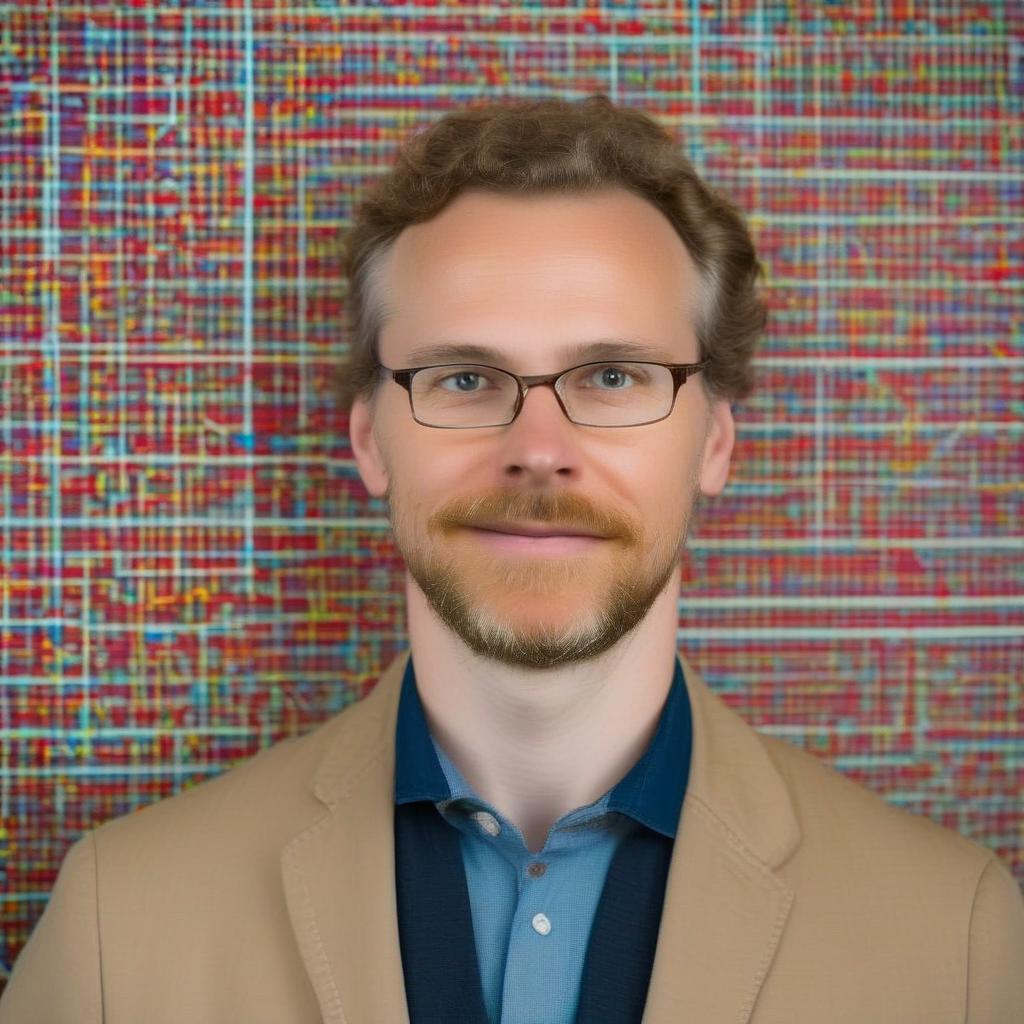}
    \end{subfigure}\hspace*{-0.25em}
    \begin{subfigure}{0.16\textwidth}
        \includegraphics[width=1\linewidth]{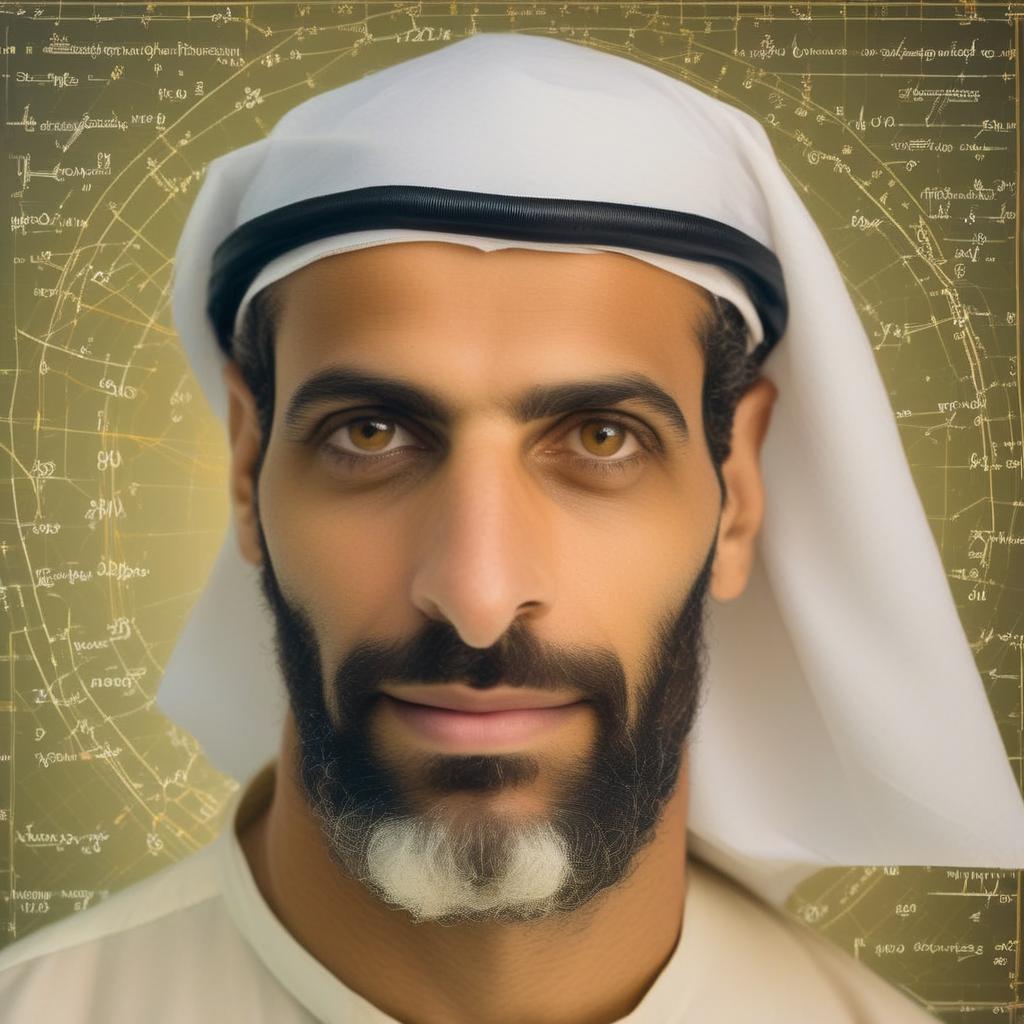}
    \end{subfigure}
    \begin{subfigure}{0.16\textwidth}
        \includegraphics[width=1\linewidth]{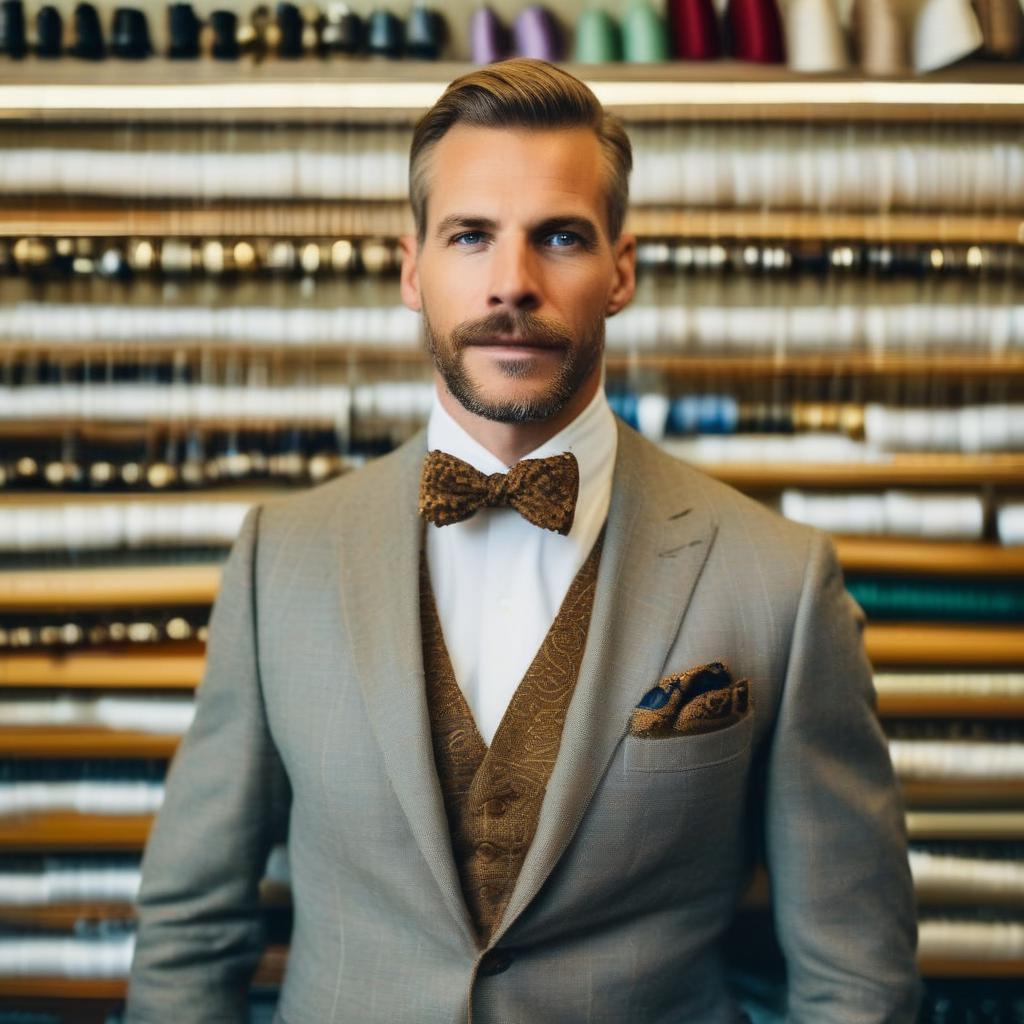}
    \end{subfigure}\hspace*{-0.25em}
    \begin{subfigure}{0.16\textwidth}
        \includegraphics[width=1\linewidth]{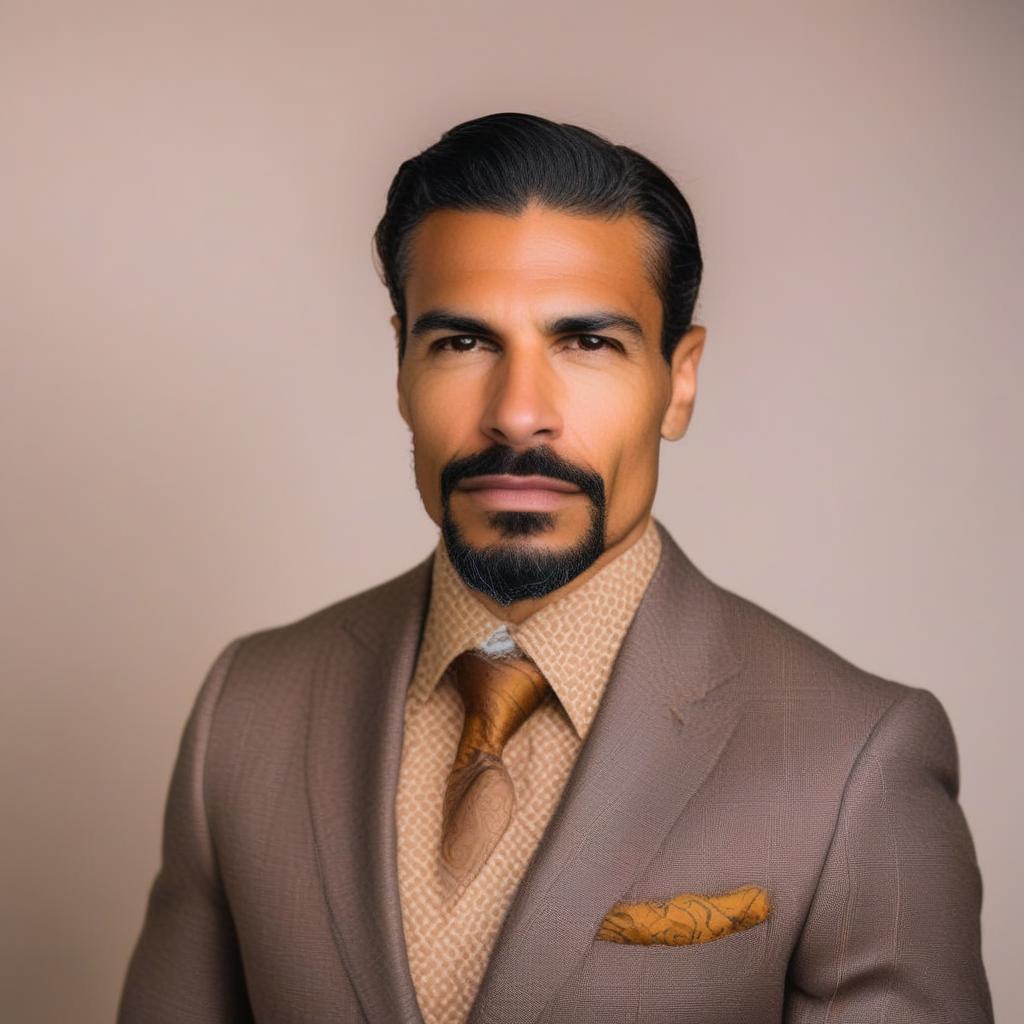}
    \end{subfigure}\hspace*{-0.25em}
    \begin{subfigure}{0.16\textwidth}
        \includegraphics[width=1\linewidth]{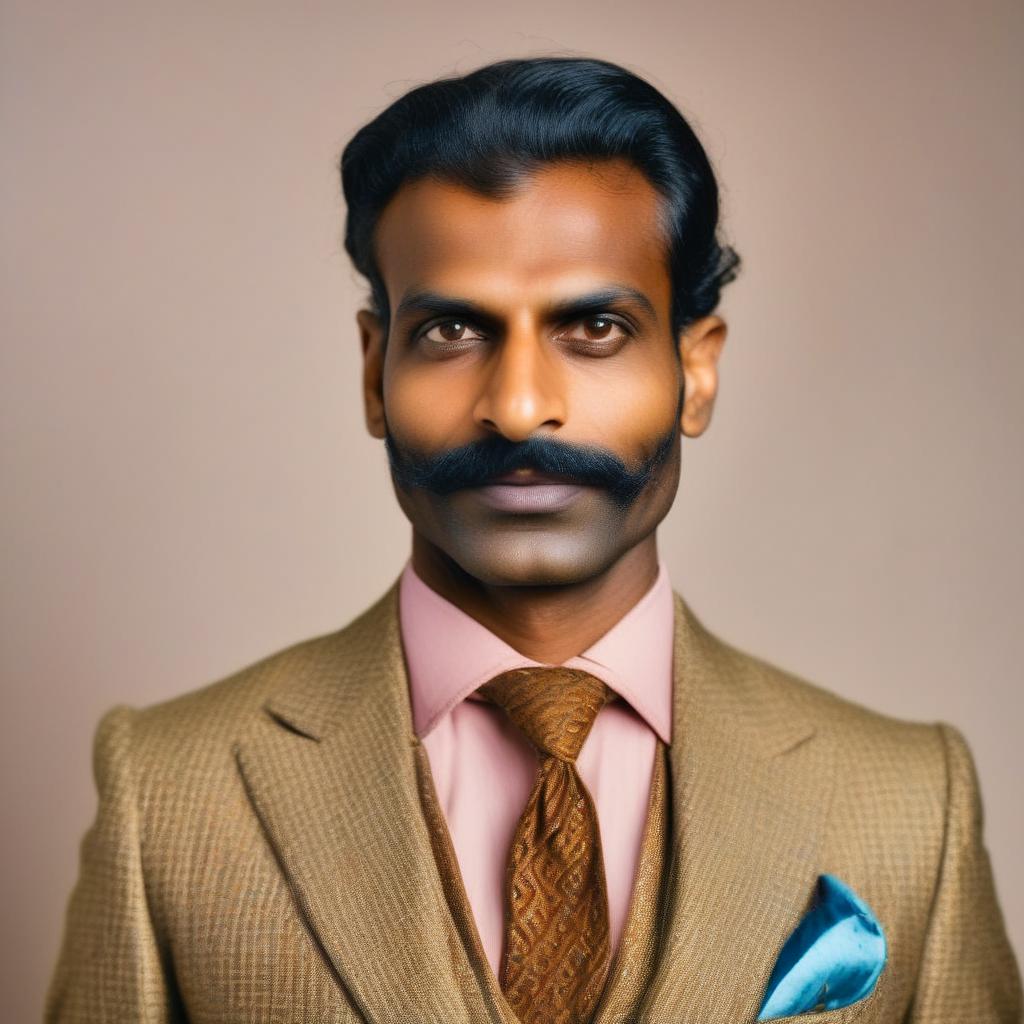}
    \end{subfigure}\hspace*{-0.25em}
    \begin{subfigure}{0.16\textwidth}
        \includegraphics[width=1\linewidth]{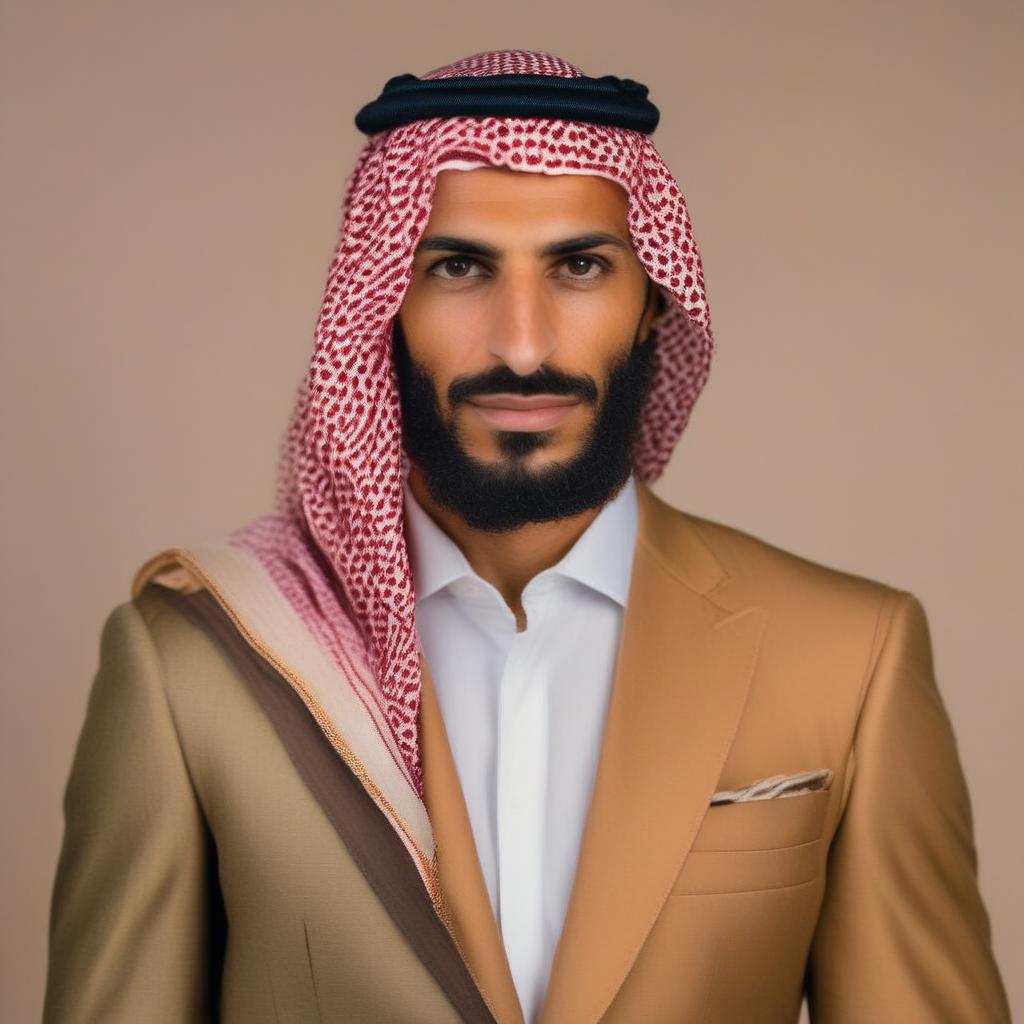}
    \end{subfigure}\hspace*{-0.25em}
    \begin{subfigure}{0.16\textwidth}
        \includegraphics[width=1\linewidth]{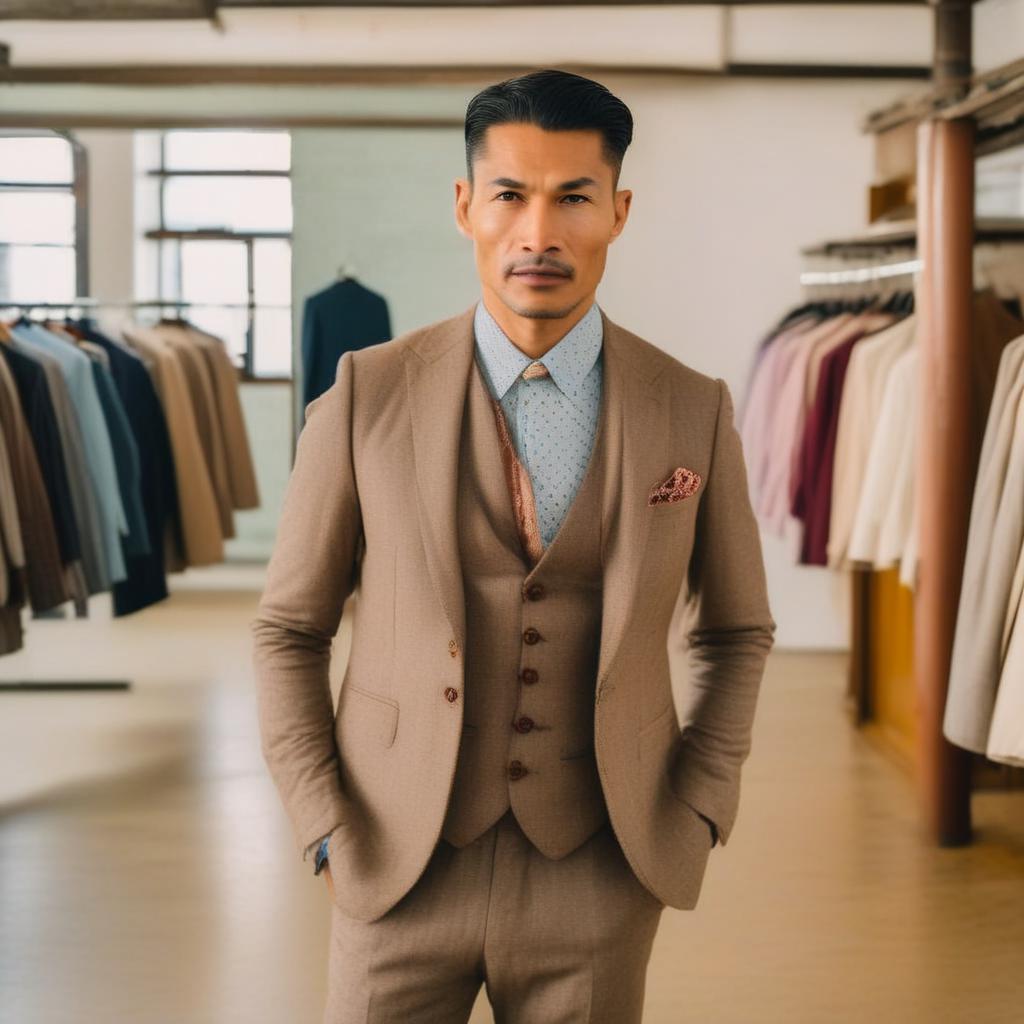}
    \end{subfigure}\hspace*{-0.25em}
    \begin{subfigure}{0.16\textwidth}
        \includegraphics[width=1\linewidth]{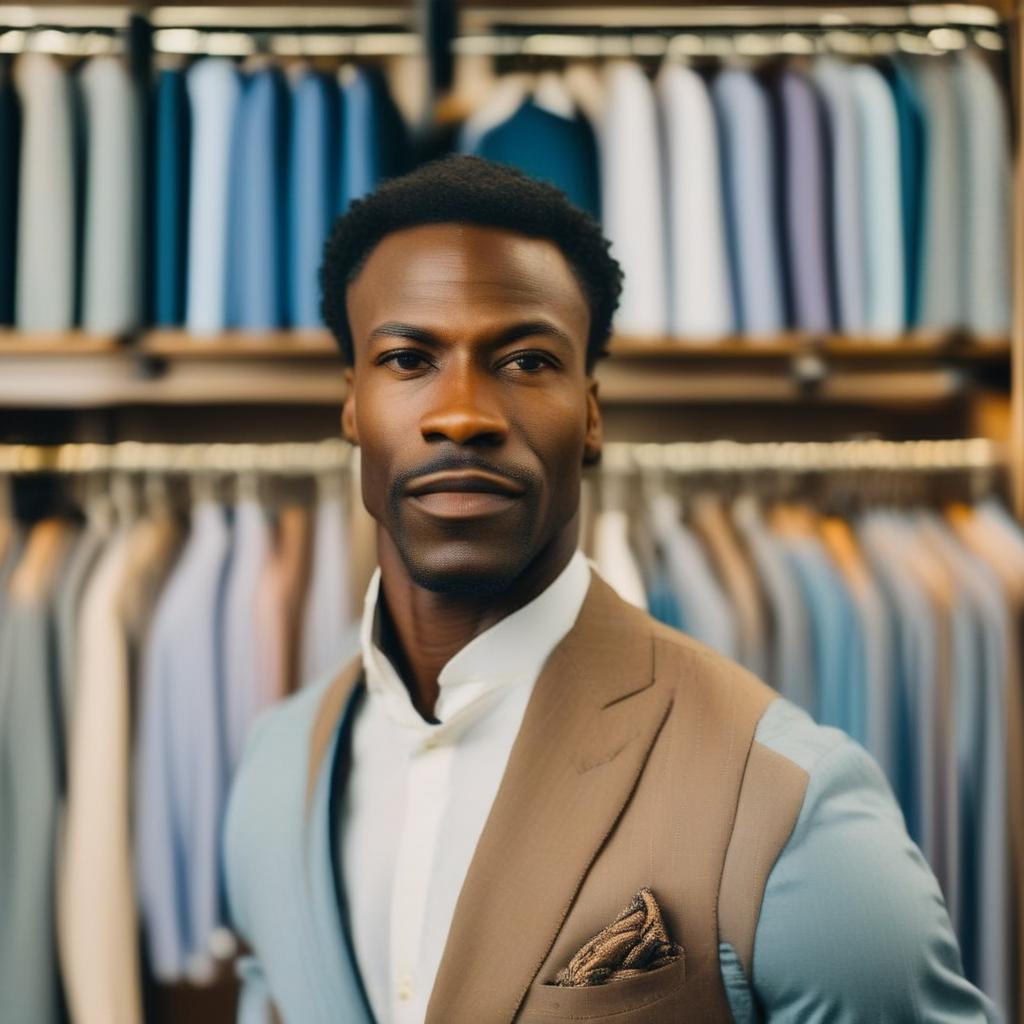}
    \end{subfigure}\hspace*{-0.25em}    
    \vspace{-10pt}
    \begin{flushleft} \textbf{b} \end{flushleft}
    \begin{subfigure}{0.16\textwidth}
        \includegraphics[width=1\linewidth]{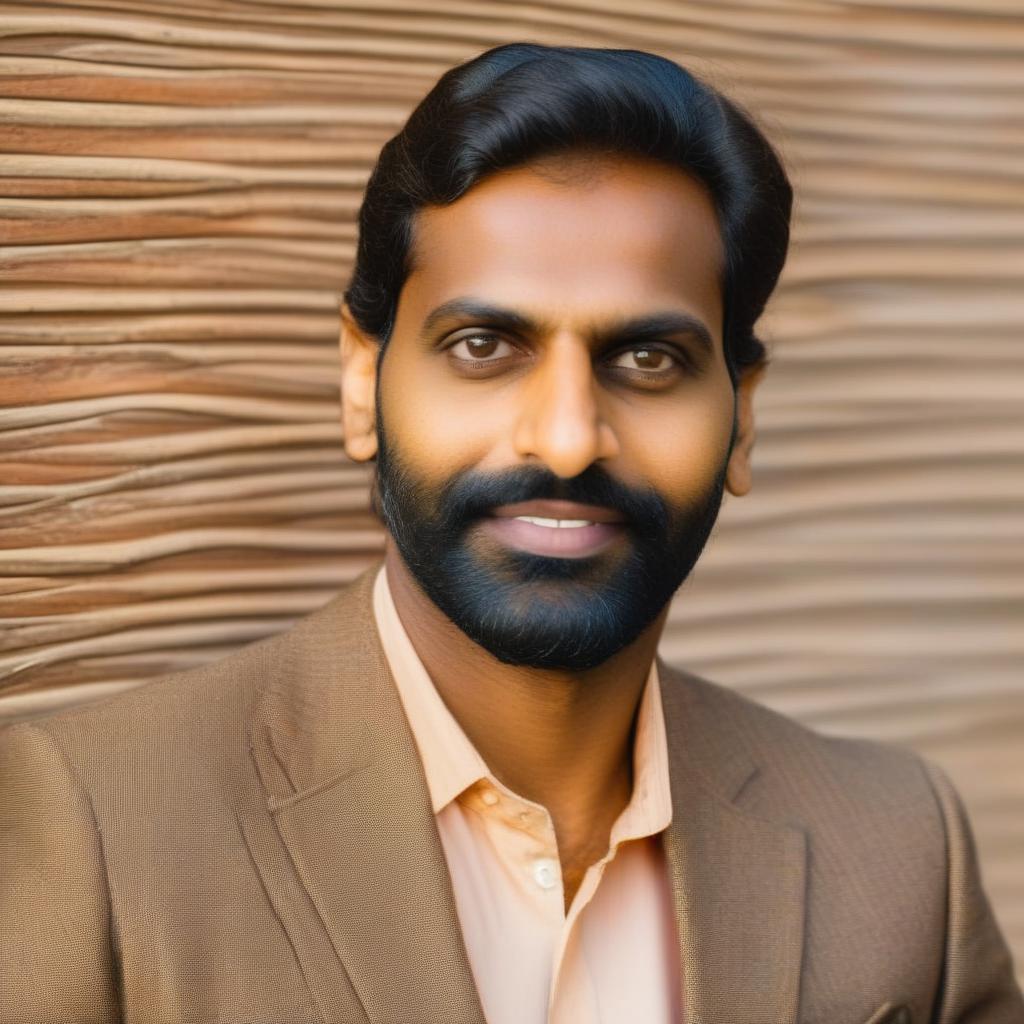}
    \end{subfigure}\hspace*{-0.25em}
    \begin{subfigure}{0.16\textwidth}
        \includegraphics[width=1\linewidth]{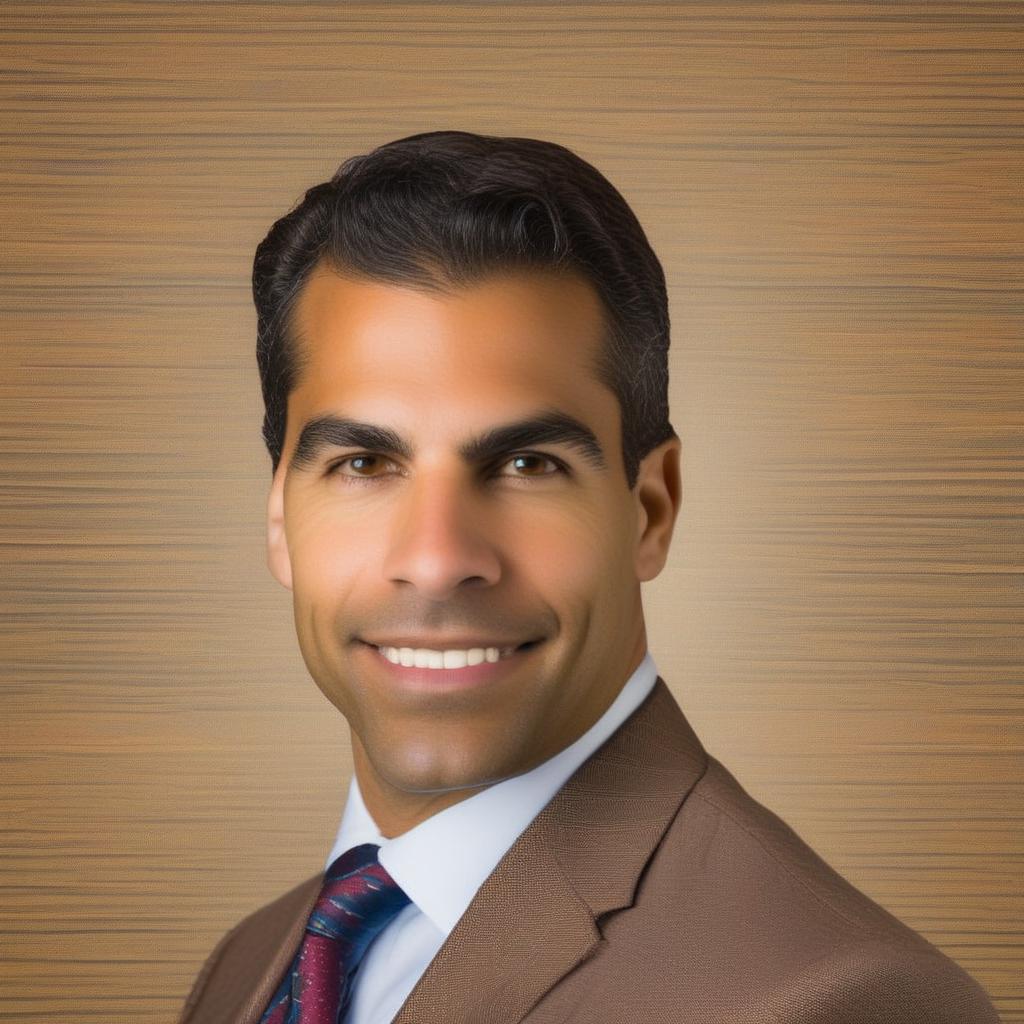}
    \end{subfigure}\hspace*{-0.25em}
    \begin{subfigure}{0.16\textwidth}
        \includegraphics[width=1\linewidth]{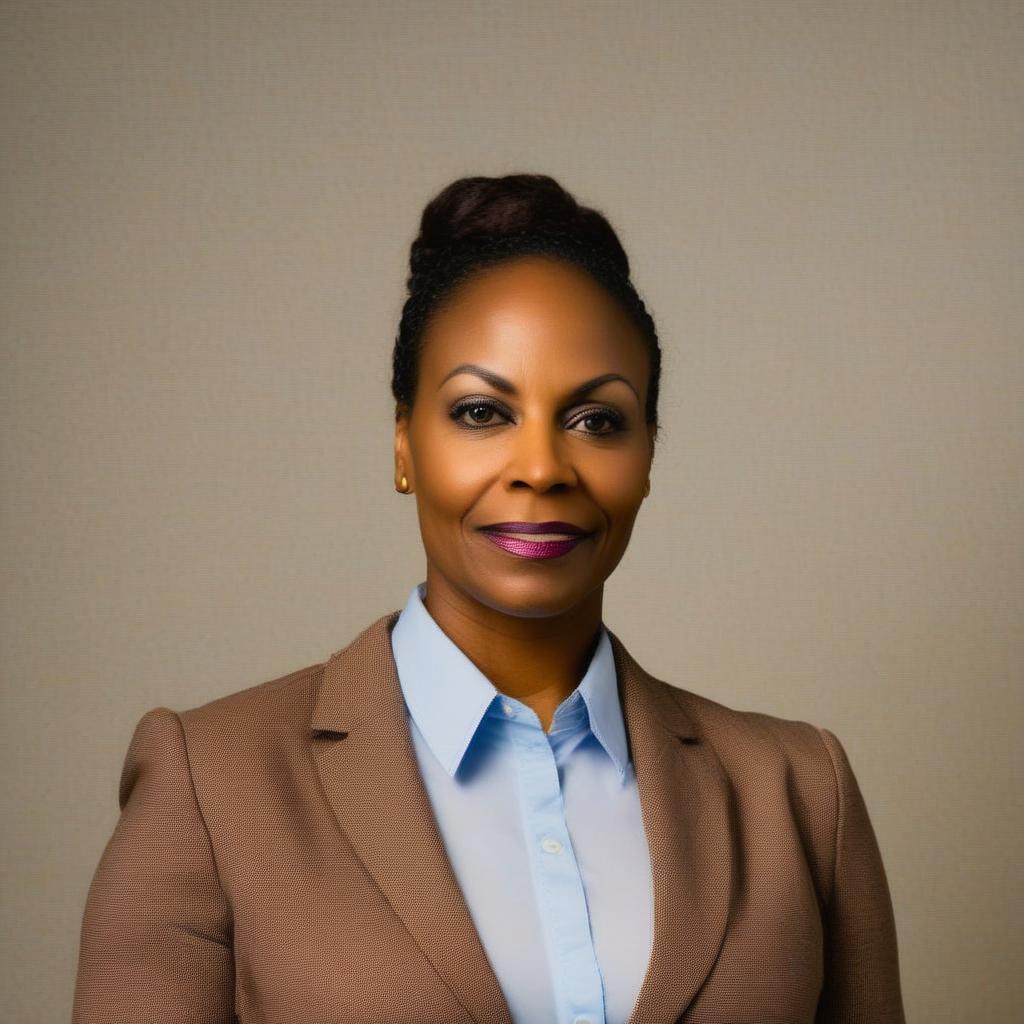}
    \end{subfigure}\hspace*{-0.25em} 
    \begin{subfigure}{0.16\textwidth}
        \includegraphics[width=1\linewidth]{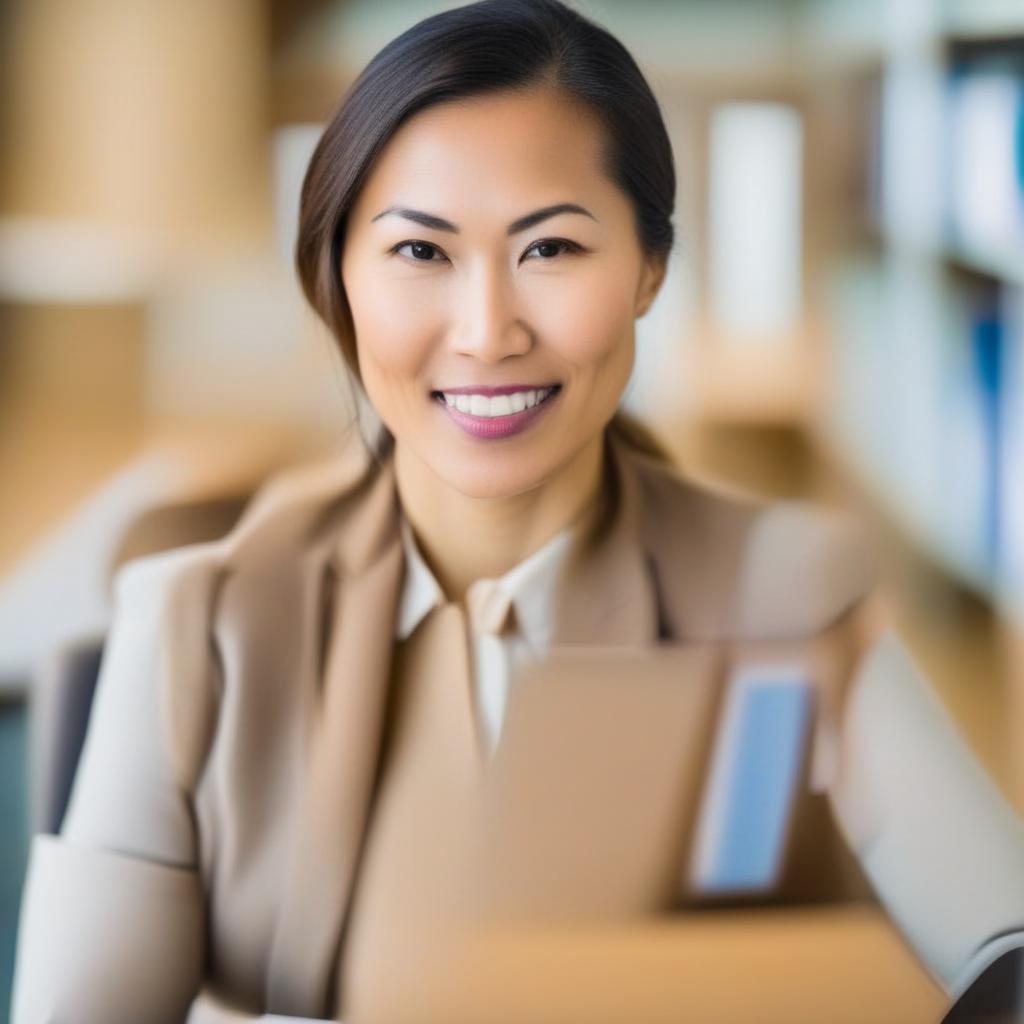}
    \end{subfigure}\hspace*{-0.25em}
    \begin{subfigure}{0.16\textwidth}
        \includegraphics[width=1\linewidth]{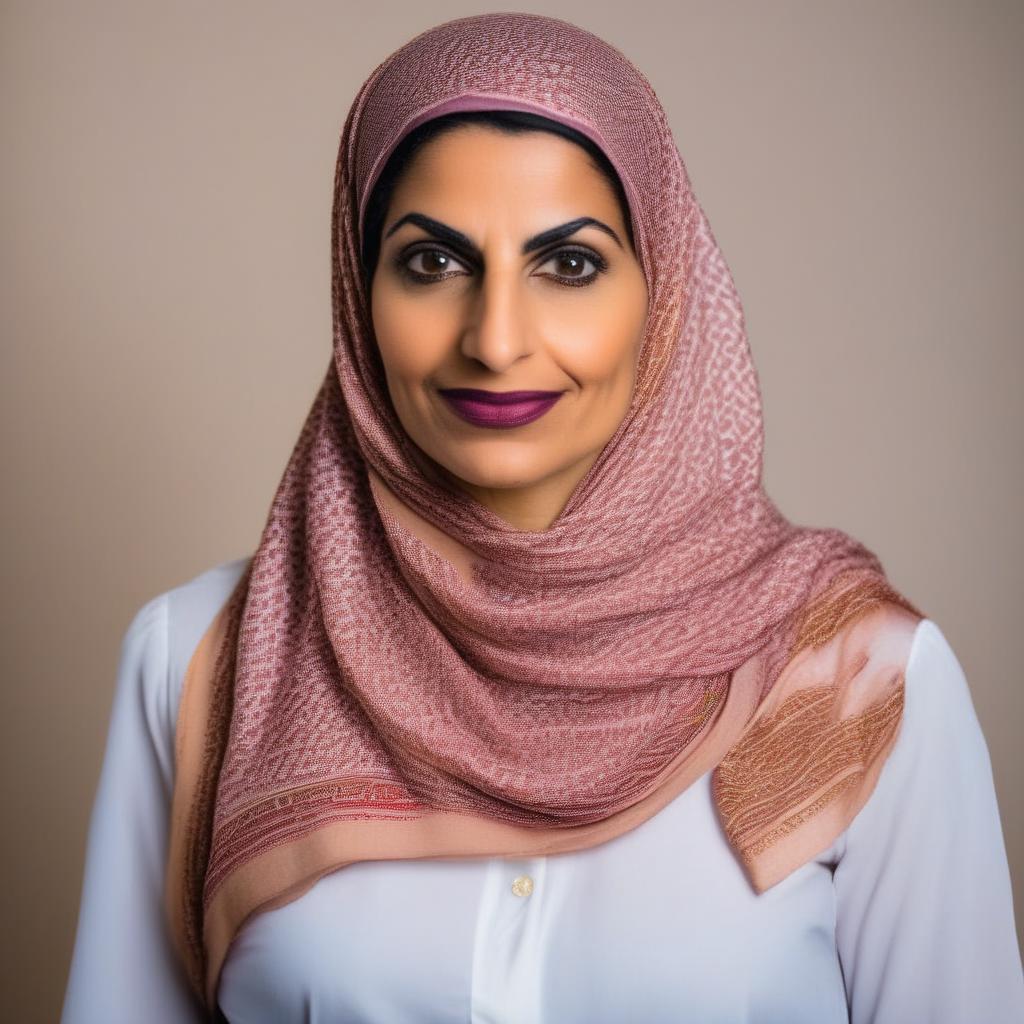}
    \end{subfigure}\hspace*{-0.25em}
    \begin{subfigure}{0.16\textwidth}
        \includegraphics[width=1\linewidth]{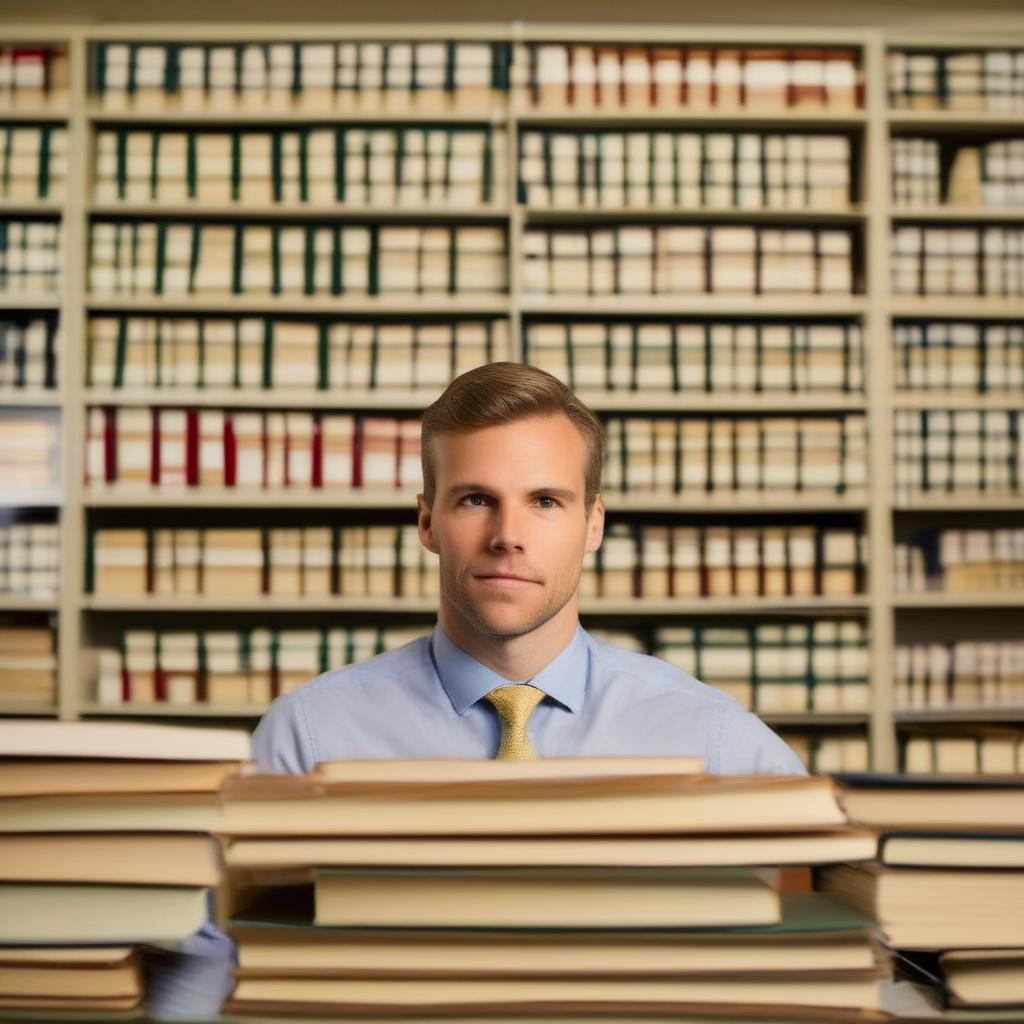}
    \end{subfigure} 
    \begin{subfigure}{0.16\textwidth}
        \includegraphics[width=1\linewidth]{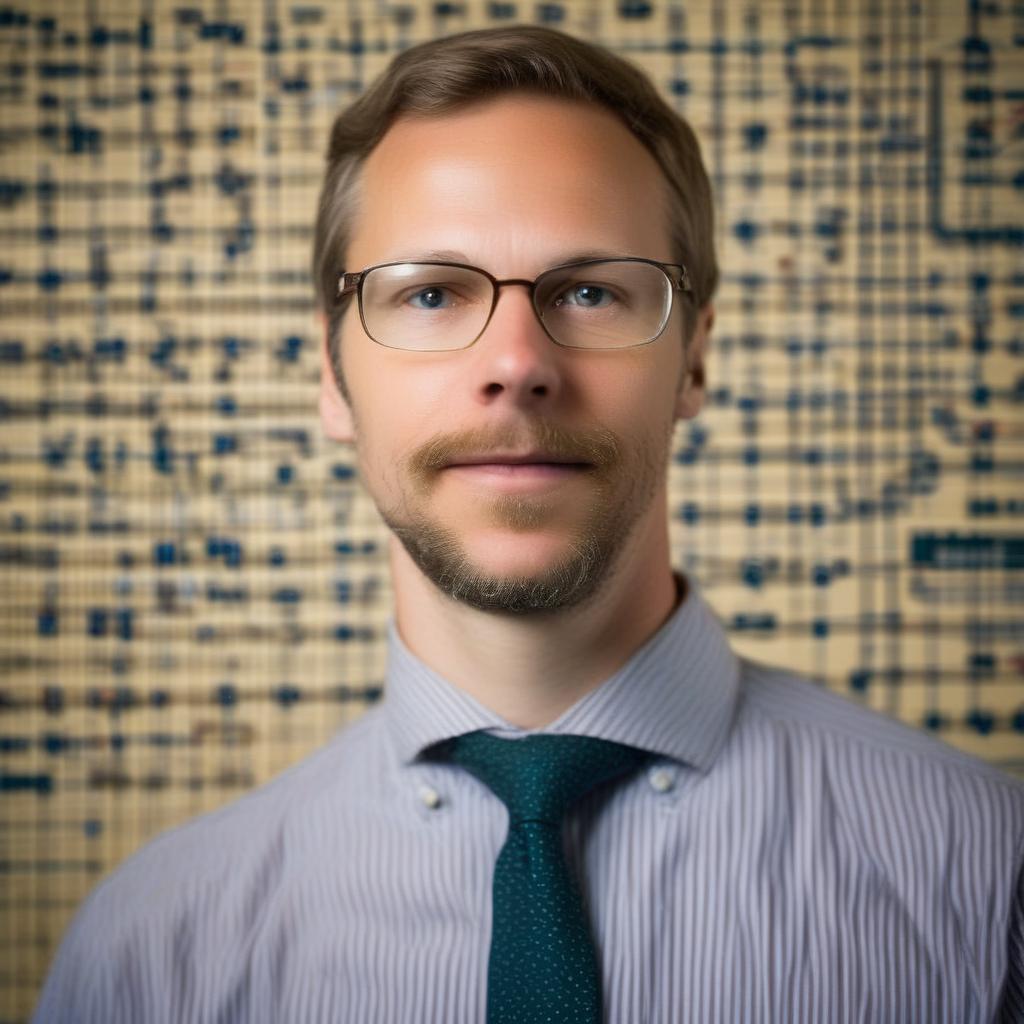}
    \end{subfigure}\hspace*{-0.25em}
    \begin{subfigure}{0.16\textwidth}
        \includegraphics[width=1\linewidth]{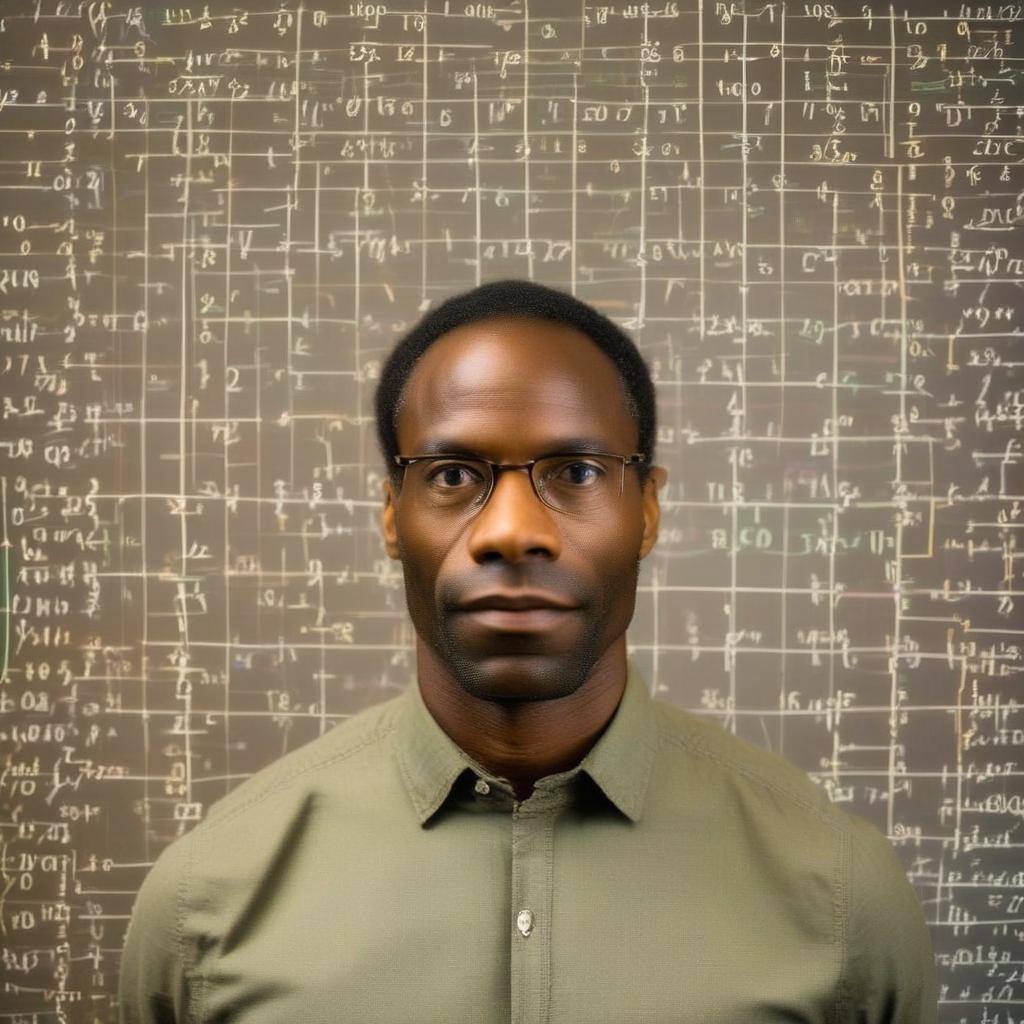}
    \end{subfigure}\hspace*{-0.25em}
    \begin{subfigure}{0.16\textwidth}
        \includegraphics[width=1\linewidth]{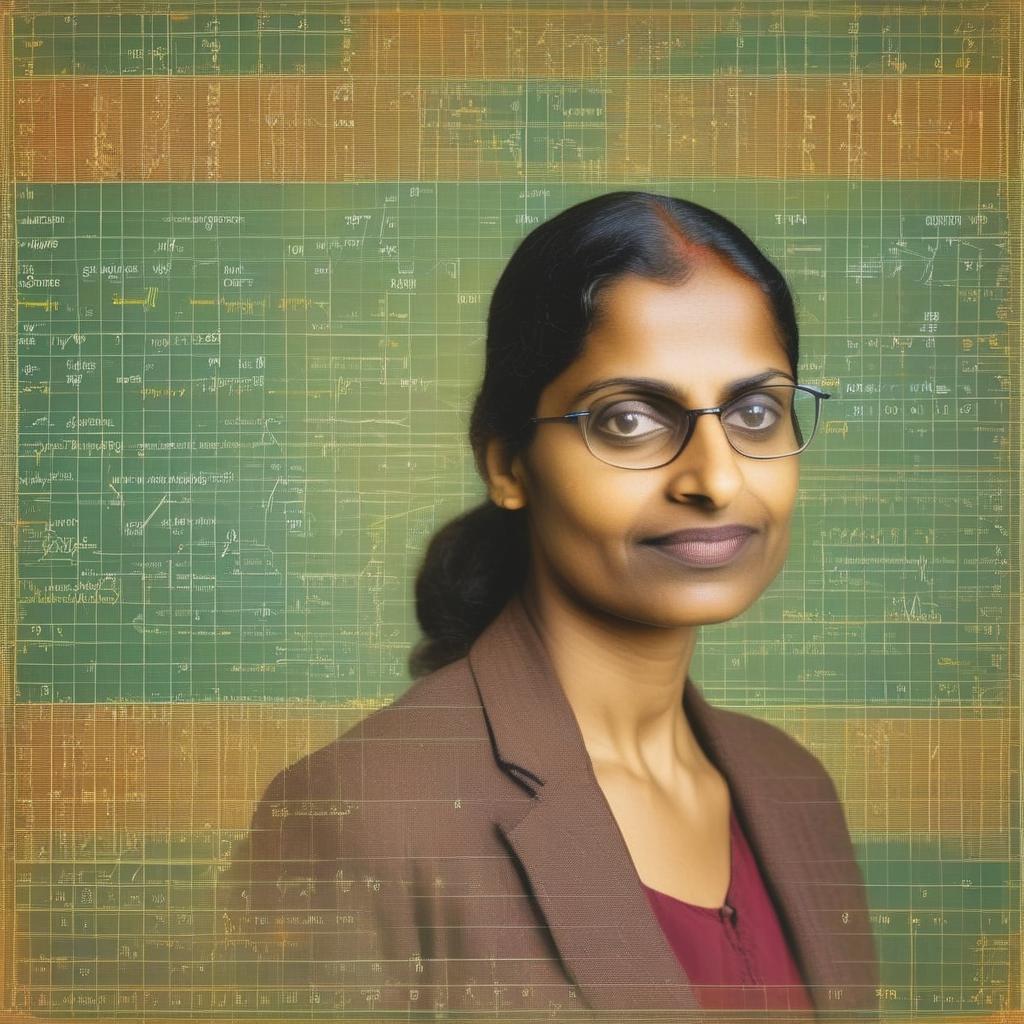}
    \end{subfigure}\hspace*{-0.25em}
    \begin{subfigure}{0.16\textwidth}
        \includegraphics[width=1\linewidth]{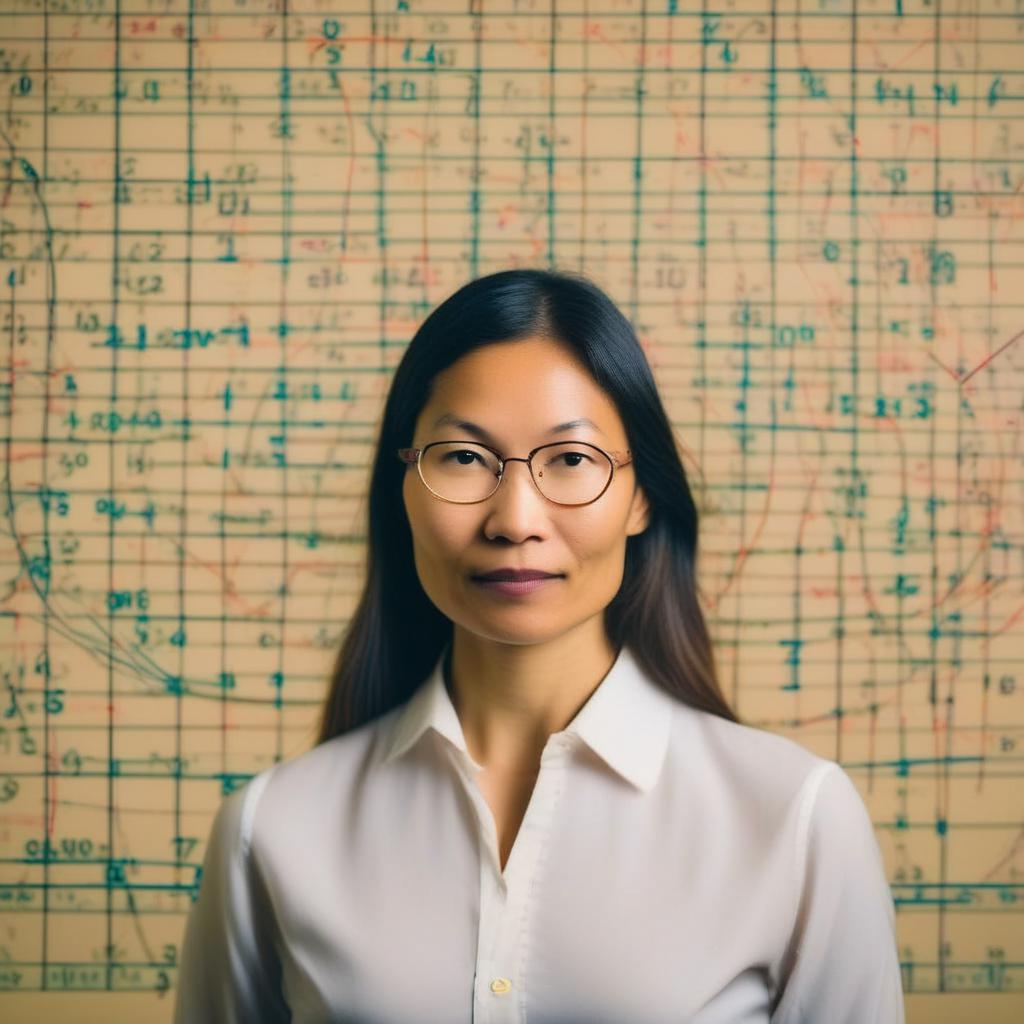}
    \end{subfigure}\hspace*{-0.25em}
    \begin{subfigure}{0.16\textwidth}
        \includegraphics[width=1\linewidth]{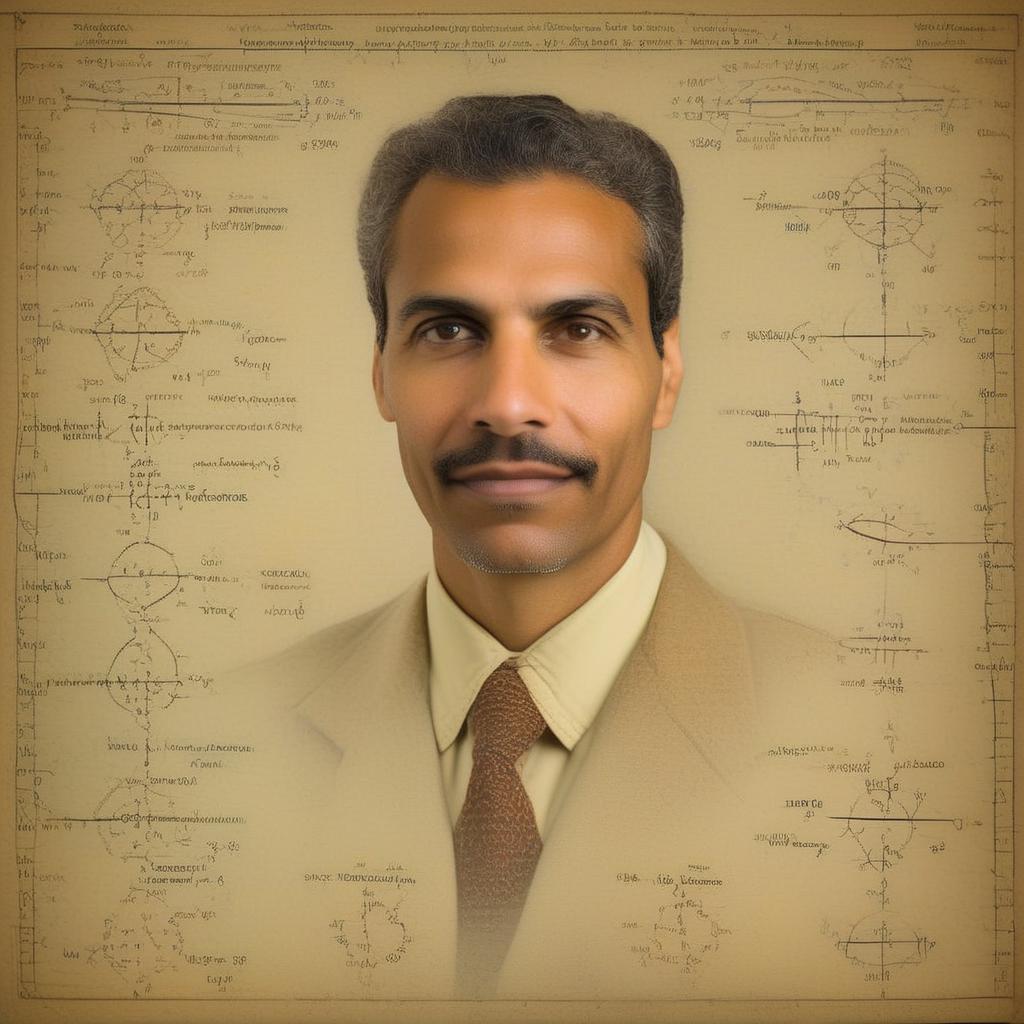}
    \end{subfigure}\hspace*{-0.25em}
    \begin{subfigure}{0.16\textwidth}
        \includegraphics[width=1\linewidth]{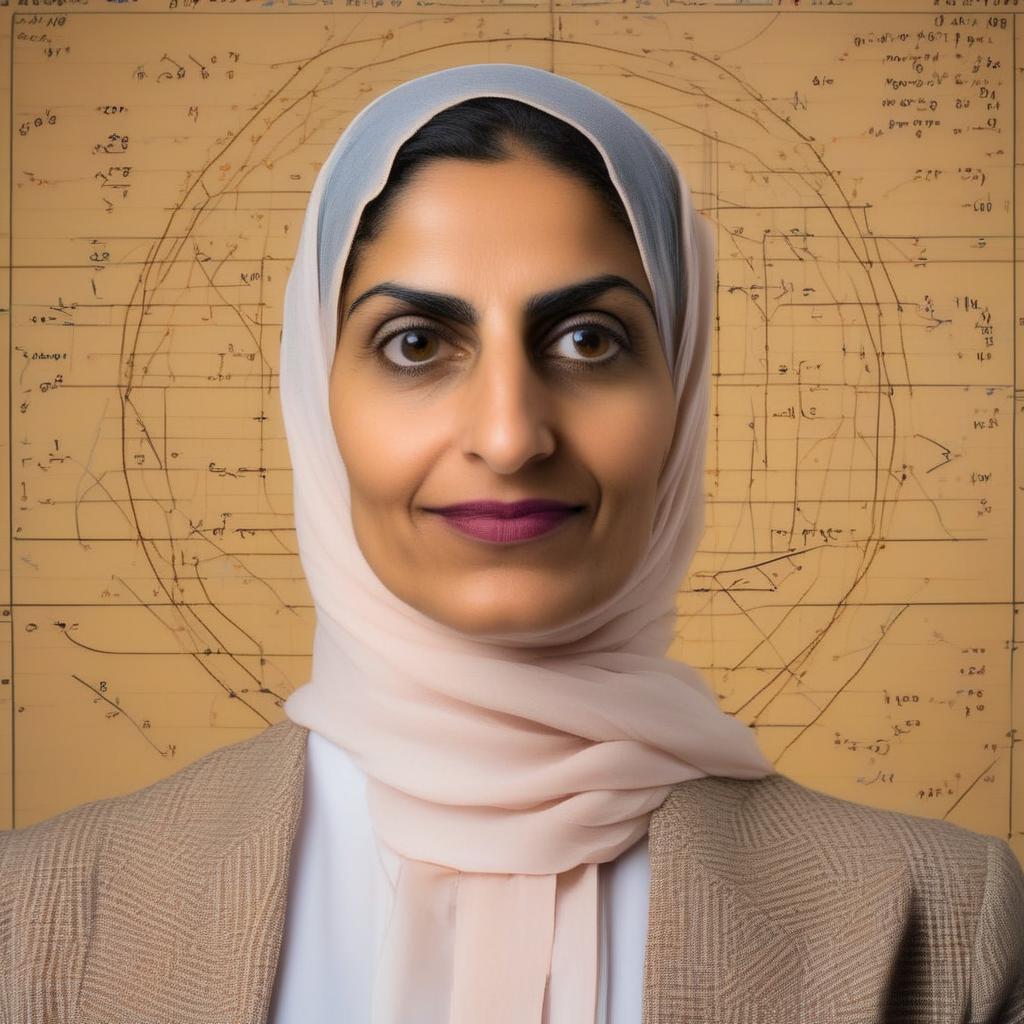}
    \end{subfigure}
    \begin{subfigure}{0.16\textwidth}
        \includegraphics[width=1\linewidth]{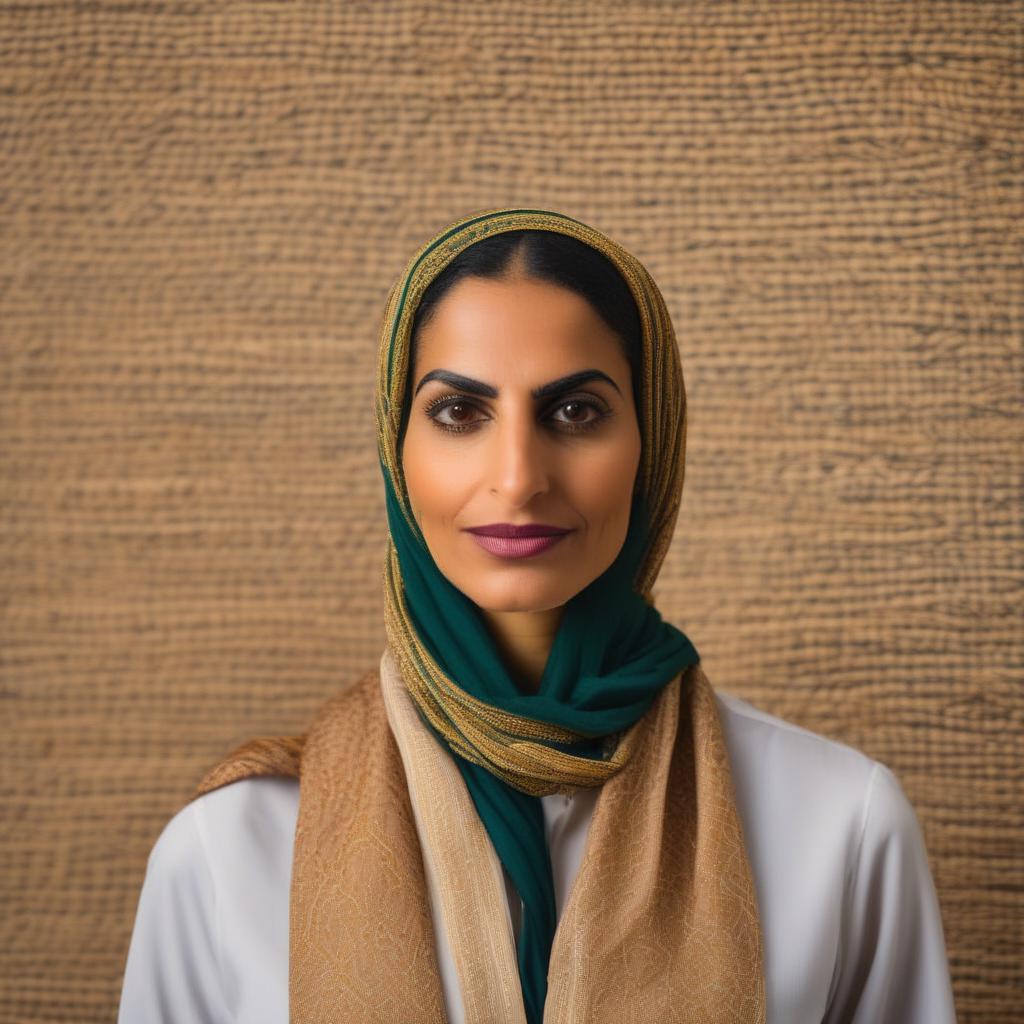}
    \end{subfigure}\hspace*{-0.25em}
    \begin{subfigure}{0.16\textwidth}
        \includegraphics[width=1\linewidth]{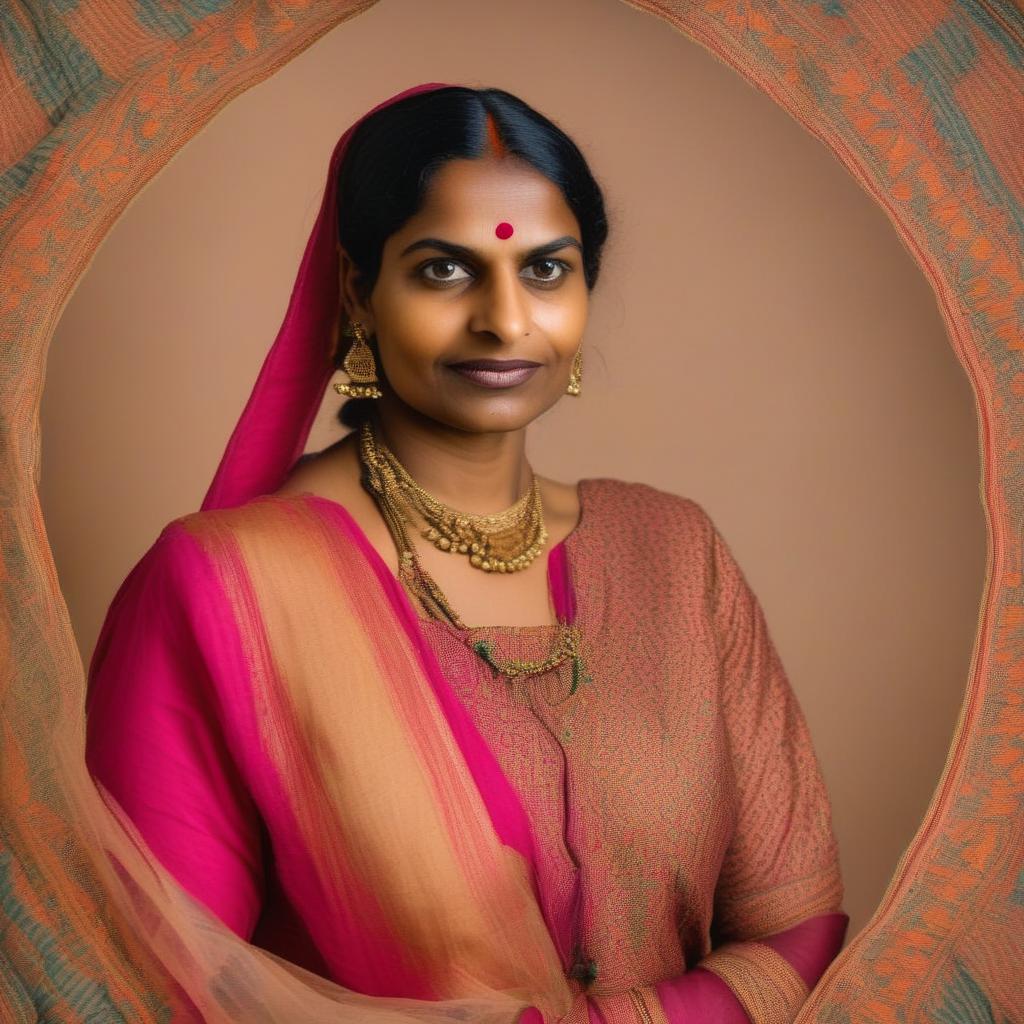}
    \end{subfigure}\hspace*{-0.25em}
    \begin{subfigure}{0.16\textwidth}
        \includegraphics[width=1\linewidth]{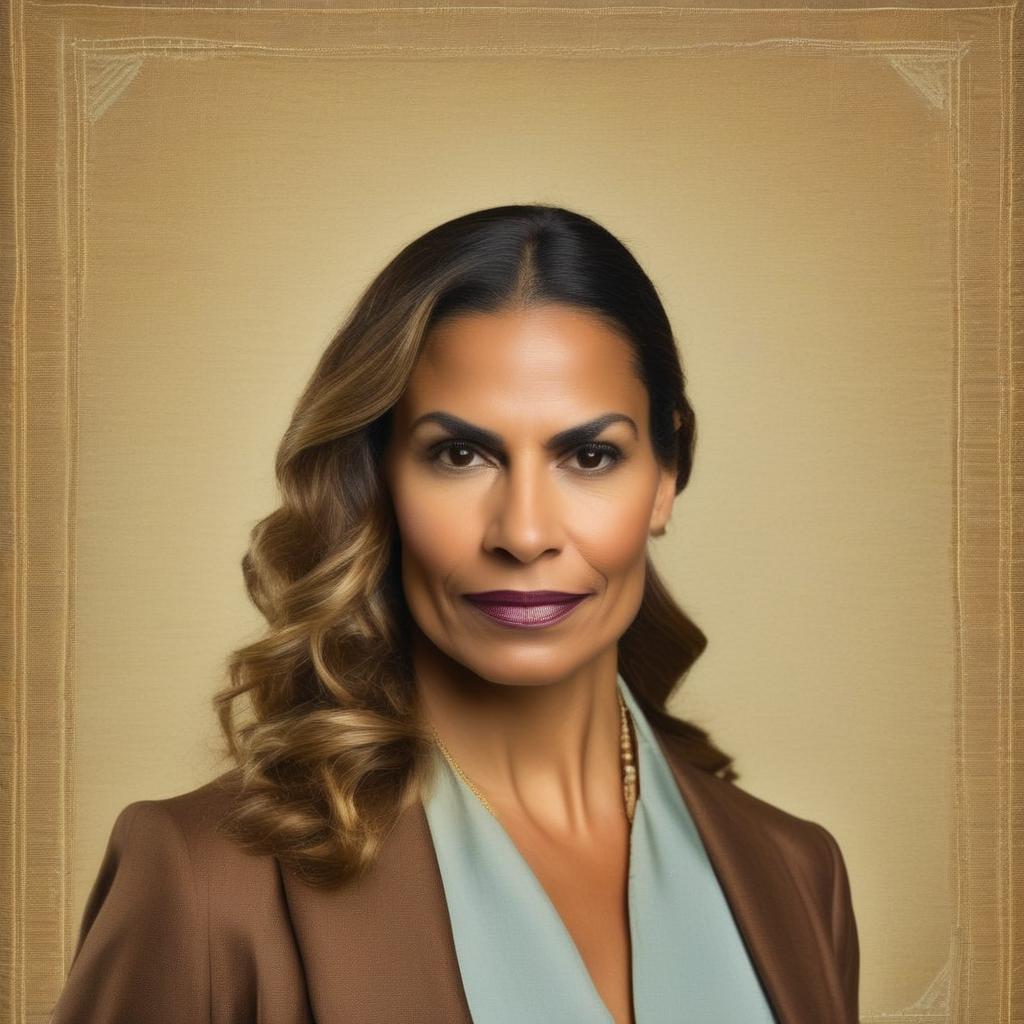}
    \end{subfigure}\hspace*{-0.25em}
    \begin{subfigure}{0.16\textwidth}
        \includegraphics[width=1\linewidth]{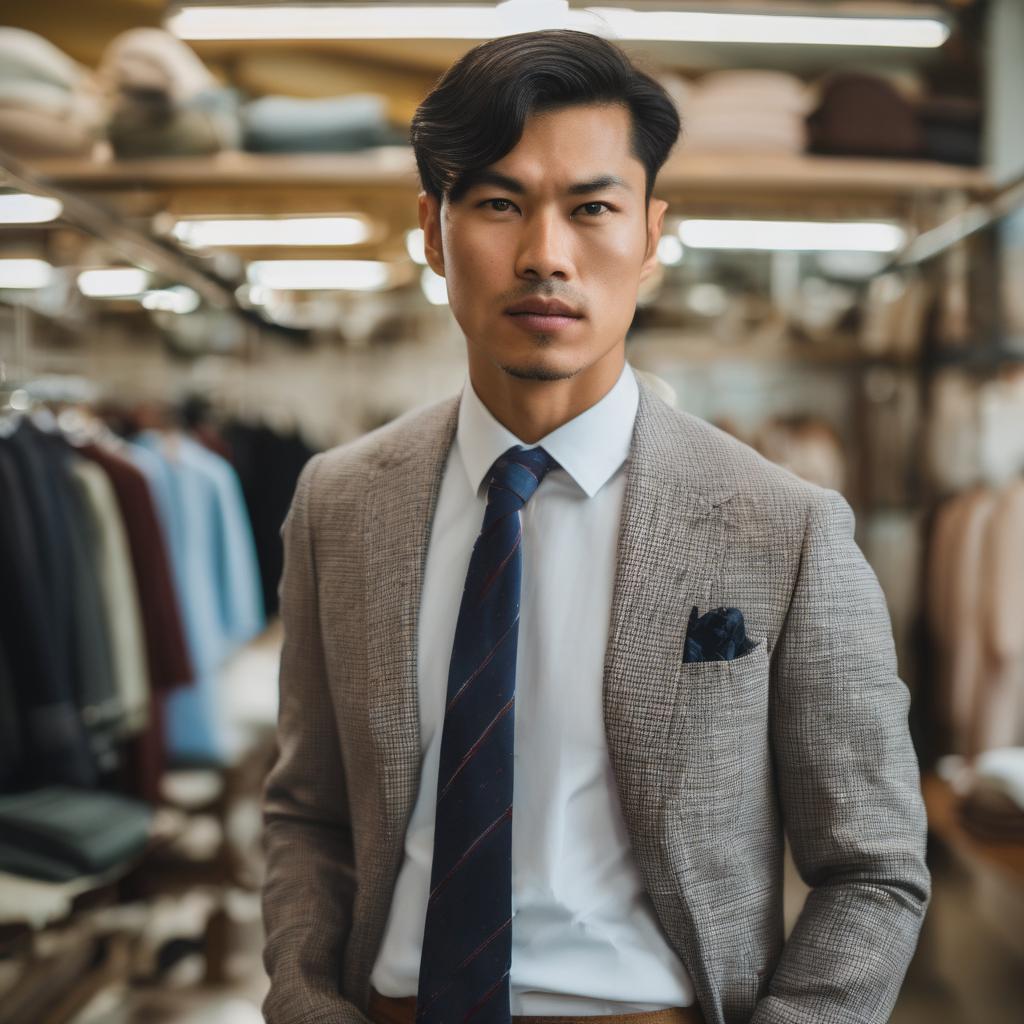}
    \end{subfigure}\hspace*{-0.25em}
    \begin{subfigure}{0.16\textwidth}
        \includegraphics[width=1\linewidth]{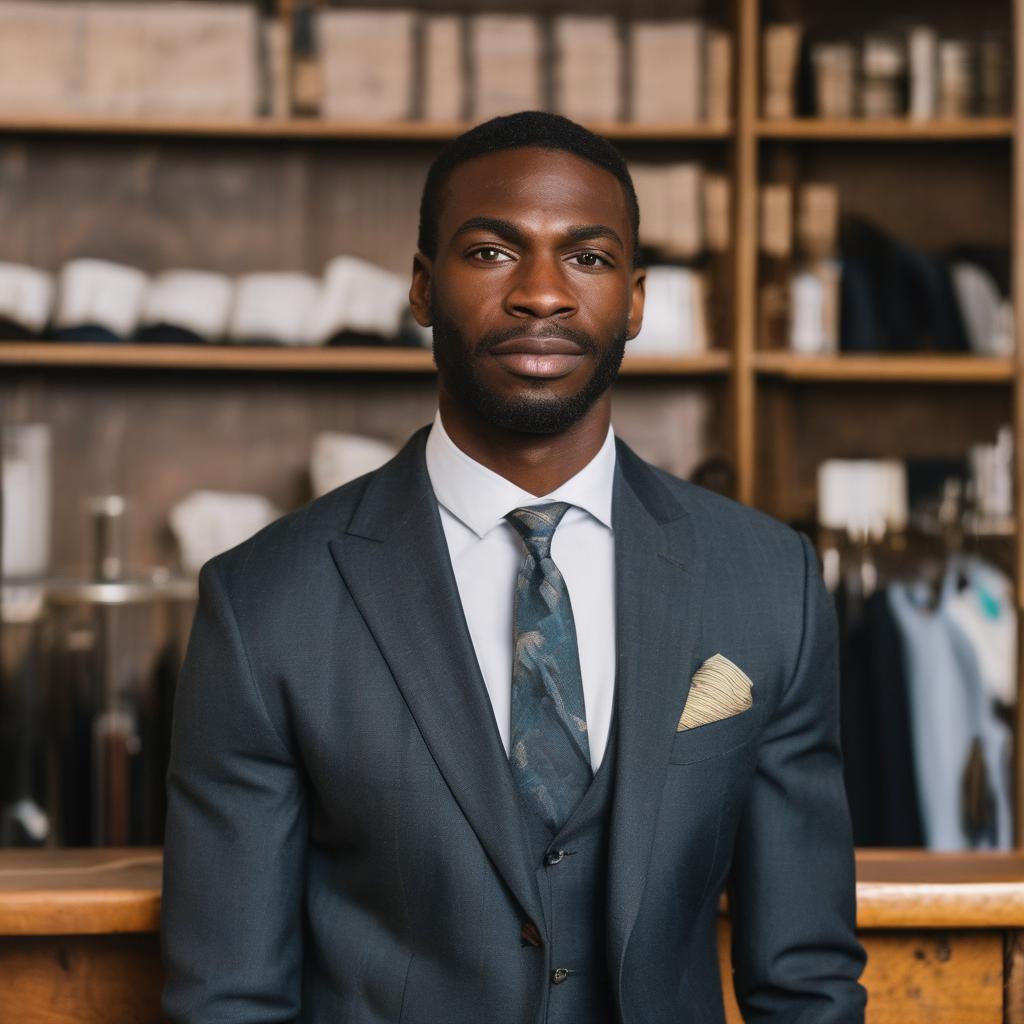}
    \end{subfigure}\hspace*{-0.25em}
    \begin{subfigure}{0.16\textwidth}
        \includegraphics[width=1\linewidth]{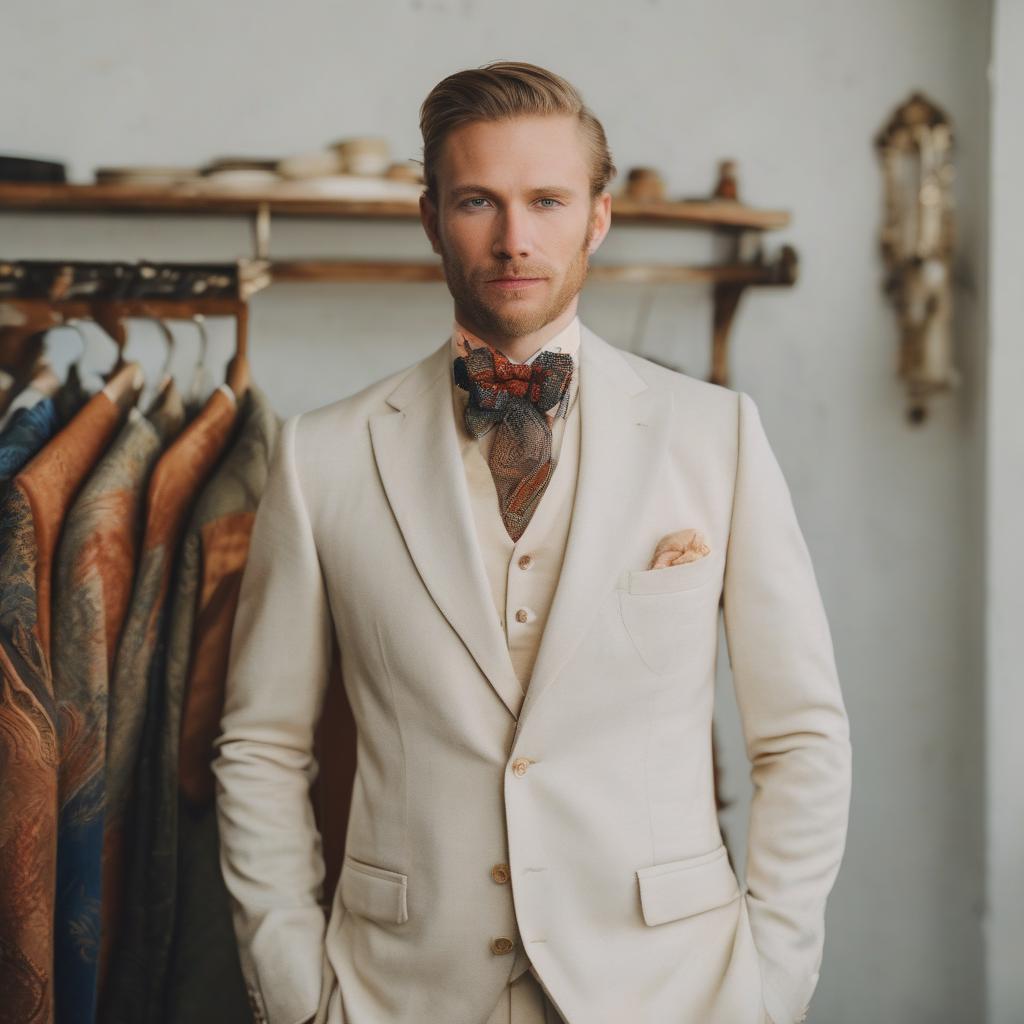}
    \end{subfigure}\hspace*{-0.25em}
\caption{
{\fontsize{11}{11}\selectfont{
\textbf{Study 2 images (gender).} \textbf{a}, 
SDXL. \textbf{b}, SDXL-Inclusive.
}}}
\label{fig:study2}
\end{figure}

\begin{figure}[htbp]
    \includegraphics[width=1\textwidth]{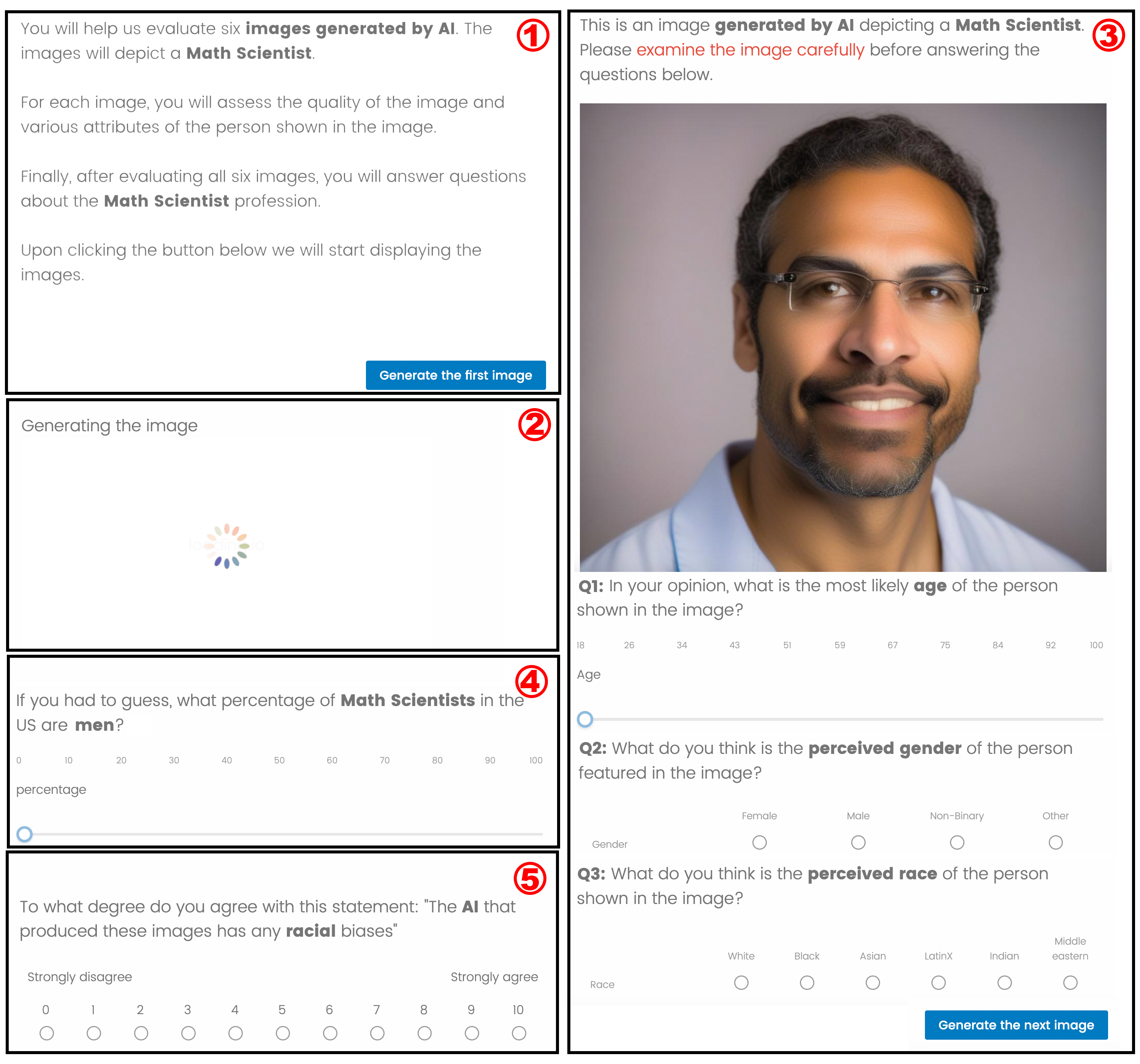}
    \caption{
    {\fontsize{11}{11}\selectfont{
    \textbf{Study 2 sample screenshots.} 
    (1) welcome screen, (2) loading screen, (3) per image questions, and (4)/(5) Final questions after seeing all six images.}}}
    \label{fig:study2_sample}
\end{figure}

\clearpage
\begin{figure}[htbp]   
\captionsetup[subfigure]{labelformat=empty}
    \centering    
    \begin{flushleft} \textbf{a} \end{flushleft}
    \begin{subfigure}{0.16\textwidth}
        \includegraphics[width=1\linewidth]{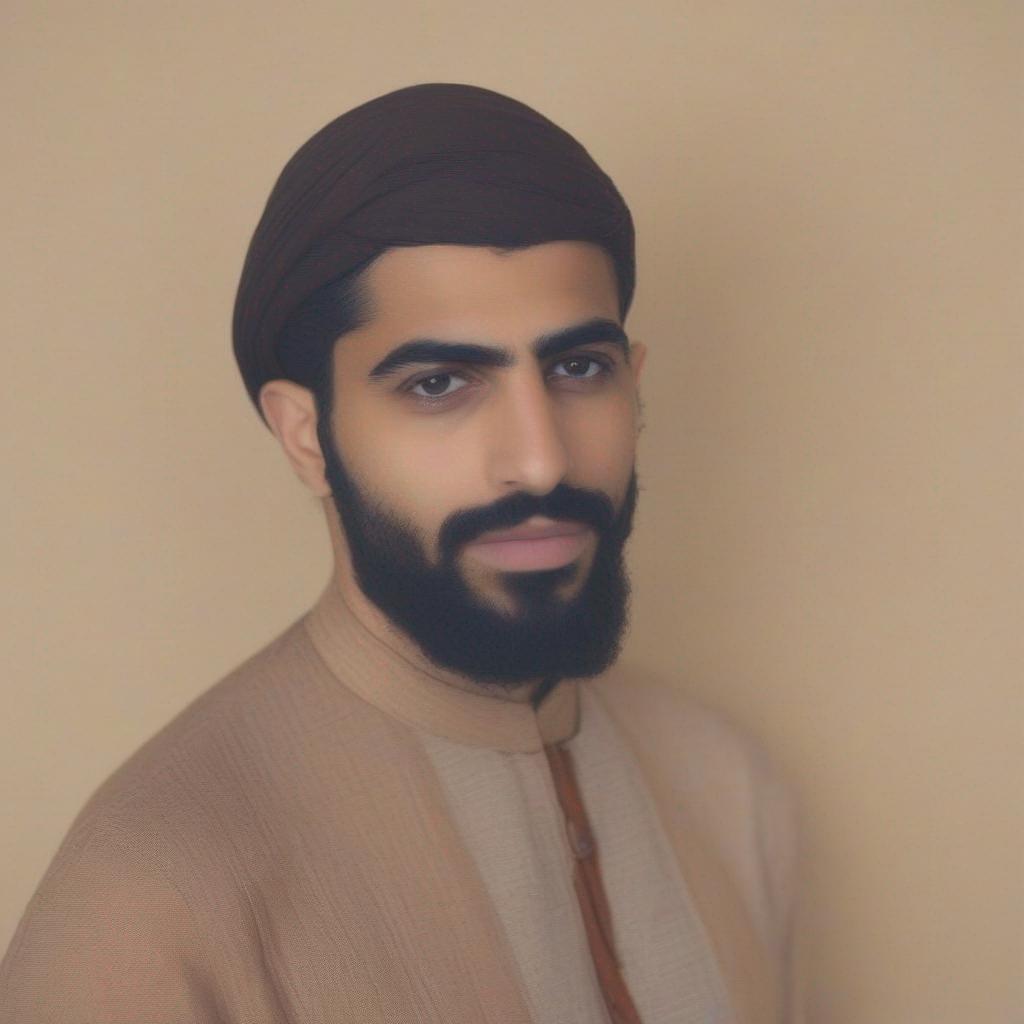}
    \end{subfigure}\hspace*{-0.25em}
    \begin{subfigure}{0.16\textwidth}
        \includegraphics[width=1\linewidth]{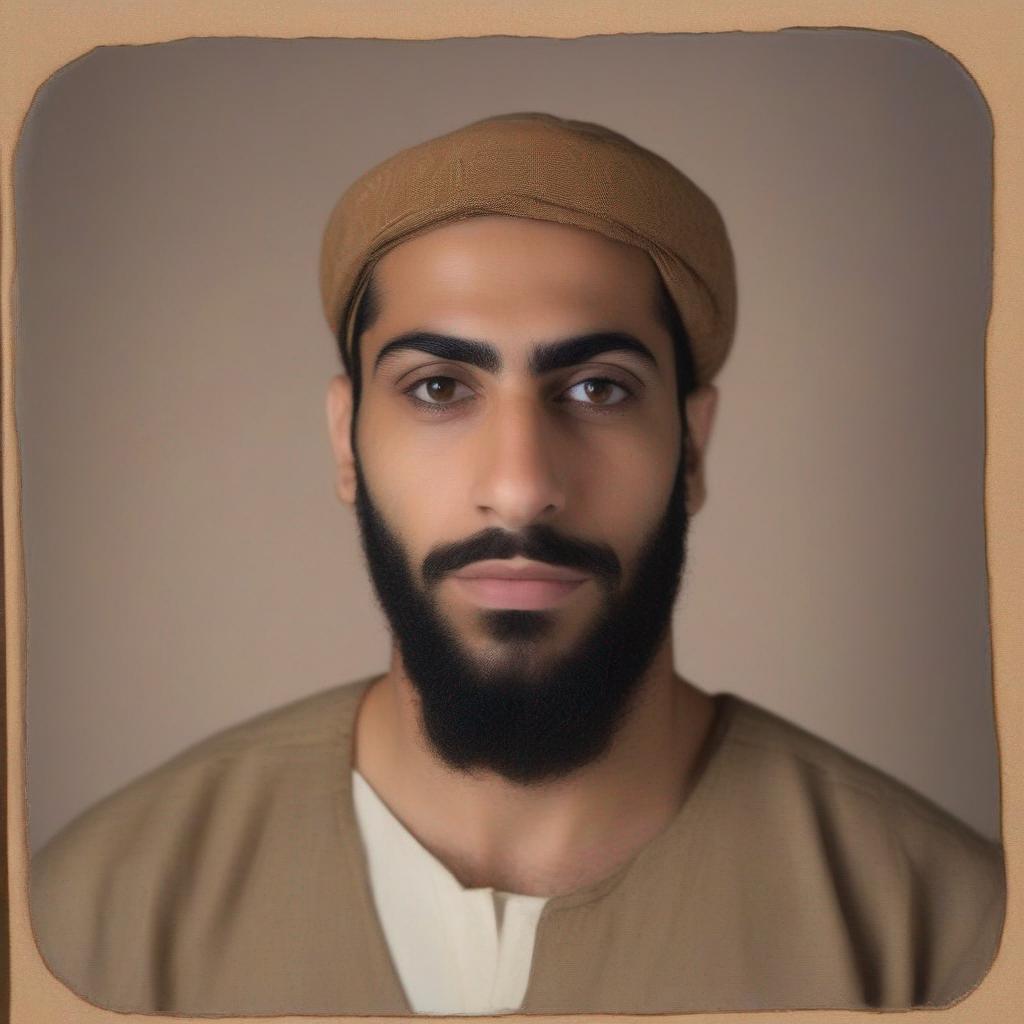}
    \end{subfigure}\hspace*{-0.25em}
    \begin{subfigure}{0.16\textwidth}
        \includegraphics[width=1\linewidth]{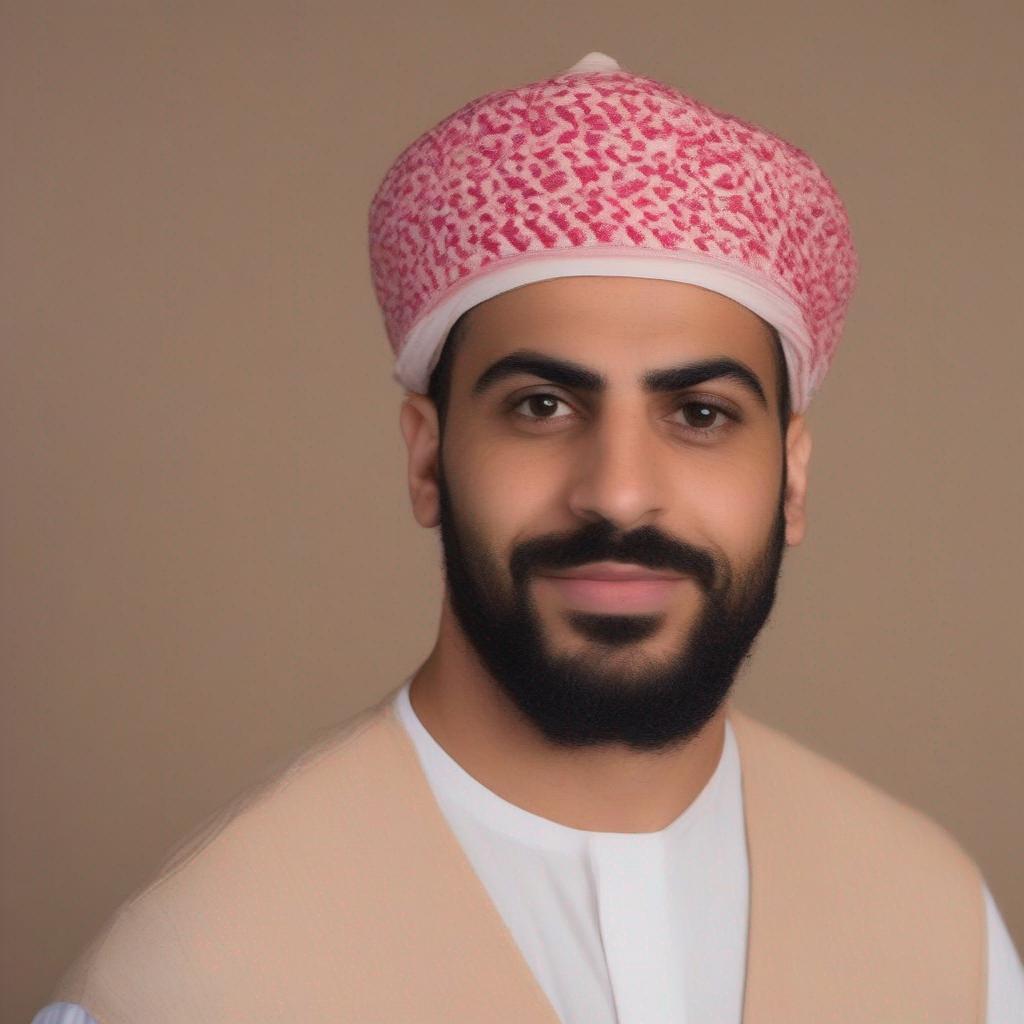}
    \end{subfigure}\hspace*{-0.25em}
    \begin{subfigure}{0.16\textwidth}
        \includegraphics[width=1\linewidth]{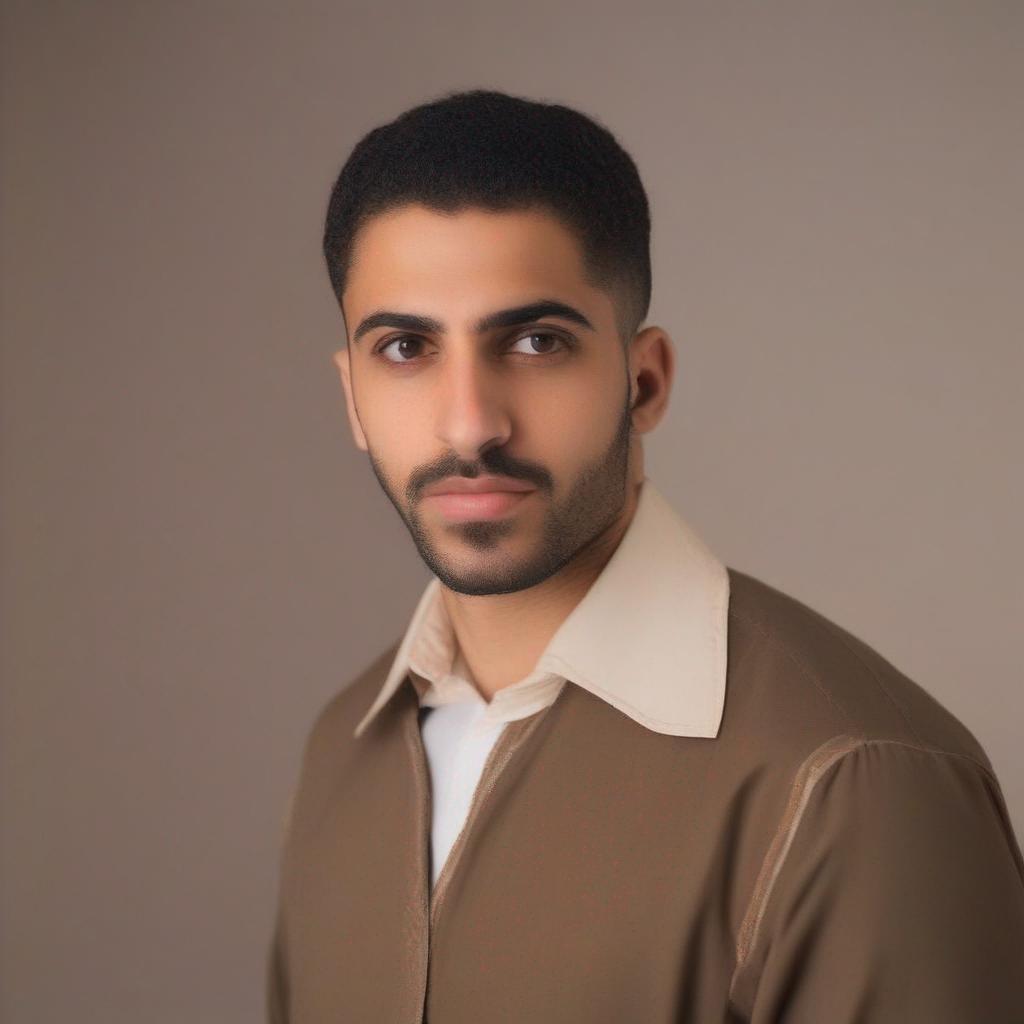}
    \end{subfigure}\hspace*{-0.25em}
    \begin{subfigure}{0.16\textwidth}
        \includegraphics[width=1\linewidth]{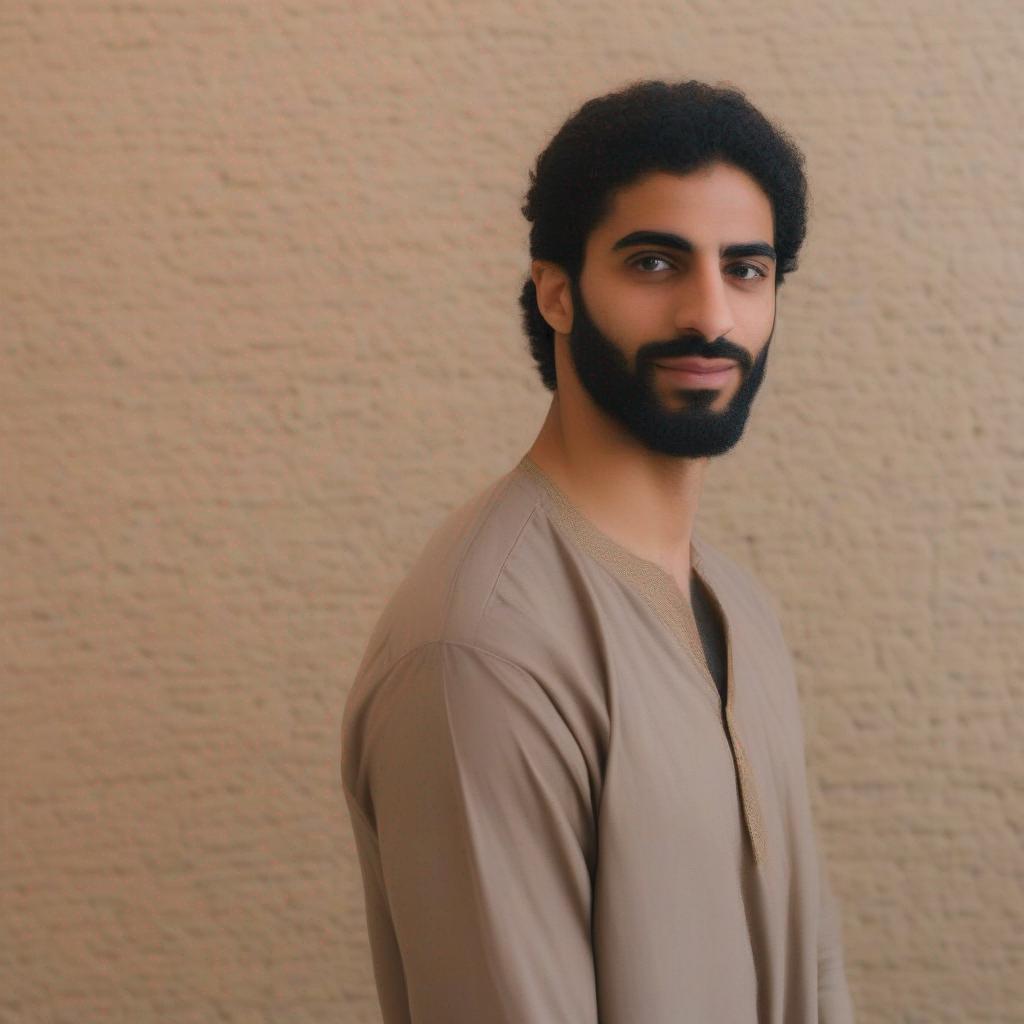}
    \end{subfigure}\hspace*{-0.25em}
    \begin{subfigure}{0.16\textwidth}
        \includegraphics[width=1\linewidth]{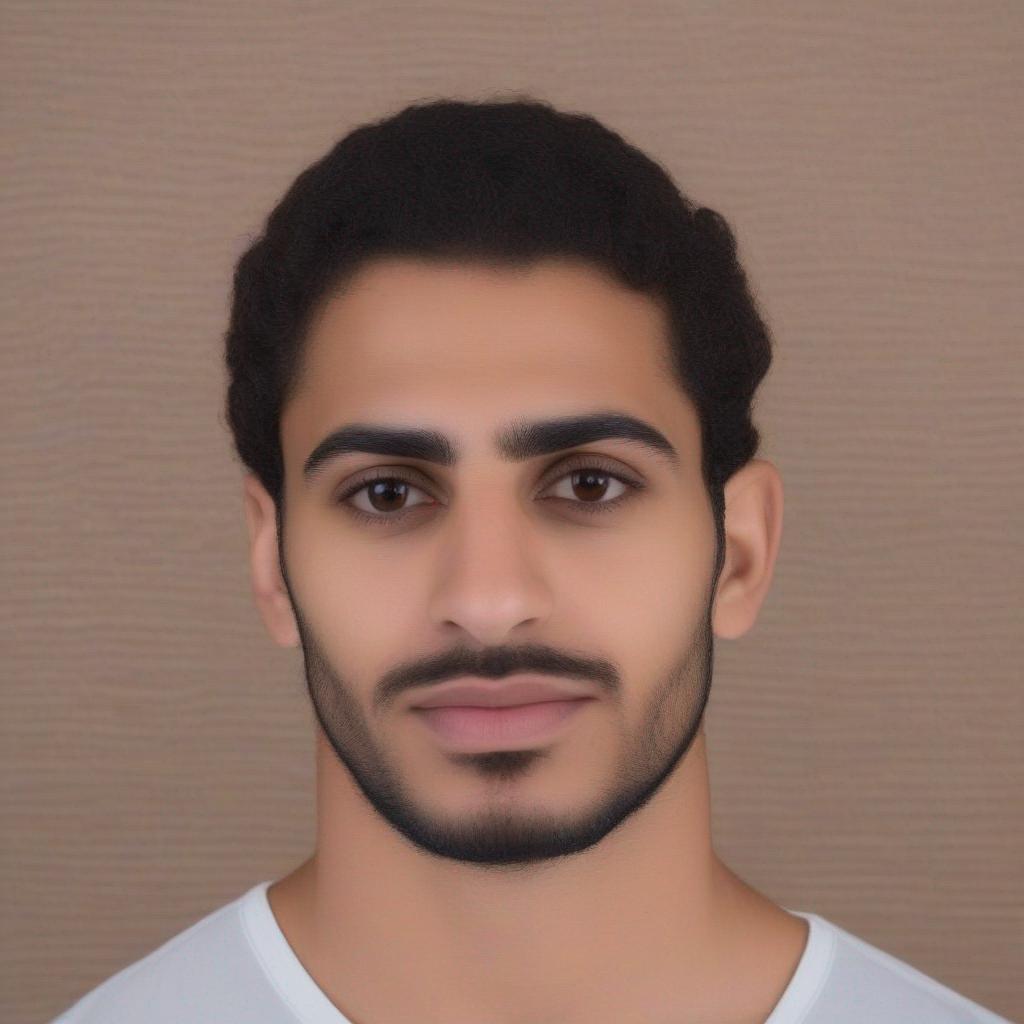}
    \end{subfigure}
    \vspace{-10pt}
    \begin{flushleft} \textbf{b} \end{flushleft}
    \begin{subfigure}{0.16\textwidth}
        \includegraphics[width=1\linewidth]{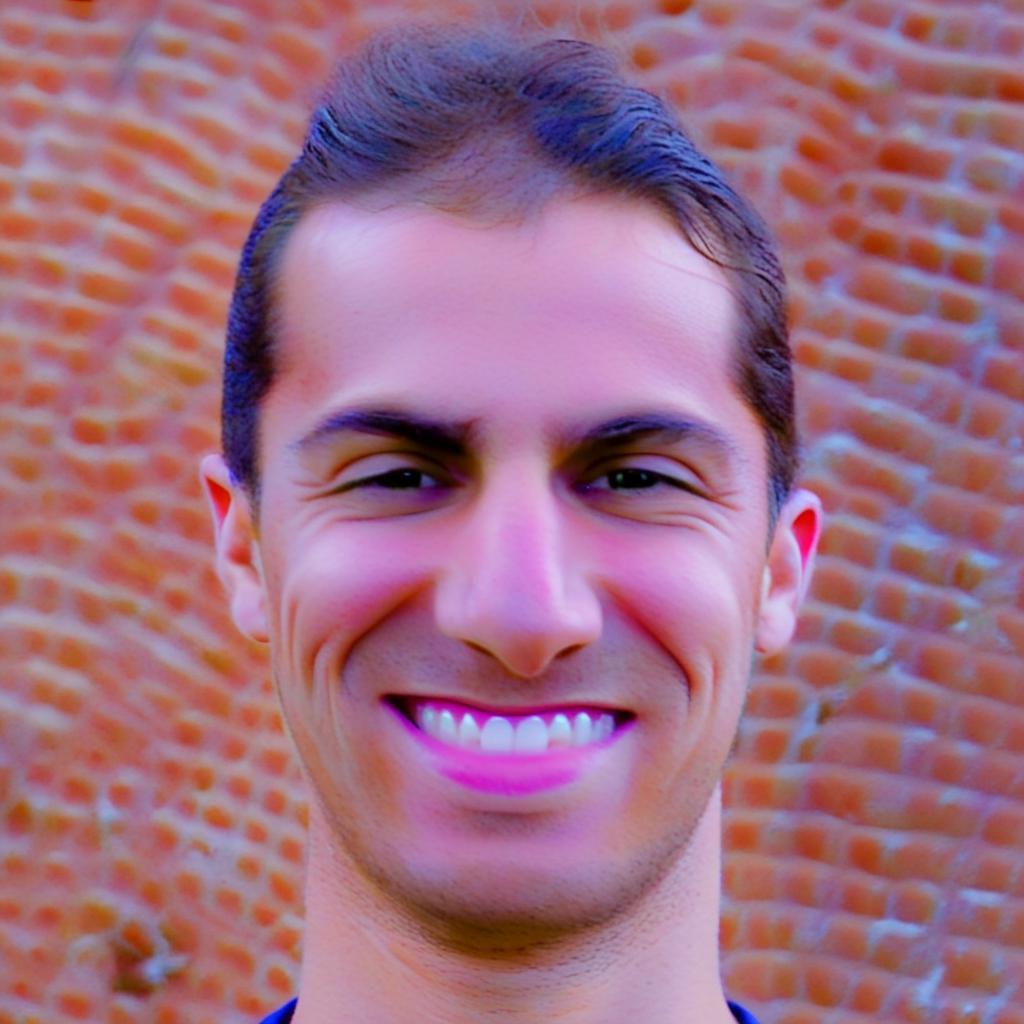}
    \end{subfigure}\hspace*{-0.25em}
    \begin{subfigure}{0.16\textwidth}
        \includegraphics[width=1\linewidth]{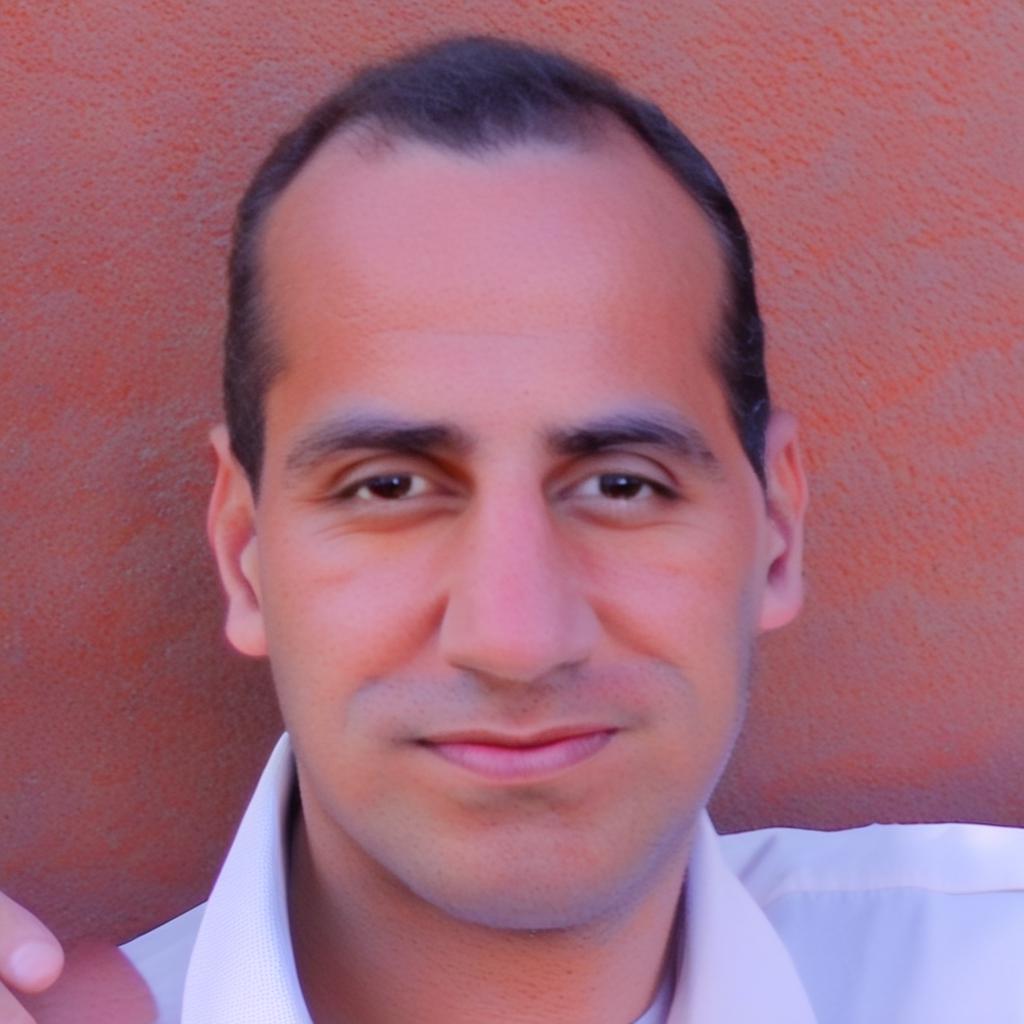}
    \end{subfigure}\hspace*{-0.25em}
    \begin{subfigure}{0.16\textwidth}
        \includegraphics[width=1\linewidth]{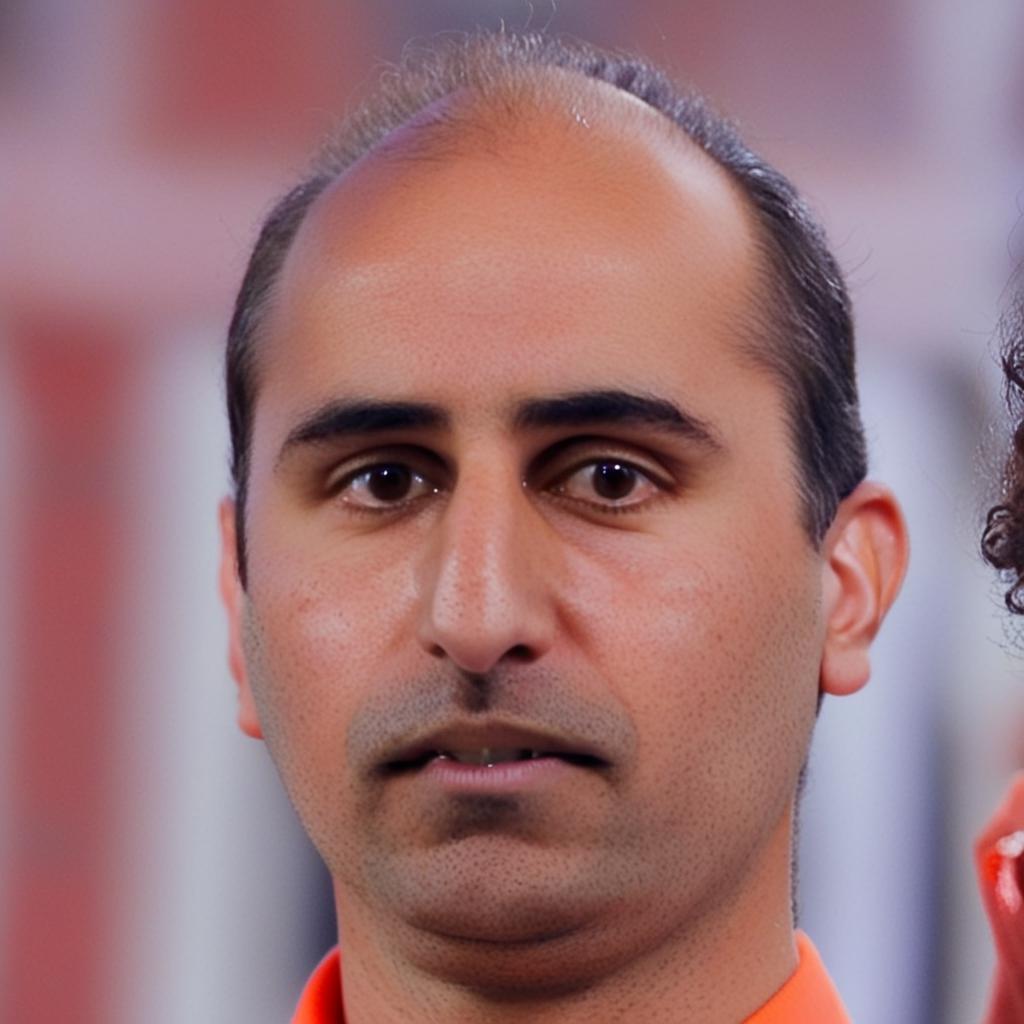}
    \end{subfigure}\hspace*{-0.25em} 
    \begin{subfigure}{0.16\textwidth}
        \includegraphics[width=1\linewidth]{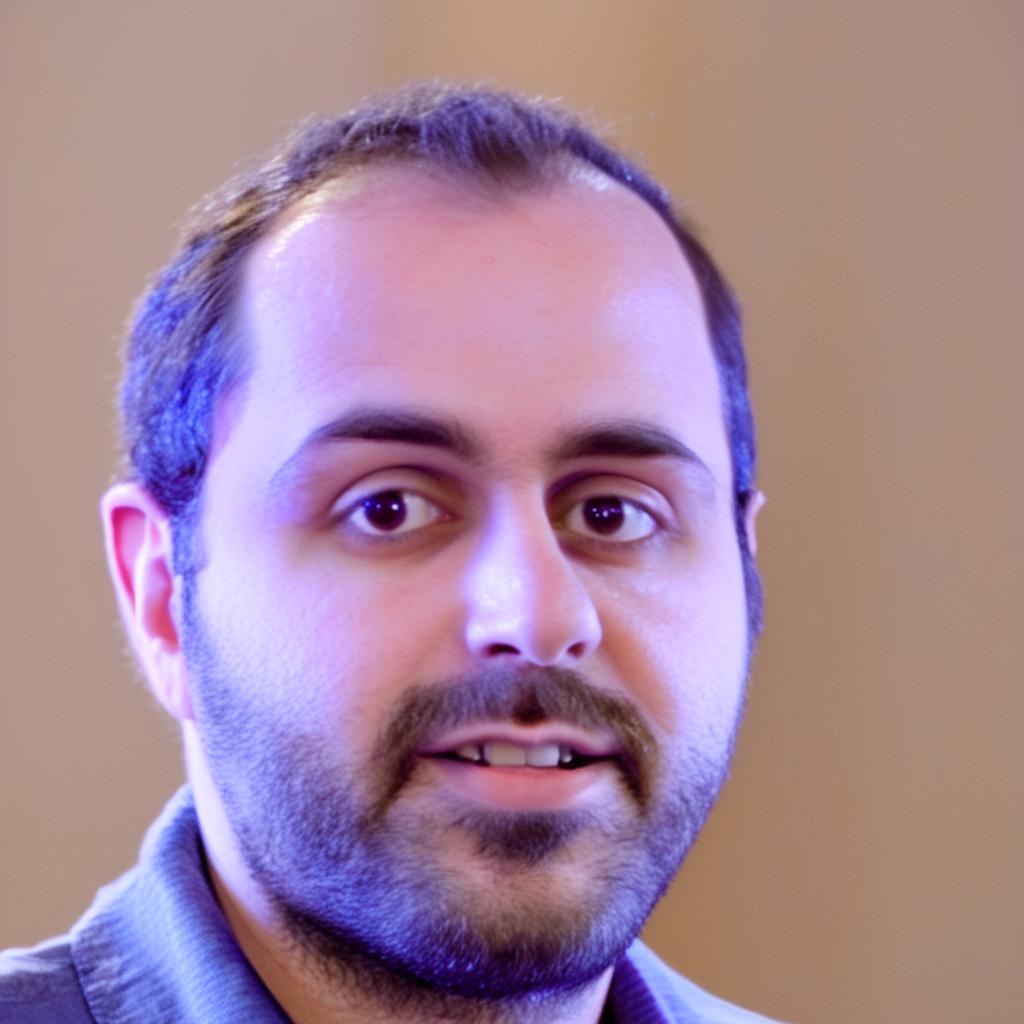}
    \end{subfigure}\hspace*{-0.25em}
    \begin{subfigure}{0.16\textwidth}
        \includegraphics[width=1\linewidth]{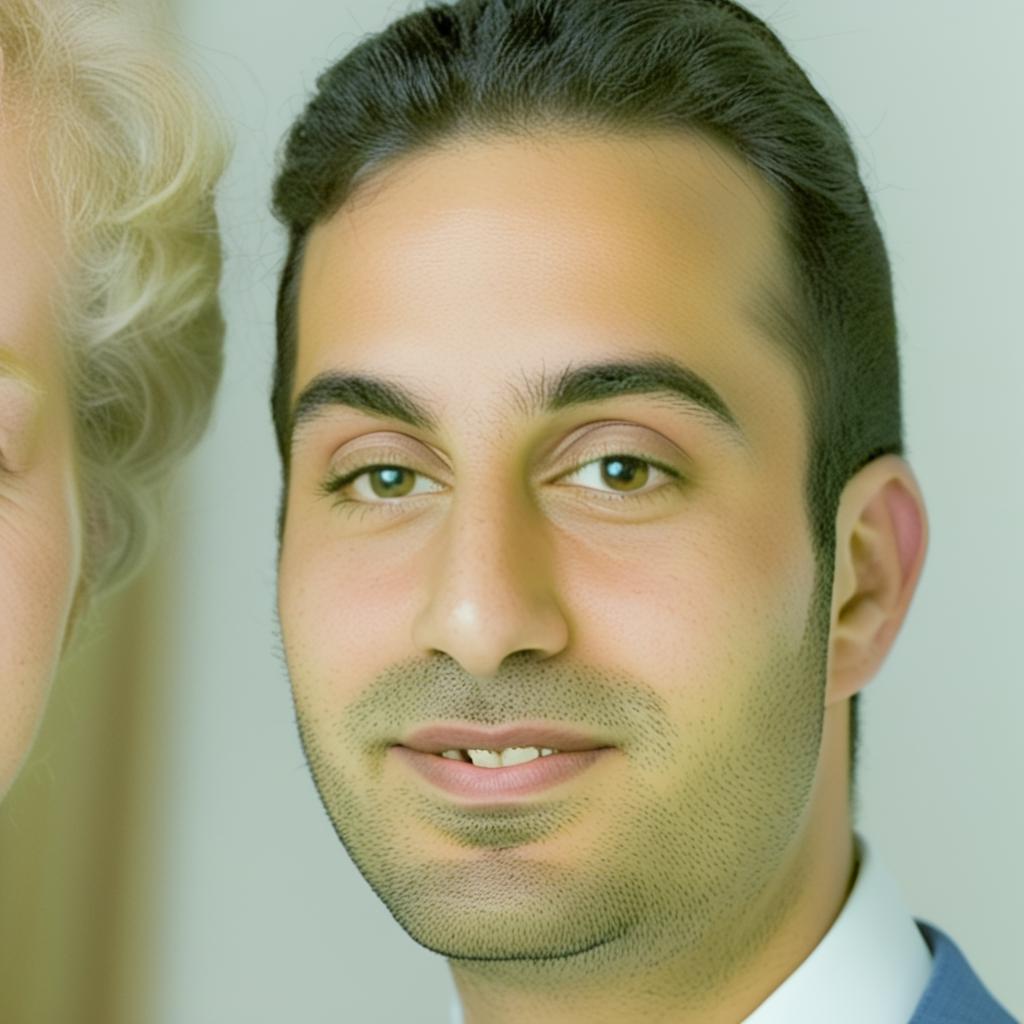}
    \end{subfigure}\hspace*{-0.25em}
    \begin{subfigure}{0.16\textwidth}
        \includegraphics[width=1\linewidth]{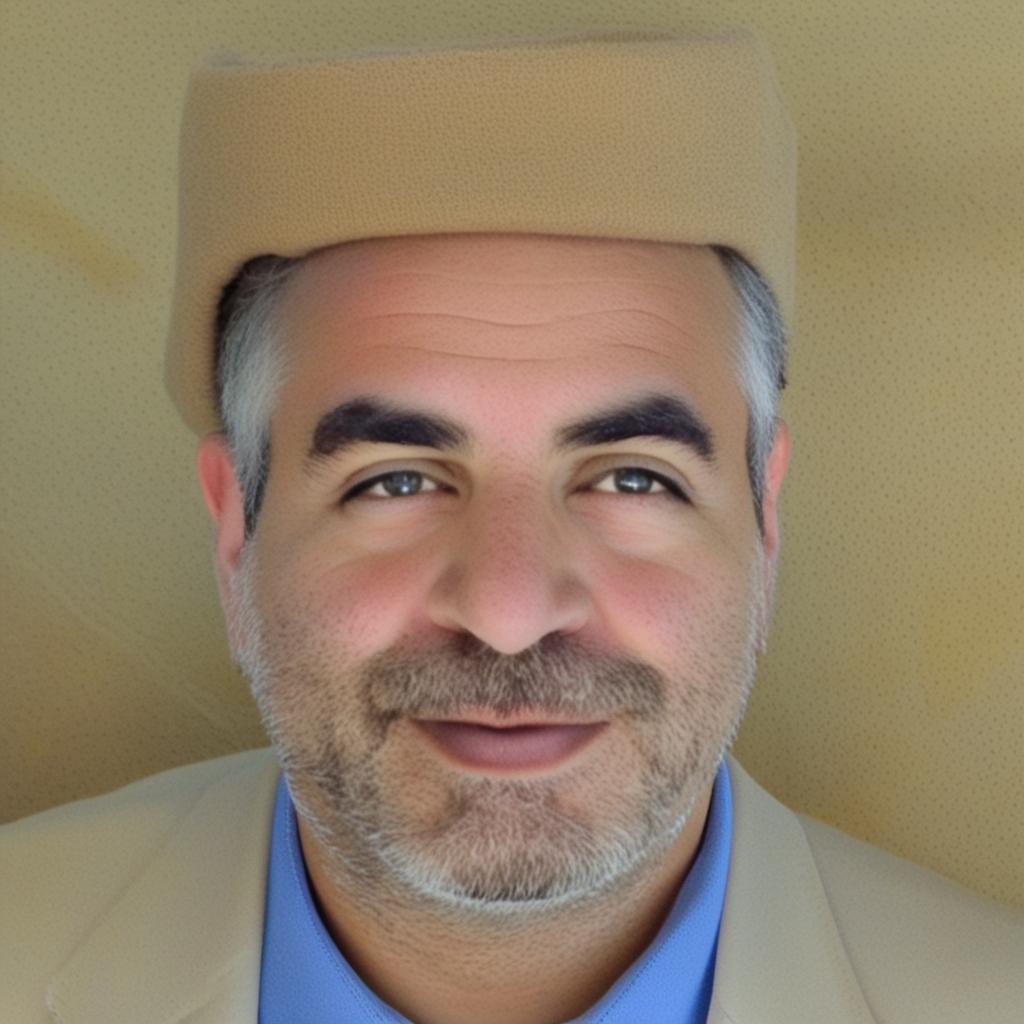}
    \end{subfigure} 
\caption{
{\fontsize{11}{11}\selectfont{
\textbf{Study 3 images (Middle eastern men).} \textbf{a}, 
SDXL. \textbf{b}, SDXL-Inclusive.
}}}
\label{fig:diversity}
\end{figure}

\begin{figure}[htbp]
    \includegraphics[width=1\textwidth]{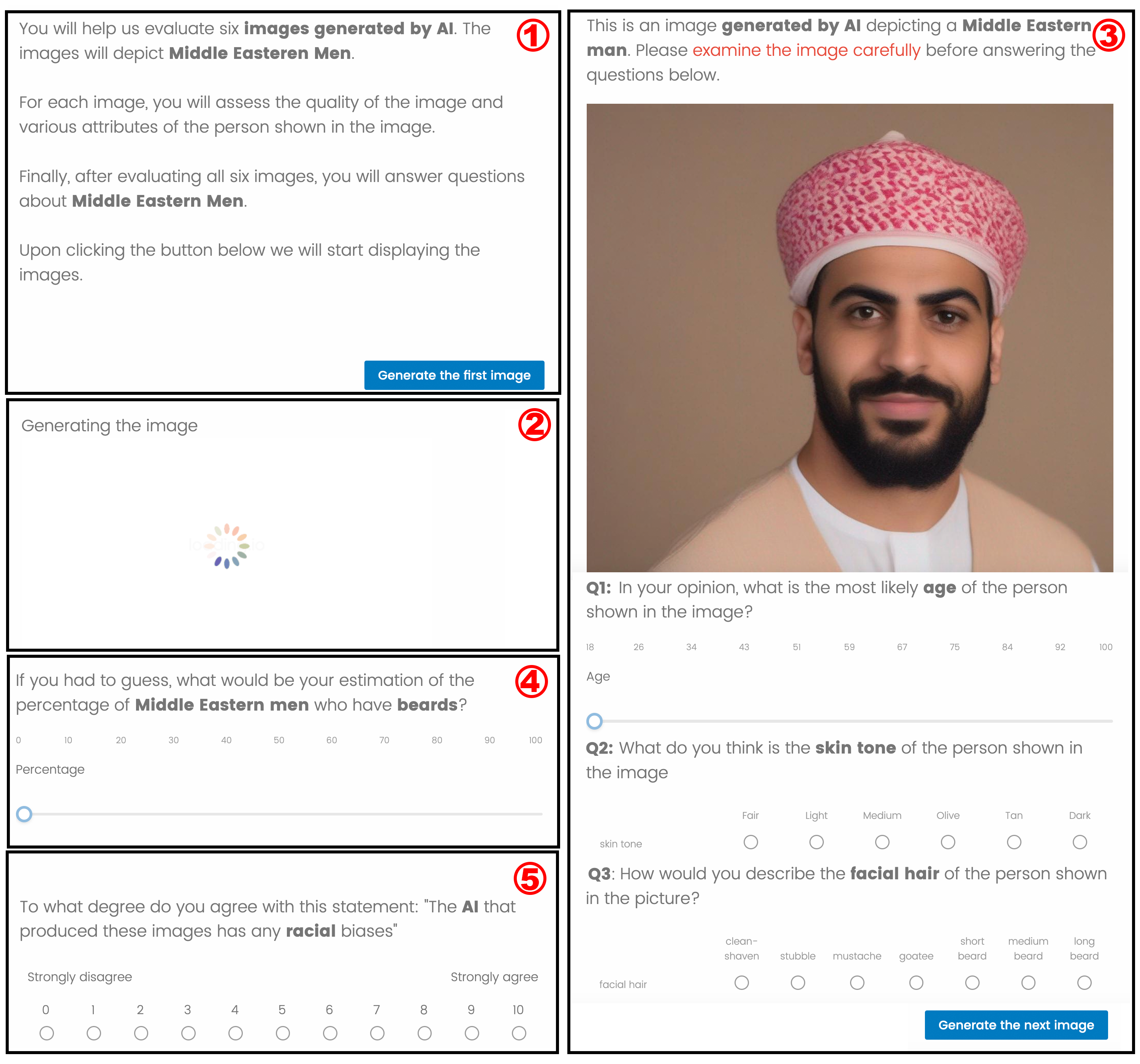}
    \caption{
    {\fontsize{11}{11}\selectfont{
    \textbf{Study 3 sample screenshots.} 
    (1) welcome screen, (2) loading screen, (3) per image questions, and (4)/(5) Final questions after seeing all six images.}}}
    \label{fig:study3_sample}
\end{figure}

\begin{figure}[htbp]   
\captionsetup[subfigure]{labelformat=empty}
    \centering    
    \begin{flushleft} \textbf{a} \end{flushleft}
    \begin{subfigure}{0.16\textwidth}
        \includegraphics[width=1\linewidth]{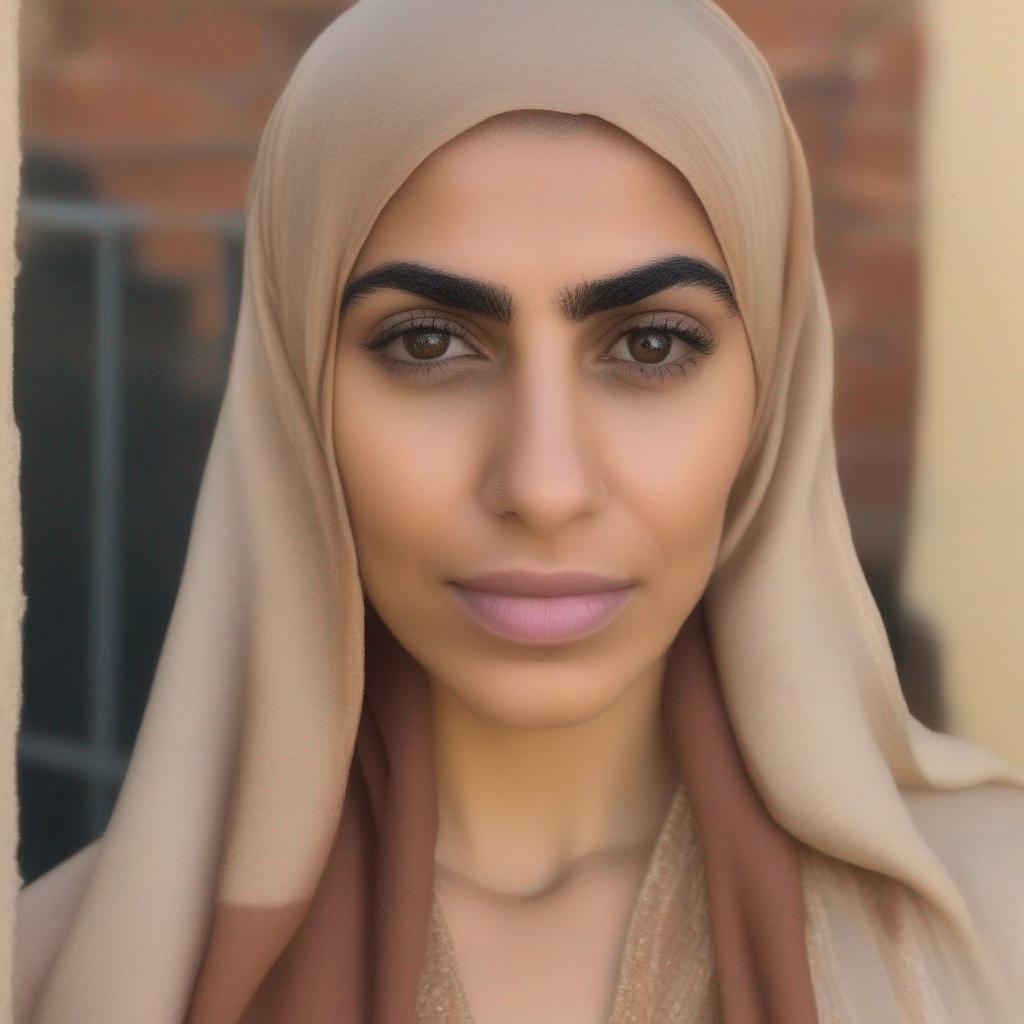}
    \end{subfigure}\hspace*{-0.25em}
    \begin{subfigure}{0.16\textwidth}
        \includegraphics[width=1\linewidth]{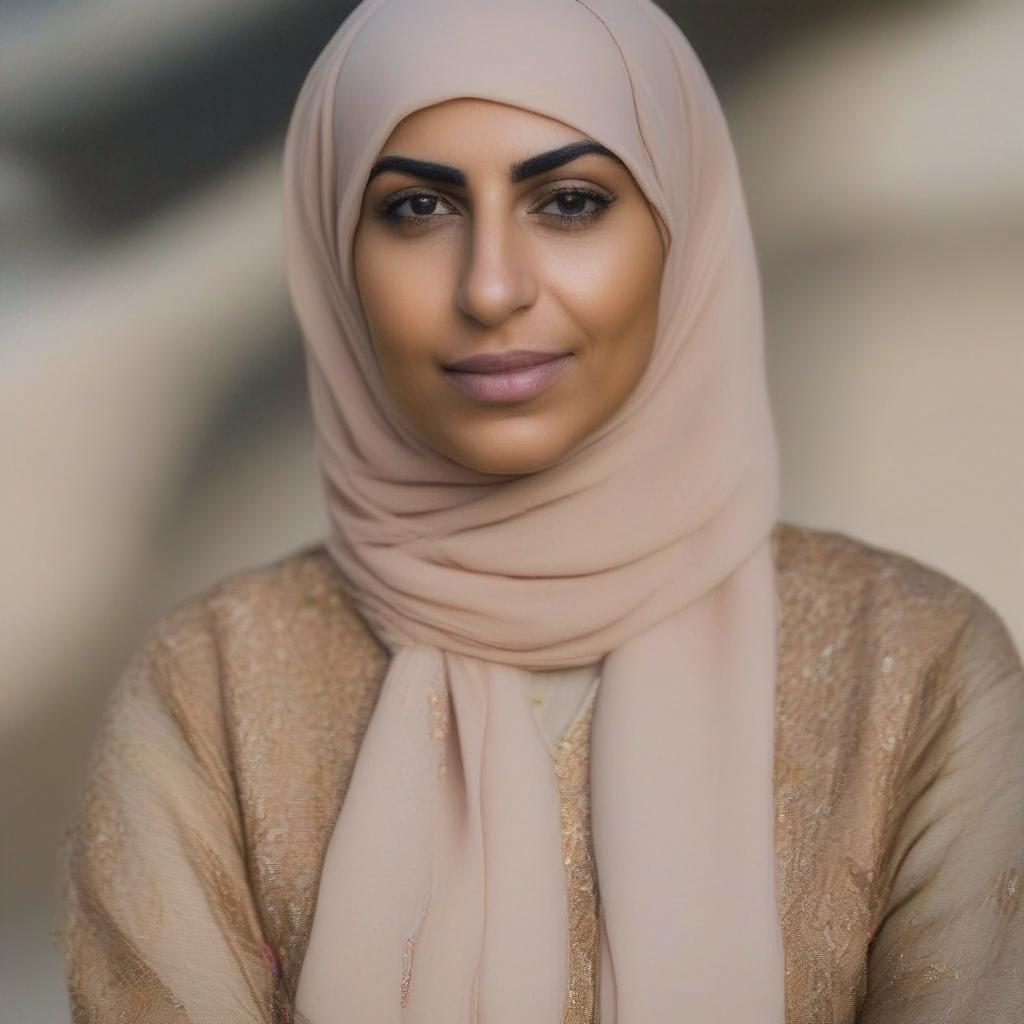}
    \end{subfigure}\hspace*{-0.25em}
    \begin{subfigure}{0.16\textwidth}
        \includegraphics[width=1\linewidth]{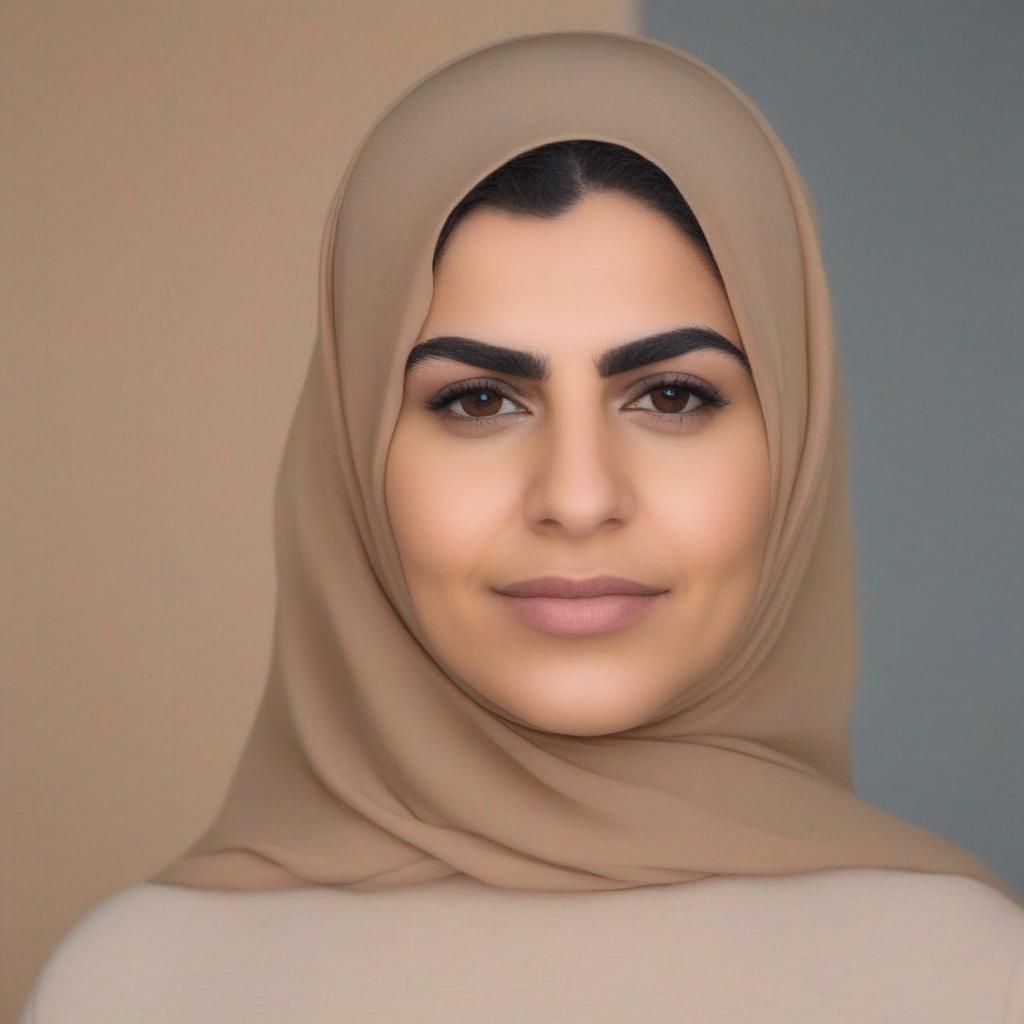}
    \end{subfigure}\hspace*{-0.25em}
    \begin{subfigure}{0.16\textwidth}
        \includegraphics[width=1\linewidth]{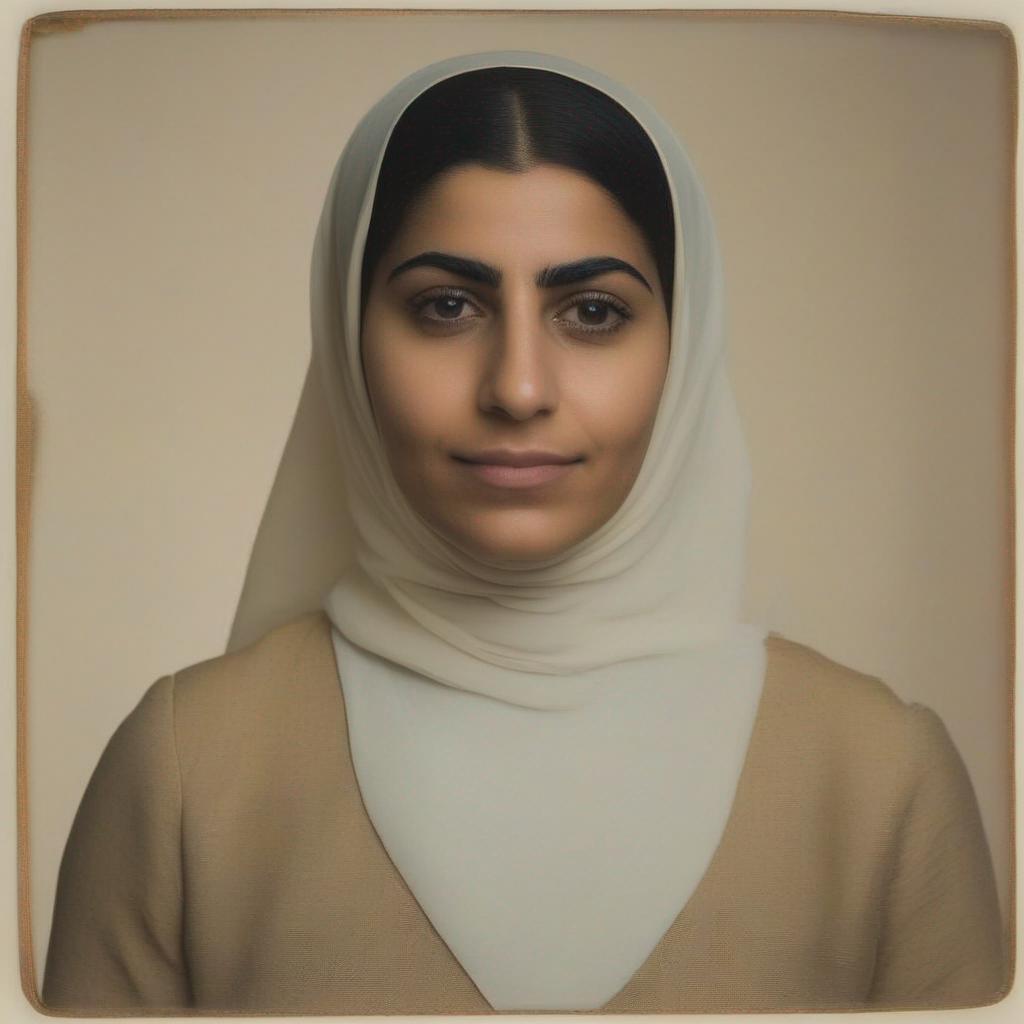}
    \end{subfigure}\hspace*{-0.25em}
    \begin{subfigure}{0.16\textwidth}
        \includegraphics[width=1\linewidth]{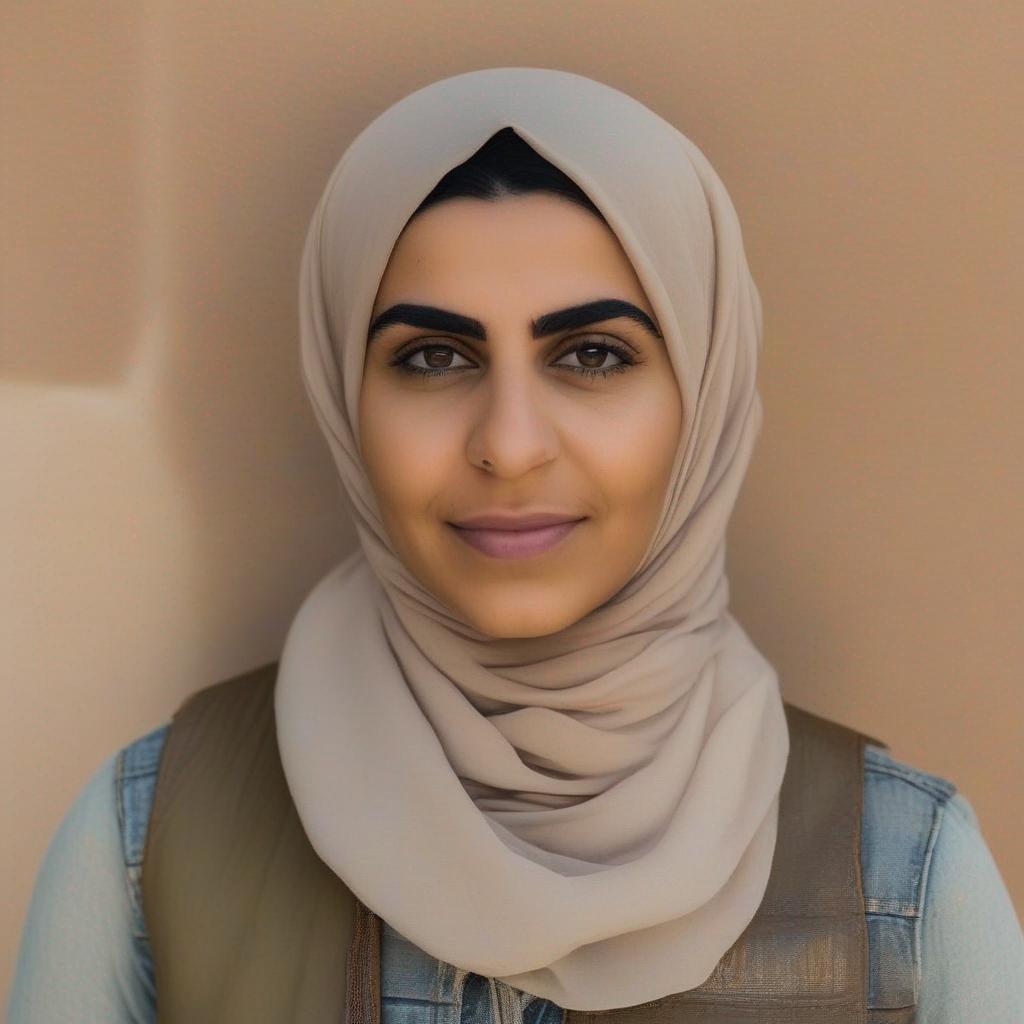}
    \end{subfigure}\hspace*{-0.25em}
    \begin{subfigure}{0.16\textwidth}
        \includegraphics[width=1\linewidth]{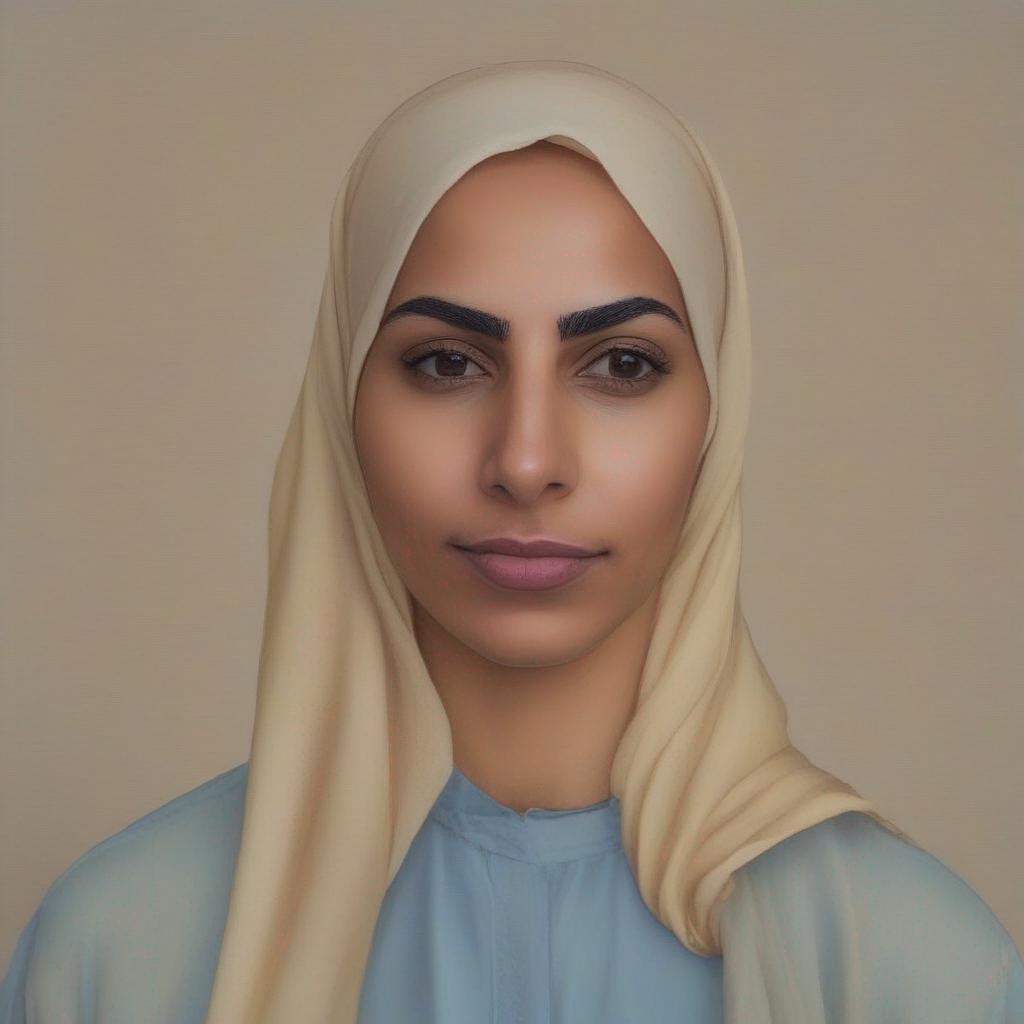}
    \end{subfigure}
    \vspace{-10pt}
    \begin{flushleft} \textbf{b} \end{flushleft}
    \begin{subfigure}{0.16\textwidth}
        \includegraphics[width=1\linewidth]{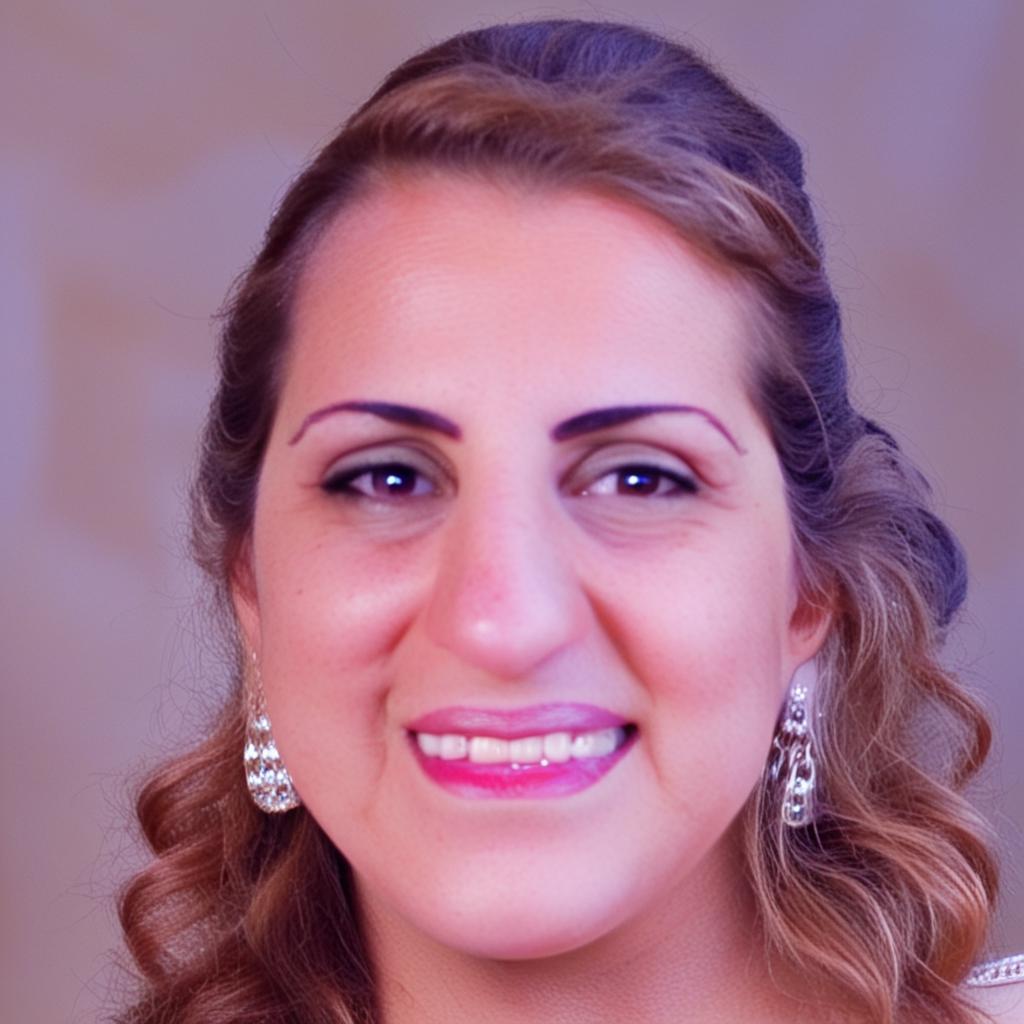}
    \end{subfigure}\hspace*{-0.25em}
    \begin{subfigure}{0.16\textwidth}
        \includegraphics[width=1\linewidth]{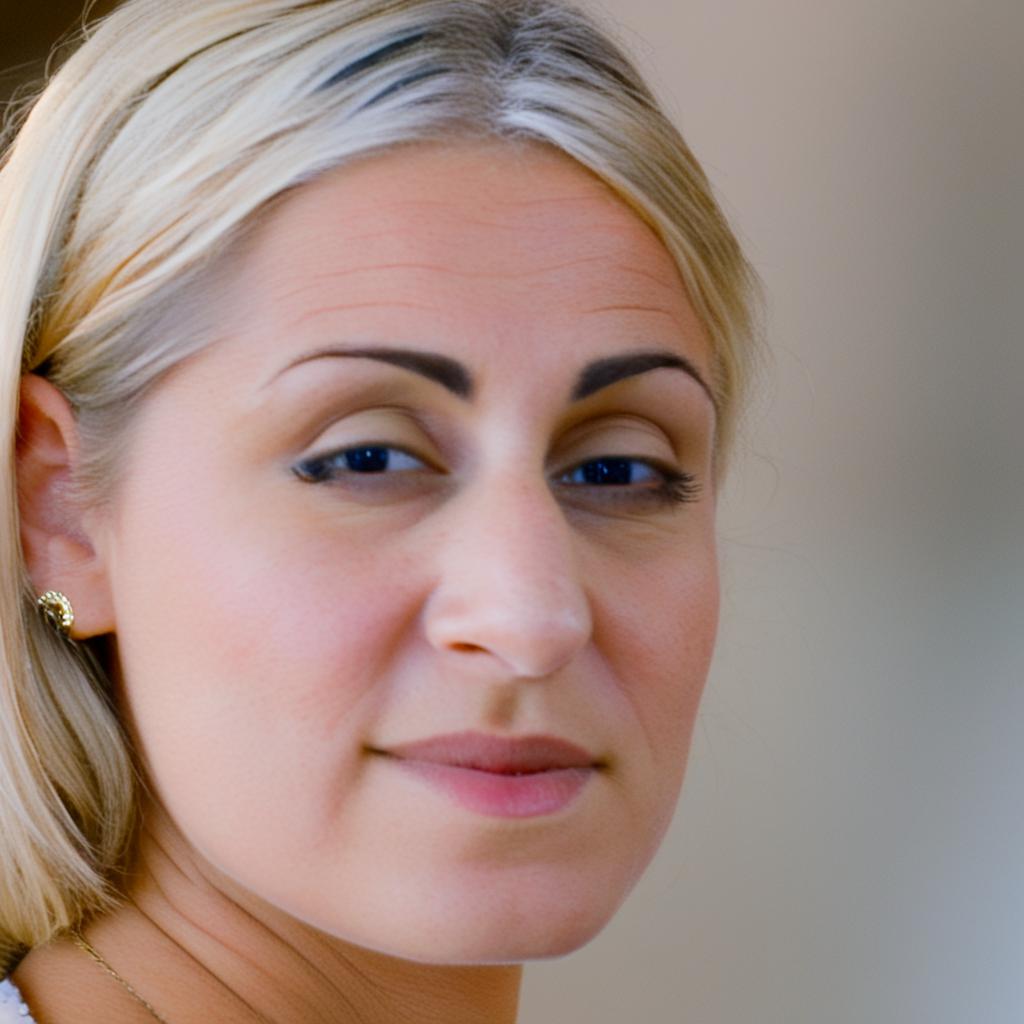}
    \end{subfigure}\hspace*{-0.25em}
    \begin{subfigure}{0.16\textwidth}
        \includegraphics[width=1\linewidth]{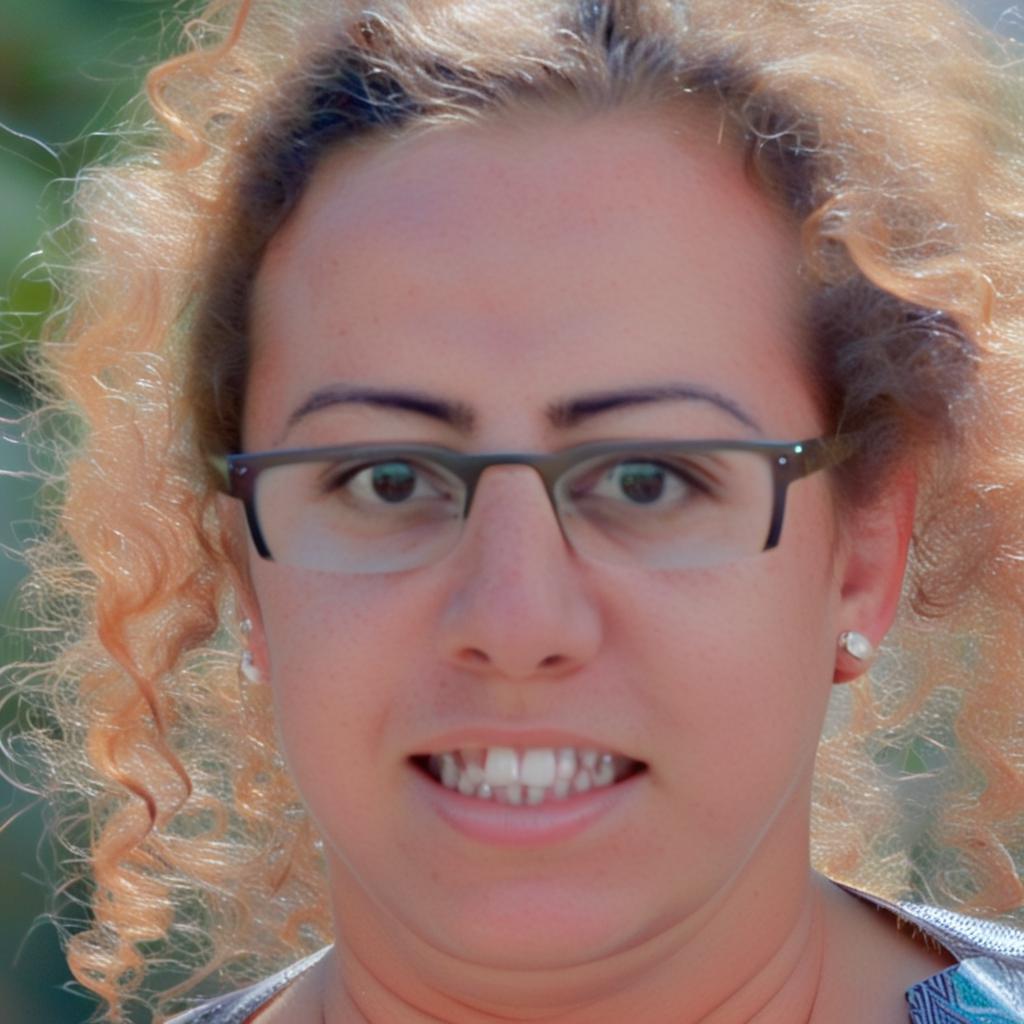}
    \end{subfigure}\hspace*{-0.25em} 
    \begin{subfigure}{0.16\textwidth}
        \includegraphics[width=1\linewidth]{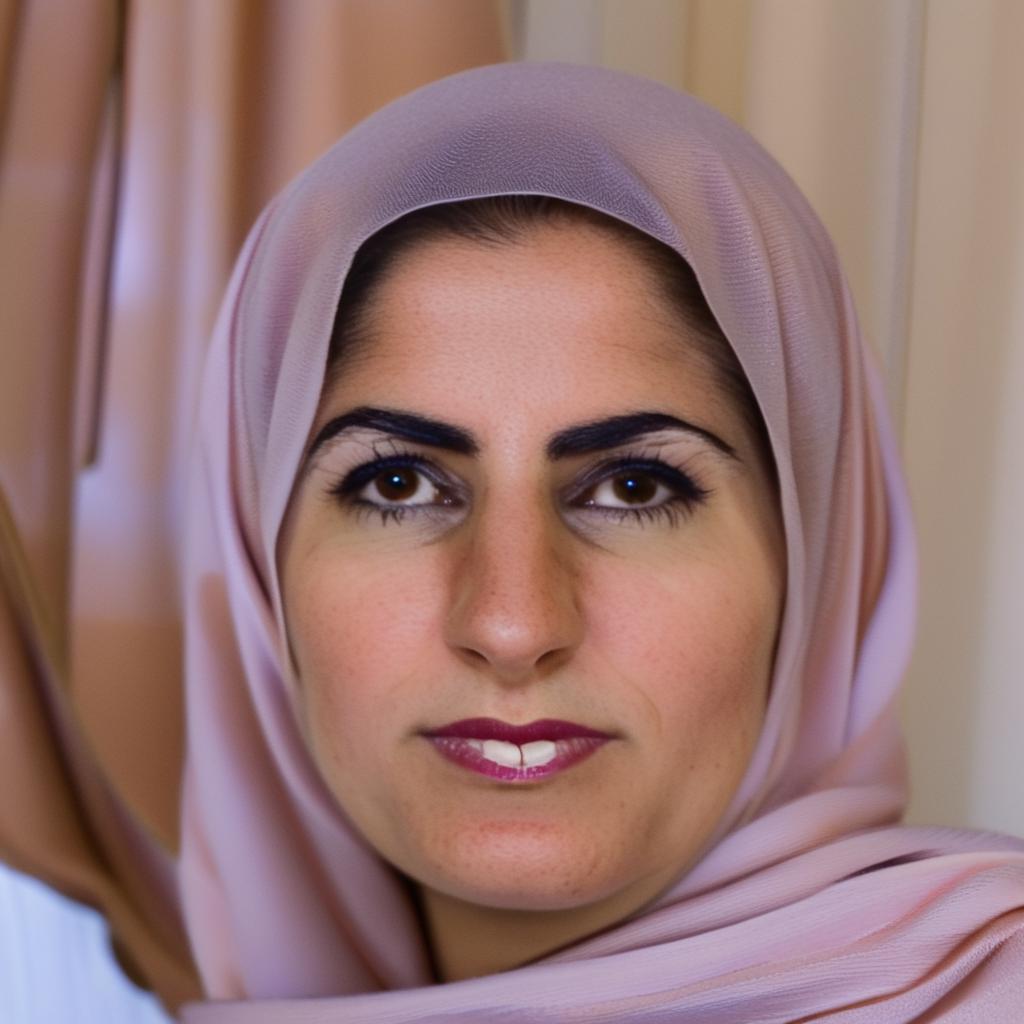}
    \end{subfigure}\hspace*{-0.25em}
    \begin{subfigure}{0.16\textwidth}
        \includegraphics[width=1\linewidth]{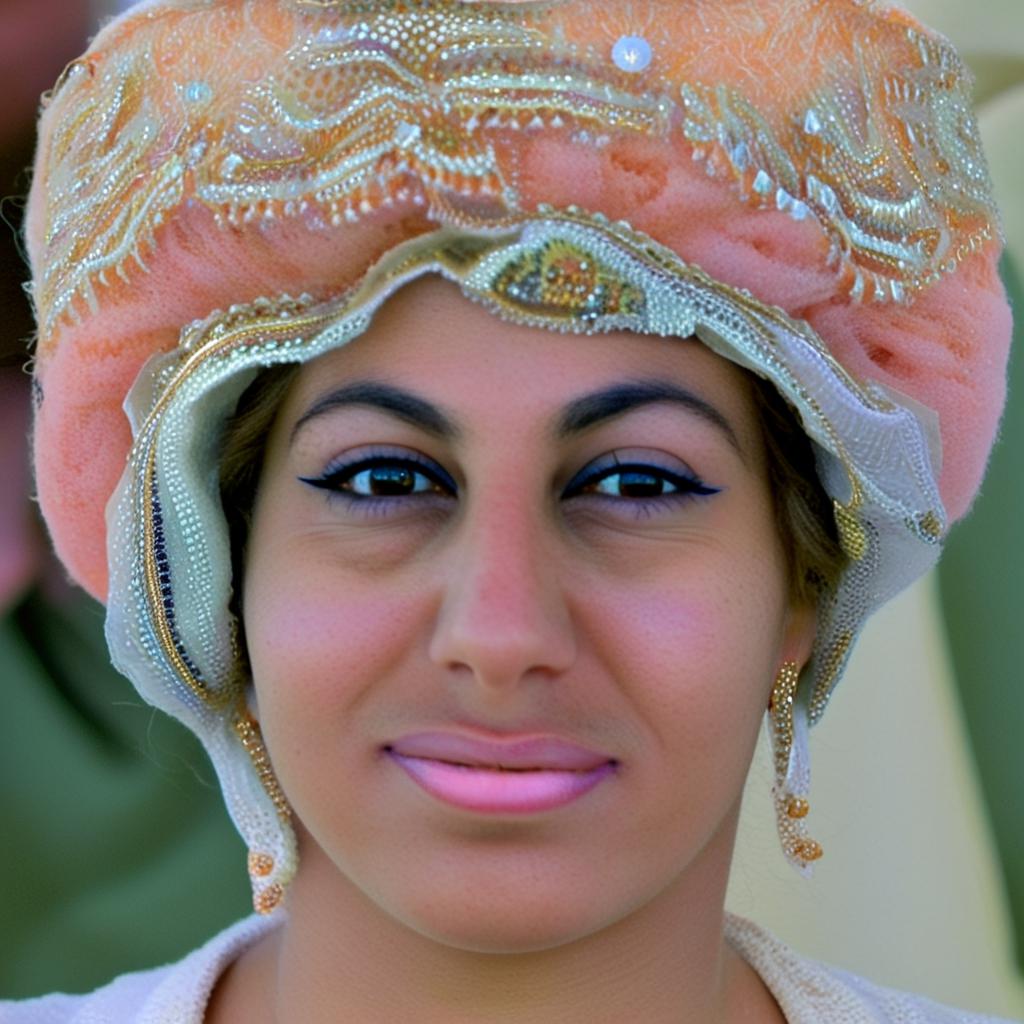}
    \end{subfigure}\hspace*{-0.25em}
    \begin{subfigure}{0.16\textwidth}
        \includegraphics[width=1\linewidth]{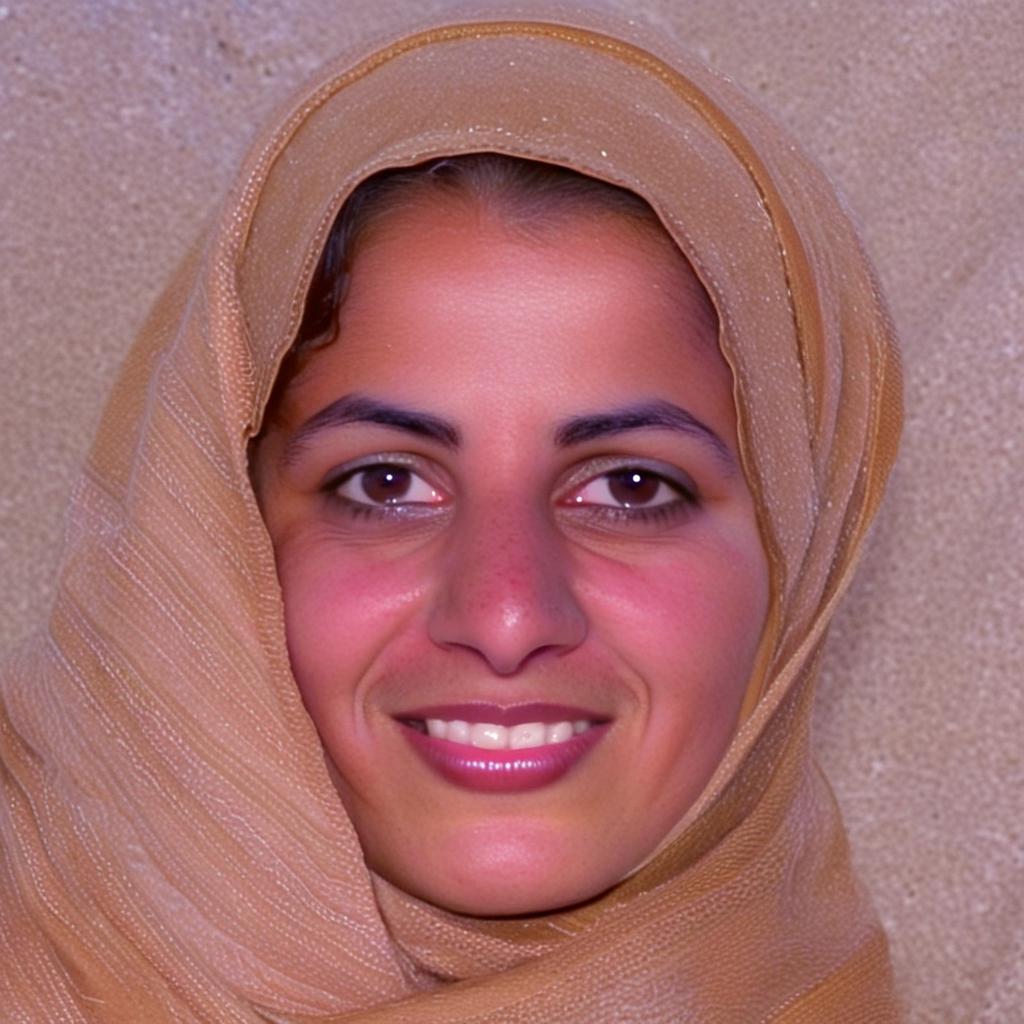}
    \end{subfigure} 
\caption{
{\fontsize{11}{11}\selectfont{
\textbf{Study 4 images (Middle eastern women).} \textbf{a}, 
SDXL. \textbf{b}, SDXL-Inclusive.
}}}
\label{fig:diversity2}
\end{figure}

\begin{figure}[htbp]
    \includegraphics[width=1\textwidth]{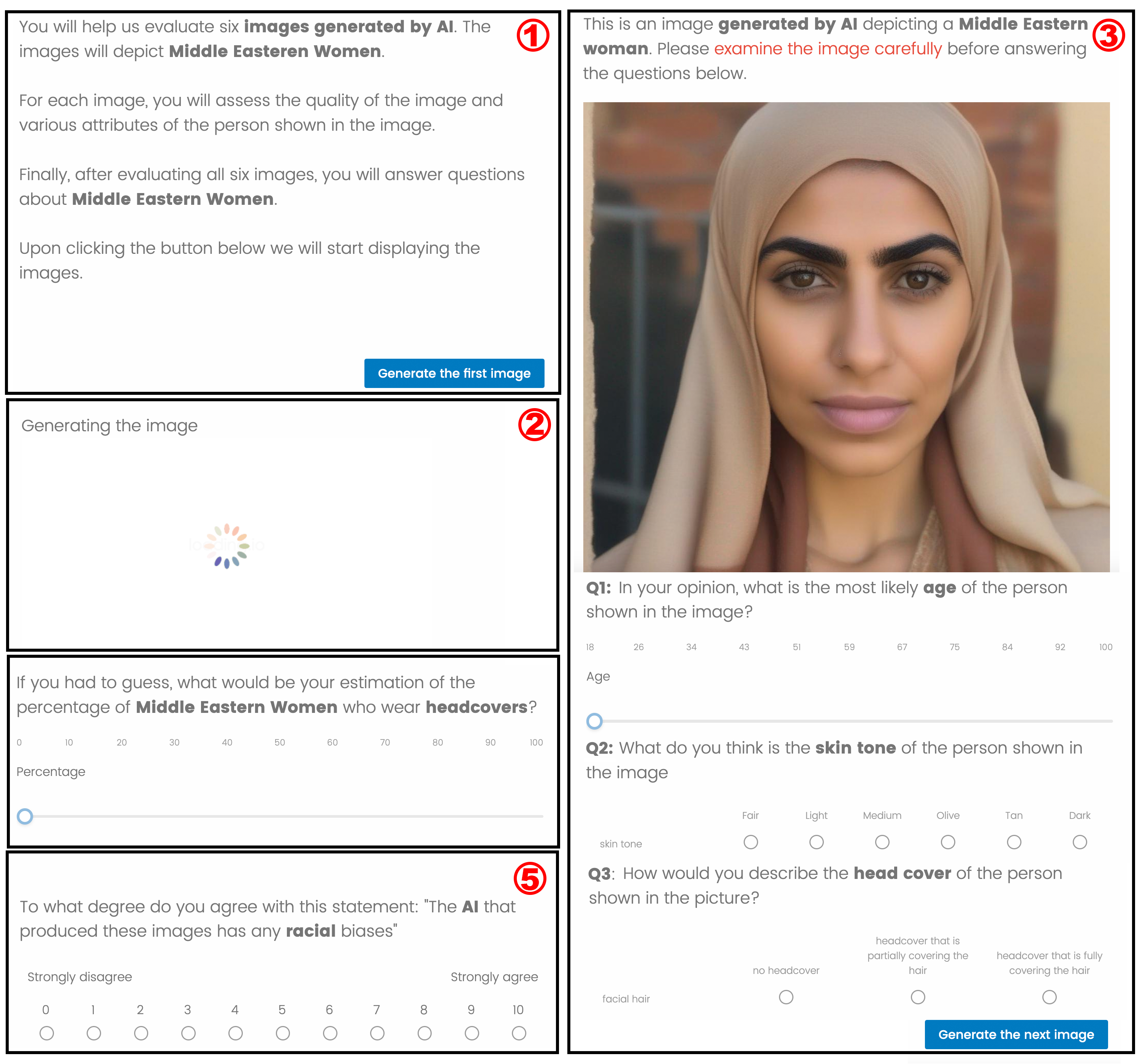}
    \caption{
    {\fontsize{11}{11}\selectfont{
    \textbf{Study 4 sample screenshots.} 
    (1) welcome screen, (2) loading screen, (3) per image questions, and (4)/(5) Final questions after seeing all six images.}}}
    \label{fig:study4_sample}
\end{figure}

\clearpage
\section{References}
\label{bibliography}
\bibliographystyle{naturemag}
\bibliography{Supplementary}